\documentclass[final]{cvpr}

\usepackage{times}
\usepackage{epsfig}
\usepackage{graphicx}
\usepackage{amsmath}
\usepackage{amssymb}
\usepackage{subcaption}
\usepackage{comment}
\usepackage{amsmath}
\usepackage{amssymb}
\usepackage{amsfonts}
\usepackage{algorithm}
\usepackage{bm}
\usepackage{multirow}
\usepackage{algpseudocode}
\usepackage{siunitx}
\usepackage{booktabs}
\renewcommand{\algorithmicrequire}{\textbf{Input:}}  

\newcommand{\nn}{\num[group-separator={,},group-minimum-digits=3]}
\newcommand{\B}{\bfseries}

\usepackage[pagebackref=true,breaklinks=true,letterpaper=true,colorlinks,bookmarks=false]{hyperref}

\def\cvprPaperID{4438} 
\def\confYear{CVPR 2021}

\begin{document}

\title{Simulating Unknown Target Models for Query-Efficient Black-box Attacks}

\author{Chen Ma,\ \ Li Chen\thanks{Corresponding author.},\ \ and Jun-Hai Yong\\
School of Software, BNRist, Tsinghua University, Beijing, China\\
{\tt\small mac16@mails.tsinghua.edu.cn, \{chenlee,yongjh\}@tsinghua.edu.cn}}

\maketitle

\begin{abstract}
Many adversarial attacks have been proposed to investigate the security issues of deep neural networks. In the black-box setting, current model stealing attacks train a substitute model to counterfeit the functionality of the target model. However, the training requires querying the target model. Consequently, the query complexity remains high, and such attacks can be defended easily. This study aims to train a generalized substitute model called ``Simulator'', which can mimic the functionality of any unknown target model. To this end, we build the training data with the form of multiple tasks by collecting query sequences generated during the attacks of various existing networks. The learning process uses a mean square error-based knowledge-distillation loss in the meta-learning to minimize the difference between the Simulator and the sampled networks. The meta-gradients of this loss are then computed and accumulated from multiple tasks to update the Simulator and subsequently improve generalization. When attacking a target model that is unseen in training, the trained Simulator can accurately simulate its functionality using its limited feedback. As a result, a large fraction of queries can be transferred to the Simulator, thereby reducing query complexity. Results of the comprehensive experiments conducted using the CIFAR-10, CIFAR-100, and TinyImageNet datasets demonstrate that the proposed approach reduces query complexity by several orders of magnitude compared to the baseline method. The implementation source code is released online\footnote{\url{https://github.com/machanic/SimulatorAttack}}.
\end{abstract}

\section{Introduction}

Deep neural networks (DNNs) are vulnerable to adversarial attacks~\cite{biggio2013evasion,goodfellow6572explaining,szegedy2014intriguing}, which add human-imperceptible perturbations to benign images for the misclassification of the \textit{target model}. The study of adversarial attacks is crucial in the implementation of robust DNNs \cite{madry2018towards}. Adversarial attacks can be categorized into two types, namely, white-box and black-box attacks. In the white-box attack setting, the target model is fully exposed to the adversary. Thus, the perturbation can be crafted easily by using gradients \cite{carlini2017towards,goodfellow6572explaining}. In the black-box attack setting, the adversary only has partial information of the target model, and adversarial examples are crafted without any gradient information. Hence, black-box attacks (\textit{i.e.,} query- and transfer-based attacks) are more practical in real-world scenarios.

\begin{figure}[t]
	\setlength{\abovecaptionskip}{0pt}%
	\setlength{\belowcaptionskip}{0pt}%
	\begin{center}
		\includegraphics[width=1\linewidth]{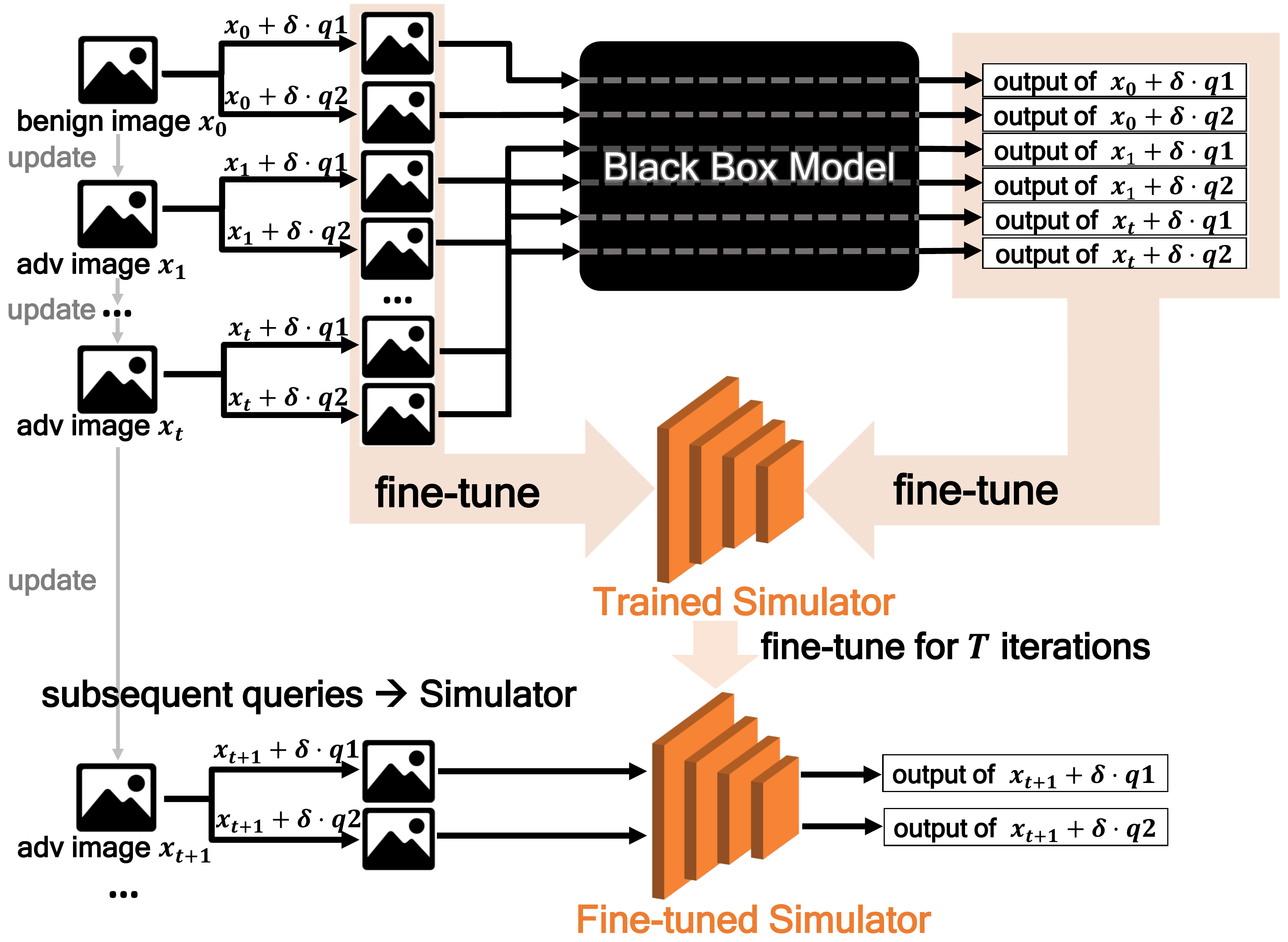}
	\end{center}
	\caption{The procedure of the Simulator Attack, where $q1$ and $q2$ are the corresponding perturbations for generating query pairs in the attack (Algorithm \ref{alg:attack}). The queries of the first $t$ iterations are fed into the target model to estimate the gradients. These queries and the corresponding outputs are collected to fine-tune the Simulator, which is trained without using the target model. The fine-tuned Simulator can accurately simulate the unknown target model, thereby transferring the queries and improving overall query efficiency.}
	\label{fig:fig1}
\end{figure}

Query-based attacks focus on estimating gradients through queries \cite{chen2017zoo,tu2019autozoom,ilyas2018blackbox,ilyas2018prior}. These attacks are considered highly effective because of their satisfactory attack success rate. However, despite their practical merits, high query complexity inevitably arises when estimating the approximate gradient with high precision, resulting in costly procedures. In addition, the queries are typically underutilized, \textit{i.e.,} the implicit but profound messages returned from the target model are overlooked, because they are abandoned after estimating the gradients. \textbf{Thus, how to make full use of the feedback of the target model to enhance the query efficiency of attacks should be thoroughly investigated.}

Transfer-based attacks generate adversarial examples by using a white-box attack method on a source model to fool the target model \cite{liu2017delving,oh2019towards,Ambra2019Why,huang2019enhancing}. Transfer-based attacks have two disadvantages: (1) they cannot achieve a high success rate, and (2) they are weak in a targeted attack. To improve transferability, model stealing attacks train a local substitute model to mimic the black-box model using a synthetic dataset, in which the labels are given by the target model through queries \cite{tramer2016stealing,papernot2017practical,orekondy2019knockoff}. In this way, the difference between the substitute and the target model is minimized, resulting in an increased attack success rate. However, such a training requires querying the target model. Consequently, the query complexity increases and such attacks can be defended easily by deploying a defense mechanism (\textit{e.g.,}~\cite{orekondy2020prediction,Lee2019DefendModelStealing}). Furthermore, the inevitable re-training to substitute a new target model is an expensive process. \textbf{Hence, how to train a substitute model without the target model requirement is worthy of further exploration.}

To eliminate the target model requirement in training, we propose a novel meta-learning-based framework to learn a generalized substitute model (\textit{i.e.,} ``Simulator'') over many different networks, thereby exploiting their characteristics to achieve fast adaptation. Once trained and fine-tuned, the Simulator can mimic the output of any target model that is unseen in training, enabling it to eventually replace the target model (Fig.~\ref{fig:fig1}). Specifically, the intermediate queries of the real black-box attack are moved to the training stage, thus allowing the Simulator to learn how to distinguish the subtle differences among queries. All the training data are reorganized into a format consisting of multiple tasks. Each task is a small data subset consisting of a query sequence of one network. In this system, a large number of tasks allow the Simulator to experience the attacks of various networks. 

\begin{figure*}[t]
	\setlength{\abovecaptionskip}{0pt}%
	\setlength{\belowcaptionskip}{0pt}%
	\begin{center}
		\includegraphics[width=1\linewidth]{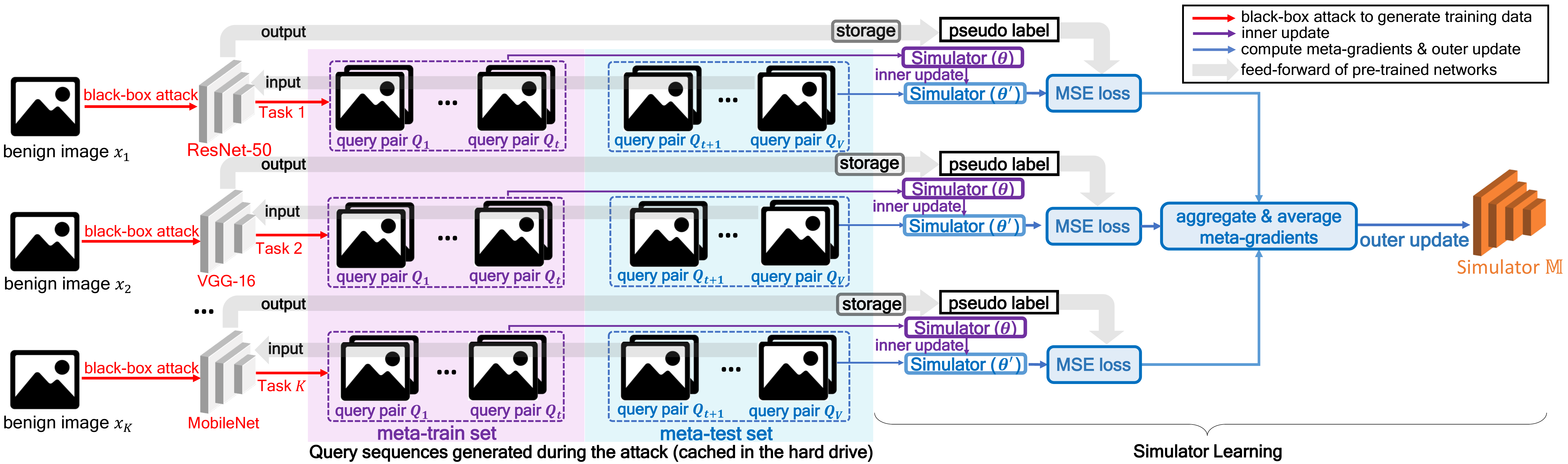}
	\end{center}
	\caption{The procedure of training the Simulator in one mini-batch. Here, the sequences of query pairs generated during the attacks are collected as training data and then reorganized into multiple tasks. Each task contains the data generated from attacking one network and is further divided into meta-train set and meta-test set. Next, the Simulator network $\mathbb{M}$ reinitializes its weights to $\theta$ at the beginning of learning each task, after which it subsequently trains on the meta-train set. After several iterations (inner update), $\mathbb{M}$ converges and its weights are updated to $\theta^\prime$. The meta-gradients of $\mathbb{M}$ are computed based on the meta-test sets of $K$ tasks and are then accumulated to update $\mathbb{M}$ (the outer update). The updated $\mathbb{M}$ is prepared for the next mini-batch learning. Finally, the learned Simulator can simulate any unknown black-box model using limited queries in the attack stage.} 	
	\label{fig:meta-train}
\end{figure*}

We propose three components to optimize the generalization. First, a query-sequence level partition strategy is adopted to divide each task into meta-train and meta-test sets (Fig. \ref{fig:meta-train}) that match the iterations of fine-tuning and simulation in the attack, respectively (Fig. \ref{fig:fig1}). Second, the mean square error (MSE)-based knowledge-distillation loss carries out the inner and outer loops of meta-learning. Finally, the meta-gradients of a batch of tasks are computed and then aggregated to update the Simulator and improve generalization. These strategies well address the problem of the target model requirement during training. In the attack (named ``Simulator Attack''), the trained Simulator is fine-tuned using the limited feedback of the unknown target model to accurately simulate its output, thereby transferring its query stress (Fig. \ref{fig:fig1}). Therefore, the feedback of the target model is fully utilized to improve query efficiency. In the proposed approach, the elimination of target models in training poses a new security threat, \textit{i.e.,} the adversary with the minimal information about the target model can also counterfeit this model for a successful attack.

In this study, we evaluate the proposed method using the CIFAR-10 \cite{krizhevsky2009learning}, CIFAR-100 \cite{krizhevsky2009learning}, and TinyImageNet~\cite{russakovsky2015imagenet} datasets and compare it with natural evolution strategies (NES)~\cite{ilyas2018blackbox}, Bandits~\cite{ilyas2018prior}, Meta Attack~\cite{du2020queryefficient}, random gradient-free (RGF)~\cite{nesterov2017random}, and prior-guided RGF (P-RGF)~\cite{cheng2019improving}. Experimental results show that the Simulator Attack can significantly reduce query complexity compared with the baseline method. 

The main contributions of this work are summarized as follows:

(1) We propose a novel black-box attack by training a generalized substitute model named ``Simulator''. The training uses a knowledge-distillation loss to carry out the meta-learning between the Simulator and the sampled networks. After training, the Simulator only requires a few queries to accurately mimic any target model that is unseen in training.

(2) We identify a new type of security threat upon eliminating the target models in training: the adversary with the minimal information about the target model can also counterfeit this model for achieving the query-efficient attack.

(3) By conducting extensive experiments using the CIFAR-10, CIFAR-100, and TinyImageNet datasets, we demonstrate that the proposed approach achieves similar success rates as those of state-of-the-art attacks but with an unprecedented low number of queries.

\section{Related Works}

\noindent\textbf{Query-based Attacks.} Black-box attacks can be divided into query- and transfer-based attacks. Query-based attacks can be further divided into score- and decision-based attacks based on how much returned information from the target model can be used by the adversary. In score-based attacks, the adversary uses the output scores of the target model to generate adversarial examples.
Most score-based attacks estimate the approximate gradient through zeroth-order optimizations~\cite{chen2017zoo,bhagoji2018practical}. Then, the adversary can optimize the adversarial example with the estimated gradient.
Although this type of approach can deliver a successful attack, it requires a large number of queries as each pixel needs two queries. Several improved methods have been introduced in the literature to reduce query complexity by using the principal components of the data \cite{bhagoji2018practical}, a latent space with reduced dimension~\cite{tu2019autozoom}, prior gradient information~\cite{ilyas2018prior,ma2020switching}, random search \cite{guo2019simple,ACFH2020square}, and active learning \cite{pengcheng2018query}. Decision-based attacks \cite{chen2019hopskipjumpattack,cheng2018queryefficient} only use the output label of the target model. In this study, we focus on the score-based attacks.

\noindent\textbf{Transfer-based Attacks.} Transfer-based attacks generate adversarial examples on a source model and then transfer them to the target model \cite{liu2017delving,Ambra2019Why,huang2019enhancing}. However, this type of attack cannot achieve a high success rate due to the large difference between the source model and the target model. Many efforts, including the use of model stealing attacks, have been made to improve the attack success rate. The original goal of model stealing attacks is to replicate the functionality of public service \cite{wang2018stealing,tramer2016stealing,milli2019model,orekondy2019knockoff}.  Papernot \etal \cite{papernot2017practical} expand the scope of use of model stealing attacks. They train a substitute model using a synthetic dataset labeled by the target model. Then, this substitute is used to craft adversarial examples. In this study, we focus on training a substitute model without using the target model.

\noindent\textbf{Meta-learning.} Meta-learning is useful in few-shot classification. It trains a meta-learner that can adapt rapidly to new environments with only a few samples. Ma~\etal\cite{MetaAdvDet} propose MetaAdvDet to detect new types of adversarial attacks with high accuracy in order to utilize meta-learning in the adversarial attack field. The Meta Attack~\cite{du2020queryefficient} trains an auto-encoder to predict the gradients of a target model to reduce the query complexity. However, its auto-encoder is only trained on natural image and gradient pairs and not on data from real attacks. Hence its prediction accuracy is not satisfied in the attack. The prediction of the large gradient map is also difficult for its lightweight auto-encoder. Thus, the Meta Attack only extracts the gradients with the top-128 values to update examples, resulting in poor performance. In comparison, the proposed Simulator in the current study is trained with knowledge-distillation loss for logits prediction; hence, the performance is not affected by the resolution of images. In addition, the training data are query sequences of black-box attacks, which are divided into meta-train set and meta-test set. The former corresponds to the fine-tuning iterations and the latter corresponds to simulation iterations in the attack. These strategies connect the training and the attack seamlessly to maximize the performance.


\section{Method}
\subsection{Task Generation}
During an attack, the trained Simulator must accurately simulate the outputs of any unknown target model when the feeding queries are only slightly different from one another. To this end, the Simulator should learn from the real attack, \textit{i.e.,} the intermediate data (query sequences and outputs) generated in the attacks of various networks. For this purpose, several classification networks $\mathbb{N}_1,\dots,\mathbb{N}_n$ are collected to construct the training tasks, creating a huge simulation environment to improve the general simulation capability (Fig. \ref{fig:meta-train}). Each task contains $V$ query pairs $Q_{1}, \dots, Q_{V}\left(Q_{i} \in \mathbb{R}^{D}, i \in \{1,\cdots,V\}\right)$, where $D$ is the image dimensionality. These pairs are generated by using Bandits to attack a randomly selected network. The data sources used by Bandits can be any image downloaded from the Internet. In this study, we use the training sets of the standard datasets with different data distributions from the tested images. Each task is divided into two subsets, namely, the meta-train set $\mathcal{D}_{mtr}$, which consists of the first $t$ query pairs $Q_1,\dots,Q_t$, and the meta-test set $\mathcal{D}_{mte}$ with the following query pairs $Q_{t+1},\dots,Q_{V}$. The former is used in the inner-update step of the training corresponding to the fine-tuning step in the attack stage. The latter corresponds to the attack iterations of using the Simulator as the substitute (Fig. \ref{fig:fig1}). This partition connects the training and attack stages seamlessly. The logits outputs of $\mathbb{N}_1,\dots,\mathbb{N}_n$ are termed as ``pseudo labels''. All query sequences and pseudo labels are cached in the hard drive to accelerate training.
\begin{algorithm}[b]
	\caption{\small Training procedure of the Simulator}
	\label{alg:trn} 
	
	\begin{algorithmic} 
		\Require
		Training dataset $D$, Bandits attack algorithm $\mathcal{A}$,
		pre-trained classification networks $\mathbb{N}_1,\dots,\mathbb{N}_n$, the Simulator network $\mathbb{M}$ and its parameters $\theta$, feed-forward function $f$ of $\mathbb{M}$, loss function $\mathcal{L}(\cdot,\cdot)$ defined in Eq. \eqref{eqn:distillation_loss}.
		\renewcommand{\algorithmicrequire}{\textbf{Parameters:}}  
		\Require  Training iterations $N$, query sequence size $V$, meta-train set size $t$, batch size $K$, inner-update learning rate $\lambda_1$, outer-update learning rate $\lambda_2$, inner-update iterations $T$.
		\Ensure The learned Simulator $\mathbb{M}$.
	\end{algorithmic}
	\begin{algorithmic}[1] 
		\For{$iter \gets 1$ to $N$}
		\State sample $K$ benign images $x_{1},\dots,x_K$ from $D$
		\For{$k \gets 1$ to $K$} \Comment{iterate over $K$ tasks}
		\State a network $\mathbb{N}_i \gets$ sample from $\mathbb{N}_1,\dots,\mathbb{N}_n$ \label{line:sample_attacked_network}
		\State $Q_1,\dots,Q_V \gets \mathcal{A}(x_k, \mathbb{N}_i)$ \Comment{query sequence} \label{line:attack_generate_train_data}
		\State $\mathcal{D}_{mtr} \gets Q_1,\dots,Q_t$
		\State $\mathcal{D}_{mte} \gets Q_{t+1},\dots,Q_V$
		\State $\bm{\mathrm{p}}_{\mathtt{train}} \gets \mathbb{N}_i(\mathcal{D}_{mtr})$
		\State $\bm{\mathrm{p}}_{\mathtt{test}} \gets \mathbb{N}_i(\mathcal{D}_{mte})$  \Comment{pseudo labels} \label{line:soft_label_1}
		\State $\theta^\prime \gets \theta$\Comment{reinitialize $\mathbb{M}$'s weights}\label{line:init_task_net} 
		\For{$j \gets 1$ to $T$}
		\State $\theta^\prime \gets \theta^\prime - \lambda_1\cdot \nabla_{\theta^\prime} \mathcal{L}\left (f_{\theta^\prime}\left(\mathcal{D}_{mtr} \right), \bm{\mathrm{p}}_{\mathtt{train}}\right )$ \label{line:inner_update}
		\EndFor
		\State $L_i \gets \mathcal{L}\left (f_{\theta^\prime}\left (\mathcal{D}_{mte}\right ),\bm{\mathrm{p}}_{\mathtt{test}}\right  )$\label{line:meta_gradient}
		\EndFor
		
		\State $\theta \gets \theta - \lambda_2\cdot \frac{1}{K}\sum_{i=1}^K \nabla_{\theta} L_i$ \Comment{the outer update}\label{line:outer_update}
		\EndFor
		\State\Return $\mathbb{M}$
	\end{algorithmic}
\end{algorithm}

\begin{algorithm}[htbp]
	\caption{\small Simulator Attack under the $\ell_p$ norm constraint}
	\label{alg:attack} 
	\begin{algorithmic} 
		\Require
		Input image $x\in \mathbb{R}^{D}$ where $D$ is the image dimensionality, true label $y$ of $x$,
		feed-forward function $f$ of target model, Simulator $\mathbb{M}$, attack objective loss $\mathcal{L}(\cdot,\cdot)$.
		\renewcommand{\algorithmicrequire}{\textbf{Parameters:}}  
		\Require
		Warm-up iterations $t$, simulator-predict interval $m$, Bandits exploration $\tau$, finite difference probe $\delta$, OCO learning rate $\eta_g$, image learning rate $\eta$.
		\Ensure $x_{\mathrm{adv}}$ that satisfies $\Vert x_{\mathrm{adv}}-x\Vert_p \le \epsilon$.
	\end{algorithmic}
	\begin{algorithmic}[1]
		\State Initialize the adversarial example $x_{\mathrm{adv}} \gets x$
		\State Initialize the gradient to be estimated $\bm{\mathrm{g}} \gets \bm{0}$
		\State Initialize $\mathbb{D} \gets deque(maxlen=t)$ \Comment{a bounded double-ended queue with maximum length of $t$, adding a full $\mathbb{D}$ leads it to drop its oldest item automatically.} 
		
		\For{$i \gets 1$ to $N$}
		\State $\bm{\mathrm{u}} \gets \mathcal{N}(\bm{0}, \frac{1}{D}\bm{\mathrm{I}})$ \Comment{the same dimension with $x$} \label{line:rnd_mat}
		\State $q1 \gets \bm{\mathrm{g}} + \tau \bm{\mathrm{u}}$,\quad $q2 \gets \bm{\mathrm{g}} - \tau \bm{\mathrm{u}}$
		\State $q1 \gets q1 / \Vert q1 \Vert_2 $,\quad $q2 \gets q2 / \Vert q2\Vert_2$ \label{line:q_normalize}
		\If{$i \leq t$ or $(i - t) \bmod m = 0$}
		\State $\hat{y}_1 \gets f(x_{\mathrm{adv}} + \delta \cdot q1)$
		\State $\hat{y}_2 \gets f(x_{\mathrm{adv}} + \delta \cdot q2)$ \label{line:warm-up}
		\State $\{x_{\mathrm{adv}} + \delta \cdot q1, \hat{y}_1, x_{\mathrm{adv}} + \delta \cdot q2, \hat{y}_2\}$ append $\mathbb{D}$ \label{line:collect_queries}
		\If{$i \geq t$} 
		\State Fine-tune $\mathbb{M}$ using $\mathbb{D}$ \Comment{fine-tune $\mathbb{M}$ every $m$ iterations after the warm-up phase.}
		\EndIf
		\Else
		\State $\hat{y}_1 \gets \mathbb{M}(x_{\mathrm{adv}} + \delta \cdot q1)$,\quad $\hat{y}_2 \gets \mathbb{M}(x_{\mathrm{adv}} + \delta \cdot q2)$
		\EndIf
		\State $\Delta_g \gets \frac{\mathcal{L}(\hat{y}_1, y) - \mathcal{L}(\hat{y}_2, y)}{\tau \delta}\bm{\mathrm{u}}$ \label{line:finite_difference}
		\If{$p=2$}
		\State $\bm{\mathrm{g}} \gets \bm{\mathrm{g}} + \eta_g\cdot \Delta_g$ 
		\State $x_{\mathrm{adv}} \gets \small{\prod}_{\mathcal{B}_2(x,\epsilon)} (x_{\mathrm{adv}} + \eta \cdot \frac{\bm{\mathrm{g}}}{\Vert \bm{\mathrm{g}} \Vert_2}) $ \Comment{$\small{\prod}_{\mathcal{B}_p(x,\epsilon)}$ denotes the $\ell_p$ norm projection under $\ell_p$ norm bound.}
		\ElsIf{$p=\infty$} \Comment{using the exponentiated gradient update \cite{ilyas2018prior} in the $\ell_\infty$ norm attack as follows.}
		\State $\bm{\mathrm{\hat{g}}} \gets \frac{\bm{\mathrm{g}}+\mathbf{1}}{2}$
		\State $\bm{\mathrm{g}} \gets \frac{\bm{\mathrm{\hat{g}}} \cdot \exp (\eta_g \cdot \Delta_g) - (\mathbf{1} - \bm{\mathrm{\hat{g}}}) \cdot \exp (- \eta_g \cdot \Delta_g)}{\bm{\mathrm{\hat{g}}} \cdot \exp (\eta_g \cdot \Delta_g) + (\mathbf{1} - \bm{\mathrm{\hat{g}}}) \cdot \exp (- \eta_g \cdot \Delta_g)}$  \label{line:prior_update}
		\State $x_{\mathrm{adv}} \gets \small{\prod}_{\mathcal{B}_\infty(x,\epsilon)} (x_{\mathrm{adv}} + \eta \cdot \text{sign}(\bm{\mathrm{g}}))$
		\EndIf
		\State $x_{\mathrm{adv}} \gets \text{Clip} (x_{\mathrm{adv}},0,1)$
		\EndFor
		\State\Return $x_{\mathrm{adv}}$
	\end{algorithmic}
\end{algorithm}
\subsection{Simulator Learning}
\label{sec:learning_Simulator}
\noindent\textbf{Initialization.} Algorithm \ref{alg:trn} and Fig. \ref{fig:meta-train} present the training procedure. In the training, we sample $K$ tasks randomly to form a mini-batch. At the beginning of learning each task, the Simulator $\mathbb{M}$ reinitializes its weights using the weights $\theta$ learned by the last mini-batch. These weights are kept for computing meta-gradients in the outer-update step.

\noindent\textbf{Meta-train.} $\mathbb{M}$ performs the gradient descent on the meta-train set $\mathcal{D}_{mtr}$ for several iterations (the inner update). This step is similar to training a student model in a knowledge distillation, which matches the fine-tuning step of the attack.

\noindent\textbf{Meta-test.} After several iterations, $\mathbb{M}$'s weights are updated to $\theta^\prime$. Then, the loss $L_i$ is computed based on meta-test set of the $i$-th task with $\theta^\prime$. Afterwards, the meta-gradient $\nabla_{\theta} L_i$ is calculated as a higher-order gradient. Then, $\nabla_{\theta} L_1,\dots, \nabla_{\theta} L_K$ of $K$ tasks are averaged as $\frac{1}{K}\sum_{i=1}^{K} \nabla_{\theta} L_i$ for updating $\mathbb{M}$ (the outer update), thus enabling $\mathbb{M}$ to learn the general simulation capability. 

\noindent\textbf{Loss Function.} In the training, we adopt a knowledge-distillation-fashioned loss to induce the Simulator to output a similar prediction with the sampled network $\mathbb{N}_i$, which we use in both the inner and outer steps. Given the two queries $Q_{i,1}$ and $Q_{i,2}$ of the $i$-th query pair $Q_i$ generated by Bandits\footnote{Bandits attack requires two queries in the finite difference for estimating a gradient. Thus, a query pair is generated in each iteration.}, where $i\in \{1,\dots,n\}$ and $n$ represents the number of query pairs in the meta-train or meta-test set. The logits outputs of the Simulator and $\mathbb{N}_{i}$ are denoted as $\hat{\bm{\mathrm{p}}}$ and $\bm{\mathrm{p}}$, respectively. The MSE loss function defined in Eq. \eqref{eqn:distillation_loss} pushes the predictions of the Simulator and the pseudo label closer. 
\begin{equation}
\label{eqn:distillation_loss}
\mathcal{L}(\hat{\bm{\mathrm{p}}}, \bm{\mathrm{p}}) = \frac{1}{n}\sum_{i=1}^{n} \big(\hat{\bm{\mathrm{p}}}_{Q_{i,1}} - \bm{\mathrm{p}}_{Q_{i,1}} \big) ^ 2 + \frac{1}{n}\sum_{i=1}^{n} \big (\hat{\bm{\mathrm{p}}}_{Q_{i,2}} - \bm{\mathrm{p}}_{Q_{i,2}}) ^ 2
\end{equation}

\subsection{Simulator Attack}
\label{sec:attack_stage}

Algorithm \ref{alg:attack} shows the Simulator Attack under the $\ell_p$ norm constraint. The query pairs of the first $t$ iterations are fed to the target model (the warm-up phase). These queries and corresponding outputs are collected into a double-ended queue $\mathbb{D}$. Then, $\mathbb{D}$ drops the oldest item once it is full, which is beneficial in terms of focusing on new queries when fine-tuning $\mathbb{M}$ using $\mathbb{D}$. After warm-up, subsequent queries are fed into the target model every $m$ iterations, and the fine-tuned $\mathbb{M}$ takes the rest. To be consistent with training, the gradient estimation steps follow that of Bandits. The attack objective loss function shown in Eq. \eqref{eqn:attack_loss} is maximized during the attack:
\begin{equation}
\begin{aligned}
\mathcal{L}(\hat{y},t) = \begin{cases}
\max_{j\neq t} \hat{y}_j - \hat{y}_t, & \text{if untargeted attack}; \\
\hat{y}_t - \max_{j\neq t} \hat{y}_j,  & \text{if targeted attack}; \\
\end{cases}
\end{aligned}
\label{eqn:attack_loss}
\end{equation}
where $\hat{y}$ represents the logits output of the Simulator or the target model, $t$ is the target class in the targeted attack or the true class in the untargeted attack, and $j$ indexes the other classes.
\subsection{Discussion}
During an attack, the Simulator must accurately simulate the outputs when feeding queries of the real attack. Thus, the Simulator is trained on the intermediate data of the real attack in a knowledge-distillation manner. None of existing meta-learning methods learn a simulator in this way, as they all focus on the few-shot classification or reinforcement learning problems. In addition, Algorithm \ref{alg:attack} alternately feeds queries to $\mathbb{M}$ and the target model to learn the latest queries. The periodic fine-tuning is crucial in achieving a high success rate when faced with a difficult attack (\textit{e.g.,} the result of the targeted attack in Fig. \ref{fig:meta_predict_interval}).

\section{Experiment}

\subsection{Experiment Setting}
\label{sec:expr_setting}
\noindent\textbf{Dataset and Target Models.} We conduct the experiments using the CIFAR-10 \cite{krizhevsky2009learning}, CIFAR-100 \cite{krizhevsky2009learning}, and TinyImageNet \cite{russakovsky2015imagenet} datasets. Following previous studies \cite{guo2019subspace}, \nn{1000} tested images are randomly selected from their validation sets for evaluation. In the CIFAR-10 and CIFAR-100 datasets, we follow Yan \etal\cite{guo2019subspace} to select the target models: (1) a 272-layer PyramidNet+Shakedrop network (PyramidNet-272) \cite{han2017deep,yamada2019shakedrop} trained using AutoAugment \cite{cubuk2019autoaugment}; (2) a model obtained via neural architecture search called GDAS \cite{dong2019searching}; (3) a WRN-28 \cite{Zagoruyko2016WRN} with 28 layers and 10 times width expansion; and (4) a WRN-40 with 40 layers. In the TinyImageNet dataset, we select ResNeXt-101 (32x4d)~\cite{Xie2017ResNext}, ResNeXt-101 (64x4d), and DenseNet-121 \cite{Huang_2017_CVPR} with a growth rate of 32.

\noindent\textbf{Method Setting.} In the training, we generate the query sequence data $Q_1,\dots,Q_{100}$ in each task. The meta-train set $\mathcal{D}_{mtr}$ contains $Q_1,\dots,Q_{50}$, and the meta-test set $\mathcal{D}_{mte}$ consists of $Q_{51},\dots,Q_{100}$. We select ResNet-34 \cite{he2016deep} as the backbone of the Simulator, which we trained for three epochs over \nn{30000} tasks. Here, 30 sampled tasks constitute a mini-batch. Training each Simulator with an NVIDIA Tesla V100 GPU lasted for 72 hours.
The fine-tune iteration number is set to 10 in the first fine-tuning and then reduced to a random number from 3 to 5 for subsequent ones. In the targeted attacks, we set the target class to $y_{adv} = (y + 1 ) \mod C$ for all attacks, where $y_{adv}$ is the target class, $y$ is the true class, and $C$ is the class number. Following previous studies~\cite{cheng2019improving,guo2019subspace}, we use the attack success rate as well as the average and median values of queries as the evaluation metrics. Table \ref{tab:default_params} presents the default parameters. 

\noindent\textbf{Pre-trained Networks.} In order to evaluate the capability of simulating unknown target models, we ensure that the selection of $\mathbb{N}_1,\dots,\mathbb{N}_n$ in Algorithm~\ref{alg:trn} is different from the target models. A total of 14 networks are selected in the CIFAR-10 and CIFAR-100 datasets, and 16 networks are selected for the TinyImageNet dataset. The details can be found in the supplementary material. In experiments involving attacks of defensive models, we re-train the Simulator by removing the data of ResNet networks. This is because the defensive models adopt a backbone of ResNet-50. 

\noindent\textbf{Compared Methods.} The compared methods include NES~\cite{ilyas2018blackbox}, Bandits~\cite{ilyas2018prior}, Meta Attack~\cite{du2020queryefficient}, RGF~\cite{nesterov2017random}, and P-RGF~\cite{cheng2019improving}. Bandits is selected as the baseline. To ensure a fair comparison, the training data (\textit{i.e.,} images and gradients) of the Meta Attack are generated by directly using the pre-trained classification networks of the present study. We translate the codes of NES, RGF, and P-RGF from the official implementations of TensorFlow into the PyTorch version for the experiments. P-RGF improves RGF query efficiency by utilizing a surrogate model, which adopts ResNet-110 \cite{he2016deep} in the CIFAR-10 and CIFAR-100 datasets and ResNet-101~\cite{he2016deep} in the TinyImageNet dataset. We exclude the experiments of RGF and P-RGF in the targeted attack experiments, because their official implements only support untargeted attacks. All methods are limited to the maximum of \nn{10000} queries in both untargeted and targeted attacks. We set the same $\epsilon$ values for all attacks, which are $4.6$ and $8/255$ in the $\ell_2$ norm attack and $\ell_\infty$ norm attack, respectively. The detailed configurations of all compared methods are provided in the supplementary material.

\begin{table}[t]
	
	\tabcolsep=0.1cm
	\centering
	 \scalebox{0.7}{
		\begin{tabular}{p{4cm}|c|p{6cm}}
			\toprule
			\makebox[4cm][c]{\textbf{name}} & \textbf{default} & \makebox[6cm][c]{\textbf{description}} \\
			\midrule
			$\lambda_1$ of the inner update & 0.01 & learning rate in the inner update. \\
			$\lambda_2$ of the outer update & 0.001 & learning rate in the outer update. \\
			maximum query times & \nn{10000} & the limitation of queries of each sample. \\
			$\epsilon$ of $\ell_2$ norm attack  & 4.6 & the maximum distortion in $\ell_2$ norm attack. \\
			$\epsilon$ of $\ell_\infty$ norm attack & 8/255 & the maximum distortion in $\ell_\infty$ norm attack. \\
			$\eta$ of $\ell_2$ norm attack & 0.1 & the image learning rate for updating image. \\
			$\eta$ of $\ell_\infty$ norm attack & 1/255 & the image learning rate for updating image. \\
			$\eta_g$ of $\ell_2$ norm attack & 0.1 & OCO learning rate for updating $\bm{\mathrm{g}}$. \\
			$\eta_g$ of $\ell_\infty$ norm attack & 1.0 & OCO learning rate for updating $\bm{\mathrm{g}}$. \\
			inner-update iterations & 12 & update iterations of learning meta-train set.\\
			simulator-predict interval & 5 & the prediction iteration's interval of $\mathbb{M}$.\\
			warm-up iterations $t$ & 10 & the first $t$ iterations of the attack. \\
			deque $\mathbb{D}$'s length & 10 & the maximum length of $\mathbb{D}$.\\
			\bottomrule
	\end{tabular}}
	\caption{The default parameters setting of Simulator Attack.}
	\label{tab:default_params}
\end{table}

\begin{table}[t]
	\small
	\tabcolsep=0.1cm
	\setlength{\abovecaptionskip}{3pt}%
	\centering
	\scalebox{0.66}{
		\begin{tabular}{c|c|c|c|c|c}
			\toprule
			\textbf{Target Model} & \textbf{Method} & \textbf{Avg. Query} & \textbf{Med. Query} & \textbf{Max Query} & \textbf{Success Rate} \\
			\midrule
			\multirow{3}{*}{PyramidNet-272} & Rnd\_init Simulator & 105 & 52 & 1470 & 100\% \\
			& Vanilla Simulator & 102 & 52 &  1374 &  100\%  \\
			& Simulator Attack & \B 92 & \B 52 & \B 834 & 100\% \\
			\bottomrule
	\end{tabular}}
	\caption{Comparison of different simulators by performing $\ell_2$ norm attack on the CIFAR-10 dataset. The Rnd\_init Simulator uses an untrained ResNet-34 as the simulator; the Vanilla Simulator uses a ResNet-34 that is trained without using meta-learning as the simulator.}
	\label{tab:ablation_meta_train}

\end{table}

\subsection{Ablation Study}
\label{sec:ablation_study}

The ablation study is conducted to validate the benefit of meta training and determine the effects of key parameters.

\begin{figure}[htbp]
	\captionsetup[sub]{font={small}}
	\centering 
	\begin{minipage}[b]{.23\textwidth}
		\includegraphics[width=\linewidth]{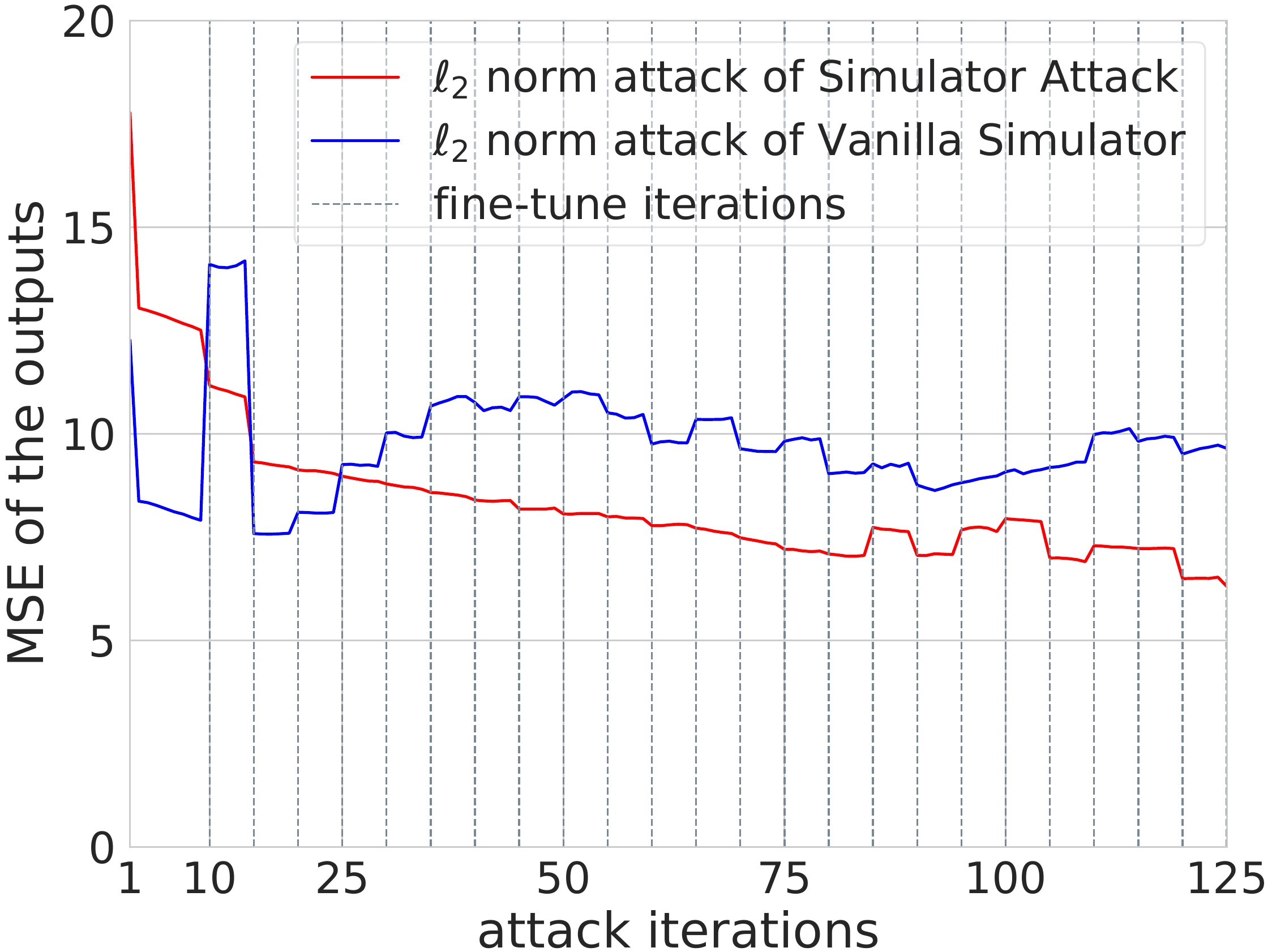}
		\subcaption{simulation's precision study}
		\label{fig:MSE_error_study}
	\end{minipage}
	\begin{minipage}[b]{.23\textwidth}
		\includegraphics[width=\linewidth]{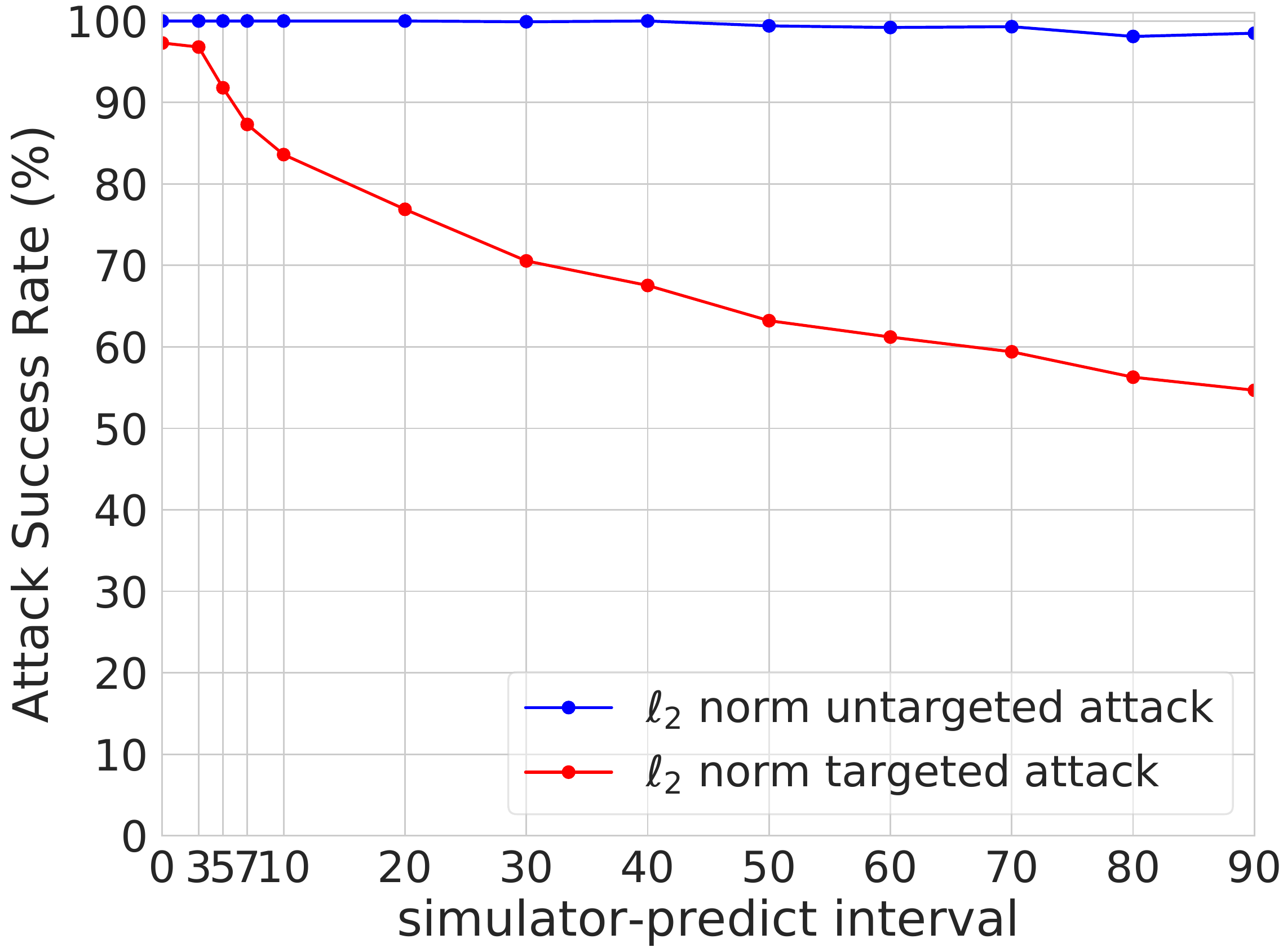}
		\subcaption{simulator-predict interval}
		\label{fig:meta_predict_interval}
	\end{minipage}
	\begin{minipage}[b]{.23\textwidth}
		\includegraphics[width=\linewidth]{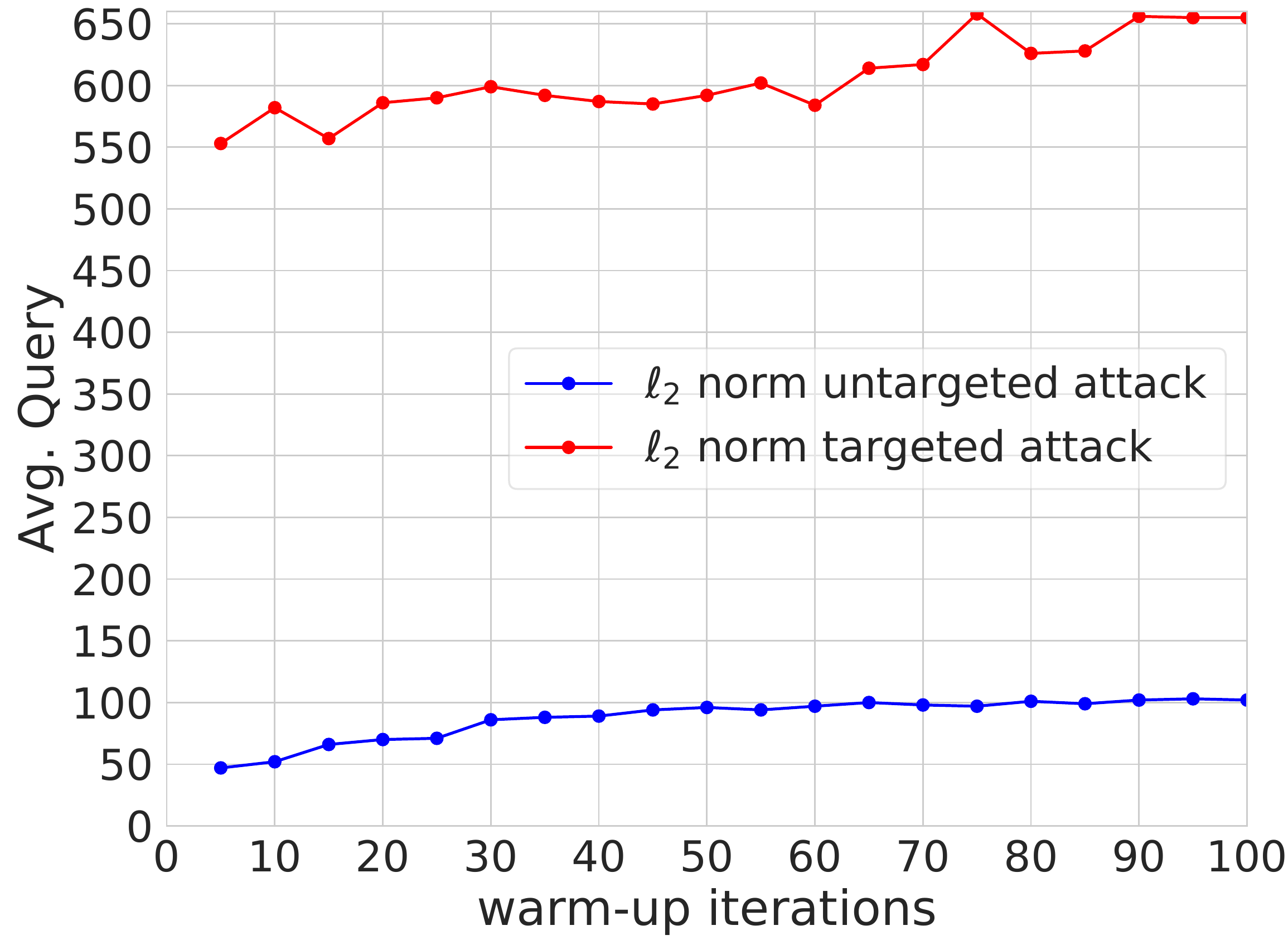}
		\subcaption{warm-up study}
		\label{fig:warm_up_study}
	\end{minipage}
	\begin{minipage}[b]{.23\textwidth}
		\includegraphics[width=\linewidth]{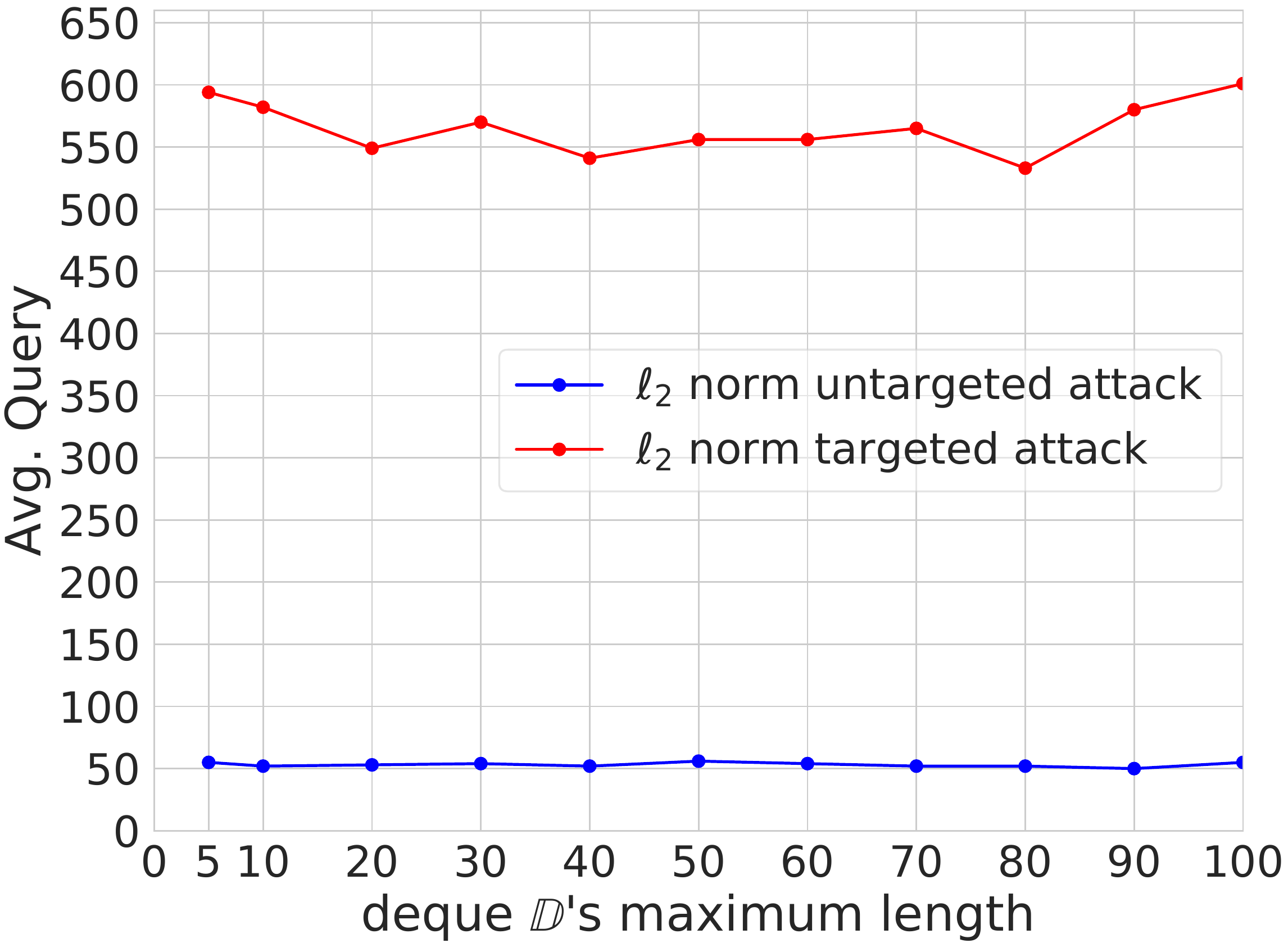}
		\subcaption{deque's maximum length}
		\label{fig:deque_study}
	\end{minipage}
	\caption{We conduct ablation studies of the simulation's precision, simulator-predict interval, warm-up iterations, and deque $\mathbb{D}$'s maximum length by attacking a WRN-28 model in the CIFAR-10 dataset. The results indicate the following: (1) the meta training is beneficial for achieving an accurate simulation (Fig. \ref{fig:MSE_error_study}), (2) a difficult attack (\textit{e.g.,} targeted attack) requires a small simulator-predict interval (Fig. \ref{fig:meta_predict_interval}), and (3) more warm-up iterations cause higher average queries (Fig. \ref{fig:warm_up_study}).}
	\label{fig:ablation_study_1}
\end{figure}

\begin{table*}[htb]
	\small
	\tabcolsep=0.1cm
	\begin{center}
	\scalebox{0.7}{
		\begin{tabular}{c|c|c|cccc|cccc|cccc}
			\toprule
			\B Dataset & \B{Norm} & \B{Attack} & \multicolumn{4}{c|}{\B{Attack Success Rate}} &  \multicolumn{4}{c|}{\B{Avg. Query}} &  \multicolumn{4}{c}{\B{Median Query}} \\
			& & & PyramidNet-272 & GDAS & WRN-28 & WRN-40 & PyramidNet-272 & GDAS & WRN-28 & WRN-40 & PyramidNet-272 & GDAS & WRN-28 & WRN-40 \\ 
			\midrule
			\multirow{12}{*}{CIFAR-10} & \multirow{6}{*}{$\ell_2$} & NES \cite{ilyas2018blackbox} & 99.5\% & 74.8\% & 99.9\% & 99.5\% & 200 & 123 & 159 & 154 & 150 & 100 & 100 & 100 \\
			& & RGF \cite{nesterov2017random} & 100\% & 100\% & 100\% & 100\% & 216 & 168 & 153 & 150 & 204 & 152 & 102 & 152 \\
			& & P-RGF \cite{cheng2019improving} & 100\% & 100\% & 100\% & 100\% & \B 64 & 40 & 76 & 73 & 62 & \B 20 & 64 & 64 \\
			& & Meta Attack \cite{du2020queryefficient} & 99.2\% & 99.4\% & 98.6\% & 99.6\% & 2359 & 1611 & 1853 & 1707 & 2211 & 1303 & 1432 & 1430 \\
			& & Bandits \cite{ilyas2018prior} & 100\% & 100\% & 100\% & 100\% & 151 & 66 & 107 & 98 & 110 & 54 & 80 & 78 \\
			& & Simulator Attack & 100\% & 100\% & 100\% & 100\% & 92 & \B 34 & \B 48 & \B 51 & \B 52 &  26 & \B 34 & \B 34 \\
			\cmidrule(rl){2-15} & \multirow{6}{*}{$\ell_\infty$} & NES \cite{ilyas2018blackbox} & 86.8\% & 71.4\% & 74.2\% & 77.5\% & 1559 & 628 & 1235 & 1209 & 600 & 300 & 400 & 400 \\
			& & RGF \cite{nesterov2017random} & 99\% & 93.8\% & 98.6\% & 98.8\% & 955 & 646 & 1178 & 928 & 668 & 460 & 663 & 612 \\
			& & P-RGF \cite{cheng2019improving} & 97.3\% & 97.9\% & 97.7\% & 98\% & \B 742 & 337 & 703 & 564 & \B 408 & 128 & 236 & 217 \\
			& & Meta Attack \cite{du2020queryefficient} & 90.6\% & 98.8\% & 92.7\% & 94.2\% & 3456 & 2034 & 2198 & 1987 & 2991 & 1694 & 1564 & 1433 \\
			& & Bandits \cite{ilyas2018prior} & 99.6\% & 100\% & 99.4\% & 99.9\% & 1015 & 391 & 611 & 542 & 560 & 166 & 224 & 228 \\
			& & Simulator Attack & 96.5\% & 99.9\% & 98.1\% & 98.8\% & 779 & \B 248 & \B 466 & \B 419 & 469 & \B 83 & \B 186 & \B 186 \\
			\midrule
			\multirow{12}{*}{CIFAR-100} & \multirow{6}{*}{$\ell_2$} & NES \cite{ilyas2018blackbox} & 92.4\% & 90.2\% & 98.4\% & 99.6\% & 118 & 94 & 102 & 105 & 100 & 50 & 100 & 100 \\
			& & RGF \cite{nesterov2017random} & 100\% & 100\% & 100\% & 100\% & 114 & 110 & 106 & 106 & 102 & 101 & 102 & 102 \\
			& & P-RGF \cite{cheng2019improving} & 100\% & 100\% & 100\% & 100\% & 54 & 46 & 54 & 73 & 62 & 62 & 62 & 62 \\
			& & Meta Attack \cite{du2020queryefficient} & 99.7\% & 99.8\% & 99.4\% & 98.4\% & 1022 & 930 & 1193 & 1252 & 783 & 781 & 912 & 913 \\
			& & Bandits \cite{ilyas2018prior} & 100\% & 100\% & 100\% & 100\% & 58 & 54 & 64 & 65 & 42 & 42 & 52 & 53 \\
			& & Simulator Attack & 100\% & 100\% & 100\% & 100\% & \B 29 & \B 29 & \B 33 & \B 34 & \B 24 & \B 24 & \B 26 & \B 26 \\
			\cmidrule(rl){2-15} & \multirow{6}{*}{$\ell_\infty$ }& NES \cite{ilyas2018blackbox} & 91.3\% & 89.7\% & 92.4\% & 89.3\% & 439 & 271 & 673 & 596 & 204 & 153 & 255 & 255 \\
			& & RGF \cite{nesterov2017random} & 99.7\% & 98.8\% & 98.9\% & 98.9\% & 385 & 420 & 544 & 619 & 256 & 255 & 357 & 357 \\
			& & P-RGF \cite{cheng2019improving} & 99.3\% & 98.2\% & 98\% & 97.8\% & 308 & 220 & 371 & 480 & 147 & 116 & 136 & 181 \\
			& & Meta Attack \cite{du2020queryefficient} & 99.7\% & 99.8\% & 97.4\% & 97.3\% & 1102 & 1098 & 1294 & 1369 & 912 & 911 & 1042 & 1040 \\
			& & Bandits \cite{ilyas2018prior} & 100\% & 100\% & 99.8\% & 99.8\% & 266 & 209 & 262 & 260 & 68 & 57 & 107 & 92 \\
			& & Simulator Attack & 100\% & 100\% & 99.9\% & 99.9\% & \B 129 & \B 124 & \B 196 & \B 209 & \B 34 & \B 28 & \B 58 & \B 54 \\
			\bottomrule
	\end{tabular}}
	\end{center}
	\caption{Experimental results of untargeted attack in CIFAR-10 and CIFAR-100 datasets.}
	\label{tab:CIFAR_untargetd_result}
\end{table*}

\begin{table*}[htb]
	\small
	\tabcolsep=0.1cm
	\centering
	\scalebox{0.7}{
		\begin{tabular}{c|c|c|cccc|cccc|cccc}
			\toprule
			\B Dataset & \B Norm & \B Attack & \multicolumn{4}{c|}{\B{Attack Success Rate}} &  \multicolumn{4}{c|}{\B{Avg. Query}} &  \multicolumn{4}{c}{\B{Median Query}} \\
			& & & PyramidNet-272 & GDAS & WRN-28 & WRN-40 & PyramidNet-272 & GDAS & WRN-28 & WRN-40 & PyramidNet-272 & GDAS & WRN-28 & WRN-40 \\ 
			\midrule
			\multirow{9}{*}{CIFAR-10} & \multirow{5}{*}{$\ell_2$} & NES \cite{ilyas2018blackbox} & 93.7\% & 95.4\% & 98.5\% & 97.7\% & 1474 & 1515 & 1043 & 1088 & 1251 & 999 & 881 & 882 \\
			& & Meta Attack \cite{du2020queryefficient} & 92.2\% & 97.2\% & 74.1\% & 74.7\% & 4215 & 3137 & 3996 & 3797 & 3842 & 2817 & 3586 & 3329 \\
			& & Bandits \cite{ilyas2018prior} & 99.7\% & 100\% & 97.3\% & 98.4\% & 852 & 718 & 1082 & 997 & 458 & 538 & 338 & 399 \\
			& & Simulator Attack (m=3) & 99.1\% & 100\% & 98.5\% & 95.6\% & 896 & 718 & 990 & 980 & 373 & \B 388 & 217 & 249 \\
			& & Simulator Attack (m=5) & 97.6\% & 99.9\% & 96.4\% & 94\% & \B 815 & \B 715 & \B 836 & \B 793 & \B 368 & 400 & \B 206 & \B 245 \\
			
			\cmidrule(rl){2-15} & \multirow{4}{*}{$\ell_\infty$} & NES \cite{ilyas2018blackbox} & 63.8\% & 80.8\% & 89.7\% & 88.8\% & 4355 & 3942 & 3046 & 3051 & 3717 & 3441 & 2535 & 2592 \\
			& & Meta Attack \cite{du2020queryefficient} & 75.6\% & 95.5\% & 59\% & 59.8\% & 4960 & 3461 & 3873 & 3899 & 4736 & 3073 & 3328 & 3586 \\
			& & Bandits \cite{ilyas2018prior} & 84.5\% & 98.3\% & 76.9\% & 79.8\% & 2830 & 1755 & 2037 & 2128 & 2081 & 1162 & 1178 & 1188 \\
			& & Simulator Attack (m=3) & 80.9\% & 97.8\% & 83.1\% & 82.2\% & 2655 & 1561 & 1855 & 1806 & 1943 & 918 & 1010 & 1018 \\
			& & Simulator Attack (m=5) & 78.7\% & 96.5\% & 80.8\% & 80.3\% & \B 2474 & \B 1470 & \B 1676 & \B 1660 & \B 1910 & \B 917 & \B 957 & \B 956 \\
			\midrule
			\multirow{9}{*}{CIFAR-100} & \multirow{5}{*}{$\ell_2$} & NES \cite{ilyas2018blackbox} & 87.6\% & 77\% & 89.3\% & 87.6\% & 1300 & 1405 & 1383 & 1424 & 1102 & 1172 & 1061 & 1049 \\
			& & Meta Attack \cite{du2020queryefficient} & 86.1\% & 88.7\% & 63.4\% & 43.3\% & 4000 & 3672 & 4879 & 4989 & 3457 & 3201 & 4482 & 4865 \\
			& & Bandits \cite{ilyas2018prior} & 99.6\% & 100\% & 98.9\% & 91.5\% & 1442 & 847 & 1645 & 2436 & 1058 & 679 & 1150 & 1584 \\
			& & Simulator Attack (m=3) & 99.3\% & 100\% & 98.6\% & 92.6\% & 921 & 724 & 1150 & 1552 & 666 & 519 & 779 & 1126 \\
			& & Simulator Attack (m=5) & 97.8\% & 99.6\% & 95.7\% & 83.9\% & \B 829 & \B 679 & \B 1000 & \B 1211 & \B 644 & \B 508 & \B 706 & \B 906 \\
			
			\cmidrule(rl){2-15} & \multirow{4}{*}{$\ell_\infty$} & NES \cite{ilyas2018blackbox} & 72.1\% & 66.8\% & 68.4\% & 69.9\% & 4673 & 5174 & 4763 & 4770 & 4376 & 4832 & 4357 & 4508 \\
			& & Meta Attack \cite{du2020queryefficient} & 80.4\% & 81.2\% & 57.6\% & 40.1\% & 4136 & 3951 & 4893 & 4967 & 3714 & 3585 & 4609 & 4737 \\
			& & Bandits \cite{ilyas2018prior} & 81.2\% & 92.5\% & 72.4\% & 56\% & 3222 & 2798 & 3353 & 3465 & 2633 & 2132 & 2766 & 2774 \\
			& & Simulator Attack (m=3) & 89.4\% & 94.2\% & 79\% & 64.3\% & 2732 & 2281 & 3078 & 3238 & 1854 & 1589 & 2185 & 2548 \\
			& & Simulator Attack (m=5) & 83.7\% & 91.4\% & 74.2\% & 60\% & \B 2410 & \B 2134 & \B 2619 & \B 2823 & \B 1754 & \B 1572 & \B 2080 & \B 2270 \\
			
			\bottomrule
	\end{tabular}}
	\caption{Experimental results of targeted attack in CIFAR-10 and CIFAR-100 datasets, where $m$ is simulator-predict interval.}
	\label{tab:CIFAR_targetd_result}
\end{table*}

\noindent\textbf{Meta Training.}
We validate the benefits of meta training by equipping with different simulators in the proposed algorithm. Simulator $\mathbb{M}$ is replaced with two networks for comparison, \textit{i.e.,} Rnd\_init Simulator: a randomly initialized ResNet-34 network without training, and Vanilla Simulator: a ResNet-34 network trained on the data of the present study but without using meta-learning. Table \ref{tab:ablation_meta_train} shows the experimental results, which indicate that the Simulator Attack is able to achieve the minimum number of queries, thereby confirming the benefit of meta training. To inspect the simulation capacity in detail, we calculate the average MSE between outputs of simulators and the target model at different attack iterations (Fig. \ref{fig:MSE_error_study}). As indicated by the results, the Simulator Attack achieves the lowest MSE at most iterations, thus exhibiting its satisfactory simulation capability.

In control experiments, we check the effects of the key parameters of the Simulator Attack by adjusting one parameter while keeping others fixed, as listed in Table \ref{tab:default_params}. The corresponding results are shown in Figs. \ref{fig:meta_predict_interval}, \ref{fig:warm_up_study}, and \ref{fig:deque_study}.

\noindent\textbf{Simulator-Predict Interval $m$.} This parameter is the iteration interval that uses Simulator $\mathbb{M}$ to make predictions. A larger $m$ results in fewer opportunities to fine-tune $\mathbb{M}$. When this happens, the Simulator cannot accurately simulate the target model in case of a difficult attack (\textit{e.g.,} the targeted attack in Fig. \ref{fig:meta_predict_interval}), resulting in a low success rate.

\noindent\textbf{Warm-up.} As shown in Fig. \ref{fig:warm_up_study}, more warm-up iterations lead to a higher average query, because more queries are fed into the target model in the warm-up phase.

\subsection{Comparisons with State-of-the-Art Methods}
\label{sec:expr_normal_models}
\noindent\textbf{Results of Attacks on Normal Models.} In this study, the normal model is the classification model without the defensive mechanism. We conduct experiments on the target models described in Section \ref{sec:expr_setting}. Tables \ref{tab:CIFAR_untargetd_result} and \ref{tab:CIFAR_targetd_result} show the results of the CIFAR-10 and CIFAR-100 datasets, respectively, whereas Tables \ref{tab:TinyImageNet_untargeted_linf_result} and \ref{tab:TinyImageNet_targeted_result} present the results of the TinyImageNet dataset. The results reveal the following: (1) the Simulator Attack can gain up to $2\times$ reduction in the average and median values of the queries compared with the baseline Bandits, and (2) the Simulator Attack can obtain significantly fewer queries and a higher attack success rate than the Meta Attack \cite{du2020queryefficient} (\textit{e.g.,} the low success rates of Meta Attack in Tables \ref{tab:TinyImageNet_untargeted_linf_result} and \ref{tab:TinyImageNet_targeted_result}). The poor performance of the Meta Attack can be attributed to its high-cost gradient estimation (specifically the use of ZOO \cite{chen2017zoo}).

\noindent\textbf{Experimental Figures.}  Tables \ref{tab:CIFAR_untargetd_result}, \ref{tab:CIFAR_targetd_result}, \ref{tab:TinyImageNet_untargeted_linf_result}, and \ref{tab:TinyImageNet_targeted_result} show the results obtained after setting the maximum number of queries to \nn{10000}. To further inspect the attack success rates at different maximum queries, we perform $\ell_\infty$ norm attacks by limiting the different maximum queries of each adversarial example. The superiority of the proposed approach in terms of attack success rate is shown in Fig. \ref{fig:query_threshold_attack_success_rate}. Meanwhile, Fig.~\ref{fig:success_rate_to_avg_query} demonstrates the average number of queries that reaches different desired success rates. Fig. \ref{fig:success_rate_to_avg_query} reveals that the proposed approach is more query-efficient than other attacks and that the gap is amplified for higher success rates.

\begin{table*}[htbp]
	\small
	\tabcolsep=0.1cm
	\setlength{\belowcaptionskip}{0pt}%
	\centering
	\scalebox{0.7}{
		\begin{tabular}{c|c|cccc|cccc|cccc}
			\toprule
			\B Dataset & \B{Attack} & \multicolumn{4}{c|}{\B{Attack Success Rate}} &  \multicolumn{4}{c|}{\B{Avg. Query}} &  \multicolumn{4}{c}{\B{Median Query}} \\
			&  & CD \cite{jia2019comdefend} & PCL \cite{mustafa2019adversarial} &  FD \cite{liu2019feature} & Adv Train \cite{madry2018towards} & CD \cite{jia2019comdefend} & PCL \cite{mustafa2019adversarial} &  FD \cite{liu2019feature} & Adv Train \cite{madry2018towards} & CD \cite{jia2019comdefend} & PCL \cite{mustafa2019adversarial} &  FD \cite{liu2019feature}  & Adv Train \cite{madry2018towards} \\ 
			\midrule
			\multirow{6}{*}{CIFAR-10} &  NES \cite{ilyas2018blackbox} & 60.4\% & 65\% & 54.5\% &  16.8\% & 1130 & 728 & 1474 & \B 858  & 400 & 150 & 450 & \B 200  \\ 
			& RGF \cite{nesterov2017random} & 48.7\% & 82.6\% & 44.4\% &  22.4\% &  2035 & 1107 & 1717 & 973  & 1071 & 306 & 768 & 510  \\
			& P-RGF \cite{cheng2019improving} & 62.8\% & 80.4\% & 65.8\% &  22.4\% &  1977 & 1006 & 1979 & 1158  & 1038 & 230 & 703 & 602  \\
			& Meta Attack \cite{du2020queryefficient} & 26.8\% & 77.7\% & 38.4\% &  18.4\% &  2468 & 1756 & 2662 & 1894  & 1302 & 1042 & 1824 & 1561  \\
			& Bandits \cite{ilyas2018prior} & 44.7\% & 84\% & 55.2\% &  34.8\% &  786 & 776 & 832 & 1941  & 100 & 126 & 114 & 759  \\
			& Simulator Attack & 54.9\% & 78.2\% & 60.8\% &  32.3\% & \B 433 & \B 641 & \B 391 & 1529  & \B 46 & \B 116 & \B 50 & 589  \\
			\midrule
			\multirow{6}{*}{CIFAR-100} & NES \cite{ilyas2018blackbox} & 78.1\% & 87.9\% & 77.6\% &  23.1\% & 892 & 429 & 1071 &\B 865  & 300 & 150 & 250 &\B 250  \\ 
			& RGF \cite{nesterov2017random} & 50.2\% & 95.5\% & 62\% &  29.2\% &  1753 & 645 & 1208 & 1009  & 765 & 204 & 408 & 510  \\
			& P-RGF \cite{cheng2019improving} & 54.2\% & 96.1\% & 73.4\% &  28.8\% &  1842 & 679 & 1169 & 1034  & 815 & 182 & 262 & 540  \\
			& Meta Attack \cite{du2020queryefficient} & 20.8\% & 93\% & 59\% &  27\% &  2084 & 1122 & 2165 & 1863  & 781 & 651 & 1043 & 1562  \\
			& Bandits \cite{ilyas2018prior} & 54.1\% & 97\% & 72.5\% &  44.9\% &  786 & 321 & 584 & 1609  & 56 & 34 & 32 & 484  \\
			& Simulator Attack & 72.9\% & 93.1\% & 80.7\% &  35.6\% & \B 330 & \B 233 & \B 250 & 1318  & \B 30 & \B 22 & \B 24 & 442  \\
			\midrule
			\multirow{6}{*}{TinyImageNet} & NES \cite{ilyas2018blackbox} & 69.5\% & 73.1\% & 33.3\% &  23.7\% & 1775 & 863 & 2908 & \B 945  & 850 & 250 & 1600 & \B 200  \\ 
			& RGF \cite{nesterov2017random} & 31.3\% & 91.8\% & 9.1\% &  34.7\% &  2446 & 1022 & 1619 & 1325  & 1377 & 408 & 765 & 612  \\
			& P-RGF \cite{cheng2019improving} & 37.3\% & 91.8\% & 25.9\% &  34.4\% &  1946 & 1065 & 2231 & 1287  & 891 & 436 & 985 & 602  \\
			& Meta Attack \cite{du2020queryefficient} & 4.5\% & 75.8\% & 3.7\% &  20.1\% &  1877 & 2585 & 4187 & 3413  & 912 & 1792 & 2602 & 2945  \\
			& Bandits \cite{ilyas2018prior} & 39.6\% & 95.8\% & 12.5\% &  49\% &  893 & 909 & 1272 & 1855  & 85 & 206 & 193 & 810  \\
			& Simulator Attack & 43\% & 84.2\% & 21.3\% &  42.5\% & \B 377 & \B 586 & \B 746 & 1631  & \B 32 & \B 148 & \B 157 & 632  \\
			\bottomrule
	\end{tabular}}
	\caption{Experimental results after performing the $\ell_\infty$ norm attacks on defensive models, where CD represents ComDefend \cite{jia2019comdefend}, FD is Feature Distillation \cite{liu2019feature}, and PCL is prototype conformity loss \cite{mustafa2019adversarial}.}
	\label{tab:defensive_model_result}
\end{table*}
\begin{table}[htb]
	\small
	\tabcolsep=0.1cm
	\centering
	\scalebox{0.7}{
		\begin{tabular}{c|ccc|ccc|ccc}
			\toprule
			\B{Attack} & \multicolumn{3}{c|}{\B{Attack Success Rate}} &  \multicolumn{3}{c|}{\B{Avg. Query}} &  \multicolumn{3}{c}{\B{Median Query}} \\
			&  D$_{121}$ & R$_{32}$ & R$_{64}$  & D$_{121}$ & R$_{32}$ & R$_{64}$ & D$_{121}$ & R$_{32}$ & R$_{64}$ \\ 
			\midrule
			NES \cite{ilyas2018blackbox} & 74.3\% & 45.3\% & 45.5\% & 1306 & 2104 & 2078 & 510 & 765 & 816 \\
			RGF \cite{nesterov2017random} & 96.4\% & 85.3\% & 87.4\% & 1146 & 2088 & 2087 & 667 & 1280 & 1305 \\
			P-RGF \cite{cheng2019improving} & 94.5\% & 83.9\% & 85.9\% & 883 & 1583 & 1581 & 448 & \B 657 & \B 690 \\
			Meta Attack \cite{du2020queryefficient} & 71.1\% & 33.8\% & 36\% & 3789 & 4101 & 4012 & 3202 & 3712 & 3649 \\
			Bandits  \cite{ilyas2018prior} & 99.2\% & 94.1\% & 95.3\% & 964 & 1737 & 1662 & 520 & 954 & 1014 \\
			Simulator Attack & 99.4\% & 96.8\% & 97.9\% & \B 811 & \B 1380 & \B 1445 & \B 431 & 850 & 878 \\
			\bottomrule
	\end{tabular}}
	\caption{Experimental results of untargeted attack under $\ell_\infty$ norm in TinyImageNet dataset. D$_{121}$: DenseNet-121, R$_{32}$: ResNeXt-101~(32$\times$4d), R$_{64}$: ResNeXt-101~(64$\times$4d).}
	\label{tab:TinyImageNet_untargeted_linf_result}
\end{table}

\begin{table}[t]
	\small
	\tabcolsep=0.1cm
	\centering
	\scalebox{0.7}{
		\begin{tabular}{c|ccc|ccc|ccc}
			\toprule
			\B{Attack} & \multicolumn{3}{c|}{\B{Attack Success Rate}} &  \multicolumn{3}{c|}{\B{Avg. Query}} &  \multicolumn{3}{c}{\B{Median Query}} \\
			& D$_{121}$ & R$_{32}$ & R$_{64}$  & D$_{121}$ & R$_{32}$ & R$_{64}$ & D$_{121}$ & R$_{32}$ & R$_{64}$ \\ 
			\midrule
			NES \cite{ilyas2018blackbox} & 88.5\% & 88\% & 88.2\% & 4625 & 4959 & 4758 & 4337 & 4703 & 4440 \\
			Meta Attack \cite{du2020queryefficient} & 24.2\% & 21\% & 18.2\% & 5420 & 5440 & 5661 & 5506 & 5249 & 5250 \\
			Bandits \cite{ilyas2018prior} &  85.1\% & 72.2\% & 72.4\% & 2724 & 3550 & 3542 & 1860 & 2700 & 2854 \\
			Simulator Attack & 89.8\% & 84.9\% & 83.9\% & \B 1959 & \B 2558 & \B 2488 & \B 1399 & \B 1966 & \B 1982 \\
			\bottomrule
	\end{tabular}}
	\caption{Experimental results of targeted attack under $\ell_2$ norm in TinyImageNet dataset. D$_{121}$: DenseNet-121, R$_{32}$: ResNeXt-101~(32$\times$4d), R$_{64}$: ResNeXt-101~(64$\times$4d).}
	\label{tab:TinyImageNet_targeted_result}
\end{table}
\begin{figure}[t]
	\setlength{\abovecaptionskip}{0pt}
	\setlength{\belowcaptionskip}{0pt}
	\captionsetup[sub]{font={scriptsize}}
	\centering 
	\begin{minipage}[b]{.23\textwidth}
		\includegraphics[width=\linewidth]{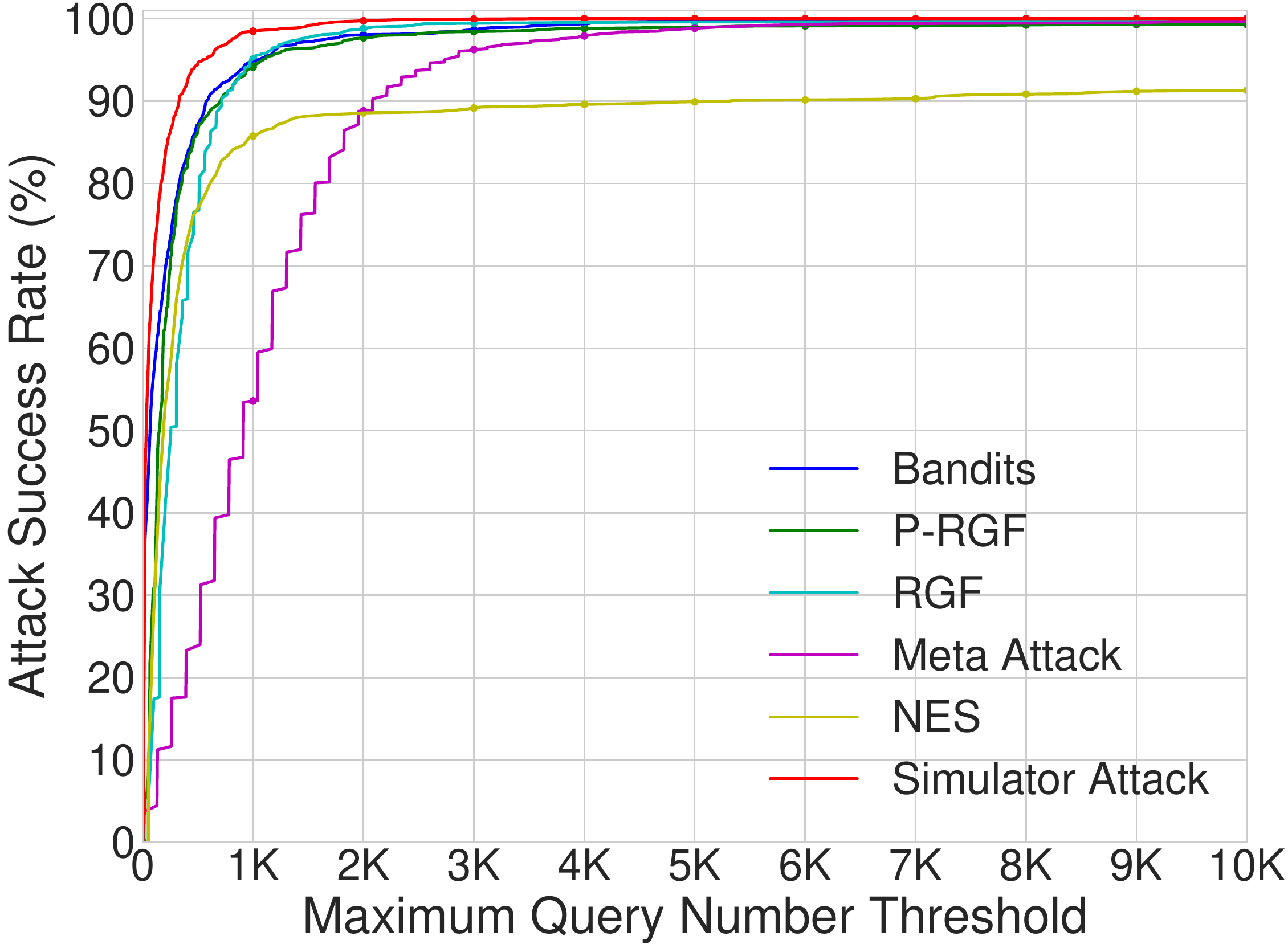}
		\subcaption{PyramidNet-272 in CIFAR-100}
		\label{fig:query_atk_rate_pyramidnet272_CIFAR-100}
	\end{minipage}
	\begin{minipage}[b]{.23\textwidth}		\includegraphics[width=\linewidth]{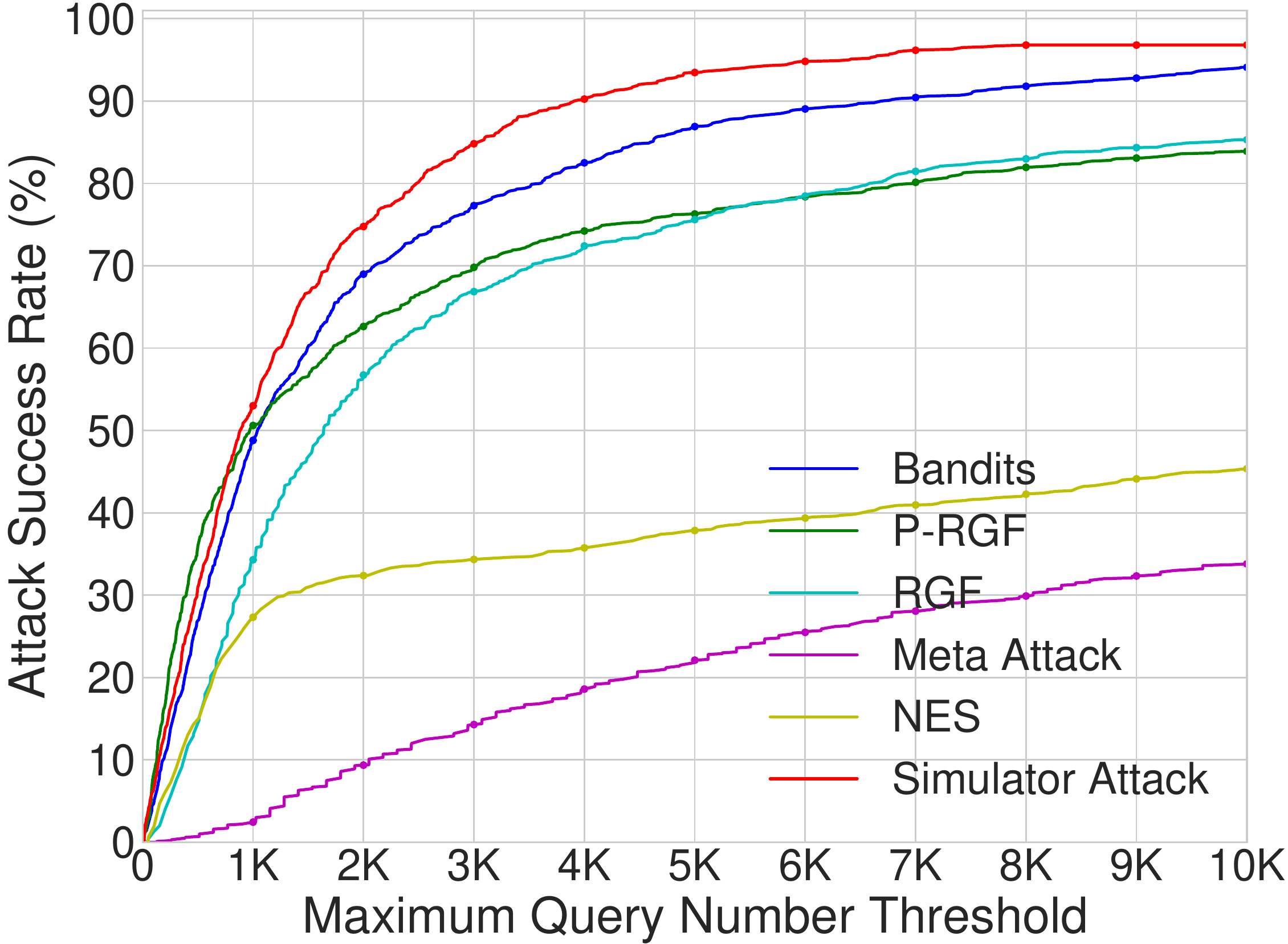}
		\subcaption{R$_{32}$ in TinyImageNet}
		\label{fig:query_atk_rate_resnext64_4_TinyImageNet}
	\end{minipage}

	\caption{Comparison of the attack success rate at different limited maximum queries in untargeted attack under $\ell_\infty$ norm, where R$_{32}$ indicates ResNext-101~(32$\times$4d).}
	\vspace{-0.8cm}
	\label{fig:query_threshold_attack_success_rate}
\end{figure}
\begin{figure}[htbp]
	\setlength{\abovecaptionskip}{0pt}
	\setlength{\belowcaptionskip}{0pt}
	\captionsetup[sub]{font={scriptsize}}
	\centering 
	
	\begin{minipage}[b]{.23\textwidth}
		\includegraphics[width=\linewidth]{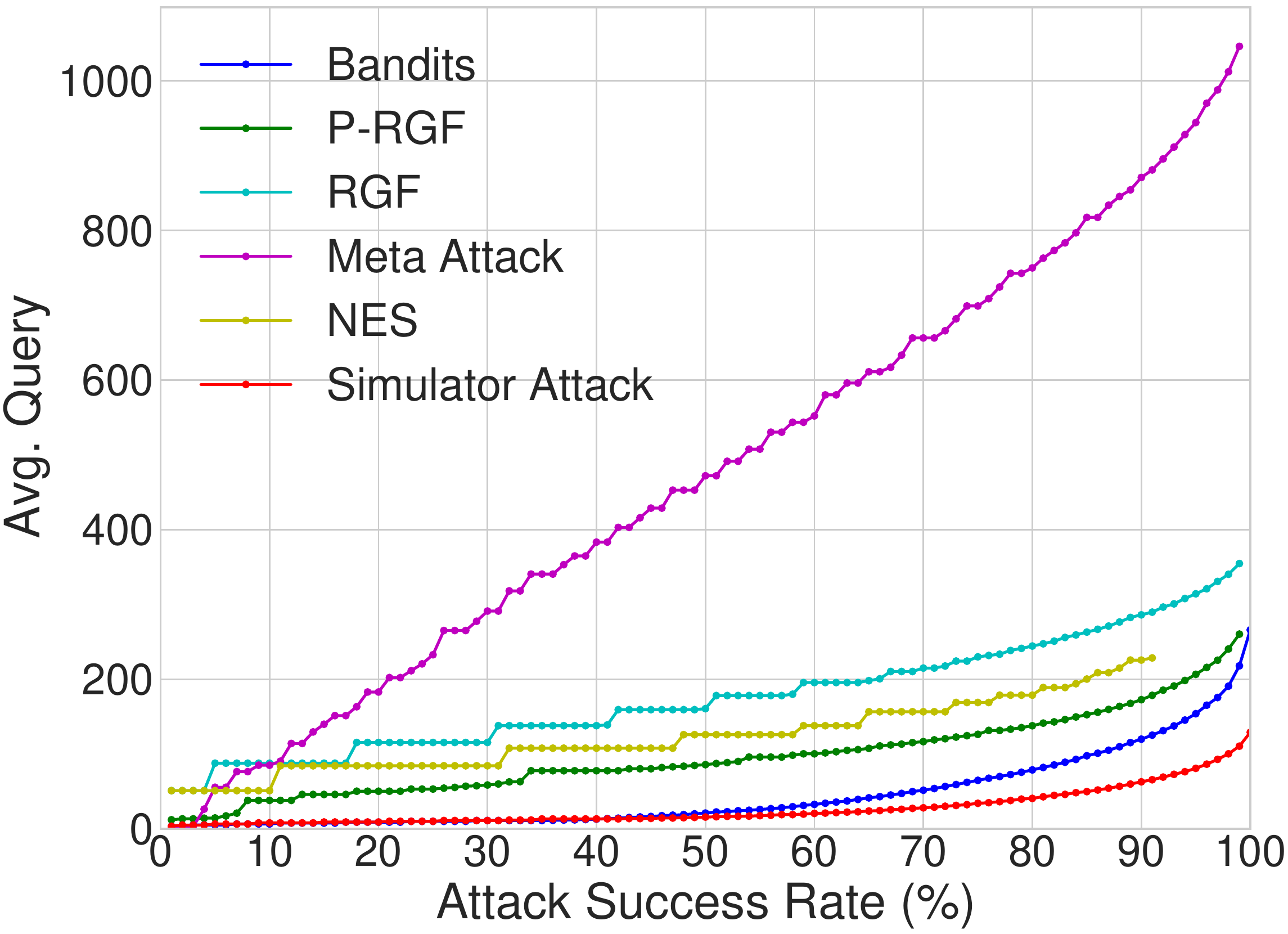}
		\subcaption{PyramidNet-272 in CIFAR-100}
		\label{fig:asr_avg_query_pyramidnet272}
	\end{minipage}
	\begin{minipage}[b]{.23\textwidth}
		\includegraphics[width=\linewidth]{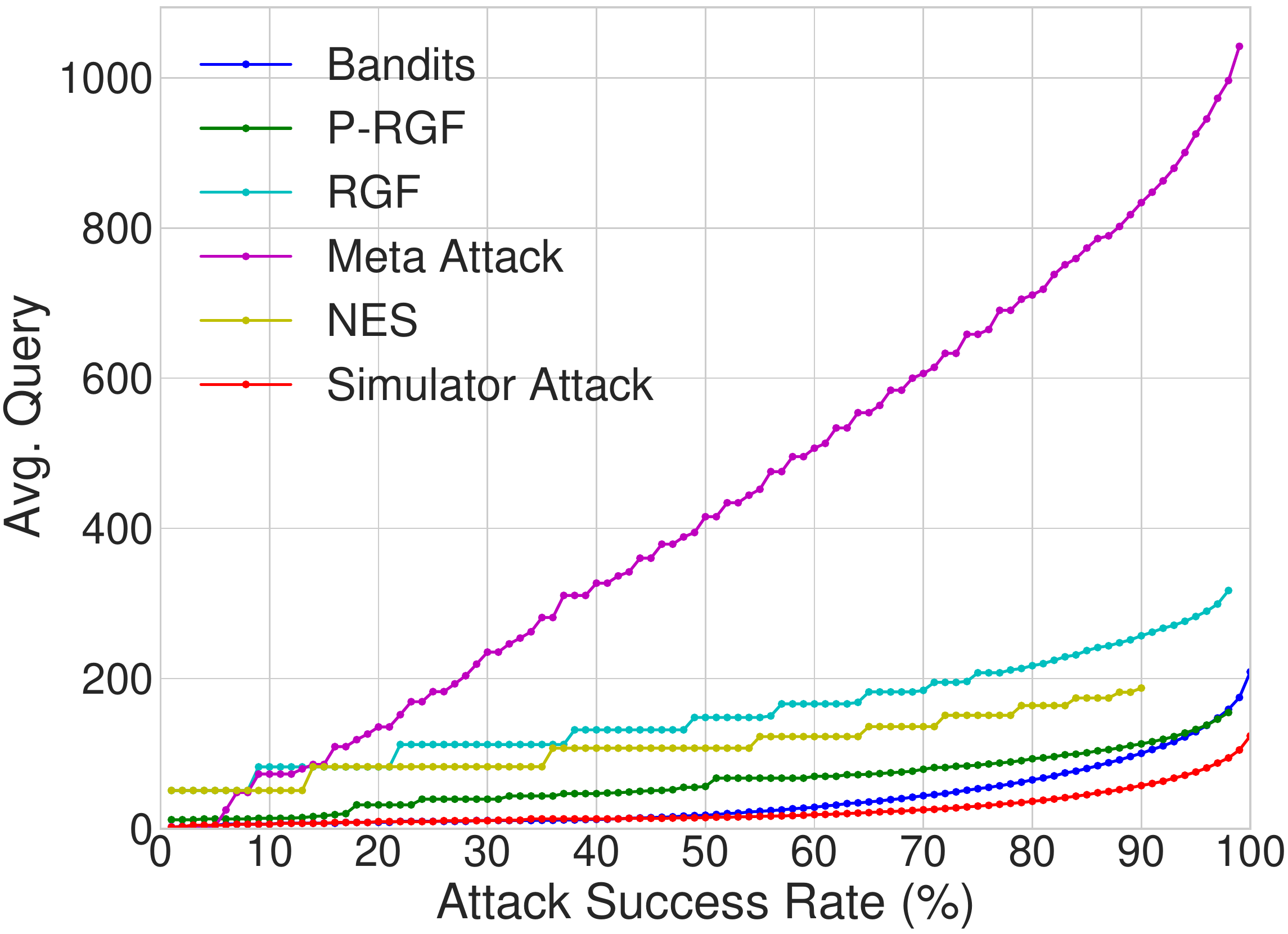}
		\subcaption{GDAS in CIFAR-100}
		\label{fig:asr_avg_query_gdas}
	\end{minipage}

	\caption{Comparisons of the average query at different success rates under the untargeted $\ell_\infty$ norm attack. More results are presented in the supplementary material.}
	\label{fig:success_rate_to_avg_query}
    \vspace{-0.5cm}
\end{figure}

\noindent\textbf{Results of Attacks on the Defensive Models.} Table \ref{tab:defensive_model_result} shows the experimental results obtained after attacking the defensive models. ComDefend (CD) \cite{jia2019comdefend} and Feature Distillation (FD) \cite{liu2019feature} are equipped with a denoiser to transform the input images to their clean versions before feeding to the target model. Prototype conformity loss (PCL) \cite{mustafa2019adversarial} introduces a new loss function to maximally separate the intermediate features of each class. Here, the PCL defensive model is obtained without using adversarial training in our experiments. Adv Train~\cite{madry2018towards} is a powerful defense method based on adversarial training. Following the results shown in Table \ref{tab:defensive_model_result}, we derive the following conclusions:

(1) Among all methods, the Simulator Attack exhibits the best performance in breaking CD, particularly outperforming the baseline method Bandits significantly.

(2) The Meta Attack demonstrates poor performance in CD and FD based on its unsatisfactory success rate. In comparison, the Simulator Attack can break this type of defensive model with a high success rate. 

(3) In experiments in which the Adv Train is attacked, the Simulator Attack consumes fewer queries to achieve a comparable success rate with Bandits.

\section{Conclusion}
In this study, we present a novel black-box attack named Simulator Attack. It focuses on training a generalized substitute model (``Simulator'') to accurately mimic any unknown target model with the aim of reducing the query complexity of the attack. To this end, the query sequences generated while attacking many different networks are used as the training data. The proposed approach uses an MSE-based knowledge-distillation loss in the inner and outer updates of meta-learning to learn the Simulator. After training, a high number of queries can be transferred to the Simulator, thereby reducing the query complexity of the attack by several orders of magnitude compared with the baseline.

\section*{Acknowledgments}
This research is supported by the National Key R\&D Program of China (2019YFB1405703) and TC190A4DA/3, the National Natural Science Foundation of China (Grant Nos. 61972221, 61572274).

\clearpage
{\small
\bibliographystyle{ieee_fullname}
\bibliography{reference}
}
\clearpage
\appendix
\twocolumn[{%
	\centering
	\LARGE \bfseries Supplementary Material \\[1.5em]

}]

\section{Experiment Settings}
\subsection{Compared Methods}

\noindent \textbf{Bandits.} Table \ref{tab:bandits_conf} shows the default hyperparameters of Bandits \cite{ilyas2018prior}, which is a subset of hyperparameters of the Simulator Attack. Specifically, the OCO learning rate is used to update the prior, which is an alias of the gradient $\bm{\mathrm{g}}$ for updating the input image.

\noindent \textbf{RGF and P-RGF.} Table \ref{tab:RGF_conf} shows the default hyperparameters of random gradient-free (RGF)~\cite{nesterov2017random} and prior-guided RGF (P-RGF)~\cite{cheng2019improving}. P-RGF improves RGF by using surrogate models (see the last row block of Table \ref{tab:RGF_conf}). The experiments of RGF and P-RGF are conducted by using the implementation of PyTorch version that is translated from the official TensorFlow version. 

\noindent \textbf{NES.} The default hyperparameters for natural evolution strategies (NES) \cite{ilyas2018blackbox} are listed in Table \ref{tab:NES_conf}. In the targeted attack, NES  uses an initial image of the target class and reduce its distortion iteratively while keeping the image residing in the adversarial region of the target class. Finally, the samples whose $\ell_p$ norm distance to the original benign image is less than a preset $\epsilon$ are considered as successful samples. Thus, the hyperparameters of NES are carefully tuned in the untargeted and targeted attack separately, so as to achieve the highest attack success rate. The experiments of NES are conducted by using the implementation of PyTorch version, which is translated from the official TensorFlow implementation.

\begin{table}[b]
	\scriptsize
	\tabcolsep=0.1cm
	\setlength{\abovecaptionskip}{0pt}%
	\setlength{\belowcaptionskip}{0pt}%
	\begin{center}
		\begin{tabular}{c|p{3cm}|c}
			\toprule
			\B Norm & \makebox[3cm][c]{\textbf{Hyperparameter}}  & \B Value \\
			\midrule
			\multirow{6}{*}{$\ell_2$} & $\delta$, finite difference probe & 0.01 \\
			& $\eta$, image learning rate  & 0.1 \\
			& $\eta_g$, OCO learning rate & 0.1 \\
			& $\tau$,  Bandits exploration & 0.3 \\
			& $\epsilon$, radius of $\ell_2$ norm ball & 4.6\\
			& maximum query times & \nn{10000} \\
			\midrule
			\multirow{6}{*}{$\ell_\infty$} & $\delta$, finite difference probe & 0.1 \\
			& $\eta$, image learning rate  & 1/255 \\
			& $\eta_g$, OCO learning rate & 1.0 \\
			& $\tau$,  Bandits exploration & 0.3 \\
			& $\epsilon$, radius of $\ell_\infty$ norm ball & 8/255\\
			& maximum query times & \nn{10000} \\
			\bottomrule
		\end{tabular}
	\end{center}
	\caption{The hyperparameters of Bandits \cite{ilyas2018prior}.}
	\label{tab:bandits_conf}
\end{table}

\begin{table}[!ht]
	\scriptsize
	\tabcolsep=0.1cm
	\setlength{\abovecaptionskip}{0pt}%
	\setlength{\belowcaptionskip}{0pt}%
	\begin{center}
		\begin{tabular}{c|p{4.5cm}|c}
			\toprule
			\B Norm & \makebox[4.5cm][c]{\textbf{Hyperparameter}} & 	\B Value \\
			\midrule
			\multirow{3}{*}{$\ell_2$} & $h$, image learning rate  & 2.0 \\
			& $\sigma$, sampling variance & 1e-4 \\
			&  $\epsilon$, radius of $\ell_2$ norm ball & 4.6 \\
			\midrule
			\multirow{3}{*}{$\ell_\infty$} & $h$, image learning rate  & 0.005 \\
			& $\sigma$, sampling variance & 1e-4 \\
			&  $\epsilon$, radius of $\ell_\infty$ norm ball & 8/255 \\
			\midrule
			$\ell_2$, $\ell_\infty$ & surrogate model used in CIFAR-10/100 & ResNet-110 \\
			$\ell_2$, $\ell_\infty$ & surrogate model used in TinyImageNet & ResNet-101 \\
			\bottomrule
		\end{tabular}
	\end{center}
	\caption{The hyperparameters of RGF~\cite{nesterov2017random} and P-RGF~\cite{cheng2019improving}, and the networks shown in the last row block are used as the surrogate models of P-RGF.}
	\label{tab:RGF_conf}
\end{table}

\begin{table}[t]
	\scriptsize
	\tabcolsep=0.1cm
	\centering
	\scalebox{1}{
		\begin{tabular}{p{5cm}|c}
			\toprule
			\B \makebox[2.8cm][c]{\textbf{Hyperparameter}} & \B Default Value \\
			\midrule
			backbone & ResNet-34 \\
			$\lambda_1$, the learning rate of the inner update & 0.01 \\
			$\lambda_2$, the learning rate of of the outer update & 0.001\\
			$\epsilon$, the maximum distortion of $\ell_2$ norm attack  & 4.6 \\
			$\epsilon$, the maximum distortion of $\ell_\infty$ norm attack  & 8/255 \\
			$\delta$, finite difference probe of $\ell_2$ norm attack &  0.01 \\
			$\delta$, finite difference probe of $\ell_\infty$ norm attack &  0.1 \\
			$\eta$, the image learning rate of $\ell_2$ norm attack & 0.1 \\
			$\eta$, the image learning rate of $\ell_\infty$ norm attack & 1/255 \\
			$\eta_g$, OCO learning rate of $\ell_2$ norm attack & 0.1 \\
			$\eta_g$, OCO learning rate of $\ell_\infty$ norm attack & 1.0 \\
			$\tau$, Bandits exploration & 0.3 \\
			inner-update iterations & 12 \\
			meta-predict interval $m$ & 5 \\
			warm-up iterations $t$ & 10 \\
			deque $\mathbb{D}$'s maximum length & 10 \\
			\bottomrule
	\end{tabular}}
	\caption{The hyperparameters of the Simulator Attack.}
	\label{tab:SimulatorAttack_params}
\end{table}
\begin{table}[htp]
	\scriptsize
	\tabcolsep=0.1cm
	\setlength{\abovecaptionskip}{0pt}%
	\setlength{\belowcaptionskip}{0pt}%
	
	\begin{center}
		\begin{tabular}{p{1.9cm}|p{2.8cm}|p{1cm}p{1cm}p{0.8cm}}
			\toprule
			\B \makebox[2.8cm][c]{\textbf{Dataset}} & \B \makebox[2.8cm][c]{\textbf{Network}} &\multicolumn{3}{c}{\B{Model Details}} \\
			&   & Params(M) & MACs(G) & Layers  \\
			\midrule
			\multirow{4}{2cm}{CIFAR-10} & PyramidNet-272 & 26.21 & 4.55 & 272 \\
			& GDAS  & 3.02 & 0.41 & 20  \\
			& WRN-28 & 36.48 & 5.25 & 28 \\
			& WRN-40 & 55.84 & 8.08 & 40  \\
			\midrule
			\multirow{4}{2cm}{CIFAR-100} & PyramidNet-272 & 26.29 & 4.55 & 272 \\
			& GDAS  & 3.14 & 0.41 & 20 \\
			& WRN-28 & 36.54 & 5.25 & 28 \\
			& WRN-40 & 55.90 & 8.08 & 40 \\
			\midrule
			\multirow{3}{2cm}{TinyImageNet} & DenseNet-121 & 7.16 & 0.23 &121  \\
			& ResNeXt-101 (32$\times$4d) & 42.54 & 0.65 & 101 \\
			& ResNeXt-101 (64$\times$4d) & 81.82 & 1.27 & 101 \\
			\bottomrule
		\end{tabular}
	\end{center}
	\caption{The details of black-box target models which are used for evaluating attack methods, where MAC is the multiply–accumulate operation count.}
	\label{tab:target_models}
\end{table}
\begin{table}[htbp]
	\scriptsize
	\tabcolsep=0.1cm
	\begin{center}
		\begin{tabular}{c|c|c|p{4.1cm}|c}
			\toprule
			\B Dataset & \B Attack & \B Norm & \makebox[4.1cm][c]{\textbf{Hyperparameter}} & \B Value \\
			\midrule
			\multirow{16}{*}{CIFAR-10} & \multirow{4}{*}{Untargeted} & \multirow{2}{*}{$\ell_2$} & $\epsilon$, radius of $\ell_2$ norm ball & 4.6 \\
			& & & $h$, image learning rate  & 2.0 \\
			\cmidrule(rl){3-5} & & \multirow{2}{*}{$\ell_\infty$} & $\epsilon$, radius of $\ell_\infty$ norm ball & 8/255 \\
			& & & $h$, image learning rate  & 1e-2 \\
			\cmidrule(rl){2-5} & \multirow{12}{*}{Targeted} & \multirow{6}{*}{$\ell_2$} & $\epsilon_{0}$, initial distance from the source image & 20.0 \\
			& & & $\epsilon$, final radius of $\ell_2$ norm ball & 4.6 \\
			& & & $\delta_{\epsilon_0}$, initial rate of decaying $\epsilon$ & 1.0 \\
			& & & $\delta_{\epsilon_{\text{min}}}$, the minimum rate of decaying $\epsilon$ & 0.1 \\
			& & & $h_{\text{max}}$, the maximum image learning rate & 2.0 \\
			& & & $h_{\text{min}}$, the minimum image learning rate & 5e-5 \\
			\cmidrule(rl){3-5}& & \multirow{6}{*}{$\ell_\infty$} & $\epsilon_{0}$, initial distance from the source image & 1.0 \\
			& & & $\epsilon$, final radius of $\ell_\infty$ norm ball & 8/255 \\
			& & & $\delta_{\epsilon_0}$, initial rate of decaying $\epsilon$ & 0.1\\
			& & & $\delta_{\epsilon_{\text{min}}}$, the minimum rate of decaying $\epsilon$ & 0.01 \\
			& & & $h_{\text{max}}$, the maximum image learning rate & 0.1 \\
			& & & $h_{\text{min}}$, the minimum image learning rate & 0.01 \\
			\midrule
			\multirow{16}{*}{CIFAR-100} & \multirow{4}{*}{Untargeted} & \multirow{2}{*}{$\ell_2$} & $\epsilon$, radius of $\ell_2$ norm ball & 4.6 \\
			& & & $h$, image learning rate  & 2.0 \\
			\cmidrule(rl){3-5} & & \multirow{2}{*}{$\ell_\infty$} & $\epsilon$, radius of $\ell_\infty$ norm ball & 8/255 \\
			& & & $h$, image learning rate  & 1e-2 \\
			\cmidrule(rl){2-5} & \multirow{12}{*}{Targeted} & \multirow{6}{*}{$\ell_2$} & $\epsilon_{0}$, initial distance from the source image & 20.0 \\
			& & & $\epsilon$, final radius of $\ell_2$ norm ball & 4.6 \\
			& & & $\delta_{\epsilon_0}$, initial rate of decaying $\epsilon$ & 1.0 \\
			& & & $\delta_{\epsilon_{\text{min}}}$, the minimum rate of decaying $\epsilon$ & 0.3 \\
			& & & $h_{\text{max}}$, the maximum image learning rate & 1.0 \\
			& & & $h_{\text{min}}$, the minimum image learning rate & 5e-5 \\
			\cmidrule(rl){3-5}& & \multirow{6}{*}{$\ell_\infty$} & $\epsilon_{0}$, initial distance from the source image & 1.0 \\
			& & & $\epsilon$, final radius of $\ell_\infty$ norm ball & 8/255 \\
			& & & $\delta_{\epsilon_0}$, initial rate of decaying $\epsilon$ & 0.1\\
			& & & $\delta_{\epsilon_{\text{min}}}$, the minimum rate of decaying $\epsilon$ & 0.01 \\
			& & & $h_{\text{max}}$, the maximum image learning rate & 0.1 \\
			& & & $h_{\text{min}}$, the minimum image learning rate & 0.01 \\
			\midrule
			\multirow{16}{*}{TinyImageNet} & \multirow{4}{*}{Untargeted} & \multirow{2}{*}{$\ell_2$} & $\epsilon$, radius of $\ell_2$ norm ball & 4.6 \\
			& & & $h$, image learning rate  & 2.0 \\
			\cmidrule(rl){3-5} & & \multirow{2}{*}{$\ell_\infty$} & $\epsilon$, radius of $\ell_\infty$ norm ball & 8/255 \\
			& & & $h$, image learning rate  & 1e-2 \\
			\cmidrule(rl){2-5} & \multirow{12}{*}{Targeted} & \multirow{6}{*}{$\ell_2$} & $\epsilon_{0}$, initial distance from the source image & 40.0 \\
			& & & $\epsilon$, final radius of $\ell_2$ norm ball & 4.6 \\
			& & & $\delta_{\epsilon_0}$, initial rate of decaying $\epsilon$ & 1.0 \\
			& & & $\delta_{\epsilon_{\text{min}}}$, the minimum rate of decaying $\epsilon$ & 0.1 \\
			& & & $h_{\text{max}}$, the maximum image learning rate & 2.0 \\
			& & & $h_{\text{min}}$, the minimum image learning rate & 0.5 \\
			\cmidrule(rl){3-5}& & \multirow{6}{*}{$\ell_\infty$} & $\epsilon_{0}$, initial distance from the source image & 1.0 \\
			& & & $\epsilon$, final radius of $\ell_\infty$ norm ball & 8/255 \\
			& & & $\delta_{\epsilon_0}$, initial rate of decaying $\epsilon$ & 0.1\\
			& & & $\delta_{\epsilon_{\text{min}}}$, the minimum rate of decaying $\epsilon$ & 1e-3 \\
			& & & $h_{\text{max}}$, the maximum image learning rate & 0.1 \\
			& & & $h_{\text{min}}$, the minimum image learning rate & 0.01 \\
			\bottomrule
		\end{tabular}
	\end{center}
	\caption{The hyperparameters of NES \cite{ilyas2018blackbox}, where the sampling variance $\sigma$ for gradient estimation is set to 1e-3, and the number of samples per draw is set to 50.}
	\label{tab:NES_conf}
\end{table}

\noindent \textbf{Meta Attack.} The default hyperparameters of the Meta Attack \cite{du2020queryefficient} are listed in Table \ref{tab:meta_attack_conf}. Specifically, the meta interval $m$ is set to 3 in two cases, namely, the targeted attack and all the experiments of TinyImageNet dataset. In other cases, the meta interval $m$ is set to 5. The gradients of training data are generated by using the classification networks listed in Table \ref{tab:pretrained_networks}. The Meta Attack uses the official PyTorch implementation to conduct $\ell_2$ norm attack experiments, and we add the additional code in the official implementation to enable it to support the $\ell_\infty$ norm attack.

\noindent \textbf{Simulator Attack.} The default hyperparameters of the proposed method are listed in Table \ref{tab:SimulatorAttack_params}. Those hyperparameters that are also used in Bandits are set to the same values as Bandits.

\begin{table}[!t]
	\scriptsize
	\tabcolsep=0.1cm
	\centering
	\begin{tabular}{c|c|c|p{4cm}|c}
		\toprule
		\B	Dataset &  \B Attack &  \B Norm & \makebox[4cm][c]{\textbf{Hyperparameter}} & \B Value \\
		\midrule
		\multirow{20}{*}{CIFAR-10/100} & \multirow{10}{*}{Untargeted} & \multirow{5}{*}{$\ell_2$} & $h$, image learning rate  & 1e-2 \\
		& & & top-$q$ coordinates for estimating gradient & 125 \\
		& & & $m$, meta interval & 5 \\
		& & & use\_tanh, change-of-variables method & true \\
		& & &  $\epsilon$, radius of $\ell_2$ norm ball & 4.6 \\
		\cmidrule(rl){3-5}& & \multirow{5}{*}{$\ell_\infty$} & $h$, image learning rate  & 1e-2 \\
		& & & top-$q$ coordinates for estimating gradient & 125 \\
		& & & $m$, meta interval & 5 \\
		& & & use\_tanh, change-of-variables method & false \\
		& & &  $\epsilon$, radius of $\ell_\infty$ norm ball & 8/255 \\
		\cmidrule(rl){2-5}
		& \multirow{10}{*}{Targeted} & \multirow{5}{*}{$\ell_2$} & $h$, image learning rate  & 1e-2 \\
		& & & top-$q$ coordinates for estimating gradient & 125 \\
		& & & $m$, meta interval & 3  \\
		& & & use\_tanh, change-of-variables method  & true \\
		& & &  $\epsilon$, radius of $\ell_2$ norm ball & 4.6 \\
		\cmidrule(rl){3-5}& & \multirow{5}{*}{$\ell_\infty$} & $h$, image learning rate  & 1e-2 \\
		& & & top-$q$ coordinates for estimating gradient & 125 \\
		& & & $m$, meta interval & 3 \\
		& & & use\_tanh, change-of-variables method & false \\
		& & &  $\epsilon$, radius of $\ell_\infty$ norm ball & 8/255 \\
		\midrule
		\multirow{20}{*}{TinyImageNet} & \multirow{10}{*}{Untargeted} & \multirow{5}{*}{$\ell_2$} & $h$, image learning rate  & 1e-2 \\
		& & & top-$q$ coordinates for estimating gradient & 125 \\
		& & & $m$, meta interval & 3 \\
		& & & use\_tanh, change-of-variables method & true \\
		& & &  $\epsilon$, radius of $\ell_2$ norm ball & 4.6 \\
		\cmidrule(rl){3-5}& & \multirow{5}{*}{$\ell_\infty$} & $h$, image learning rate  & 1e-2 \\
		& & & top-$q$ coordinates for estimating gradient & 125 \\
		& & & $m$, meta interval & 3 \\
		& & & use\_tanh, change-of-variables method & false \\
		& & &  $\epsilon$, radius of $\ell_\infty$ norm ball & 8/255 \\
		\cmidrule(rl){2-5}
		& \multirow{10}{*}{Targeted} & \multirow{5}{*}{$\ell_2$} & $h$, image learning rate  & 1e-2 \\
		& & & top-$q$ coordinates for estimating gradient & 125 \\
		& & & $m$, meta interval & 3  \\
		& & & use\_tanh, change-of-variables method  & true \\
		& & &  $\epsilon$, radius of $\ell_2$ norm ball & 4.6 \\
		\cmidrule(rl){3-5}& & \multirow{5}{*}{$\ell_\infty$} & $h$, image learning rate  & 1e-2 \\
		& & & top-$q$ coordinates for estimating gradient & 125 \\
		& & & $m$, meta interval & 3 \\
		& & & use\_tanh, change-of-variables method & false \\
		& & &  $\epsilon$, radius of $\ell_\infty$ norm ball & 8/255 \\
		\bottomrule
	\end{tabular}
	\caption{The hyperparameters of the Meta Attack, where the binary step is set to 1, and the solver of gradient estimation adopts the Adam optimizer \cite{kingma2015adam}.}
	\label{tab:meta_attack_conf}
\end{table}

\subsection{Pre-trained Networks and Target Models}
\noindent \textbf{Pre-trained Networks.} In the training of the Simulator and the auto-encoder of the Meta Attack, we collect various types of classification networks to generate the training data. In our experiments, we select 14 networks for generating training data of CIFAR-10 and CIFAR-100 datasets, and select 16 networks for generating training data of TinyImageNet datasets. The names of these networks and their training configurations are shown in Table \ref{tab:pretrained_networks}.

\noindent \textbf{Target Models.} To evaluate the performance of attacking unknown target models, we specify the target models to equip with completely different architectures from the pre-trained networks. The target models and their complexity are listed in Table \ref{tab:target_models}.

\begin{table*}[t]
	\small
	\tabcolsep=0.1cm
	\begin{center}
		\resizebox{1\linewidth}{!}{
			\begin{tabular}{p{2cm}|p{3cm}|p{1cm}p{1cm}p{2cm}p{1.7cm}p{1.7cm}|p{1.5cm}p{4.4cm}}
				\toprule
				\makebox[2cm][c]{\textbf{Dataset}} & \makebox[4cm][c]{\textbf{Network}} & \multicolumn{5}{c|}{\B{Training Configuration}} &\multicolumn{2}{c}{\B{Hyperparameters}} \\
				& & \makebox[1cm][c]{epochs} & \makebox[1cm][c]{lr} & \makebox[2cm][c]{lr decay epochs}  & \makebox[1.7cm][c]{lr decay rate}  & \makebox[1.7cm][c]{weight decay}   & \makebox[1.5cm][c]{layer depth}   & \makebox[3.1cm][c]{other hyperparameters}   \\
				\midrule
				\multirow{14}{2cm}{CIFAR-10/100} & AlexNet & 164 & 0.1 &  81, 122 & 0.1 & 5e-4 & 9& - \\
				& DenseNet-100 & 300  & 0.1 & 150, 225 & 0.1 & 1e-4 & 100 & growth rate:12, compression rate:2 \\
				& DenseNet-190 & 300  & 0.1 & 150, 225 & 0.1 &1e-4 &190  & growth rate:40, compression rate:2 \\
				& PreResNet-110 & 164  & 0.1 & 81, 122 & 0.1 &1e-4 & 110  & block name: BasicBlock  \\
				& ResNeXt-29 ($8\times 64d$) & 300 & 0.1 & 150, 225 & 0.1 & 5e-4 & 29  & widen factor:4, cardinality:8  \\
				& ResNeXt-29 ($16\times64d$) & 300 & 0.1 & 150, 225 & 0.1 & 5e-4 & 29 & widen factor:4, cardinality:16  \\
				& VGG-19 (BN) & 164  & 0.1 & 81, 122 & 0.1 & 5e-4 & 19 & - \\
				& ResNet-20 & 164 & 0.1 & 81, 122 & 0.1 & 1e-4 & 20 & block name: BasicBlock \\
				& ResNet-32  & 164 & 0.1 & 81, 122 & 0.1 & 1e-4 & 32 & block name: BasicBlock \\
				& ResNet-44 & 164& 0.1 & 81, 122 & 0.1 & 1e-4 & 44 & block name: BasicBlock \\
				& ResNet-50& 164 & 0.1 & 81, 122 & 0.1 & 1e-4 & 50 & block name: BasicBlock \\
				& ResNet-56 & 164 & 0.1 & 81, 122 & 0.1 & 1e-4 & 56 & block name: BasicBlock \\
				& ResNet-110 & 164 & 0.1 & 81, 122 & 0.1 & 1e-4 & 110 & block name: BasicBlock \\
				& ResNet-1202 & 164 & 0.1 & 81, 122 & 0.1 & 1e-4 & 1202 & block name: BasicBlock \\
				\midrule
				\multirow{16}{2cm}{TinyImageNet} &  VGG-11 & 300 & 1e-3 & 100, 200 & 0.1 & 1e-4 &  11 &  - \\
				&  VGG-11 (BN) & 300 & 1e-3 & 100, 200 & 0.1 & 1e-4 &  11 &  - \\
				&  VGG-13 & 300 & 1e-3 & 100, 200 & 0.1 & 1e-4 &  13 &  - \\
				&  VGG-13 (BN) & 300 & 1e-3 & 100, 200 & 0.1 & 1e-4 &  13 &  - \\
				&  VGG-16 & 300 & 1e-3 & 100, 200 & 0.1 & 1e-4 &  16 &  - \\
				&  VGG-16 (BN) & 300 & 1e-3 & 100, 200 & 0.1 & 1e-4 &  16 &  - \\
				&  VGG-19 & 300 & 1e-3 & 100, 200 & 0.1 & 1e-4 &  19 &  - \\
				&  VGG-19 (BN) & 300 & 1e-3 & 100, 200 & 0.1 & 1e-4 &  19 & - \\
				&  ResNet-18 & 300 & 1e-3 & 100, 200 & 0.1 & 1e-4 &  18 &  block name: BasicBlock \\
				&  ResNet-34 & 300 & 1e-3 & 100, 200 & 0.1 & 1e-4 &  34 & block name: BasicBlock \\
				&  ResNet-50 & 300 & 1e-3 & 100, 200 & 0.1 & 1e-4 &  50 &  block name: Bottleneck \\
				&  ResNet-101 & 300 & 1e-3 & 100, 200 & 0.1 & 1e-4 &  101 &  block name: Bottleneck \\
				&  ResNet-152 & 300 & 1e-3 & 100, 200 & 0.1 & 1e-4 &  152 &  block name: Bottleneck \\
				& DenseNet-161& 300 & 1e-3 & 100, 200 & 0.1 & 1e-4 &  161 &  growth rate: 32 \\
				& DenseNet-169 & 300 & 1e-3 & 100, 200 & 0.1 & 1e-4 &  169 &  growth rate: 32 \\
				& DenseNet-201 & 300 & 1e-3 & 100, 200 & 0.1 & 1e-4 &  201 &  growth rate: 32 \\
				\bottomrule
		\end{tabular}}
	\end{center}
	\caption{The details of pre-trained classification networks, which are $\mathbb{N}_1$,$\cdots$,$\mathbb{N}_n$ used for the generation of training data in both the Simulator Attack and the Meta Attack. All the data of ResNet networks are excluded in the experiments of attacking defensive models.}
	\label{tab:pretrained_networks}
\end{table*}

\section{Experimental Results}

\subsection{Detailed Experimental Figures}

\begin{figure*}[hb]
	\setlength{\abovecaptionskip}{0pt}
	\setlength{\belowcaptionskip}{0pt}
	\captionsetup[sub]{font={scriptsize}}
	\centering 
	\begin{minipage}[b]{.245\textwidth}
		\includegraphics[width=\linewidth]{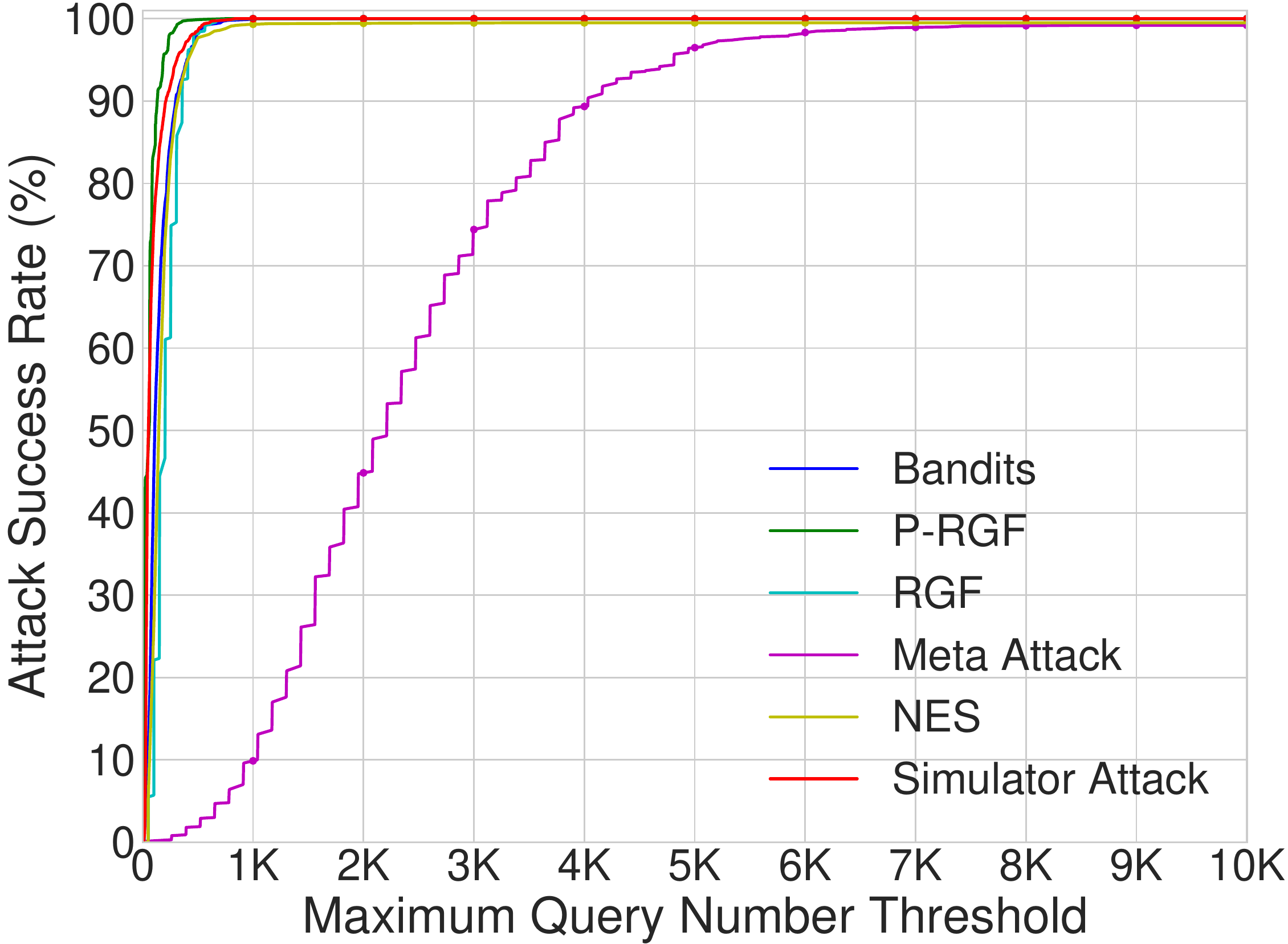}
		\subcaption{untargeted $\ell_2$ attack PyramidNet-272}
	\end{minipage}
	\begin{minipage}[b]{.245\textwidth}
		\includegraphics[width=\linewidth]{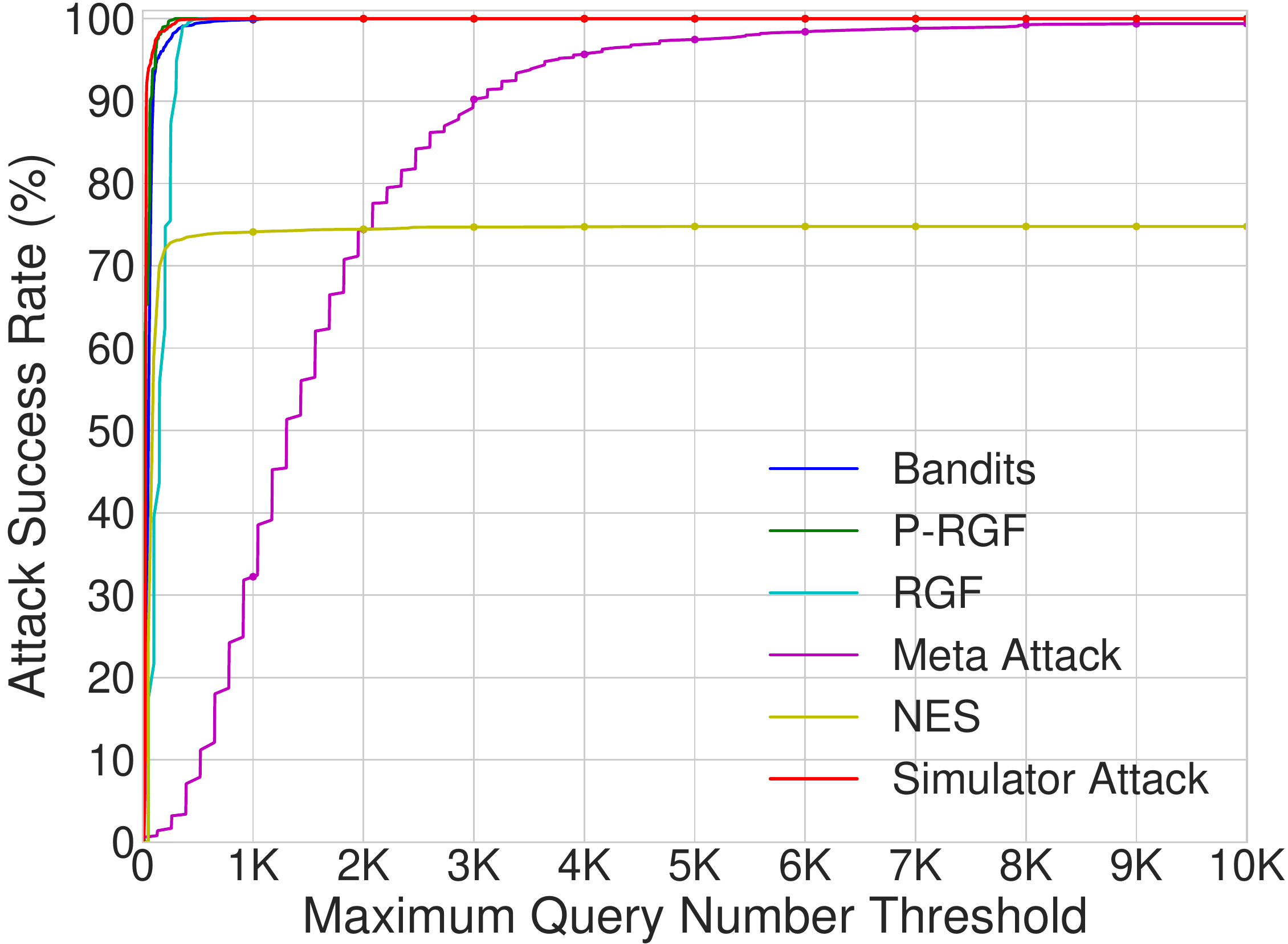}
		\subcaption{untargeted $\ell_2$ attack GDAS}
	\end{minipage}
	\begin{minipage}[b]{.245\textwidth}
		\includegraphics[width=\linewidth]{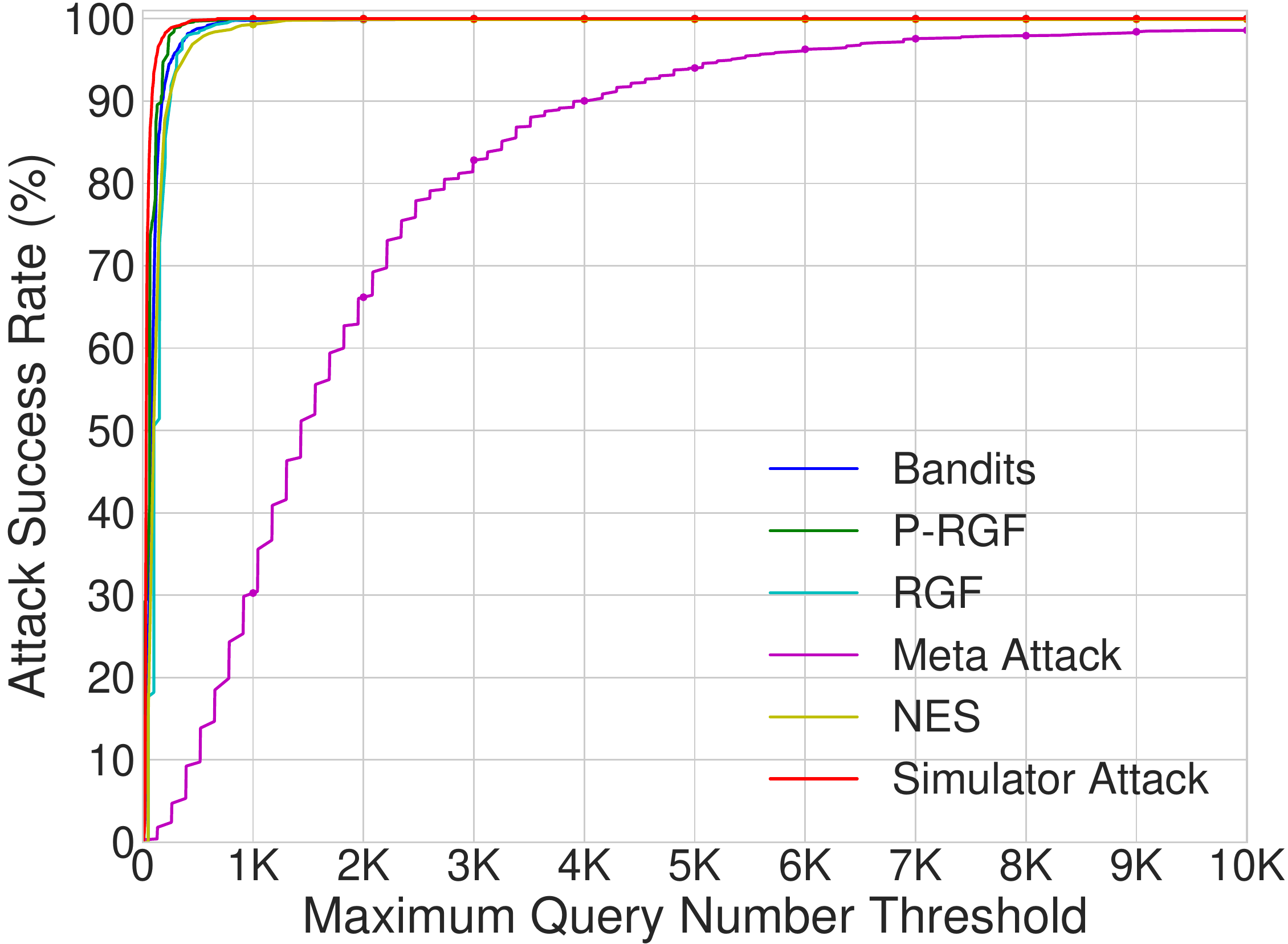}
		\subcaption{untargeted $\ell_2$ attack WRN-28}
	\end{minipage}
	\begin{minipage}[b]{.245\textwidth}
		\includegraphics[width=\linewidth]{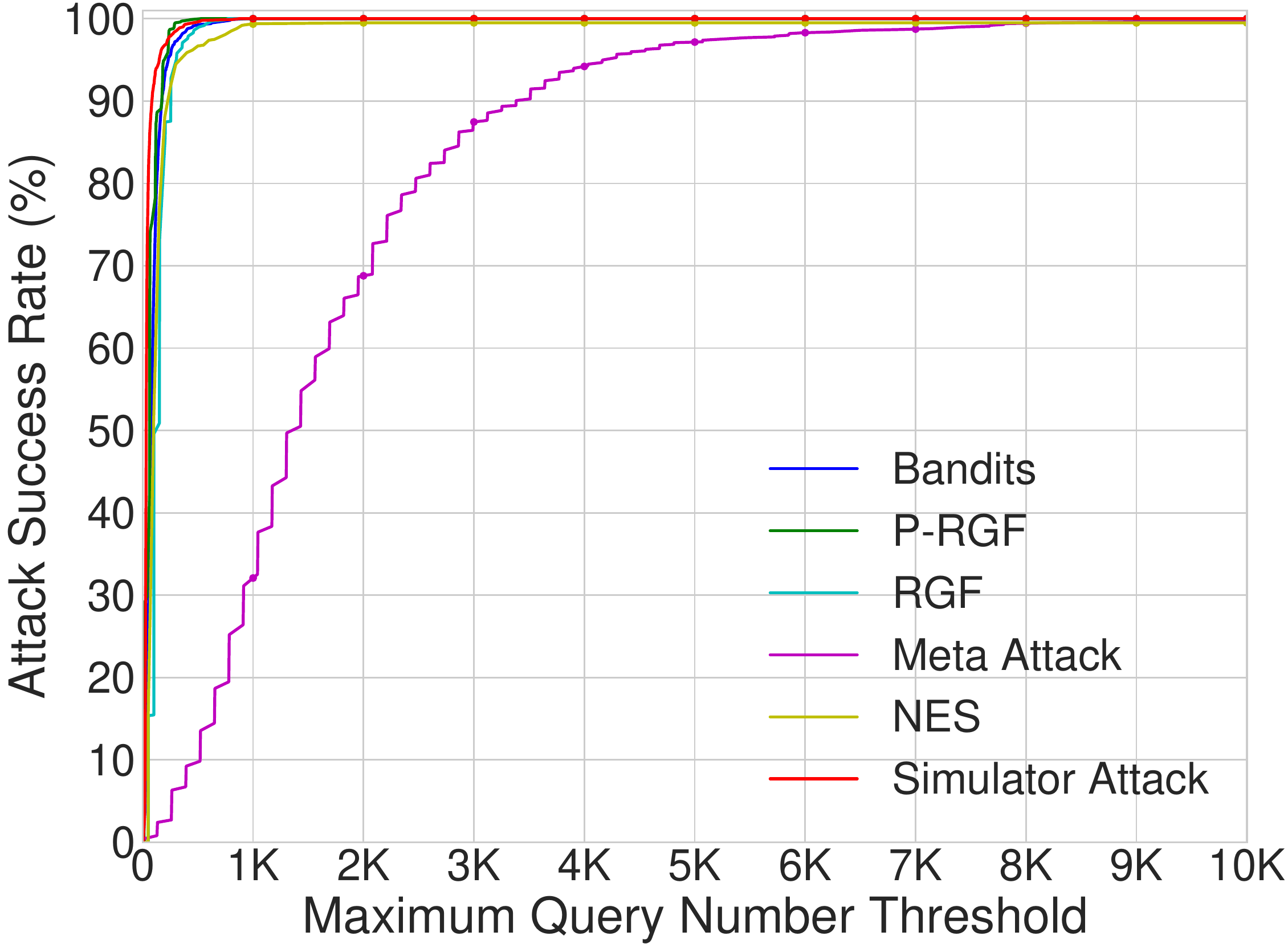}
		\subcaption{untargeted $\ell_2$ attack WRN-40}
	\end{minipage}
	\begin{minipage}[b]{.245\textwidth}
		\includegraphics[width=\linewidth]{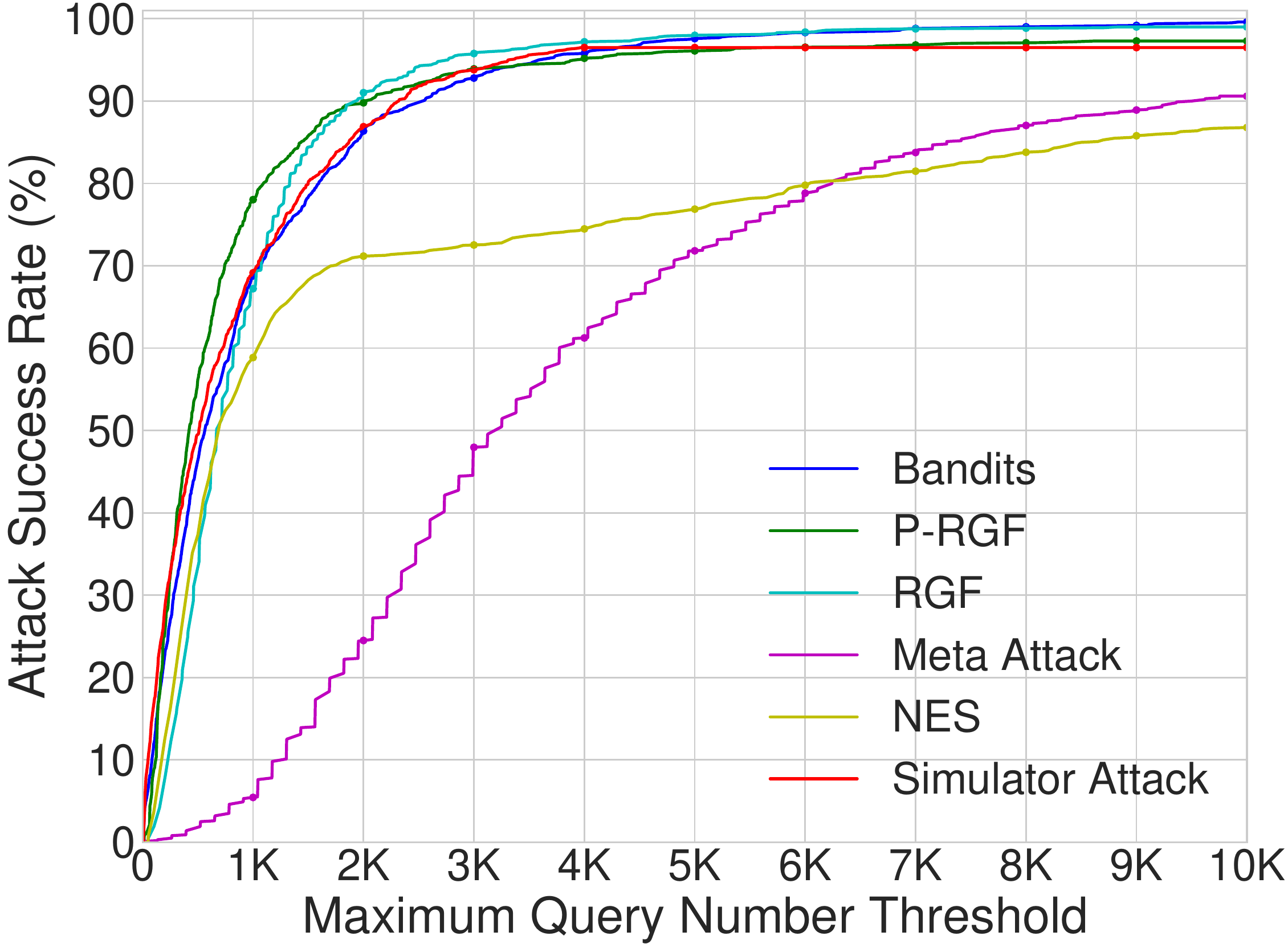}
		\subcaption{untargeted $\ell_\infty$ attack PyramidNet-272}
	\end{minipage}
	\begin{minipage}[b]{.245\textwidth}
		\includegraphics[width=\linewidth]{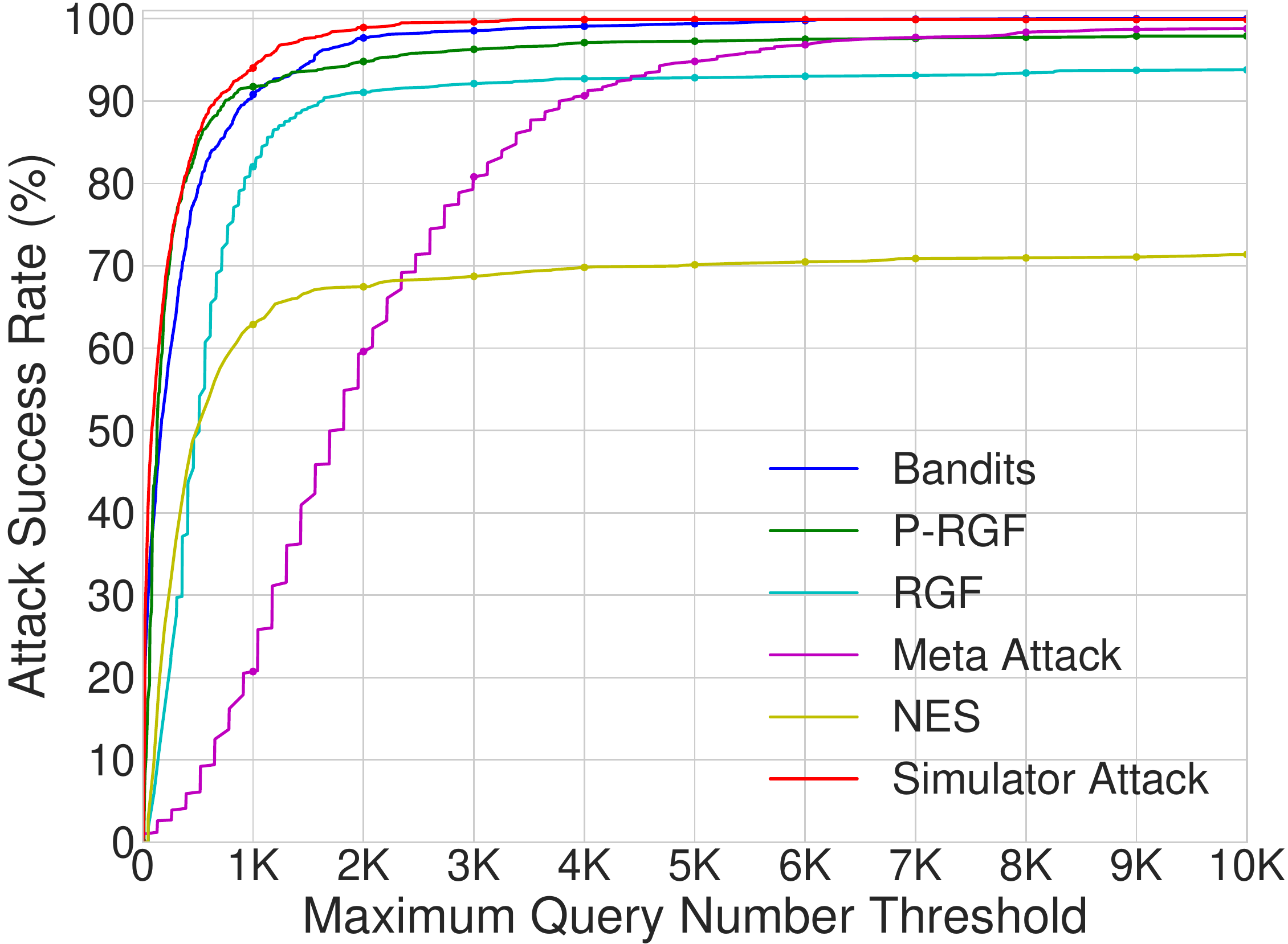}
		\subcaption{untargeted $\ell_\infty$ attack GDAS}
	\end{minipage}
	\begin{minipage}[b]{.245\textwidth}
		\includegraphics[width=\linewidth]{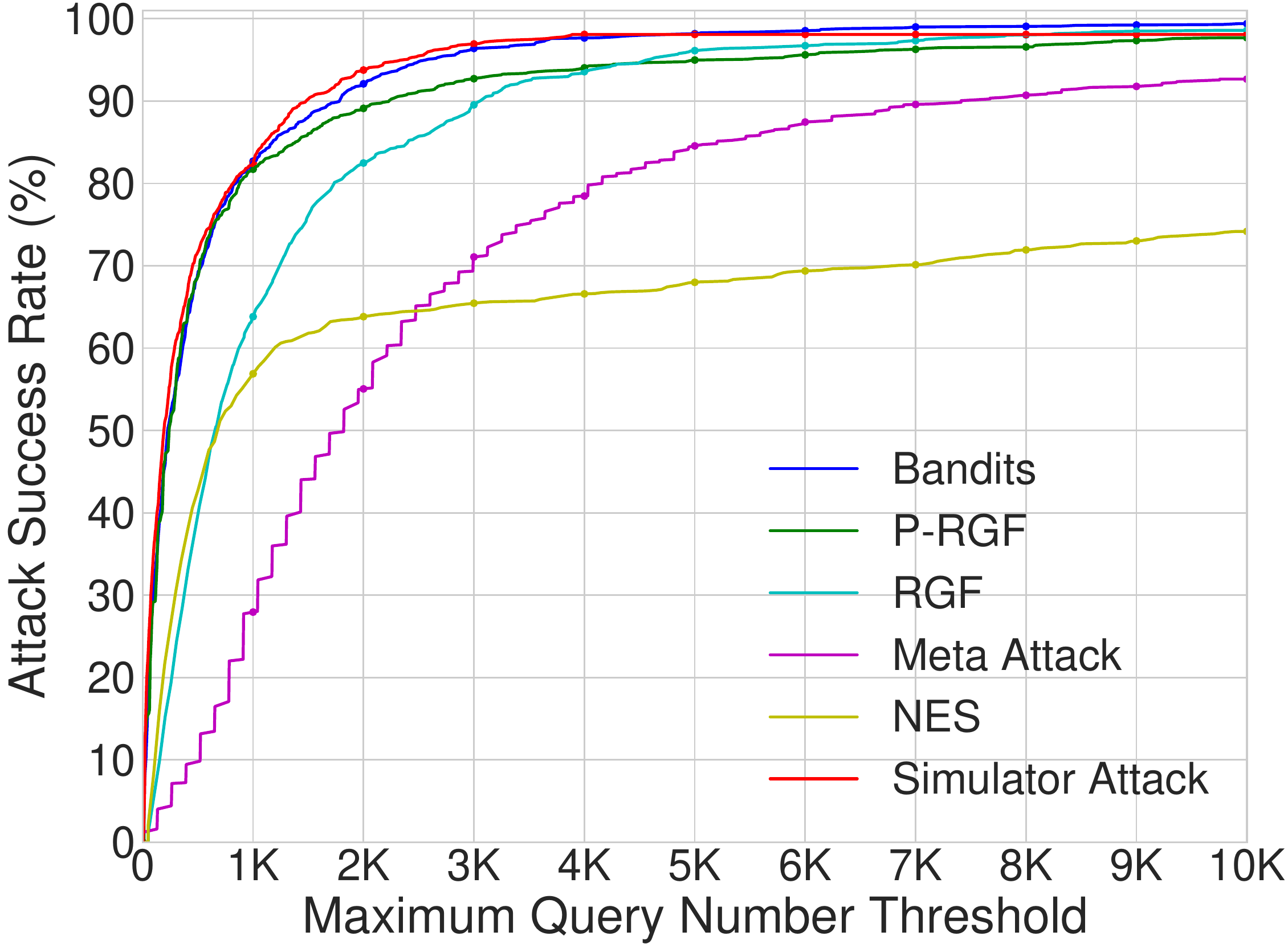}
		\subcaption{untargeted $\ell_\infty$ attack WRN-28}
	\end{minipage}
	\begin{minipage}[b]{.245\textwidth}
		\includegraphics[width=\linewidth]{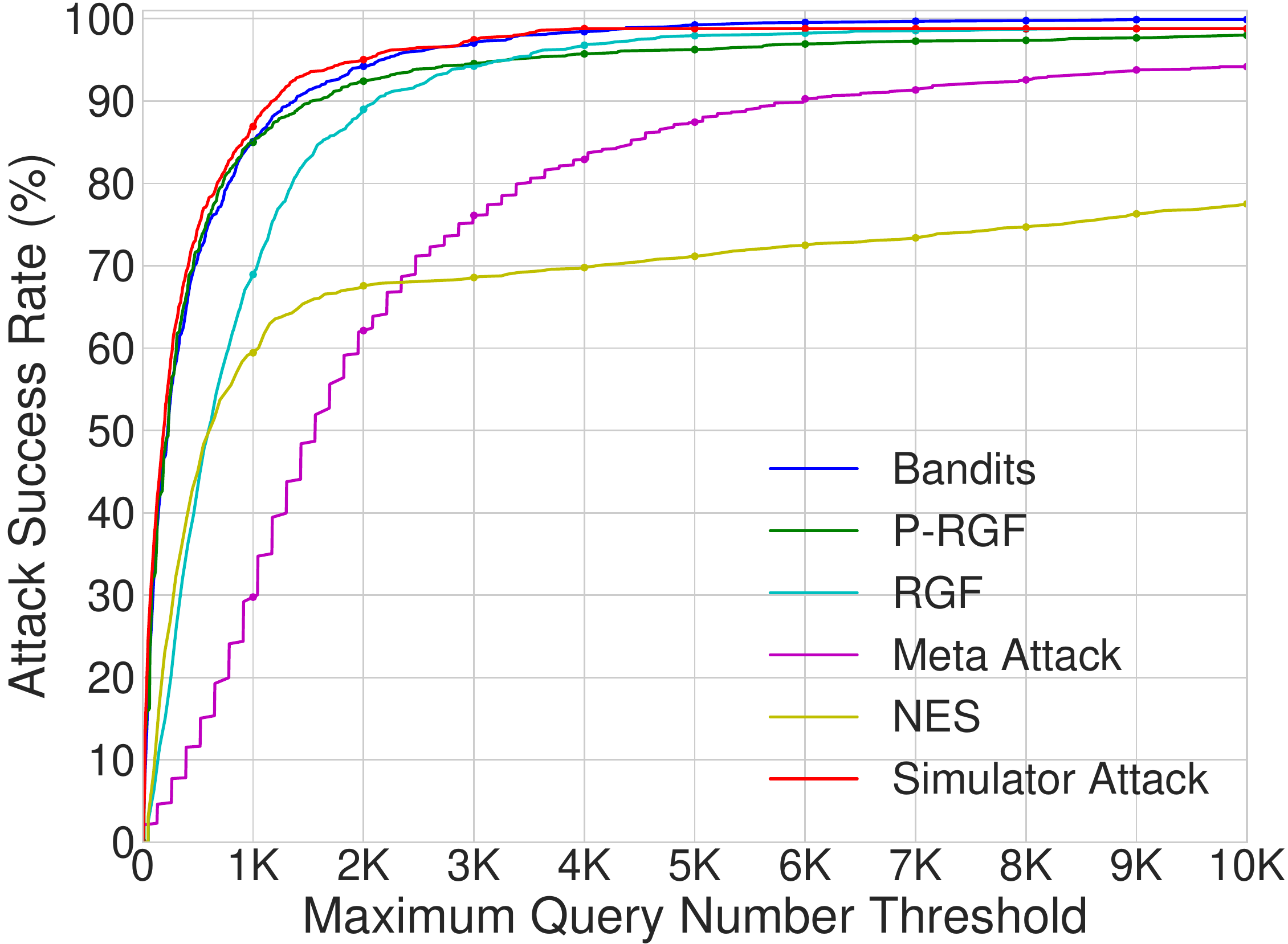}
		\subcaption{untargeted $\ell_\infty$ attack WRN-40}
	\end{minipage}
	\begin{minipage}[b]{.245\textwidth}
		\includegraphics[width=\linewidth]{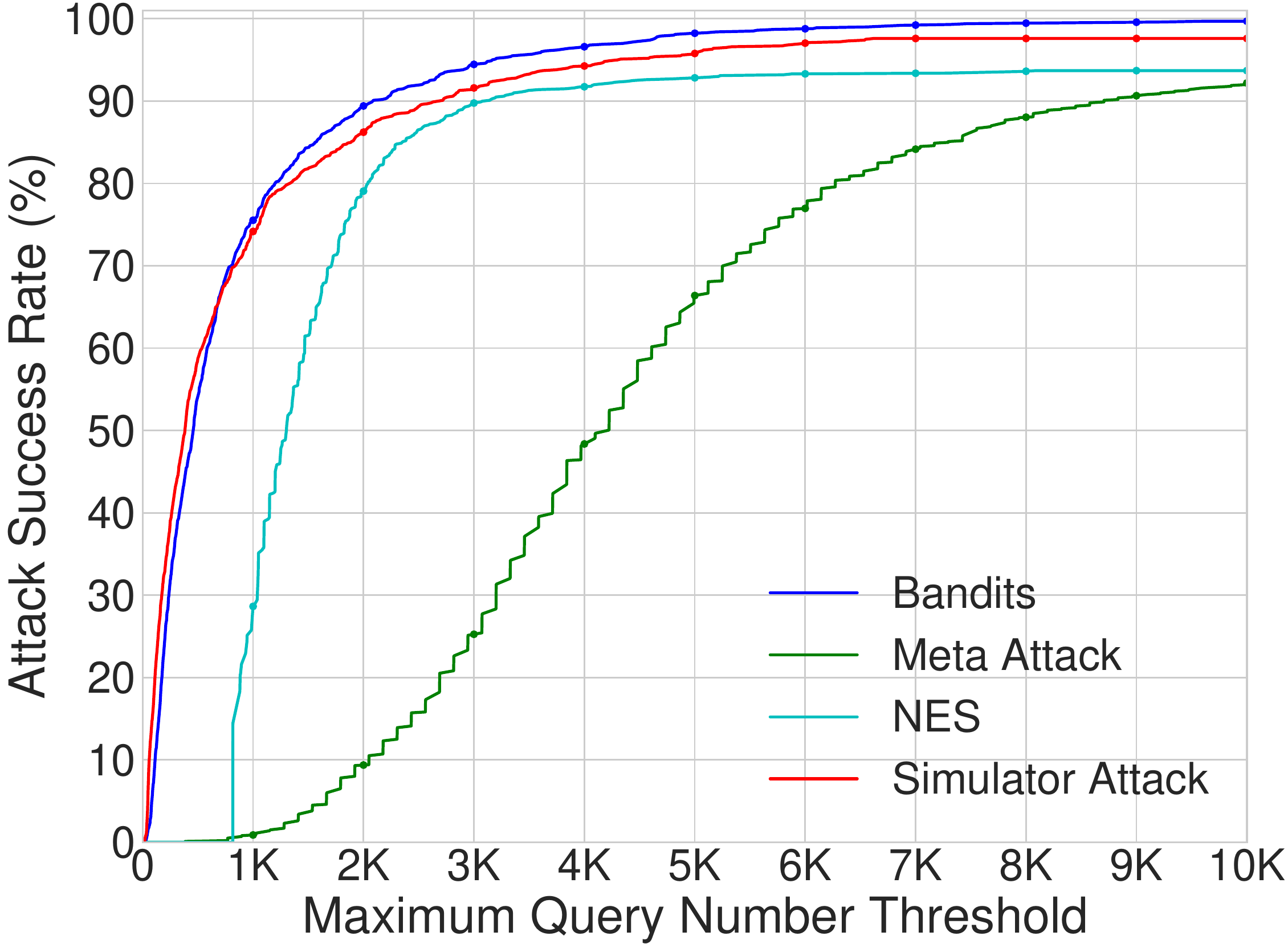}
		\subcaption{targeted $\ell_2$ attack PyramidNet-272}
	\end{minipage}
	\begin{minipage}[b]{.245\textwidth}
		\includegraphics[width=\linewidth]{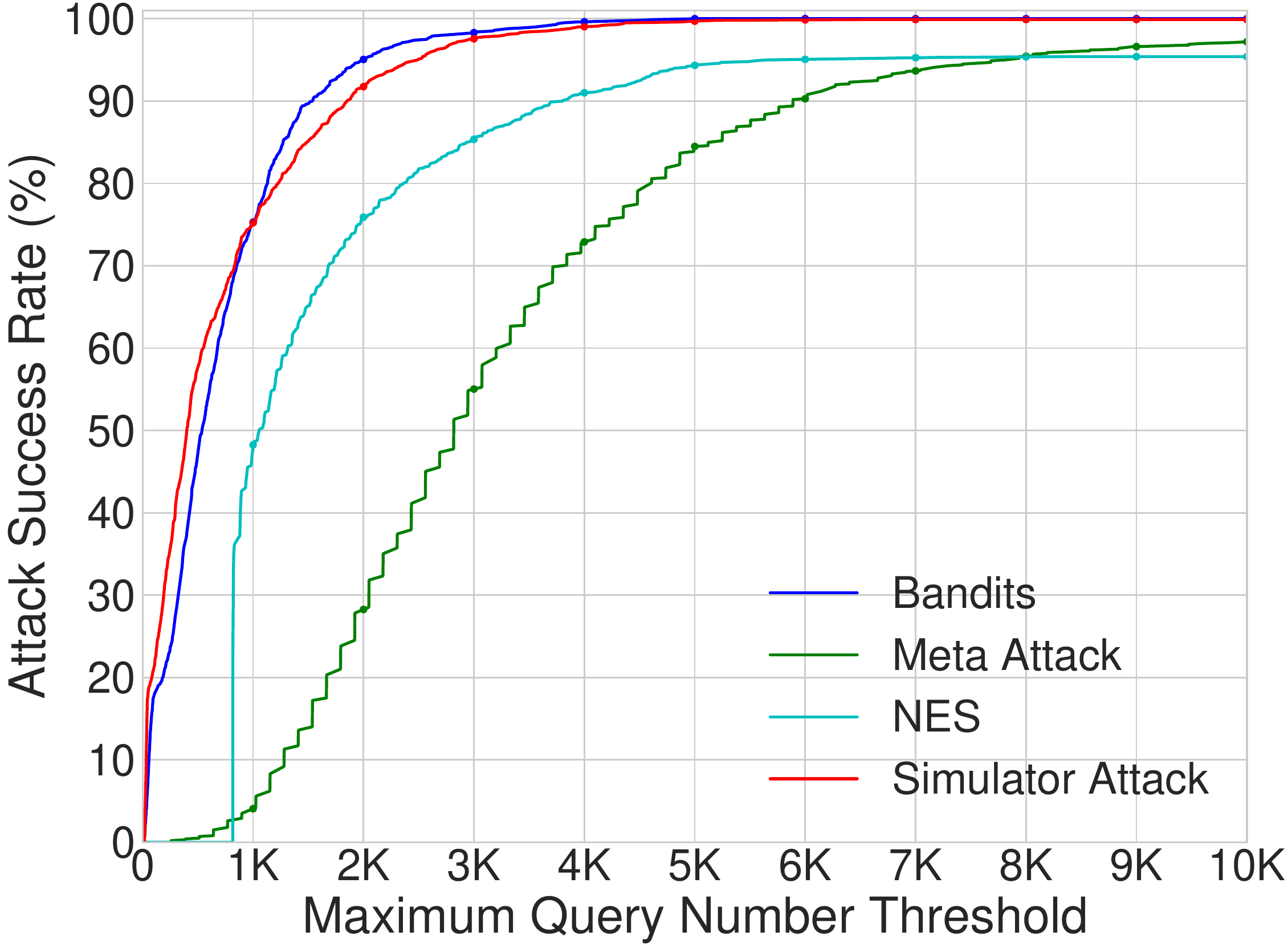}
		\subcaption{targeted $\ell_2$ attack GDAS}
	\end{minipage}
	\begin{minipage}[b]{.245\textwidth}
		\includegraphics[width=\linewidth]{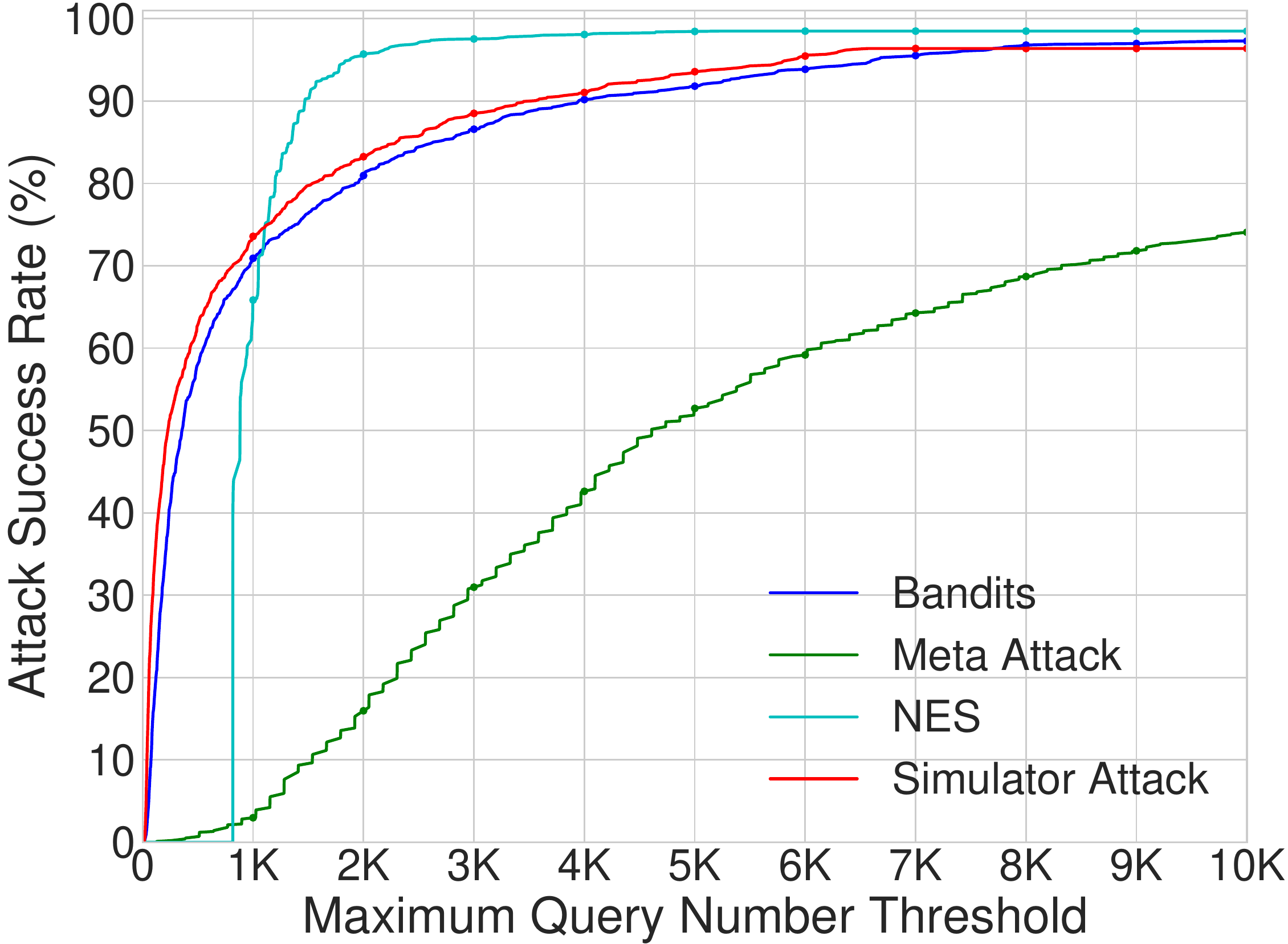}
		\subcaption{targeted $\ell_2$ attack WRN-28}
	\end{minipage}
	\begin{minipage}[b]{.245\textwidth}
		\includegraphics[width=\linewidth]{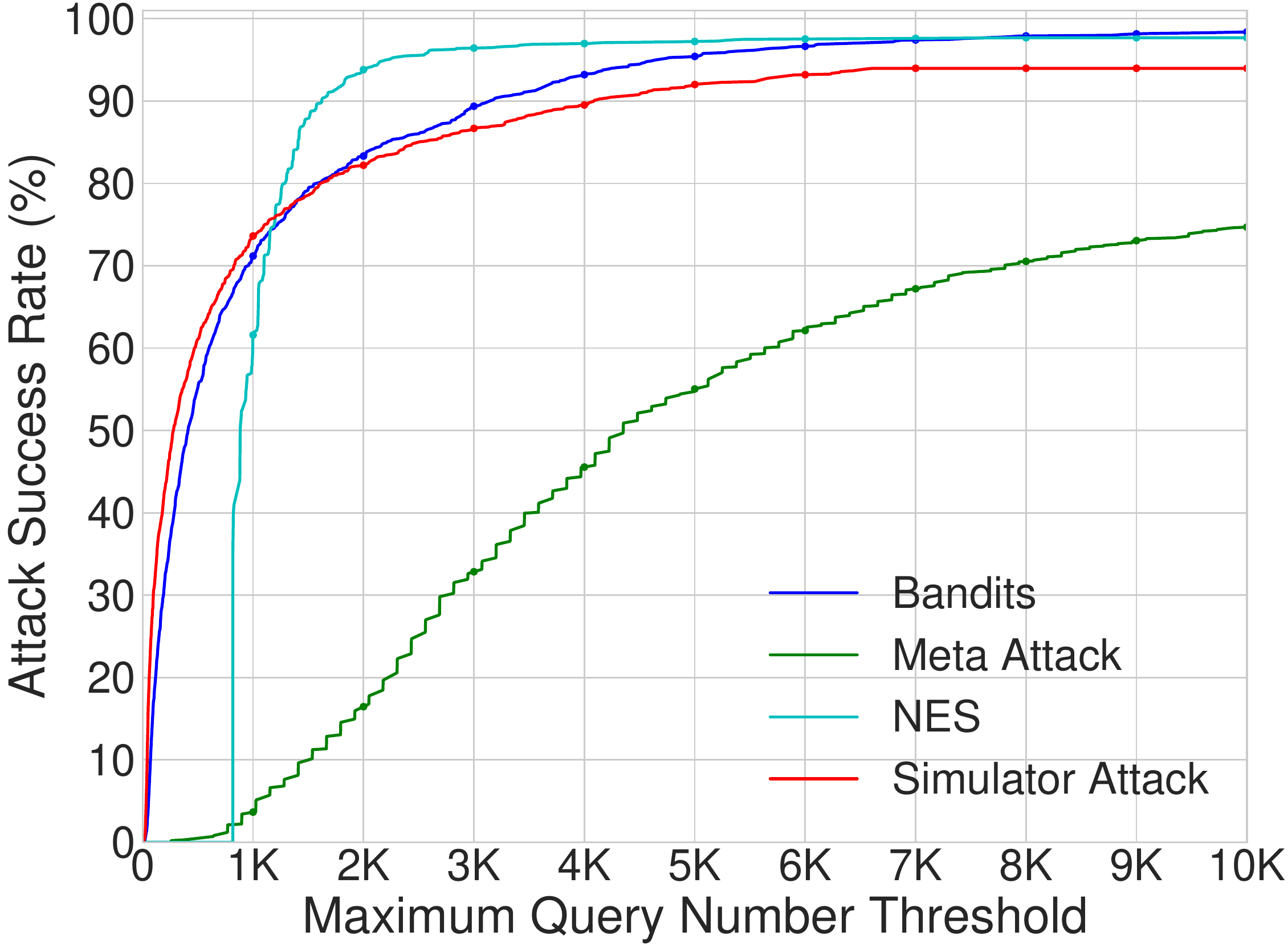}
		\subcaption{targeted $\ell_2$ attack WRN-40}
	\end{minipage}
	\begin{minipage}[b]{.245\textwidth}
		\includegraphics[width=\linewidth]{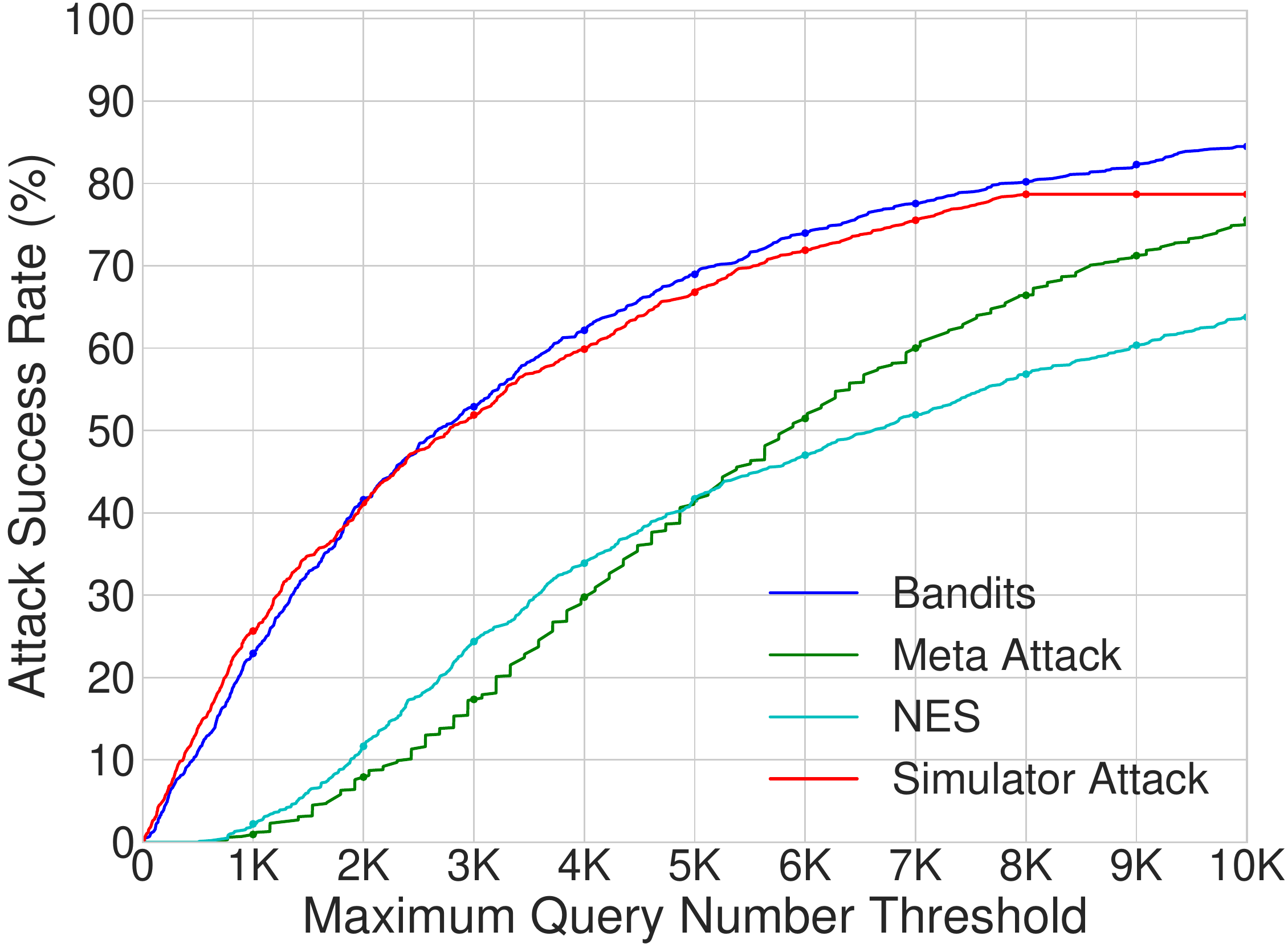}
		\subcaption{targeted $\ell_\infty$ attack PyramidNet-272}
	\end{minipage}
	\begin{minipage}[b]{.245\textwidth}
		\includegraphics[width=\linewidth]{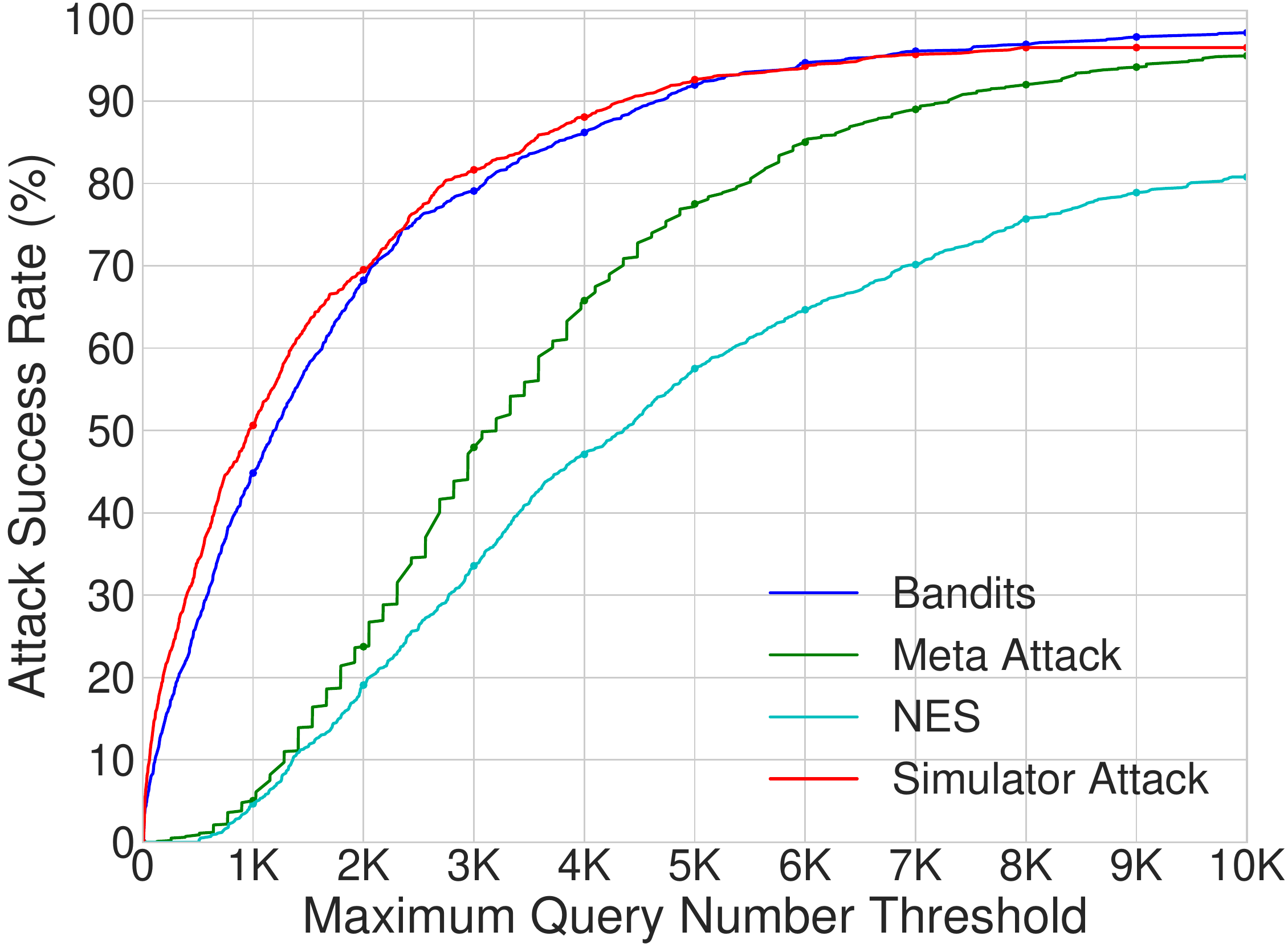}
		\subcaption{targeted $\ell_\infty$ attack GDAS}
	\end{minipage}
	\begin{minipage}[b]{.245\textwidth}
		\includegraphics[width=\linewidth]{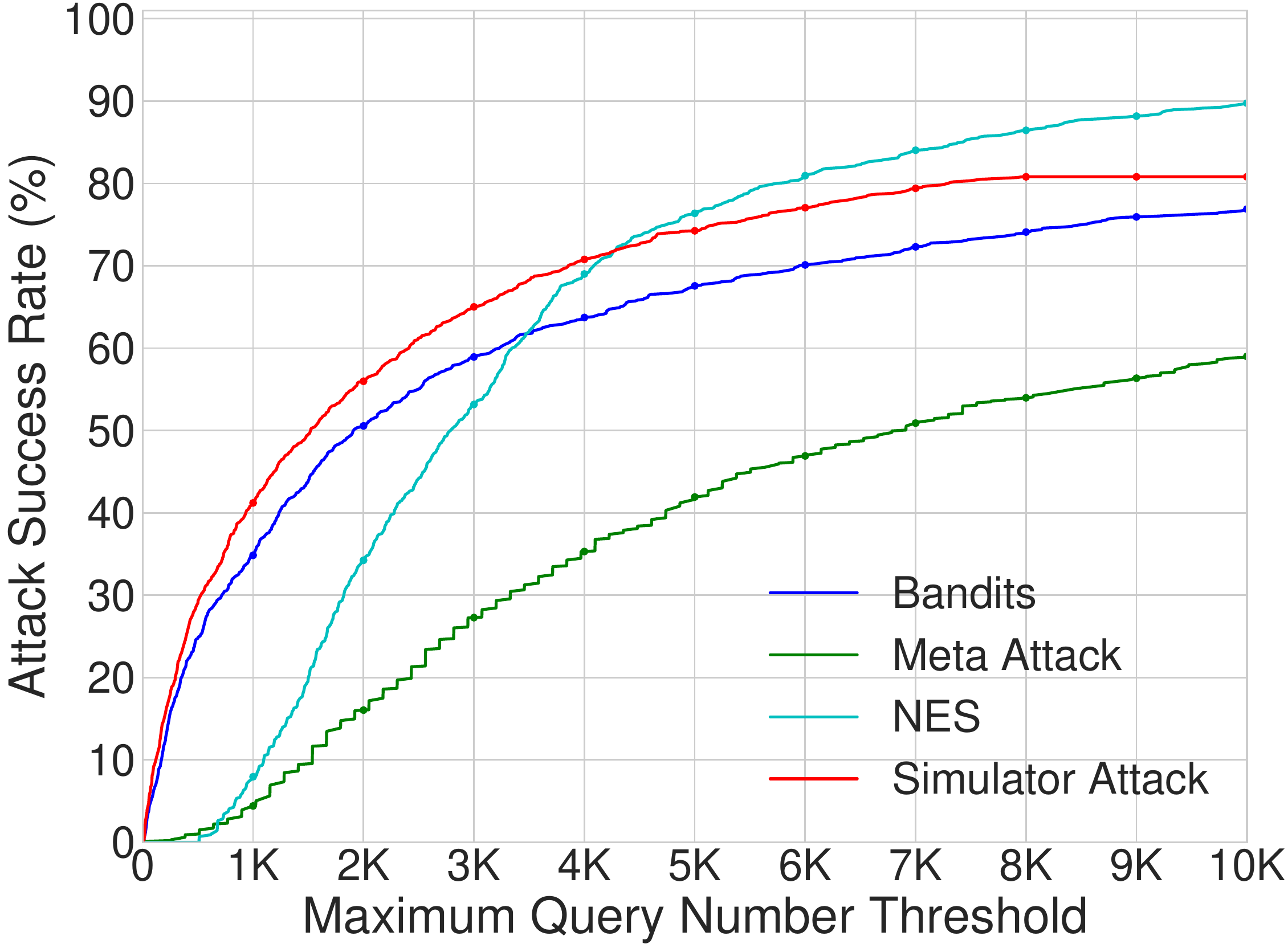}
		\subcaption{targeted $\ell_\infty$ attack WRN-28}
	\end{minipage}
	\begin{minipage}[b]{.245\textwidth}
		\includegraphics[width=\linewidth]{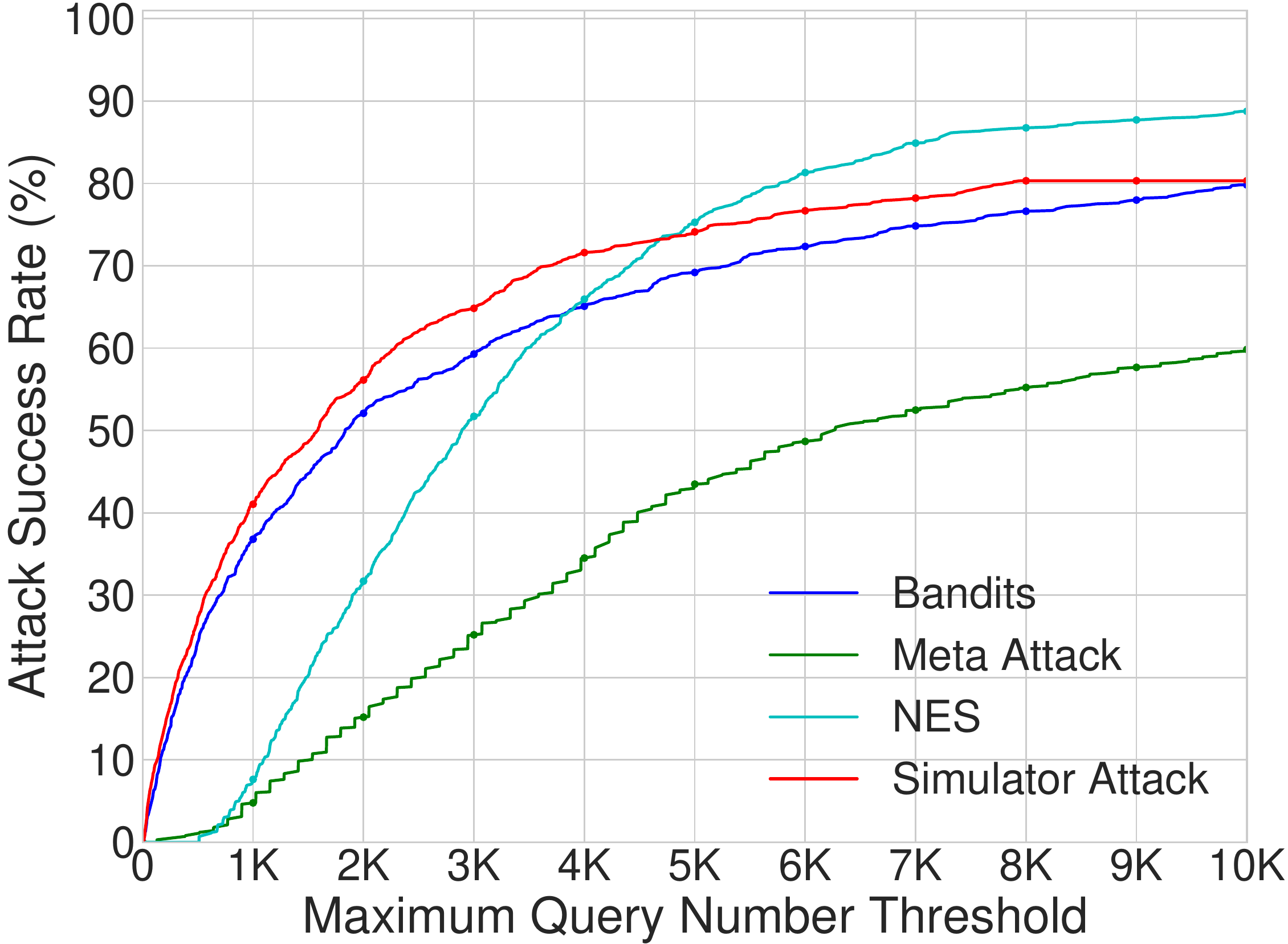}
		\subcaption{targeted $\ell_\infty$ attack WRN-40}
	\end{minipage}
	\caption{Comparisons of attack success rates at different limited maximum queries in CIFAR-10 dataset.}
	\label{fig:query_to_attack_success_rate_CIFAR-10}
\end{figure*}

\begin{figure*}[hb]
	\setlength{\abovecaptionskip}{0pt}
	\setlength{\belowcaptionskip}{0pt}
	\captionsetup[sub]{font={scriptsize}}
	\centering 
	\begin{minipage}[b]{.245\textwidth}
		\includegraphics[width=\linewidth]{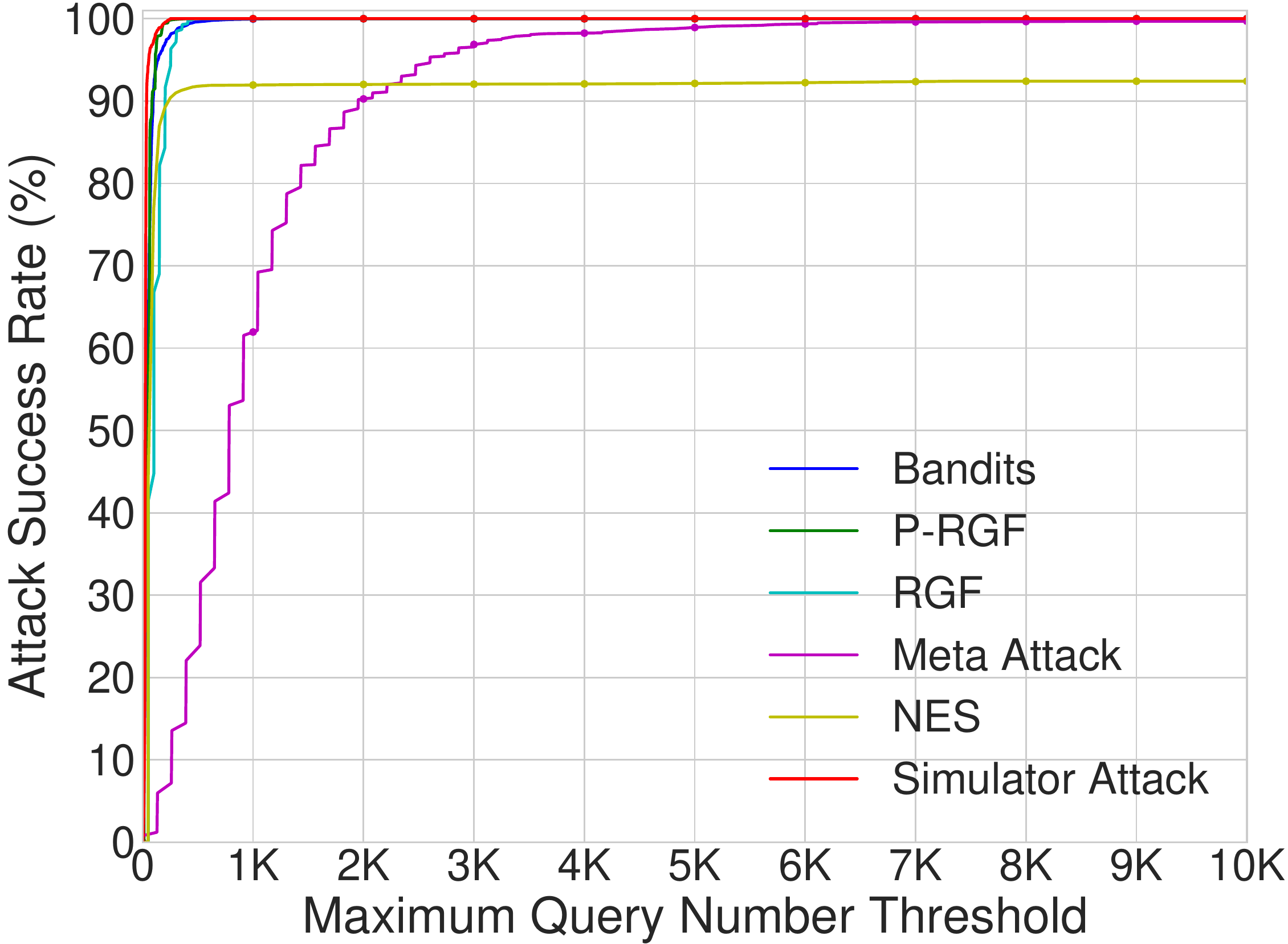}
		\subcaption{untargeted $\ell_2$ attack PyramidNet-272}
	\end{minipage}
	\begin{minipage}[b]{.245\textwidth}
		\includegraphics[width=\linewidth]{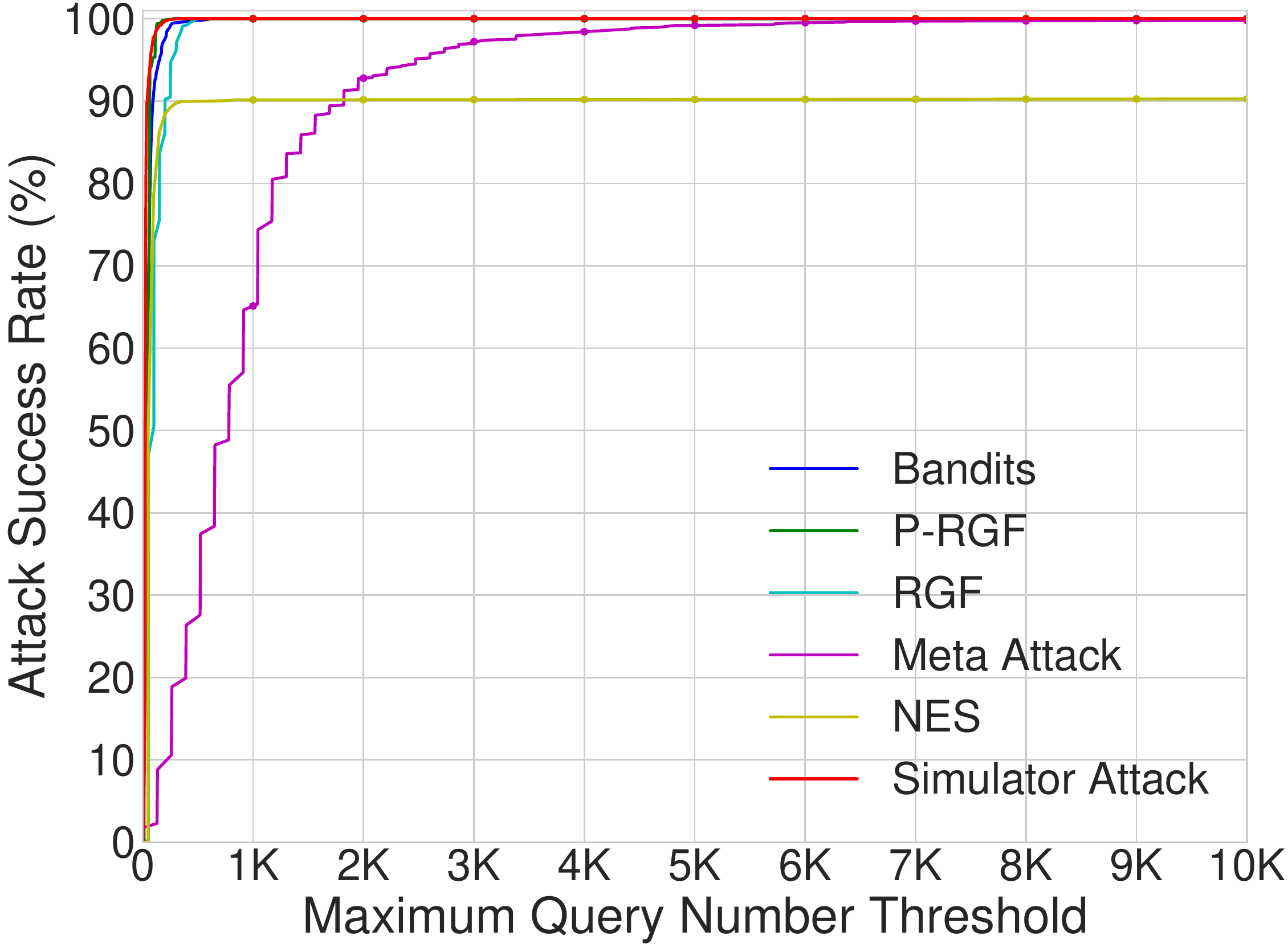}
		\subcaption{untargeted $\ell_2$ attack GDAS}
	\end{minipage}
	\begin{minipage}[b]{.245\textwidth}
		\includegraphics[width=\linewidth]{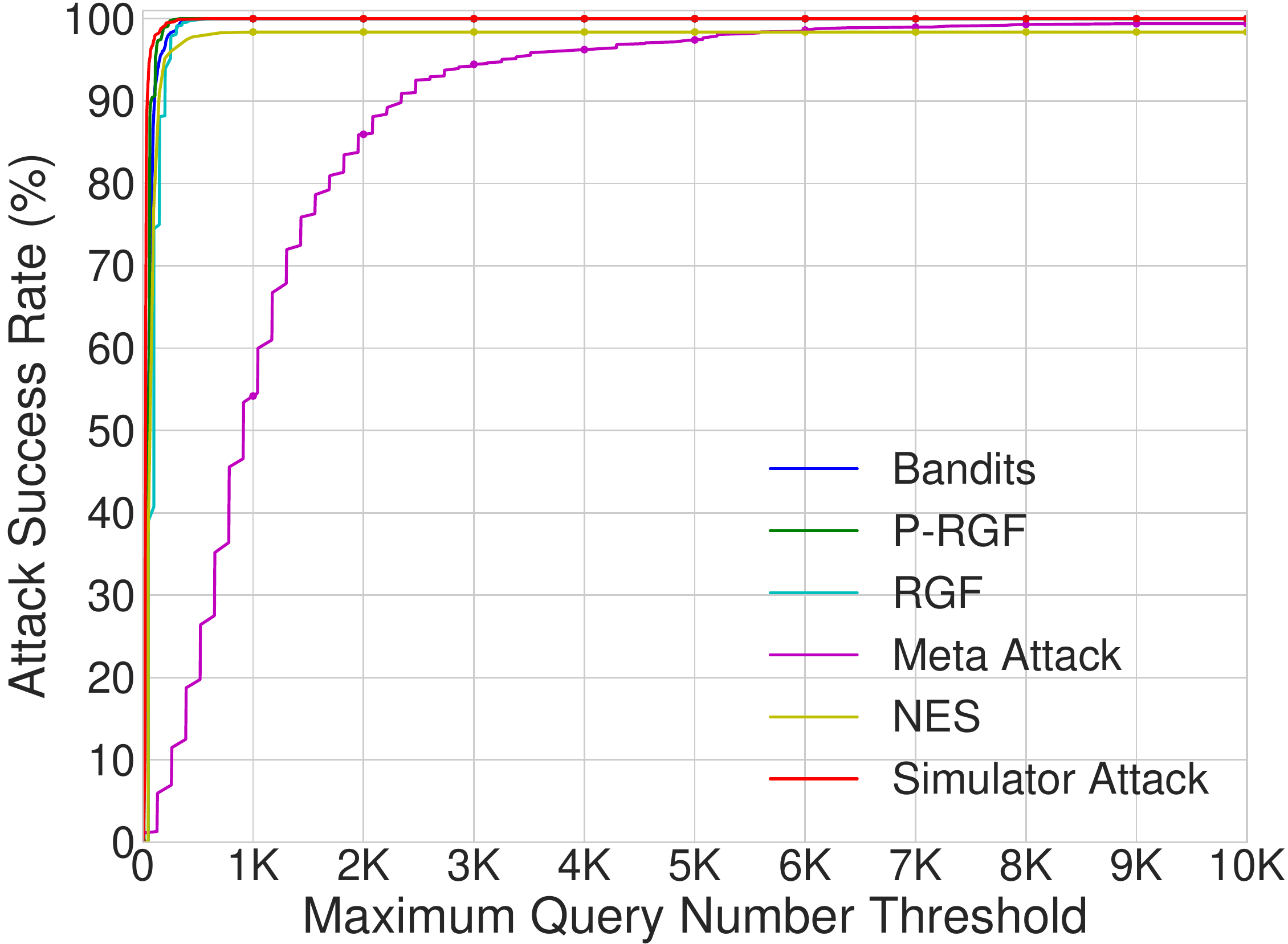}
		\subcaption{untargeted $\ell_2$ attack WRN-28}
	\end{minipage}
	\begin{minipage}[b]{.245\textwidth}
		\includegraphics[width=\linewidth]{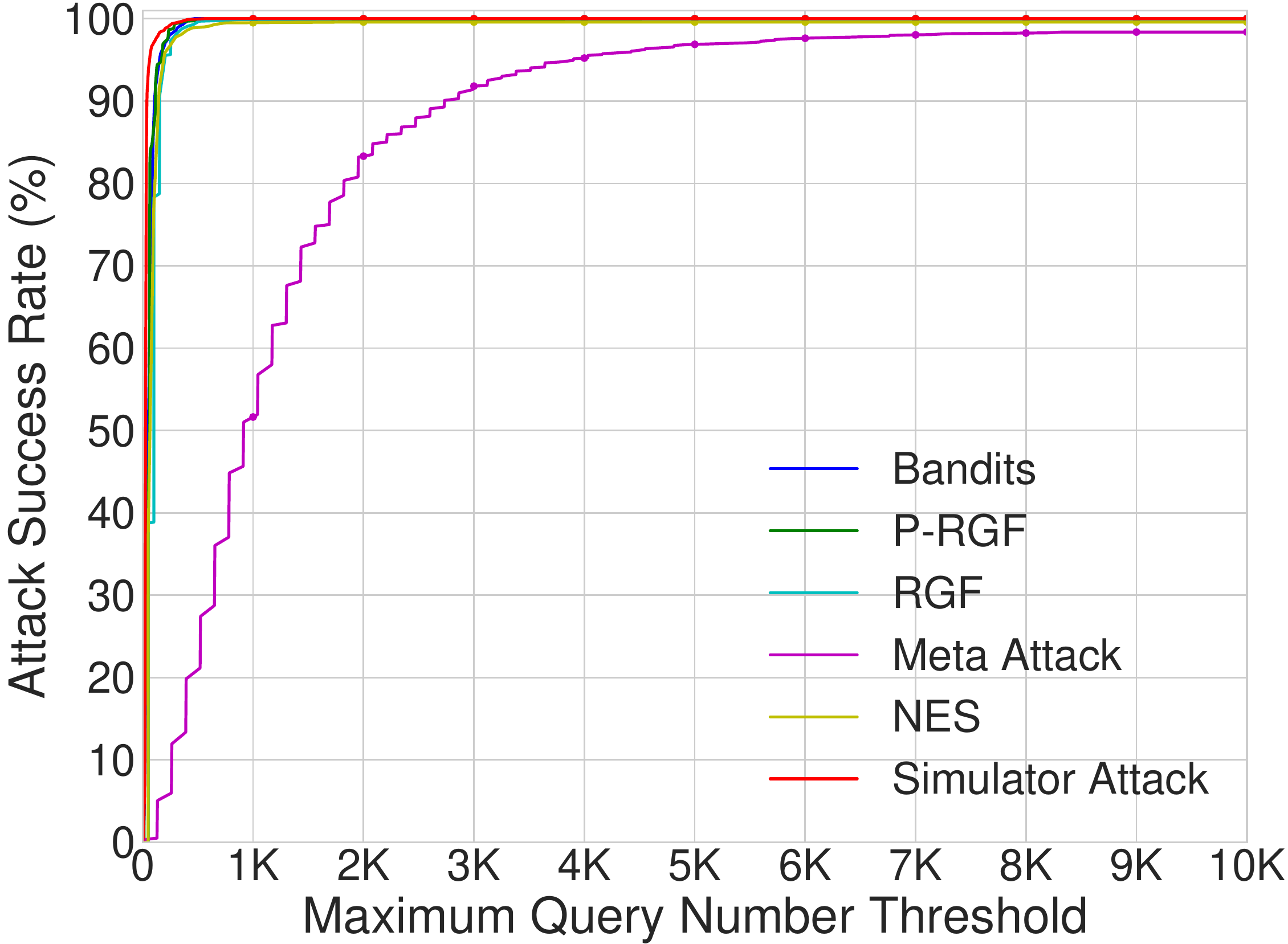}
		\subcaption{untargeted $\ell_2$ attack WRN-40}
	\end{minipage}
	\begin{minipage}[b]{.245\textwidth}
		\includegraphics[width=\linewidth]{figure/fig/query_threshold_attack_success_rate/CIFAR-100_pyramidnet272_linf_untargeted_attack_query_threshold_success_rate_dict.pdf}
		\subcaption{untargeted $\ell_\infty$ attack PyramidNet-272}
	\end{minipage}
	\begin{minipage}[b]{.245\textwidth}
		\includegraphics[width=\linewidth]{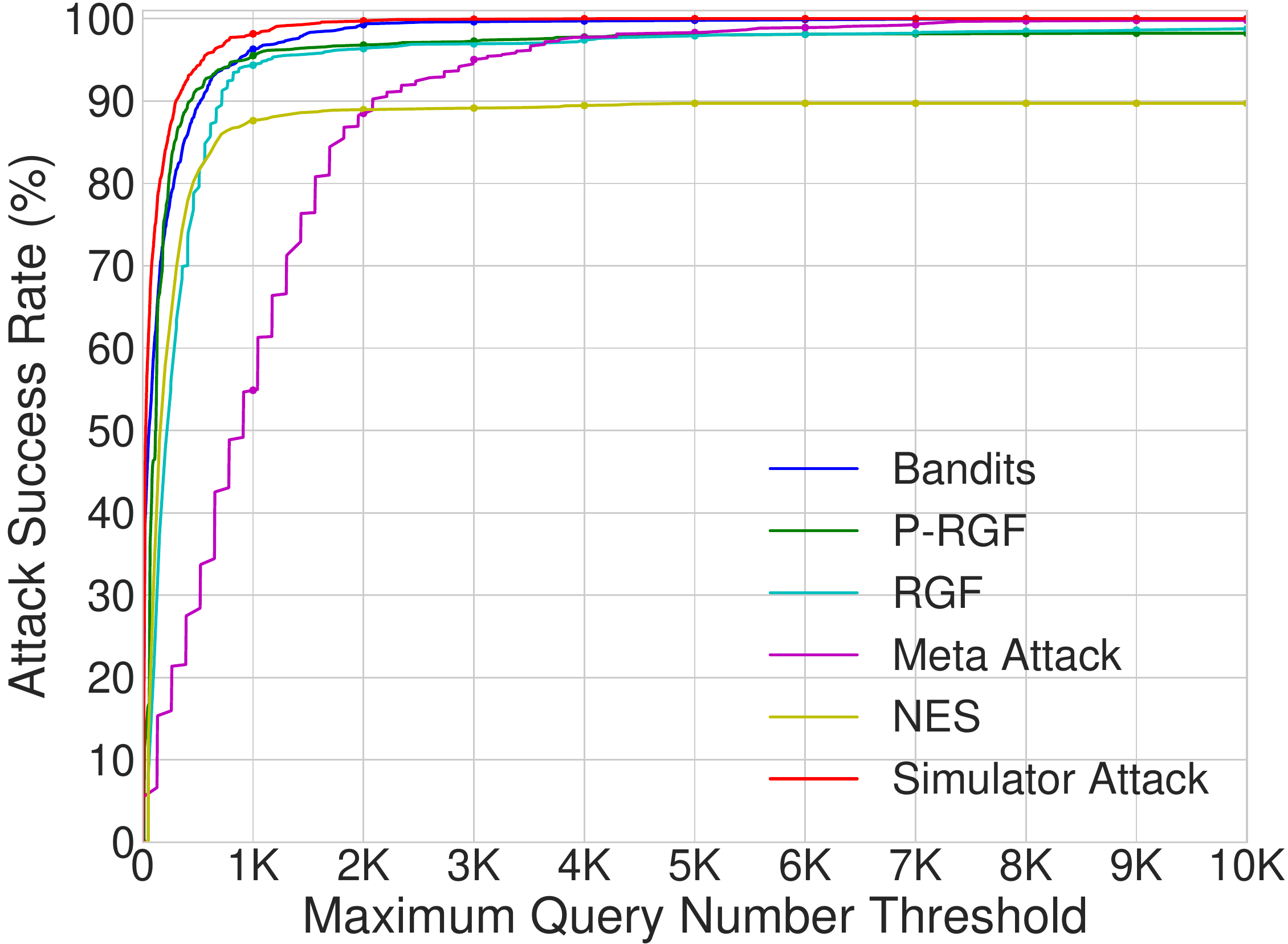}
		\subcaption{untargeted $\ell_\infty$ attack GDAS}
	\end{minipage}
	\begin{minipage}[b]{.245\textwidth}
		\includegraphics[width=\linewidth]{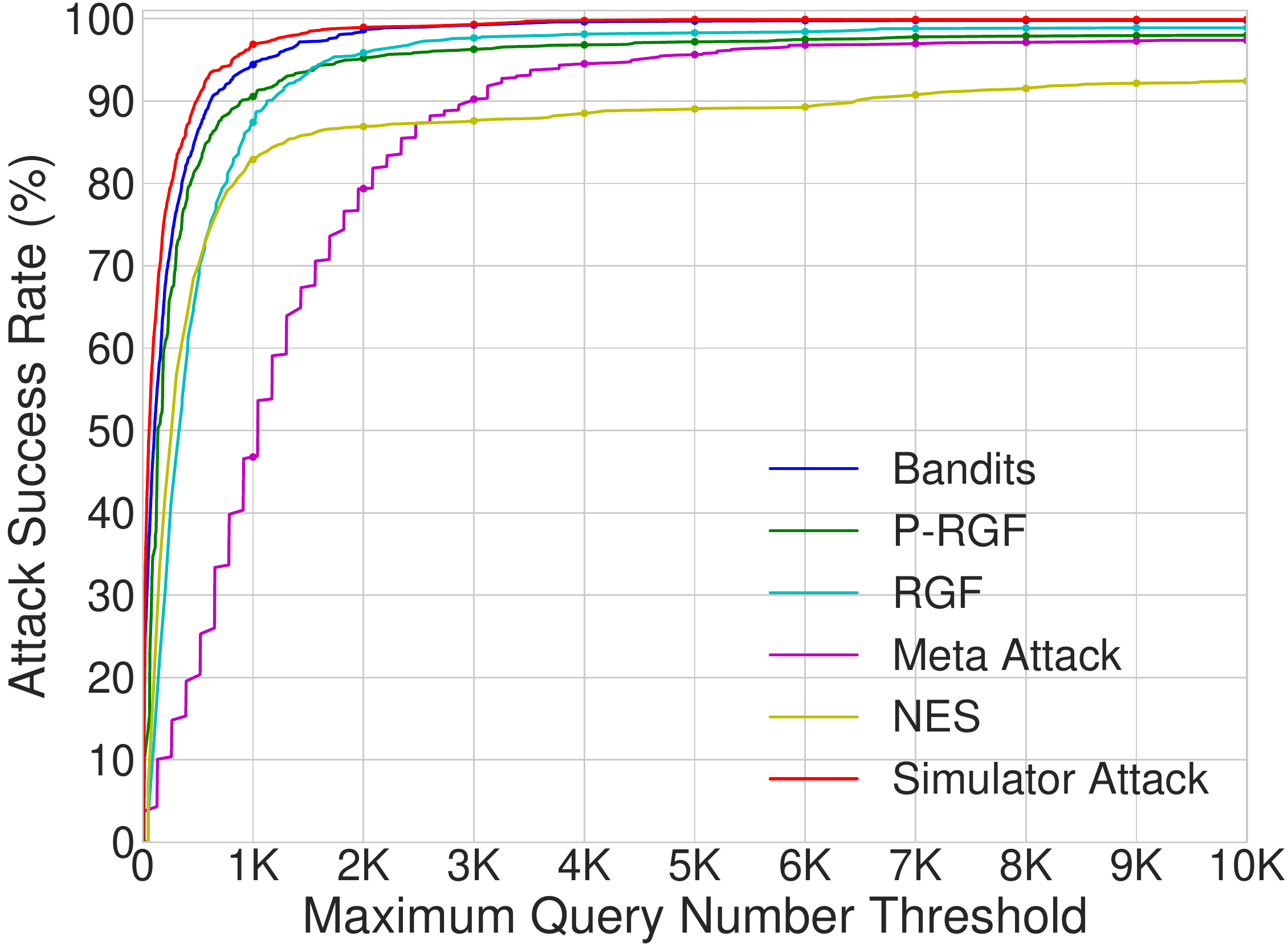}
		\subcaption{untargeted $\ell_\infty$ attack WRN-28}
	\end{minipage}
	\begin{minipage}[b]{.245\textwidth}
		\includegraphics[width=\linewidth]{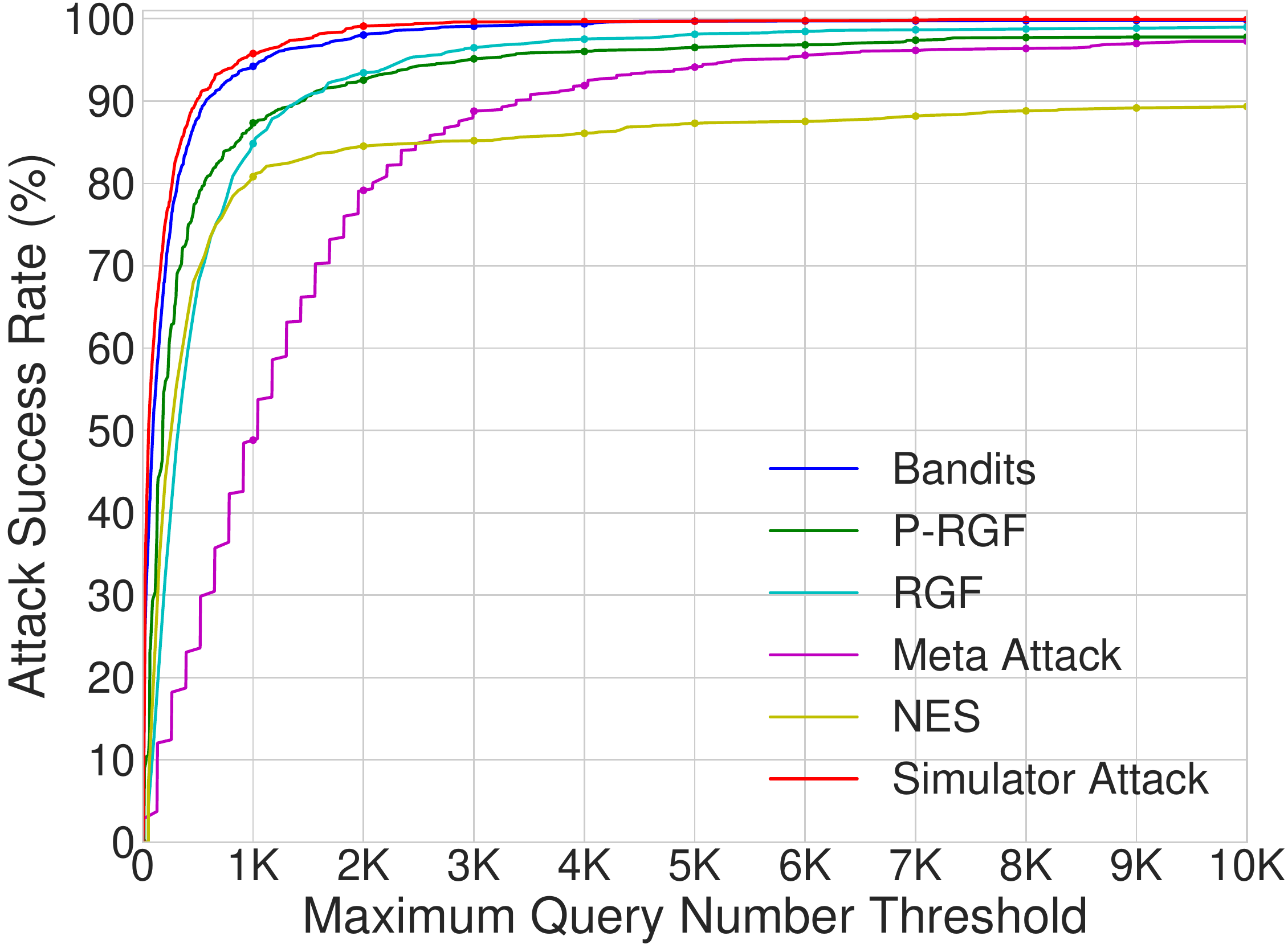}
		\subcaption{untargeted $\ell_\infty$ attack WRN-40}
	\end{minipage}
	\begin{minipage}[b]{.245\textwidth}
		\includegraphics[width=\linewidth]{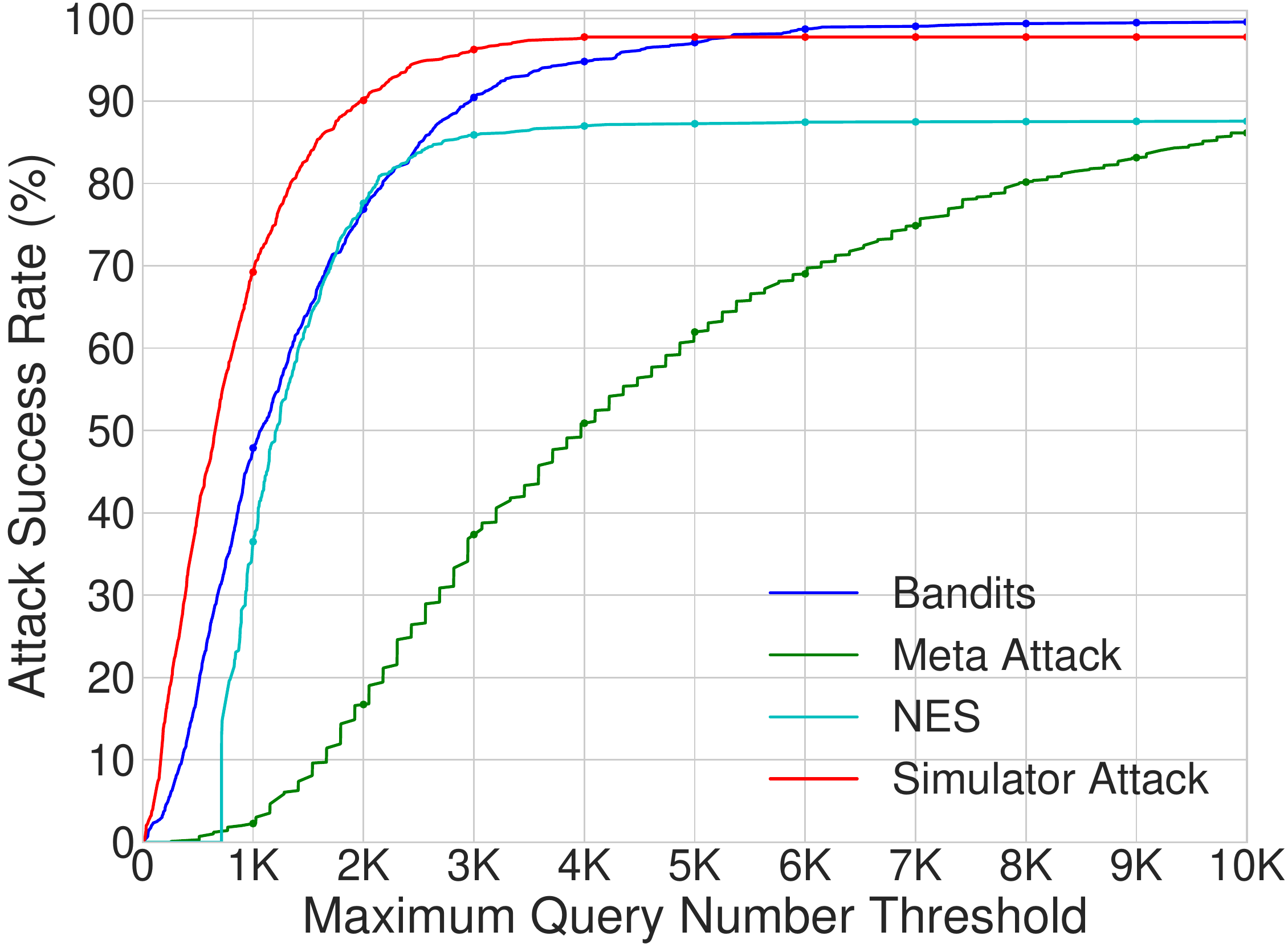}
		\subcaption{targeted $\ell_2$ attack PyramidNet-272}
	\end{minipage}
	\begin{minipage}[b]{.245\textwidth}
		\includegraphics[width=\linewidth]{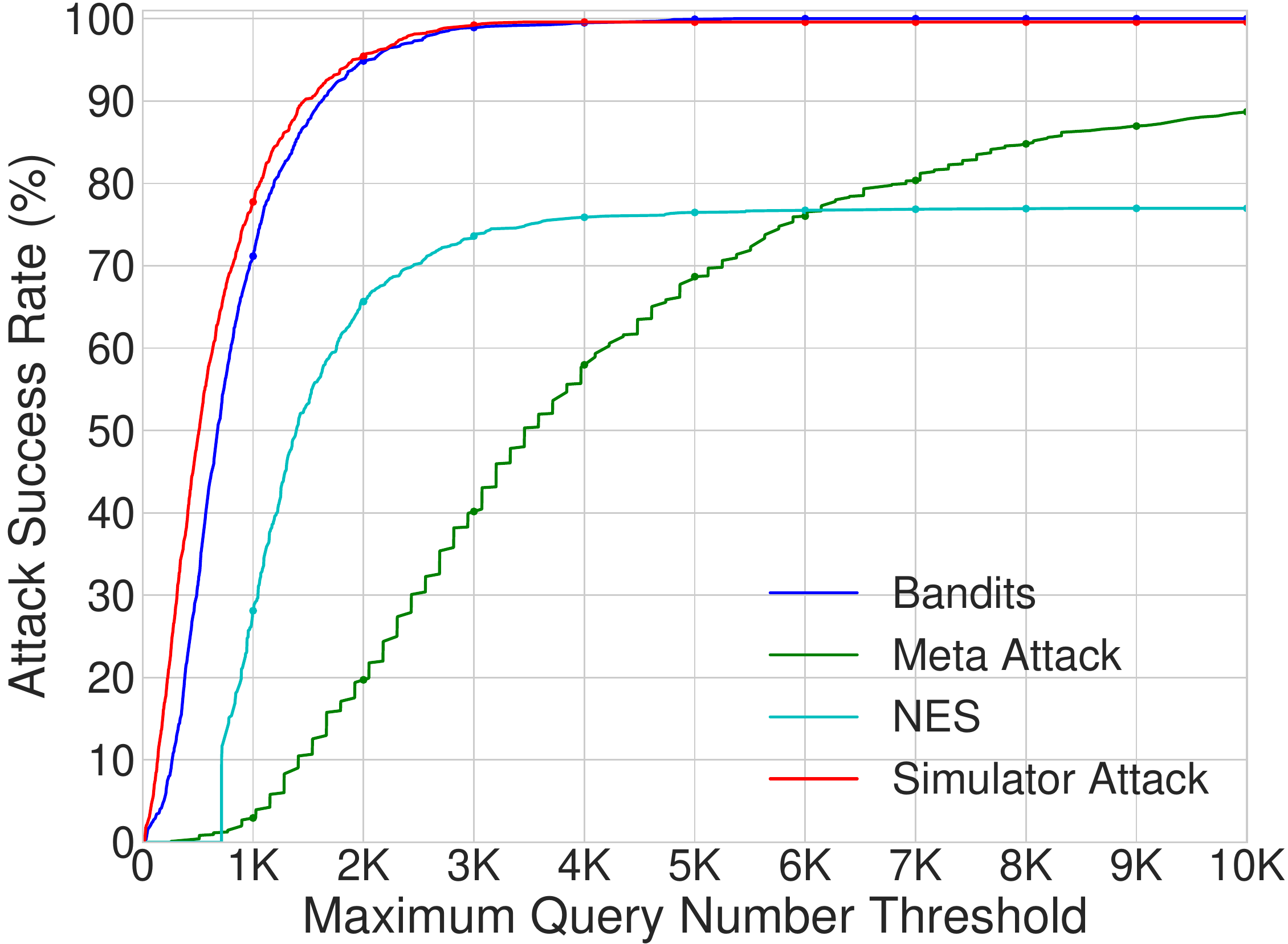}
		\subcaption{targeted $\ell_2$ attack GDAS}
	\end{minipage}
	\begin{minipage}[b]{.245\textwidth}
		\includegraphics[width=\linewidth]{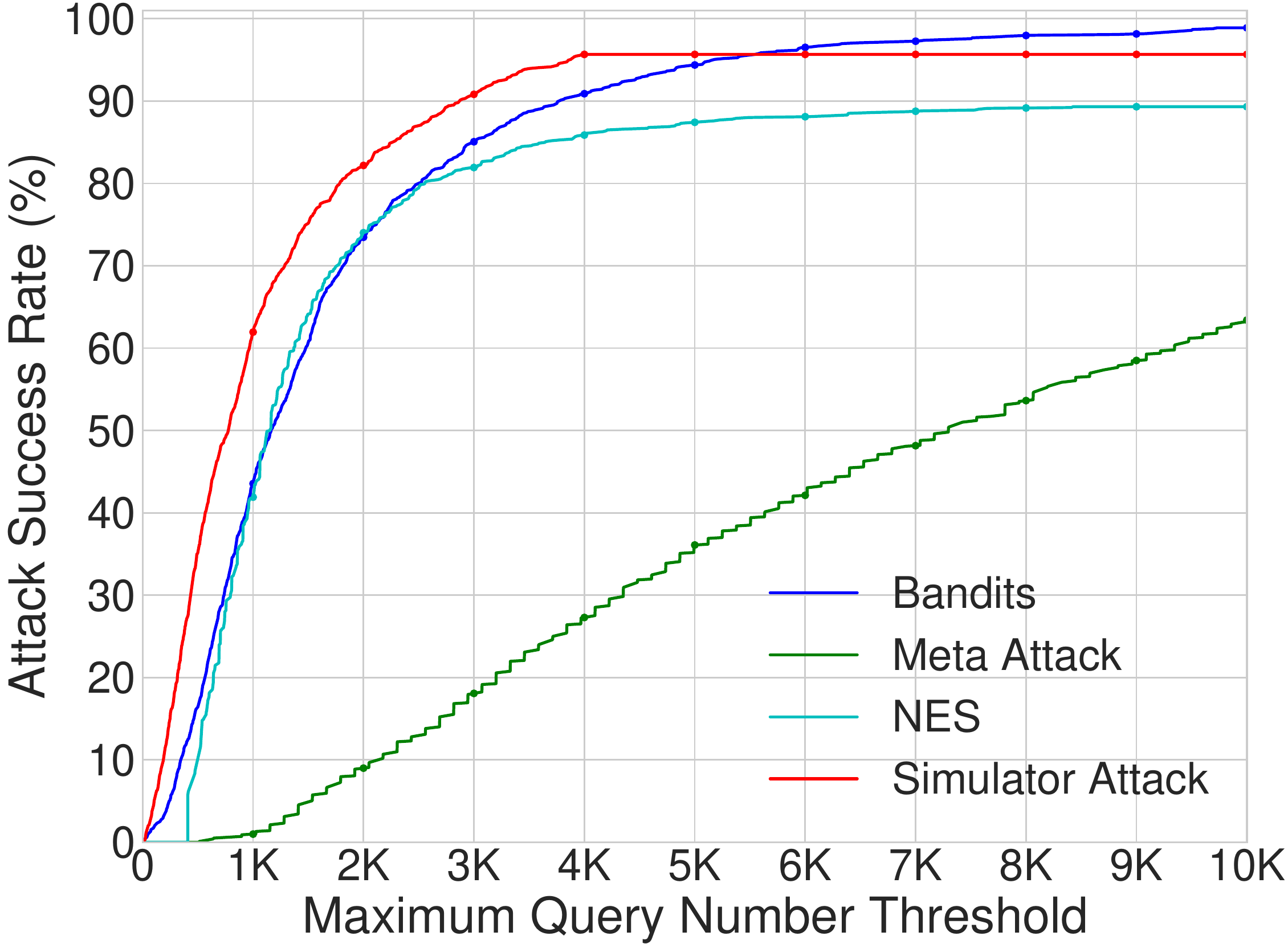}
		\subcaption{targeted $\ell_2$ attack WRN-28}
	\end{minipage}
	\begin{minipage}[b]{.245\textwidth}
		\includegraphics[width=\linewidth]{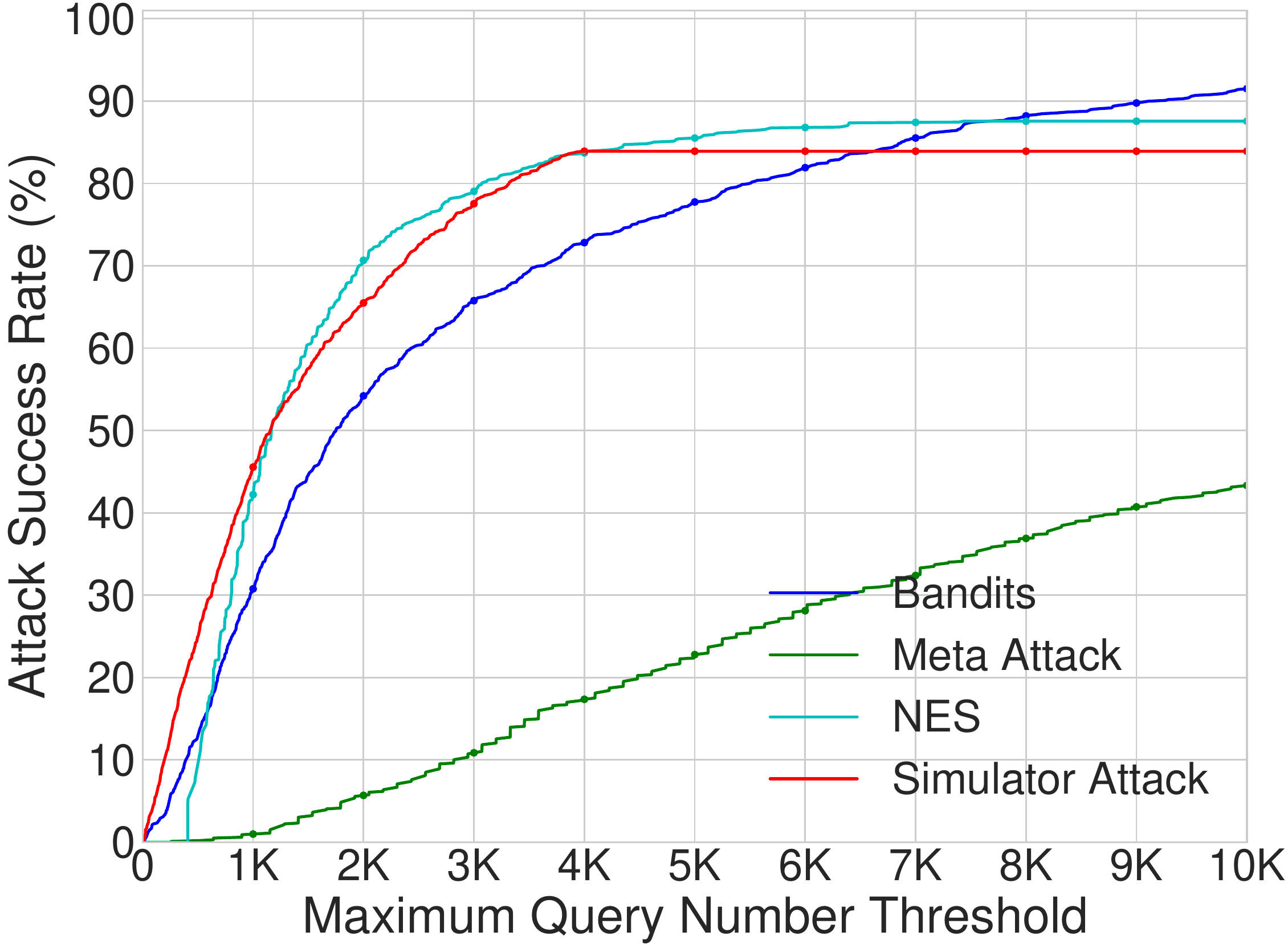}
		\subcaption{targeted $\ell_2$ attack WRN-40}
	\end{minipage}
	\begin{minipage}[b]{.245\textwidth}
		\includegraphics[width=\linewidth]{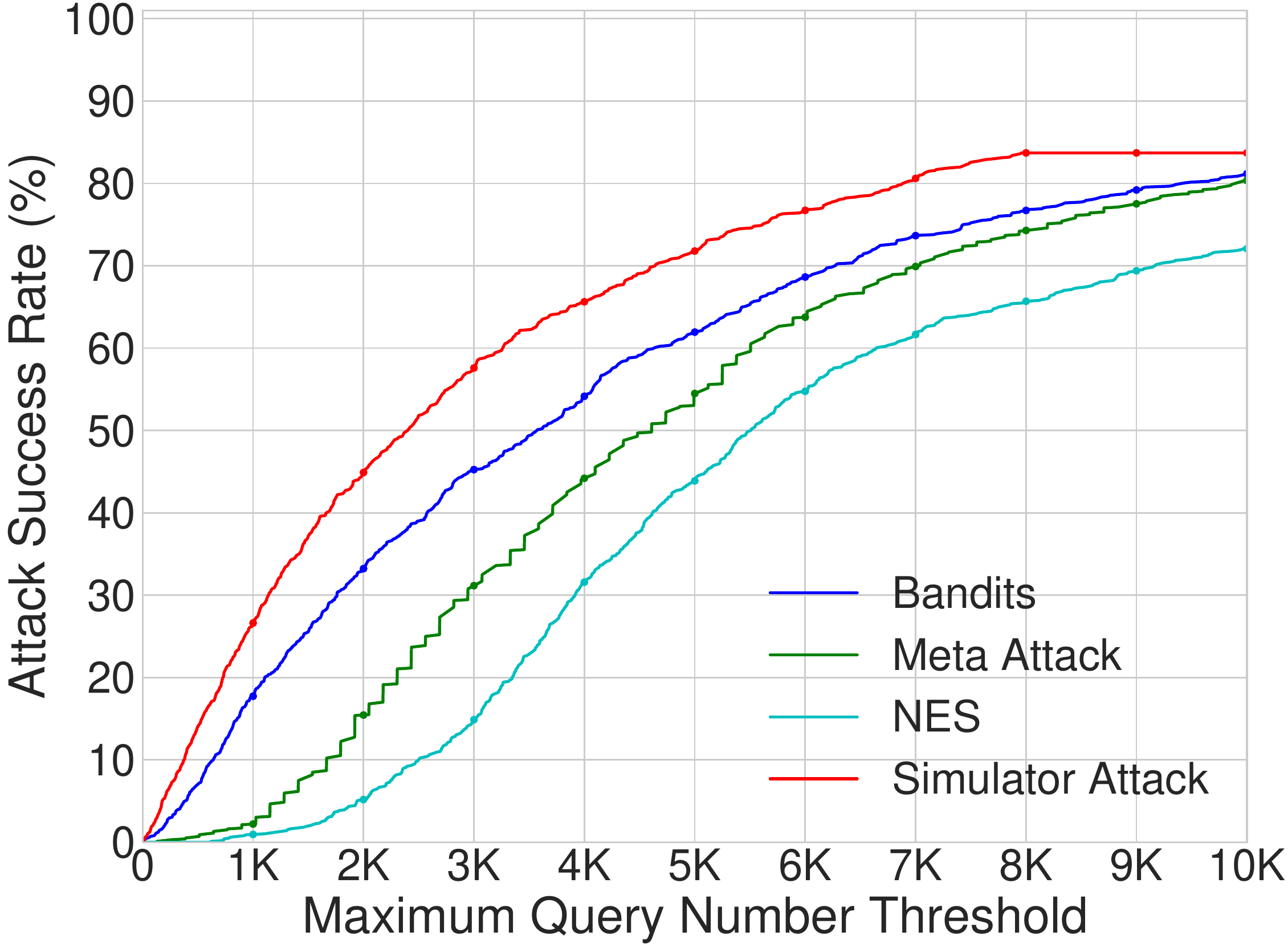}
		\subcaption{targeted $\ell_\infty$ attack PyramidNet-272}
	\end{minipage}
	\begin{minipage}[b]{.245\textwidth}
		\includegraphics[width=\linewidth]{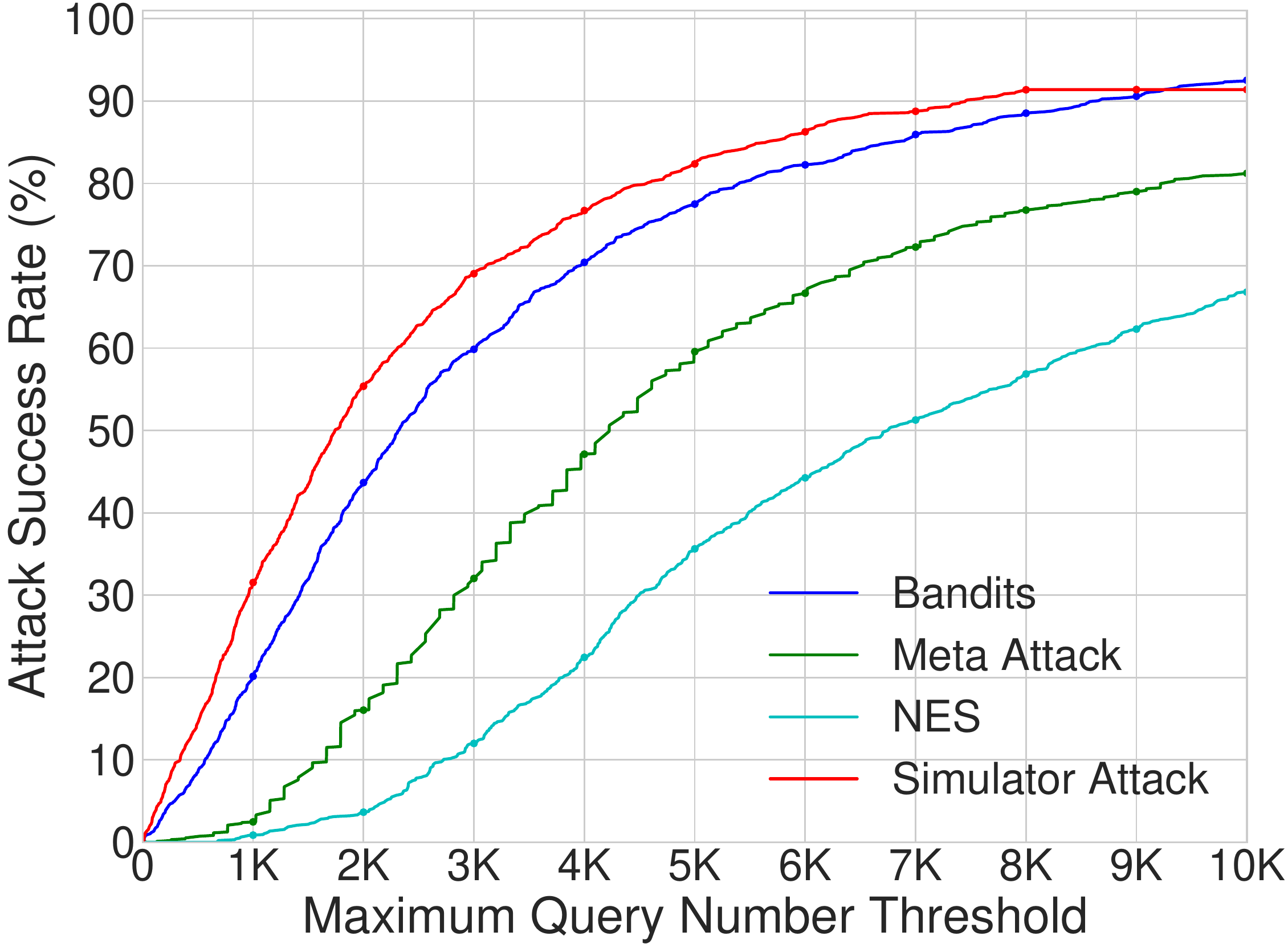}
		\subcaption{targeted $\ell_\infty$ attack GDAS}
	\end{minipage}
	\begin{minipage}[b]{.245\textwidth}
		\includegraphics[width=\linewidth]{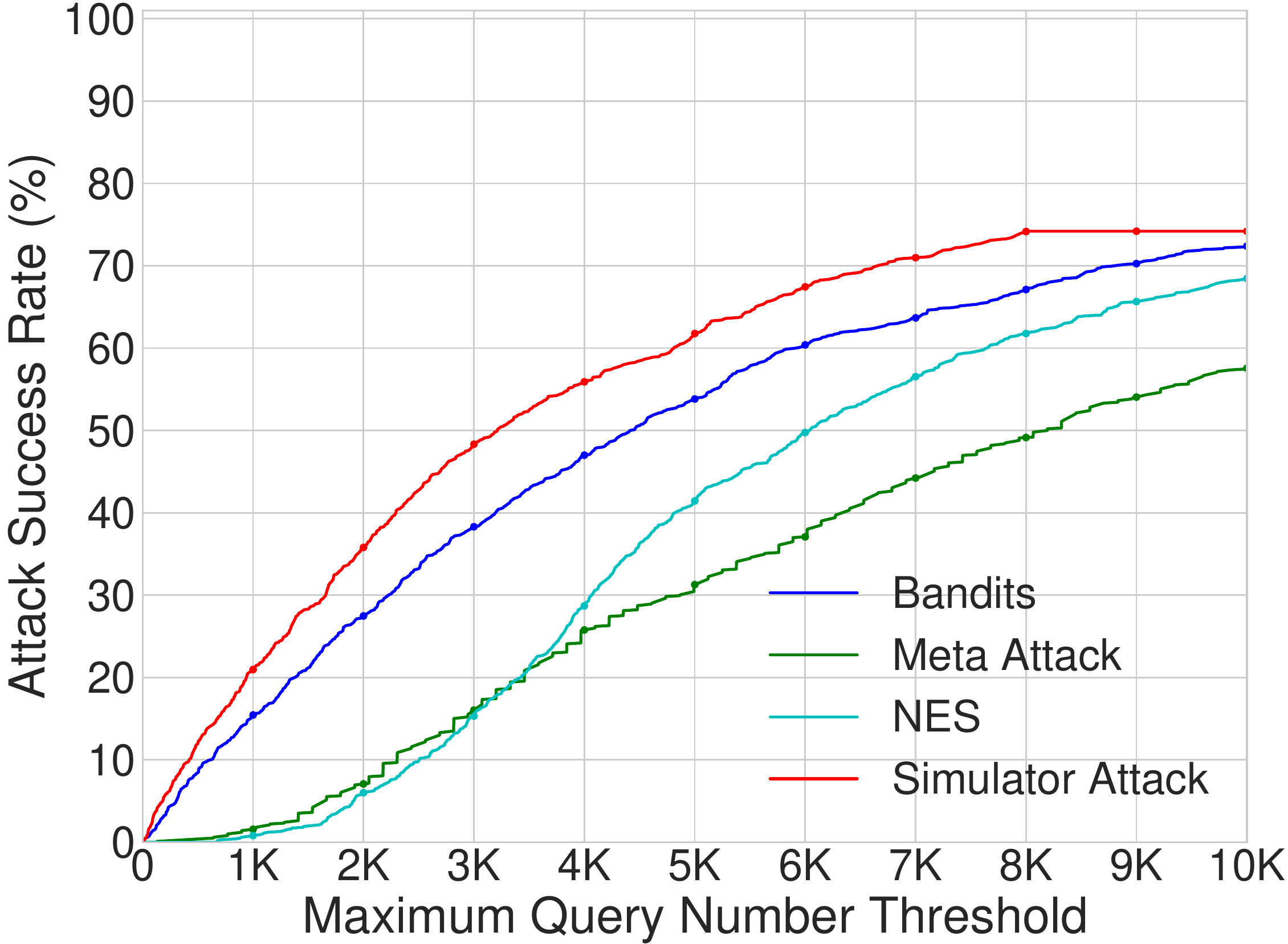}
		\subcaption{targeted $\ell_\infty$ attack WRN-28}
	\end{minipage}
	\begin{minipage}[b]{.245\textwidth}
		\includegraphics[width=\linewidth]{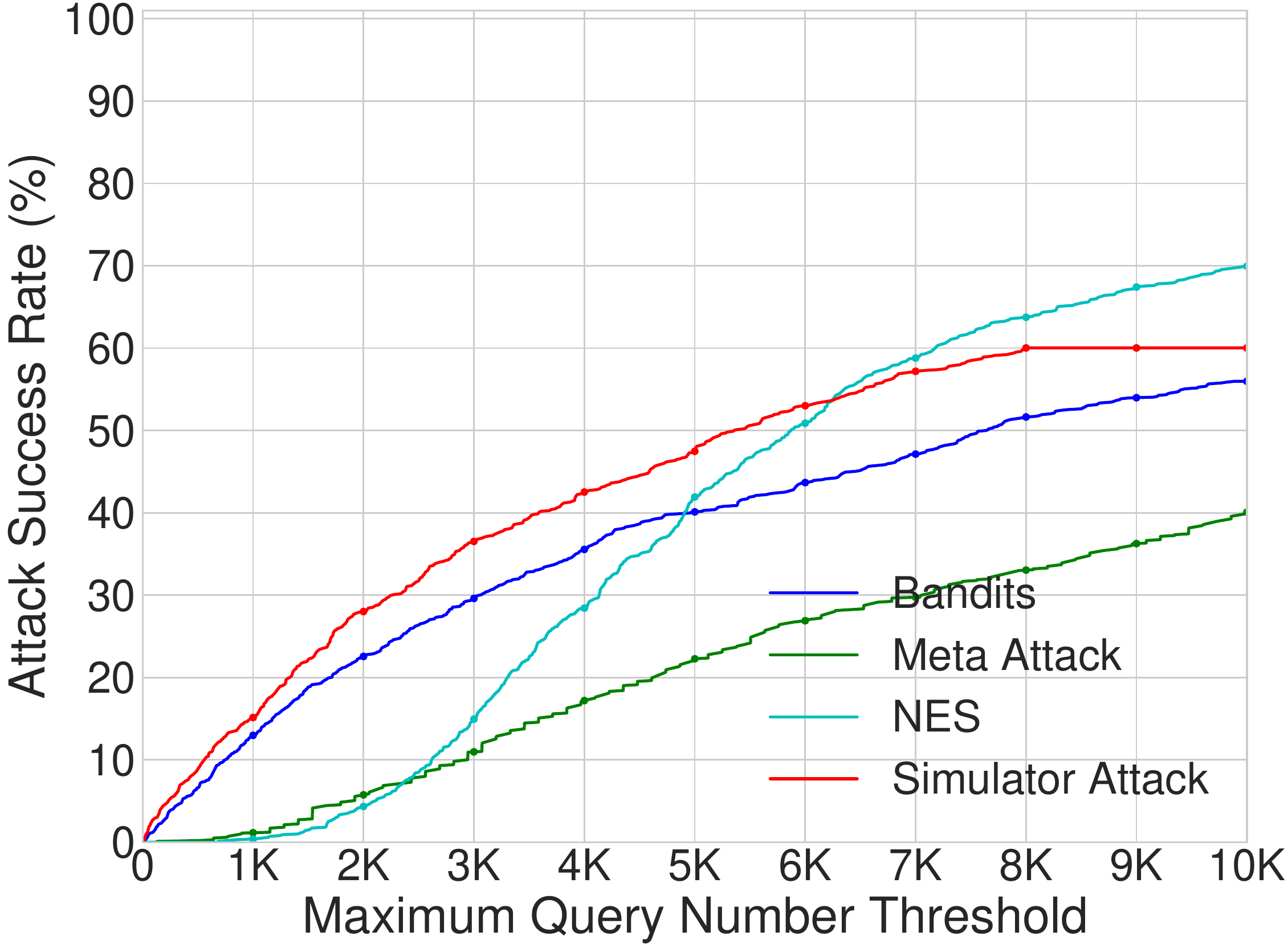}
		\subcaption{targeted $\ell_\infty$ attack WRN-40}
	\end{minipage}
	\caption{Comparisons of attack success rates at different limited maximum queries in CIFAR-100 dataset.}
	\label{fig:query_to_attack_success_rate_CIFAR-100}
\end{figure*}

\begin{figure*}[bp]
	\setlength{\abovecaptionskip}{0pt}
	\setlength{\belowcaptionskip}{0pt}
	\captionsetup[sub]{font={scriptsize}}
	\centering 
	\begin{minipage}[b]{.3\textwidth}
		\includegraphics[width=\linewidth]{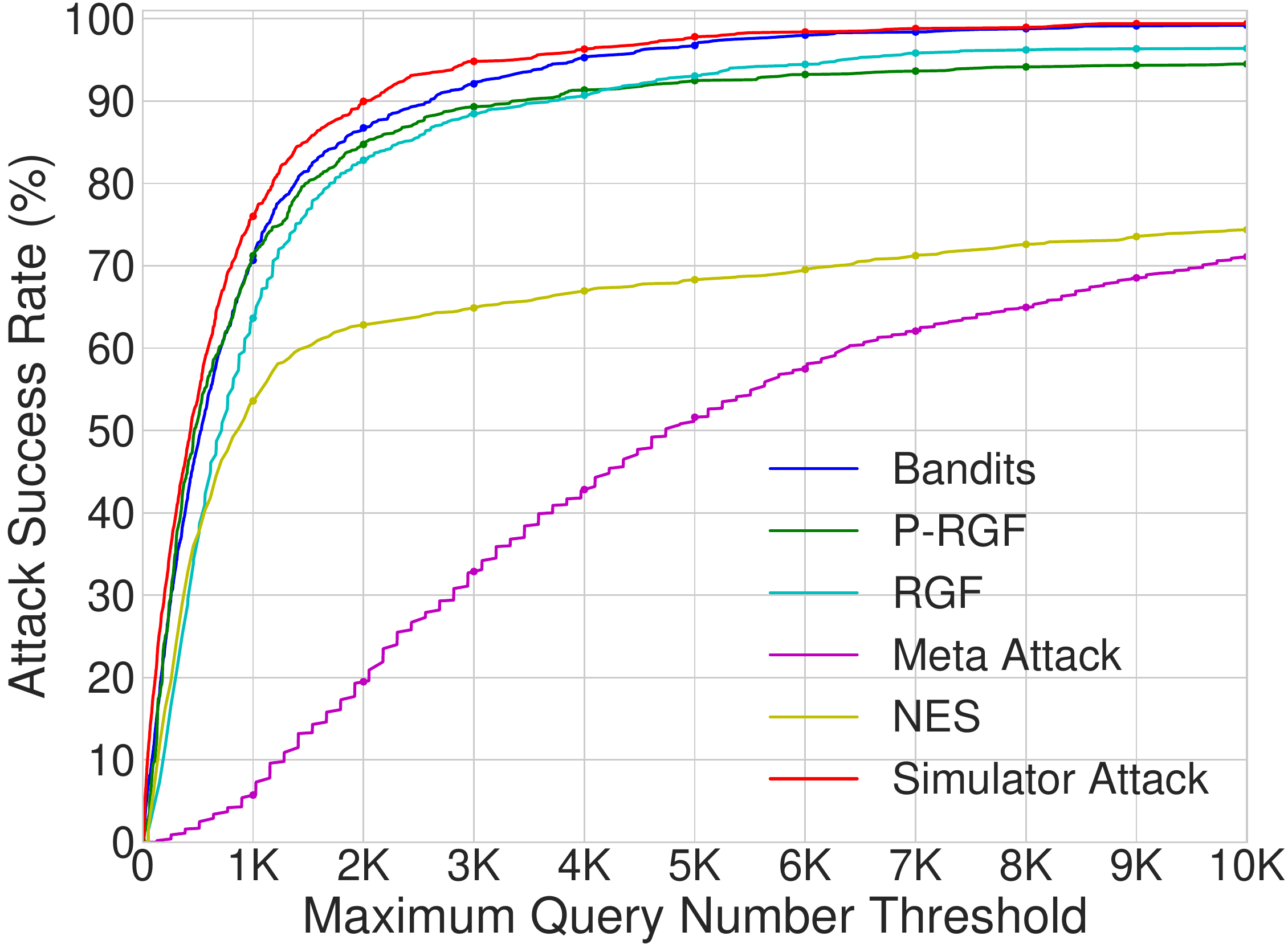}
		\subcaption{untargeted $\ell_\infty$ attack DenseNet-121}
	\end{minipage}
	\begin{minipage}[b]{.3\textwidth}
		\includegraphics[width=\linewidth]{figure/fig/query_threshold_attack_success_rate/TinyImageNet_resnext32_4_linf_untargeted_attack_query_threshold_success_rate_dict.pdf}
		\subcaption{untargeted $\ell_\infty$ attack ResNeXt-101(32$\times$4d)}
	\end{minipage}
	\begin{minipage}[b]{.3\textwidth}
		\includegraphics[width=\linewidth]{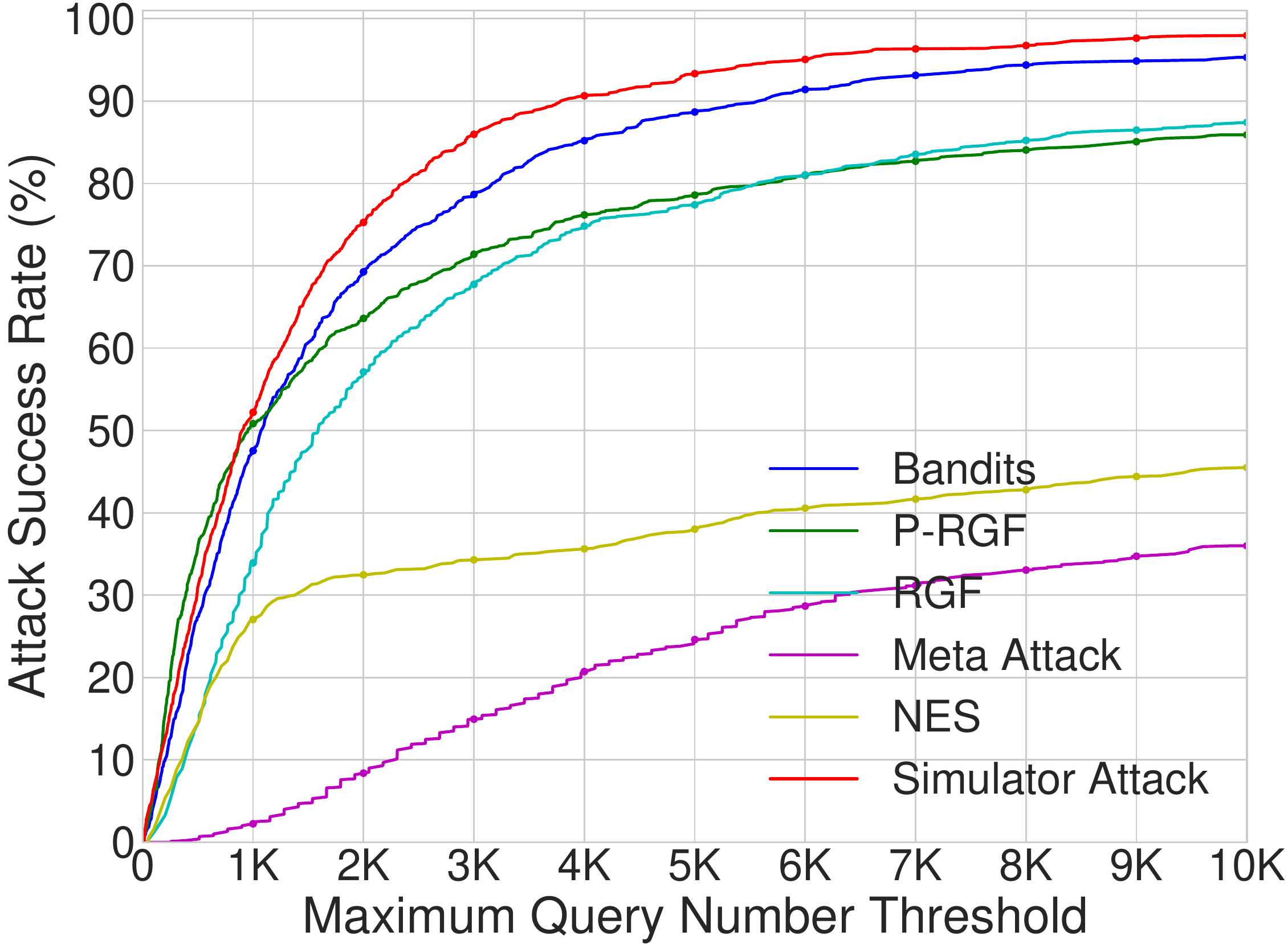}
		\subcaption{untargeted $\ell_\infty$ attack ResNeXt-101(64$\times$4d)}
	\end{minipage}
	\begin{minipage}[b]{.3\textwidth}
		\includegraphics[width=\linewidth]{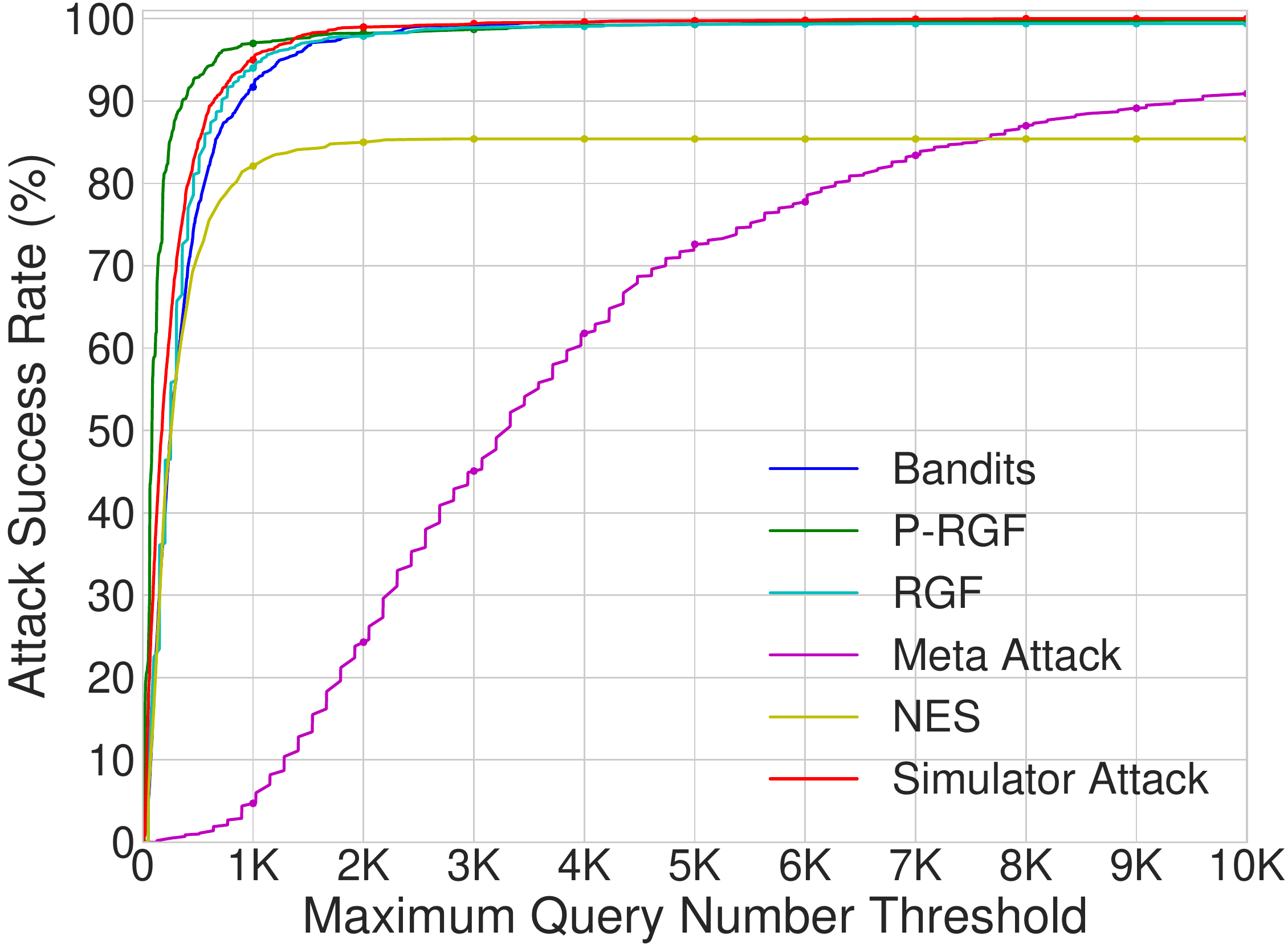}
		\subcaption{untargeted $\ell_2$ attack DenseNet-121}
	\end{minipage}
	\begin{minipage}[b]{.3\textwidth}
		\includegraphics[width=\linewidth]{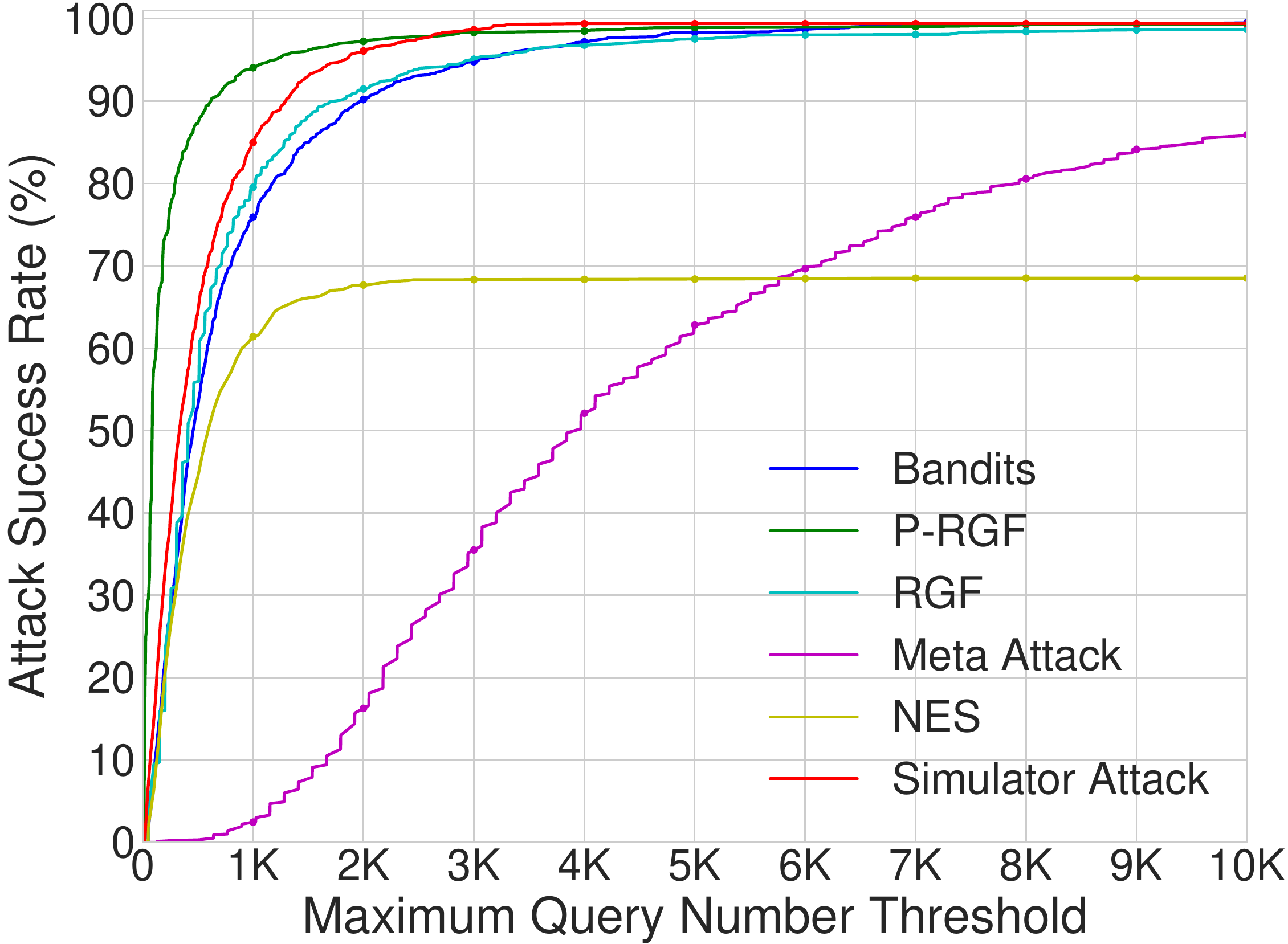}
		\subcaption{untargeted $\ell_2$ attack ResNeXt-101(32$\times$4d)}
	\end{minipage}
	\begin{minipage}[b]{.3\textwidth}
		\includegraphics[width=\linewidth]{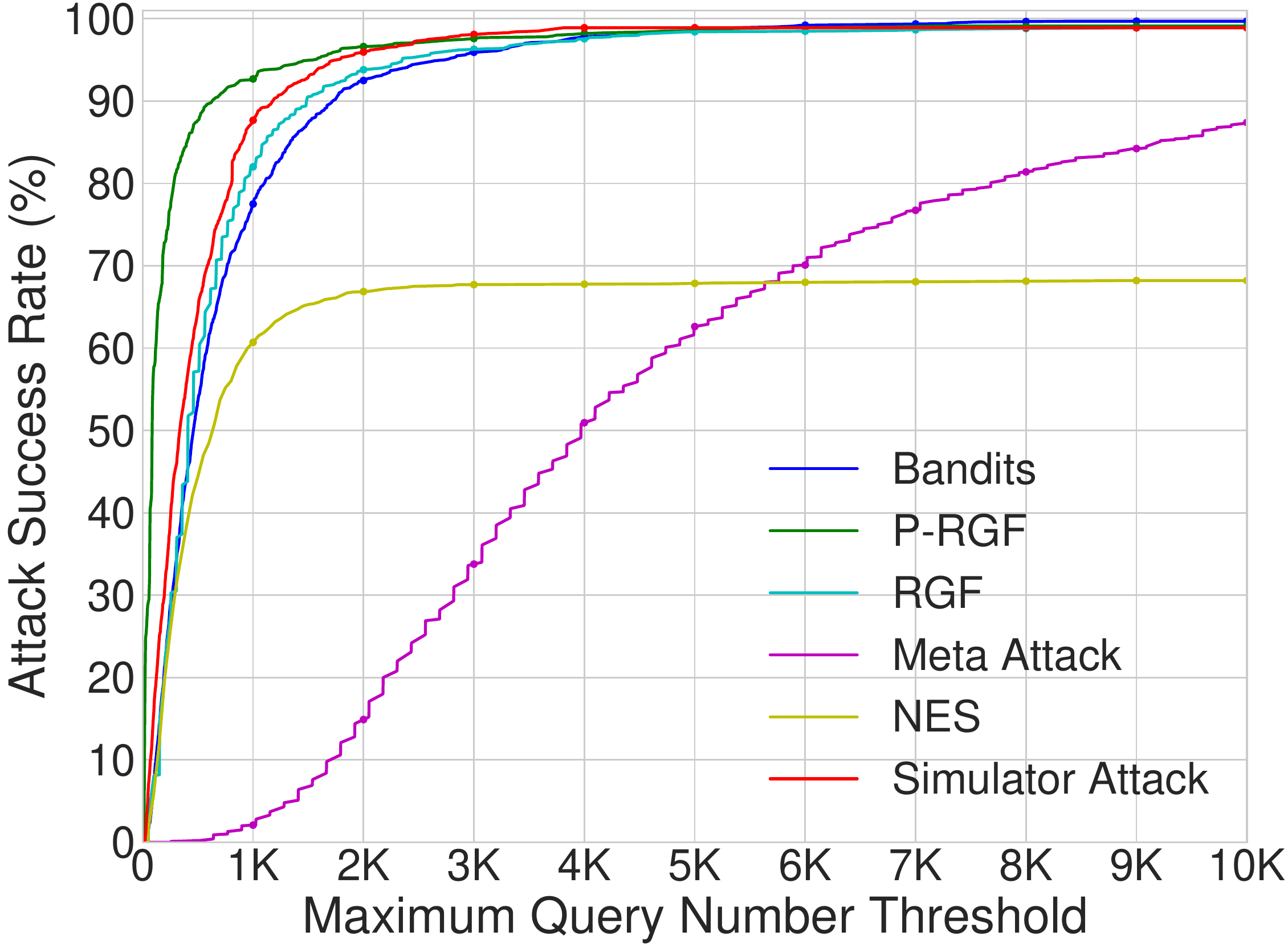}
		\subcaption{untargeted $\ell_2$ attack ResNeXt-101(64$\times$4d)}
	\end{minipage}
	\begin{minipage}[b]{.3\textwidth}
		\includegraphics[width=\linewidth]{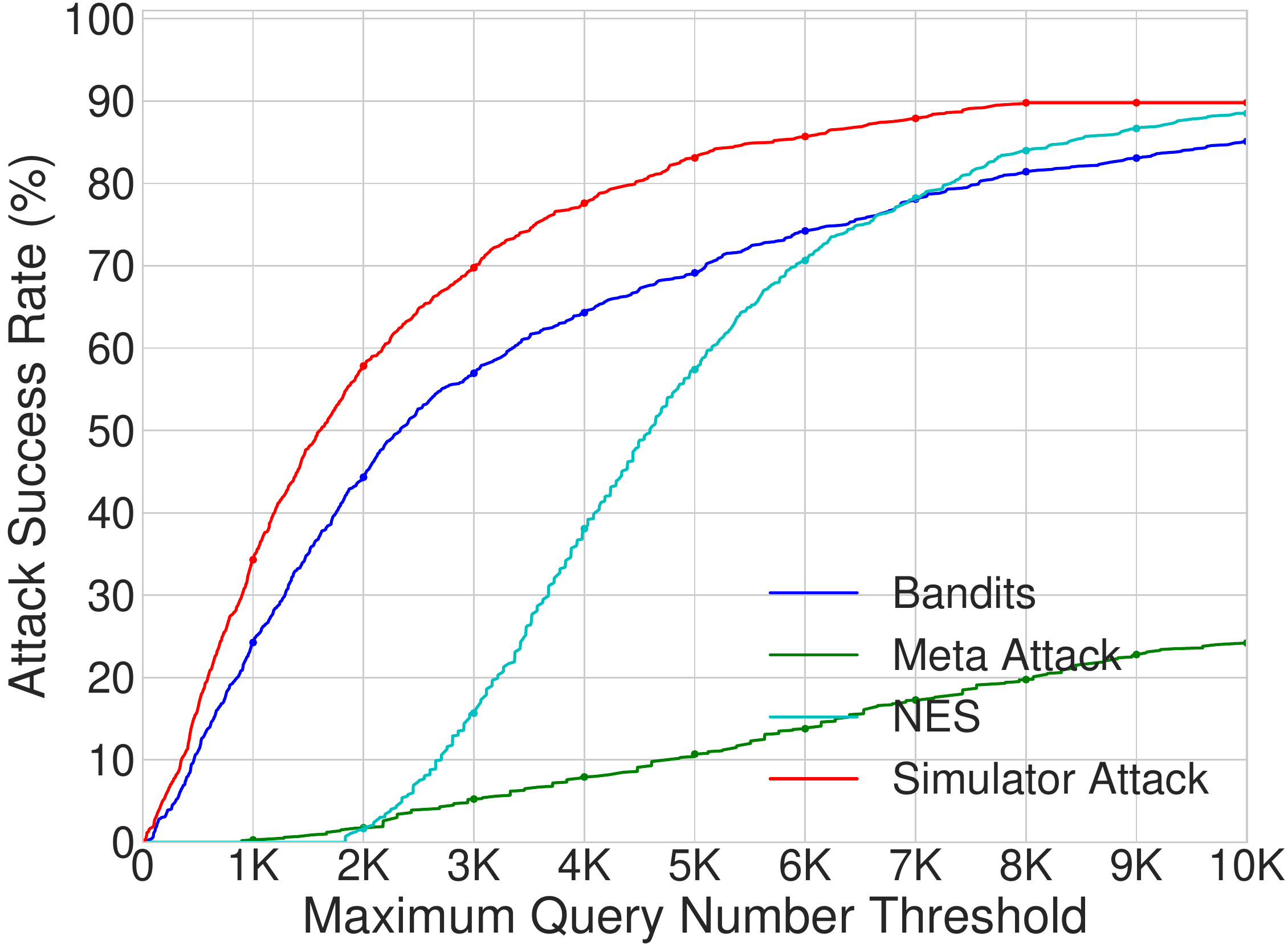}
		\subcaption{targeted $\ell_2$ attack DenseNet-121}
	\end{minipage}
	\begin{minipage}[b]{.3\textwidth}
		\includegraphics[width=\linewidth]{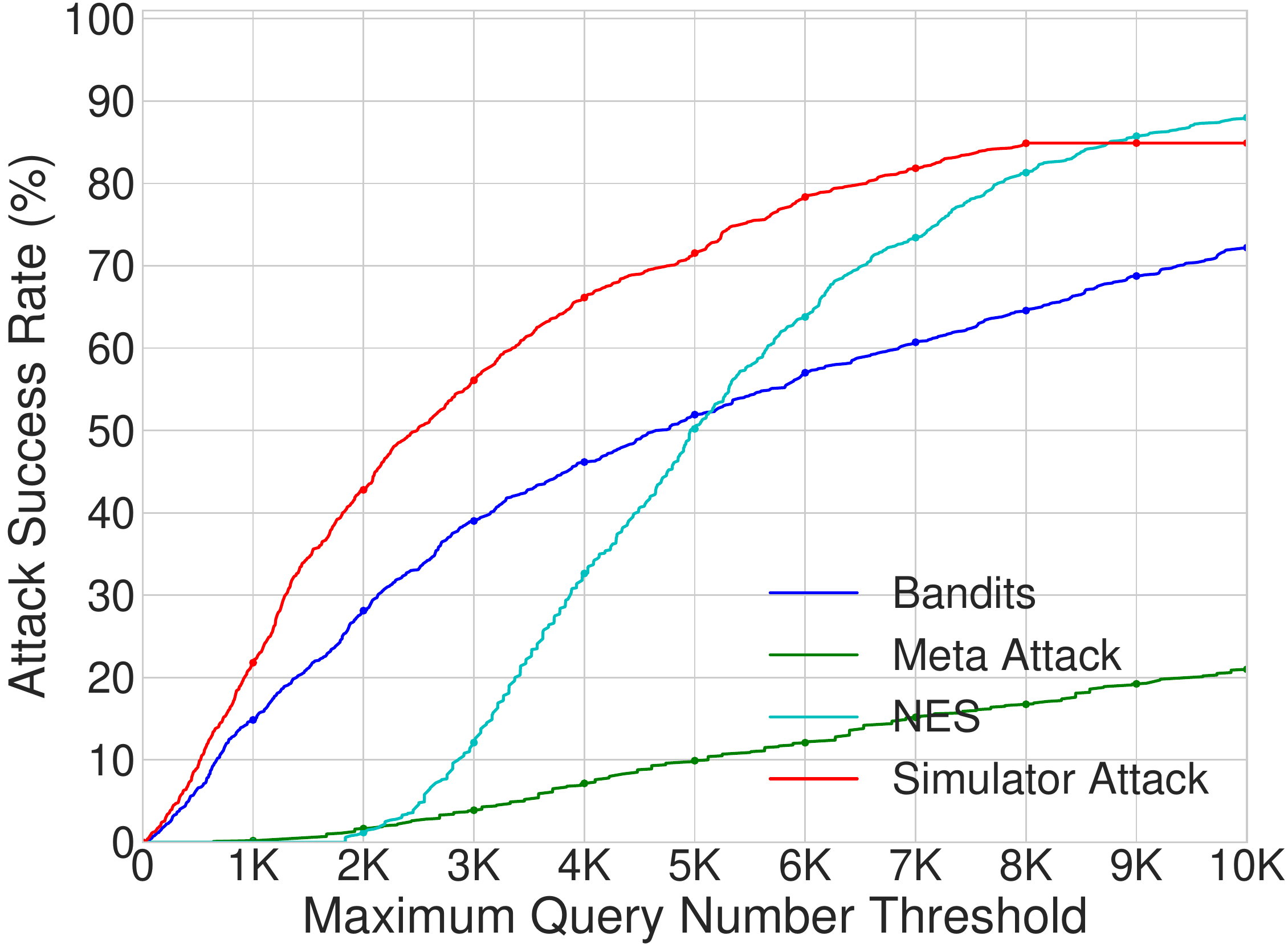}
		\subcaption{targeted $\ell_2$ attack ResNeXt-101(32$\times$4d)}
	\end{minipage}
	\begin{minipage}[b]{.3\textwidth}
		\includegraphics[width=\linewidth]{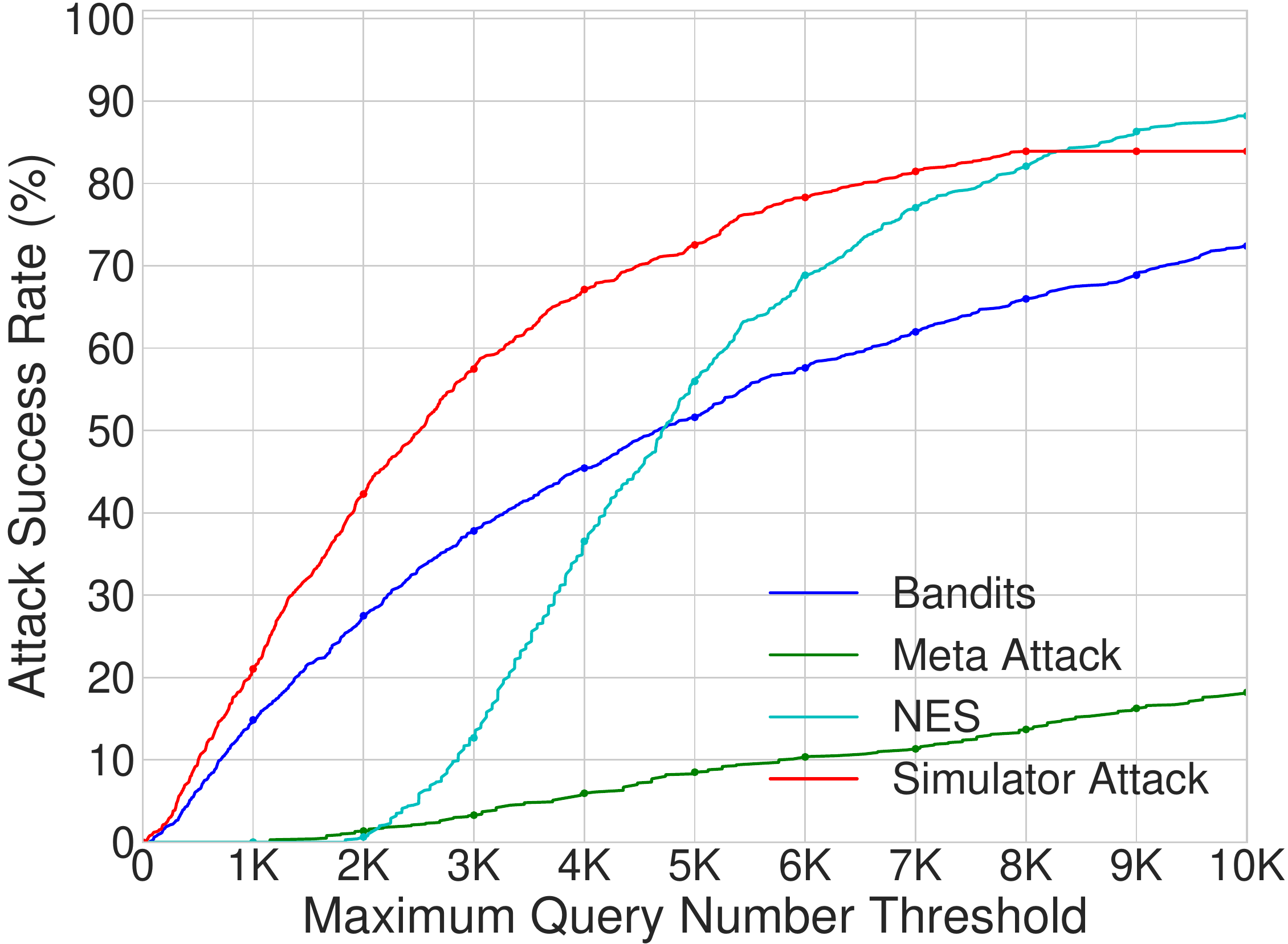}
		\subcaption{targeted $\ell_2$ attack ResNeXt-101(64$\times$4d)}
	\end{minipage}
	\caption{Comparisons of attack success rates at different limited maximum queries in TinyImageNet dataset.}
	\label{fig:query_to_attack_success_rate_TinyImageNet}
\end{figure*}

\begin{figure*}[htbp]
	\setlength{\abovecaptionskip}{0pt}
	\setlength{\belowcaptionskip}{0pt}
	\captionsetup[sub]{font={scriptsize}}
	\centering 
	\begin{minipage}[b]{.3\textwidth}
		\includegraphics[width=\linewidth]{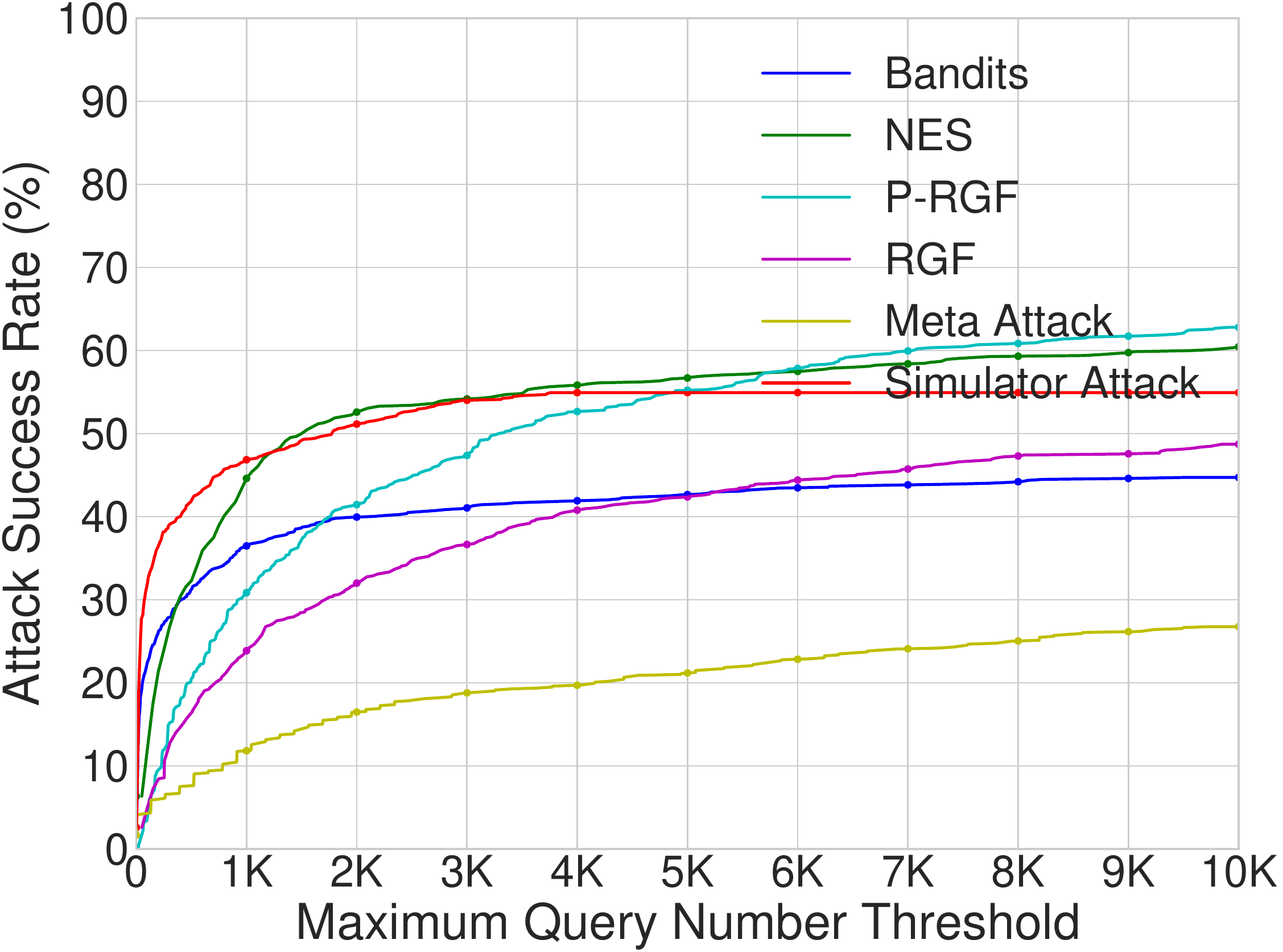}
		\subcaption{attack ComDefend in CIFAR-10}
	\end{minipage}
	\begin{minipage}[b]{.3\textwidth}
		\includegraphics[width=\linewidth]{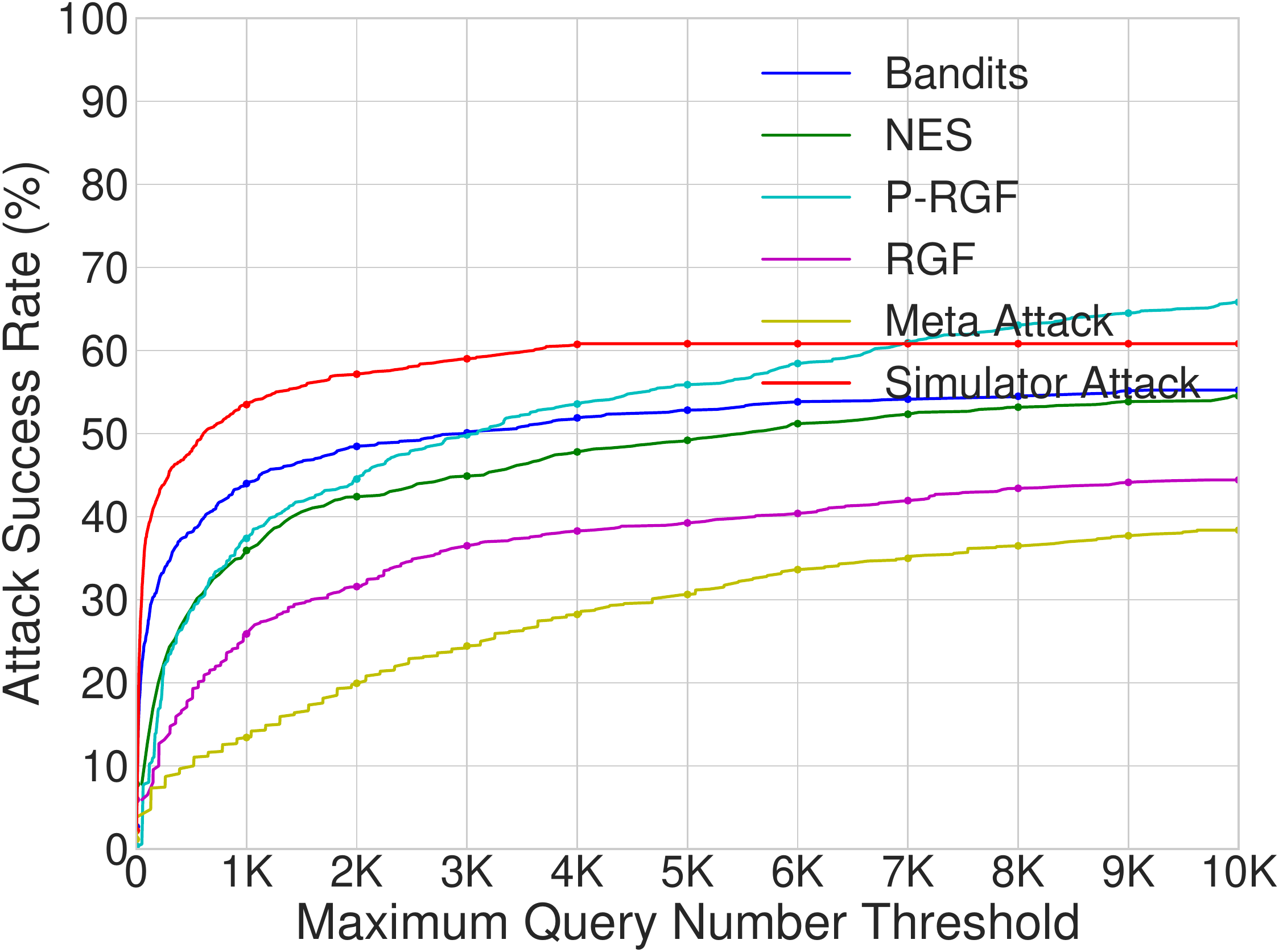}
		\subcaption{attack Feature Distillation in CIFAR-10}
	\end{minipage}
	\begin{minipage}[b]{.3\textwidth}
		\includegraphics[width=\linewidth]{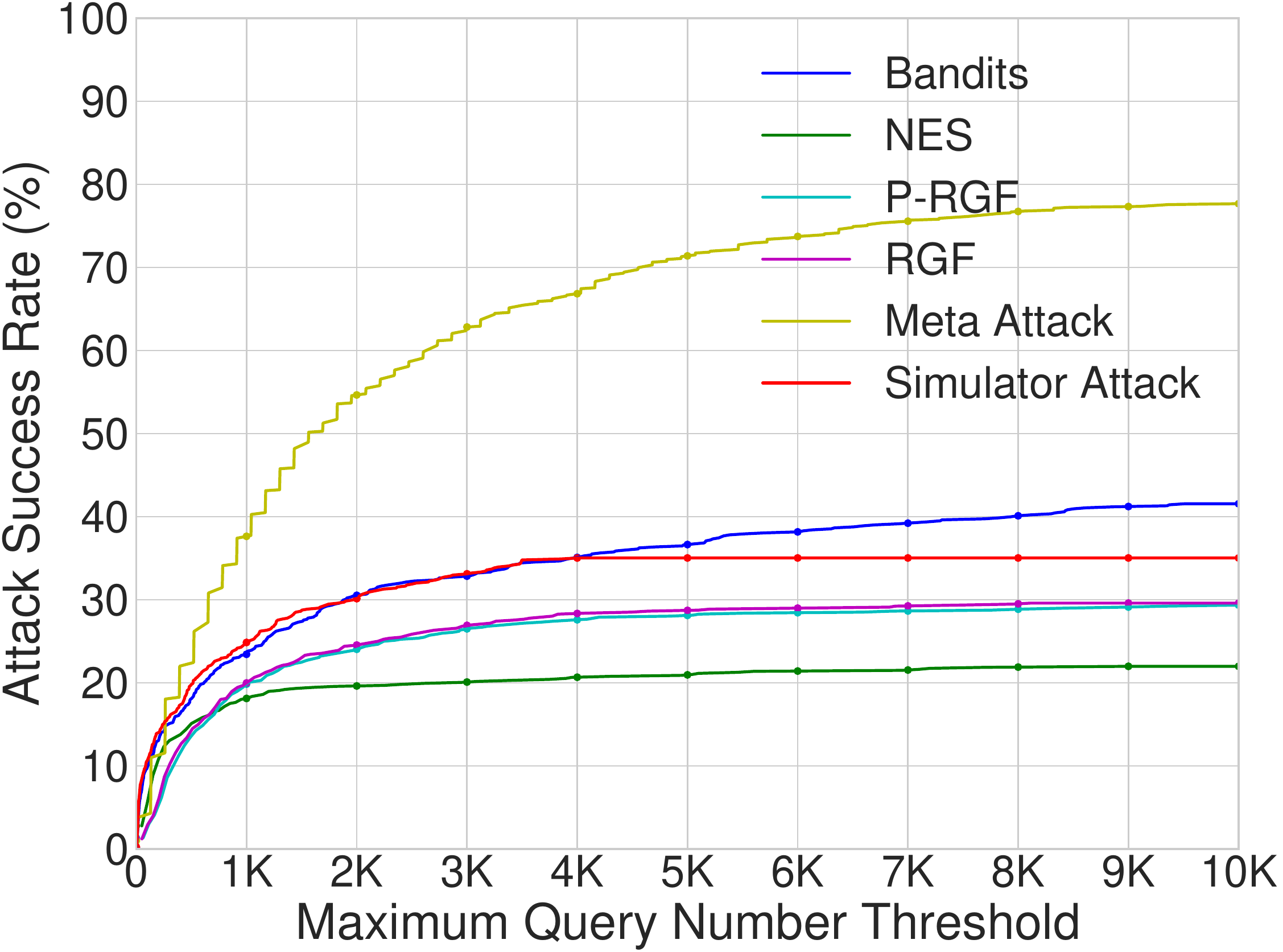}
		\subcaption{attack PCL in CIFAR-10}
	\end{minipage}
	\begin{minipage}[b]{.3\textwidth}
		\includegraphics[width=\linewidth]{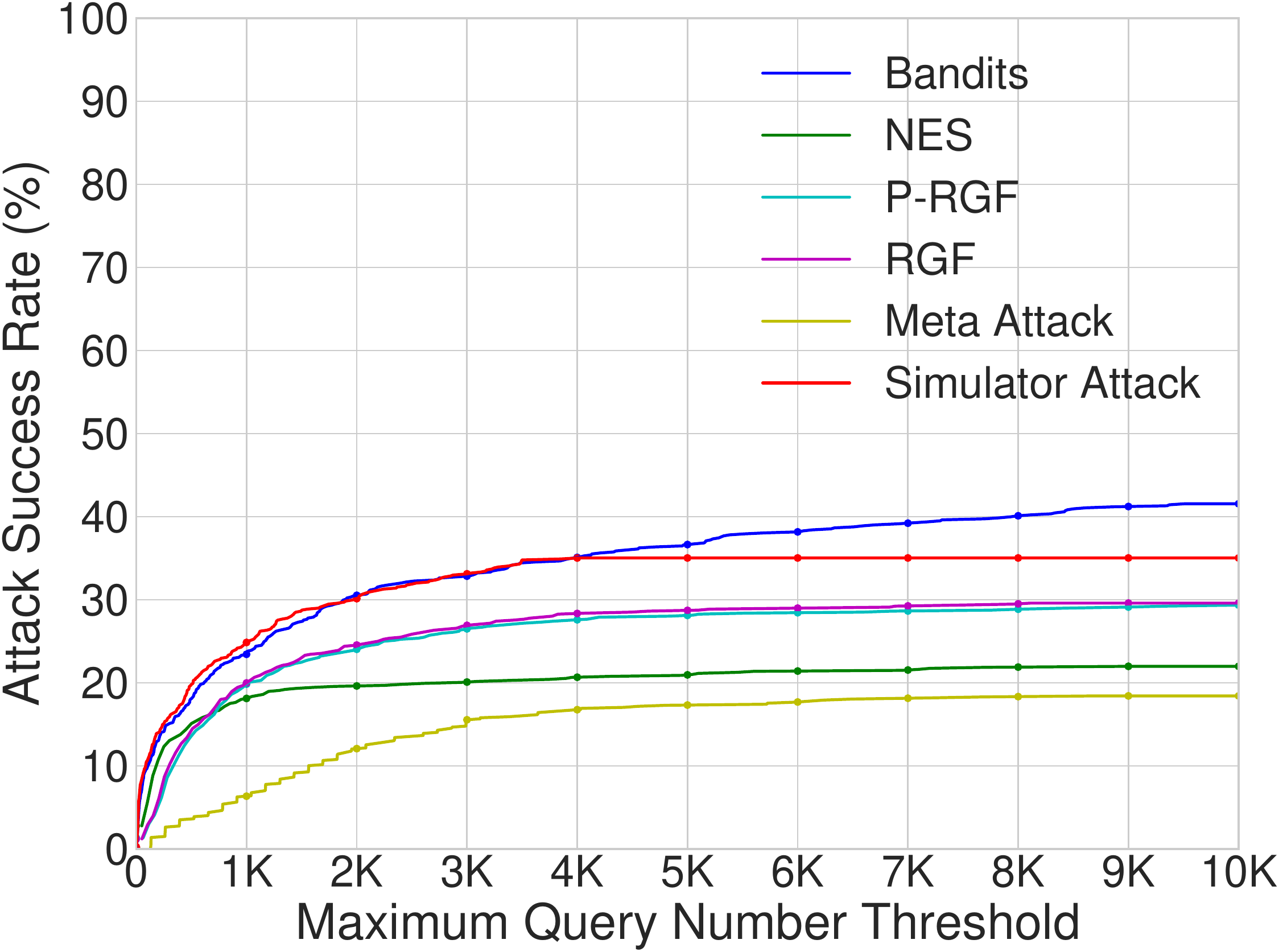}
		\subcaption{attack Adv Train in CIFAR-10}
	\end{minipage}
	\begin{minipage}[b]{.3\textwidth}
		\includegraphics[width=\linewidth]{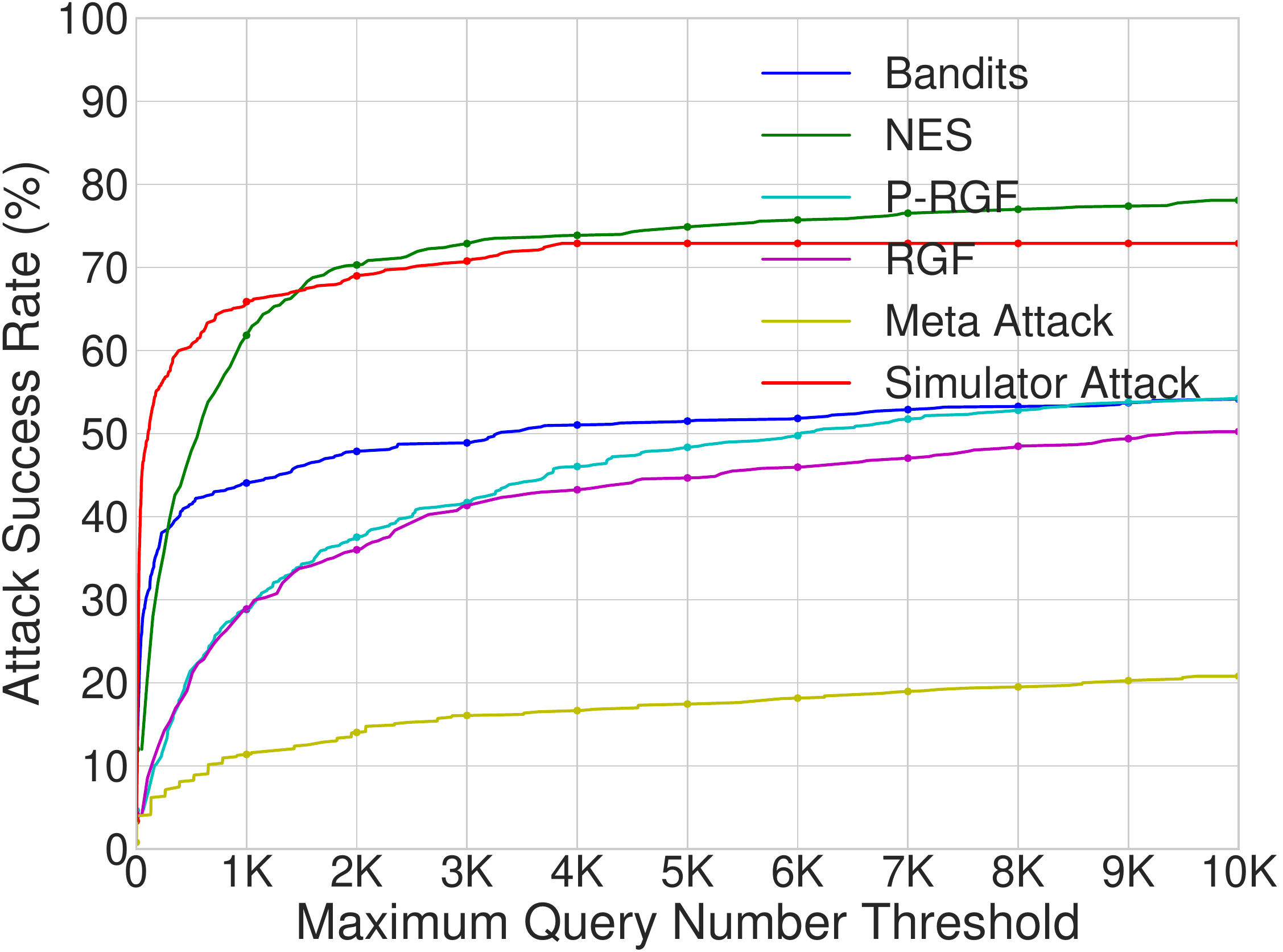}
		\subcaption{attack ComDefend in CIFAR-100}
	\end{minipage}
	\begin{minipage}[b]{.3\textwidth}
		\includegraphics[width=\linewidth]{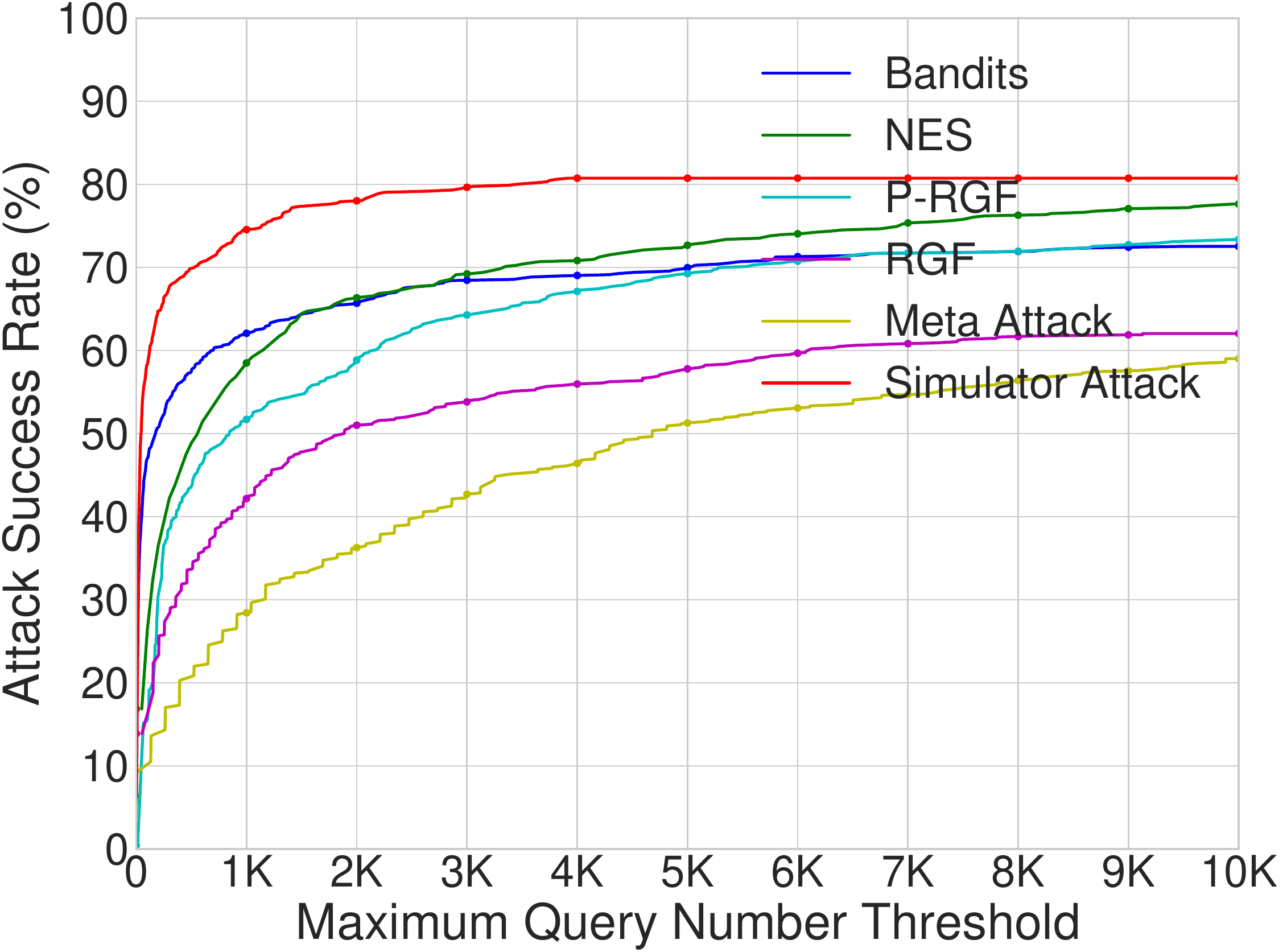}
		\subcaption{attack Feature Distillation in CIFAR-100}
	\end{minipage}
	\begin{minipage}[b]{.3\textwidth}
		\includegraphics[width=\linewidth]{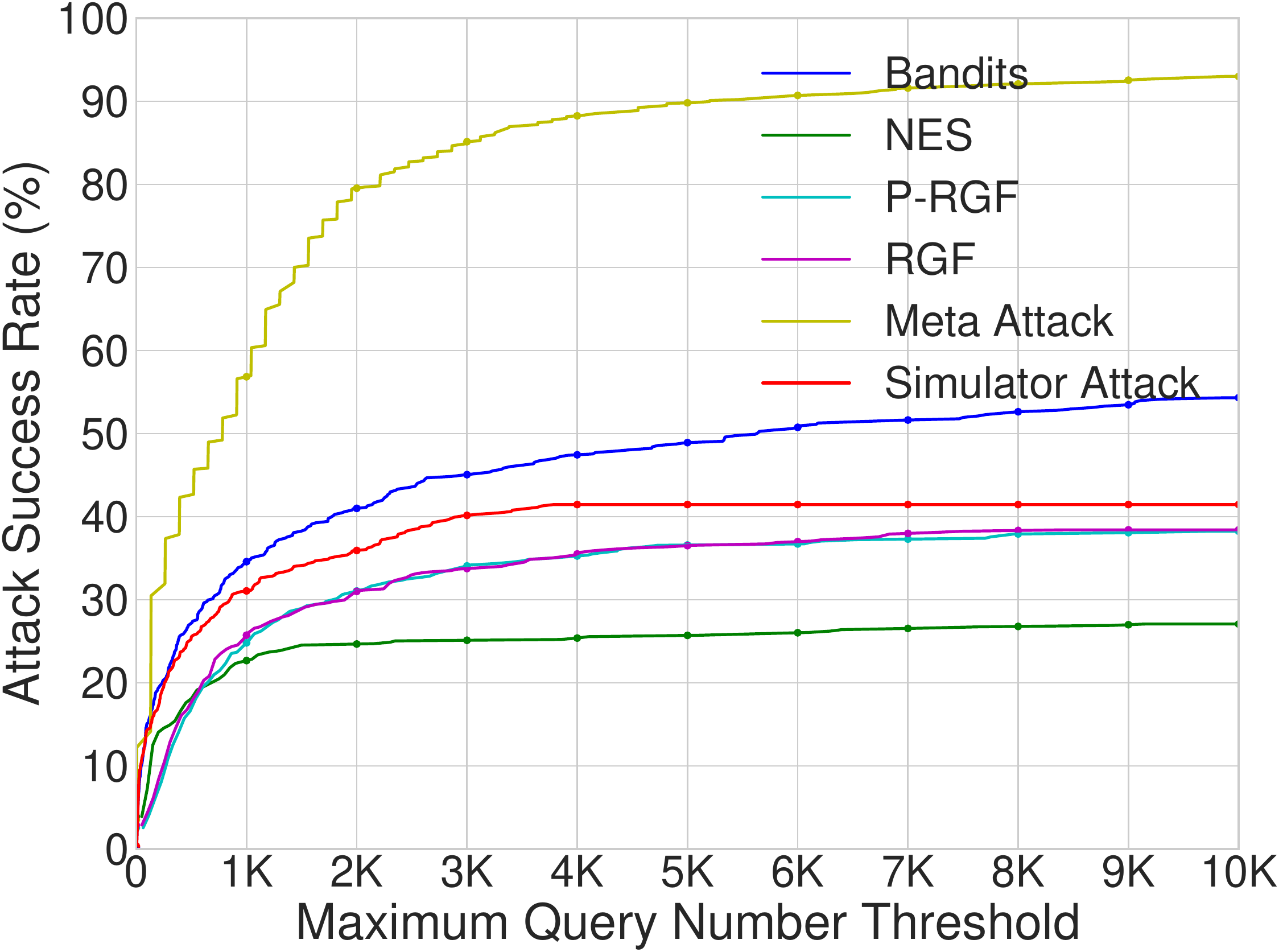}
		\subcaption{attack PCL in CIFAR-100}
	\end{minipage}
	\begin{minipage}[b]{.3\textwidth}
		\includegraphics[width=\linewidth]{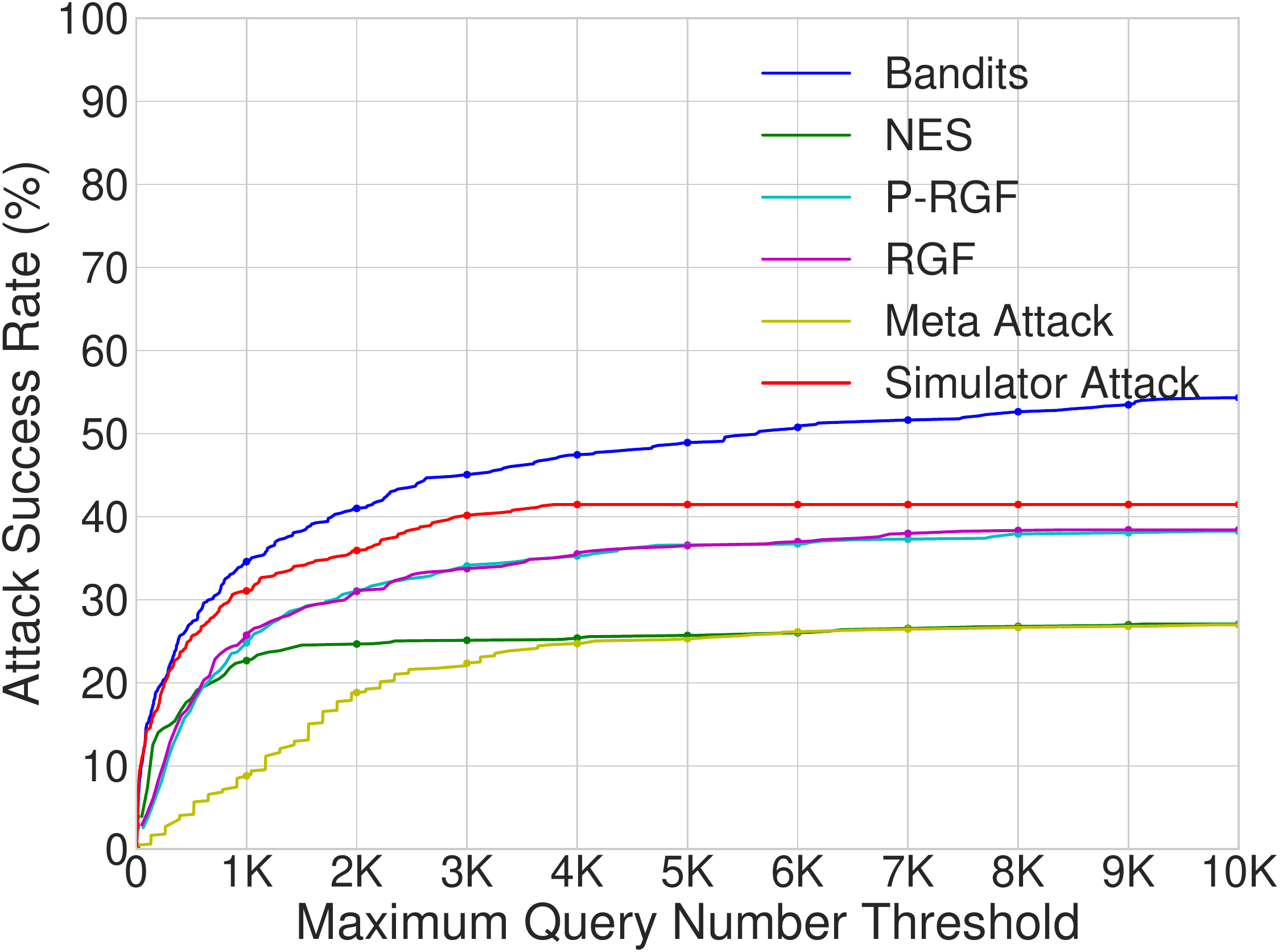}
		\subcaption{attack Adv Train in CIFAR-100}
	\end{minipage}
	\begin{minipage}[b]{.3\textwidth}
		\includegraphics[width=\linewidth]{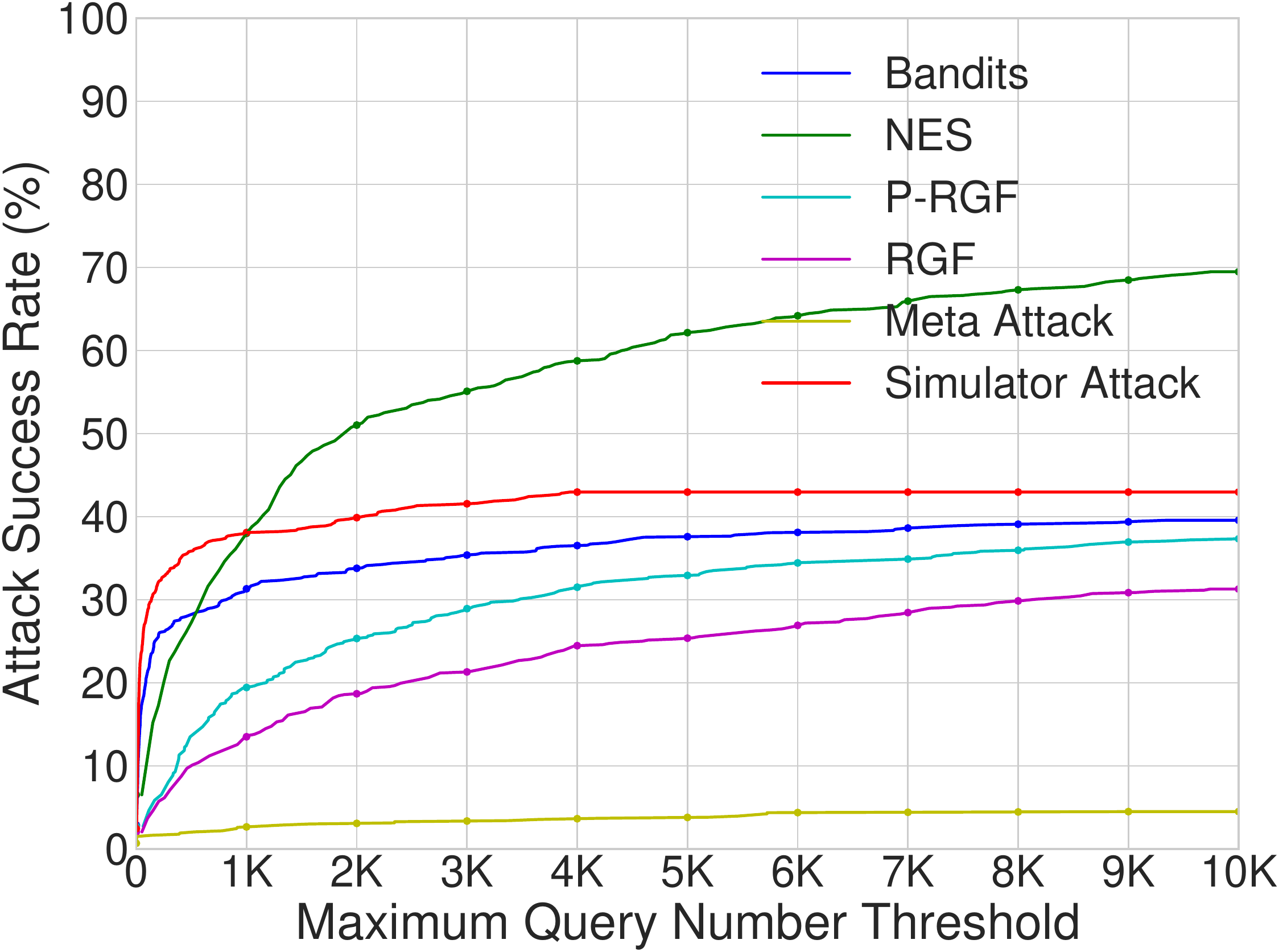}
		\subcaption{attack ComDefend in TinyImageNet}
	\end{minipage}
	\begin{minipage}[b]{.3\textwidth}
		\includegraphics[width=\linewidth]{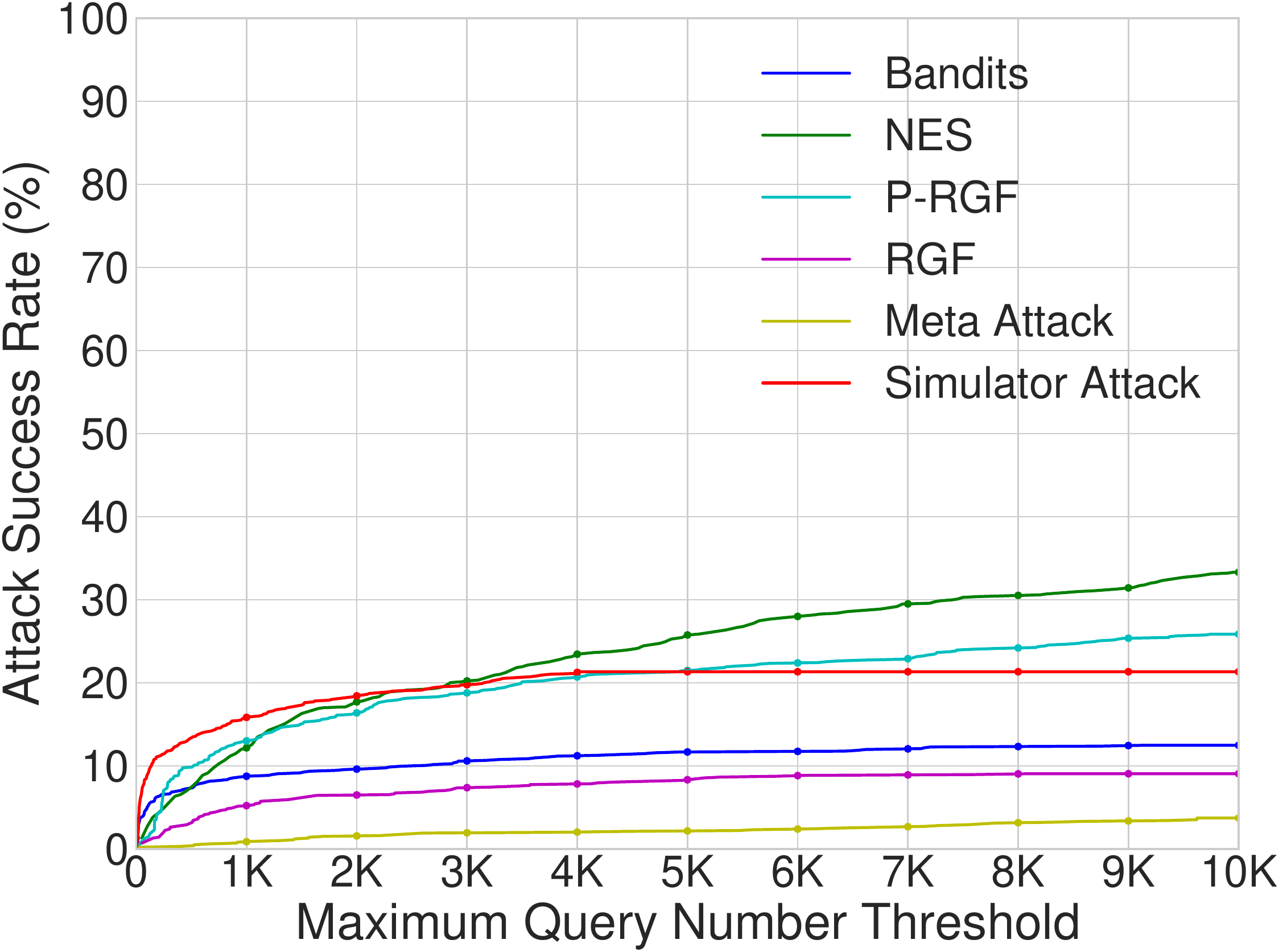}
		\subcaption{attack Feature Distillation in TinyImageNet}
	\end{minipage}
	\begin{minipage}[b]{.3\textwidth}
		\includegraphics[width=\linewidth]{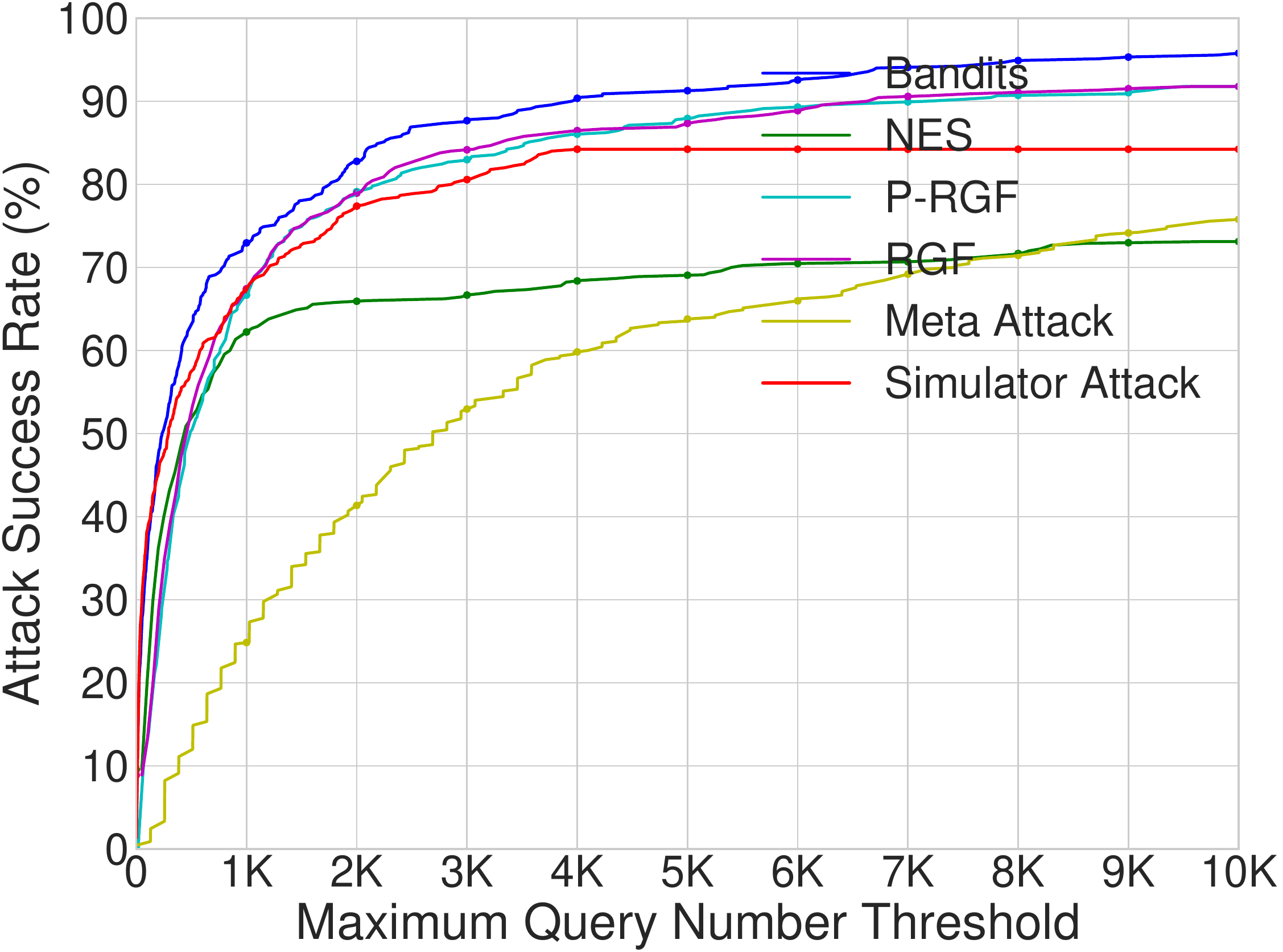}
		\subcaption{attack PCL in TinyImageNet}
	\end{minipage}
	\begin{minipage}[b]{.3\textwidth}
		\includegraphics[width=\linewidth]{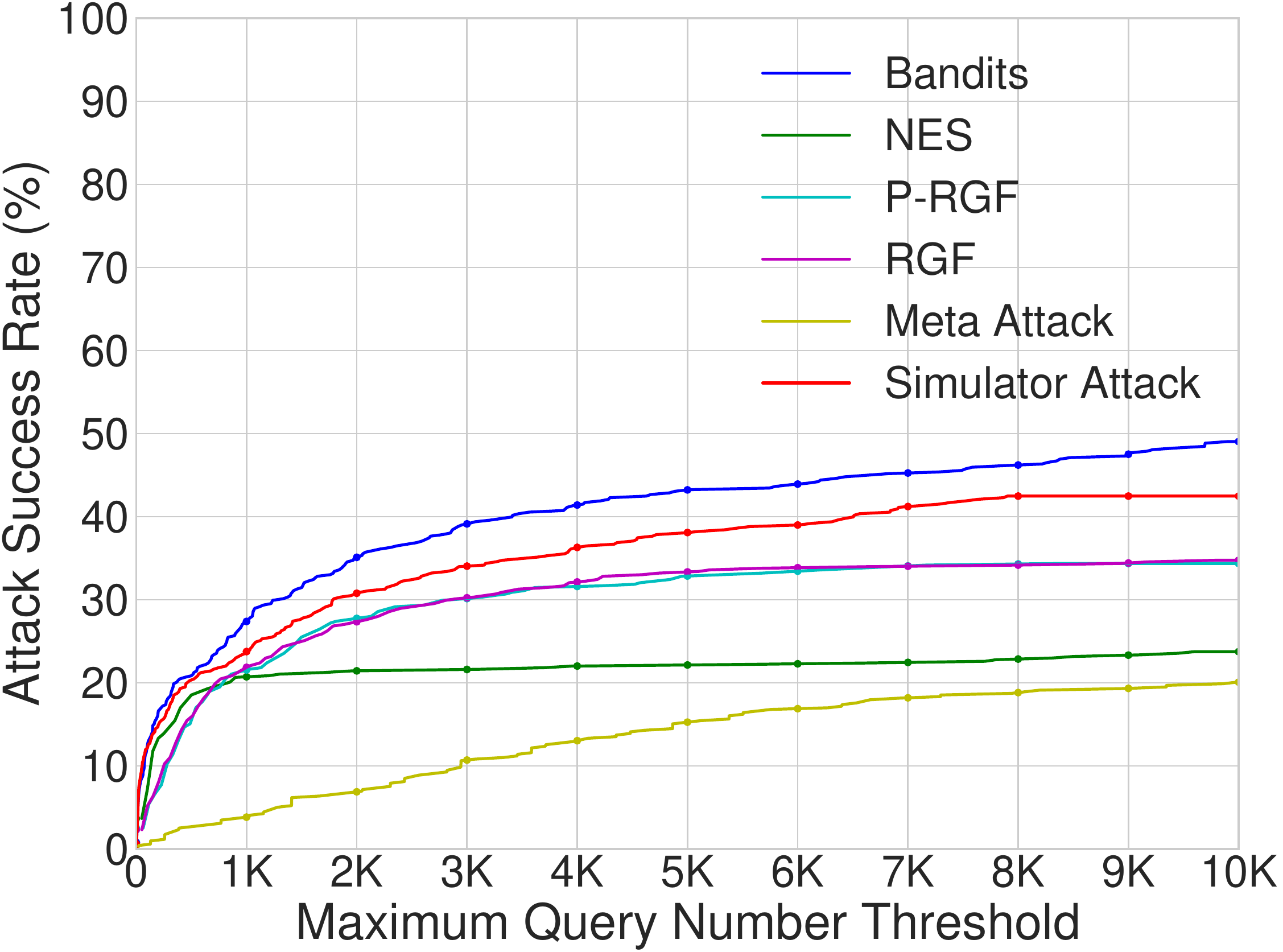}
		\subcaption{attack Adv Train in TinyImageNet}
	\end{minipage}
	
	\caption{Comparisons of attack success rates at different maximum queries on defensive models with the ResNet-50 backbone. The experimental results are obtained by performing the untargeted attacks under $\ell_\infty$ norm.}
	\label{fig:query_to_attack_success_rate_on_defensive_models}
\end{figure*}

\noindent\textbf{Attack Success Rates at Different Maximum Queries.}  We conduct experiments by limiting different maximum queries of attacks and compare their attack success rates. Figs. \ref{fig:query_to_attack_success_rate_CIFAR-10}, \ref{fig:query_to_attack_success_rate_CIFAR-100}, \ref{fig:query_to_attack_success_rate_TinyImageNet}, and \ref{fig:query_to_attack_success_rate_on_defensive_models} show the results which are obtained by attacking normal models and defensive models with different maximum number of queries. Four defensive models are adopted, namely, ComDefend \cite{jia2019comdefend}, Feature Distillation \cite{liu2019feature}, prototype conformity loss (PCL) \cite{mustafa2019adversarial} and Adv Train~\cite{madry2018towards}. The ResNet-50 \cite{he2016deep} is selected as the backbone in these defensive models.

\begin{figure*}[bp]
	\setlength{\abovecaptionskip}{0pt}
	\setlength{\belowcaptionskip}{0pt}
	\captionsetup[sub]{font={scriptsize}}
	\centering 
	\begin{minipage}[b]{.245\textwidth}
		\includegraphics[width=\linewidth]{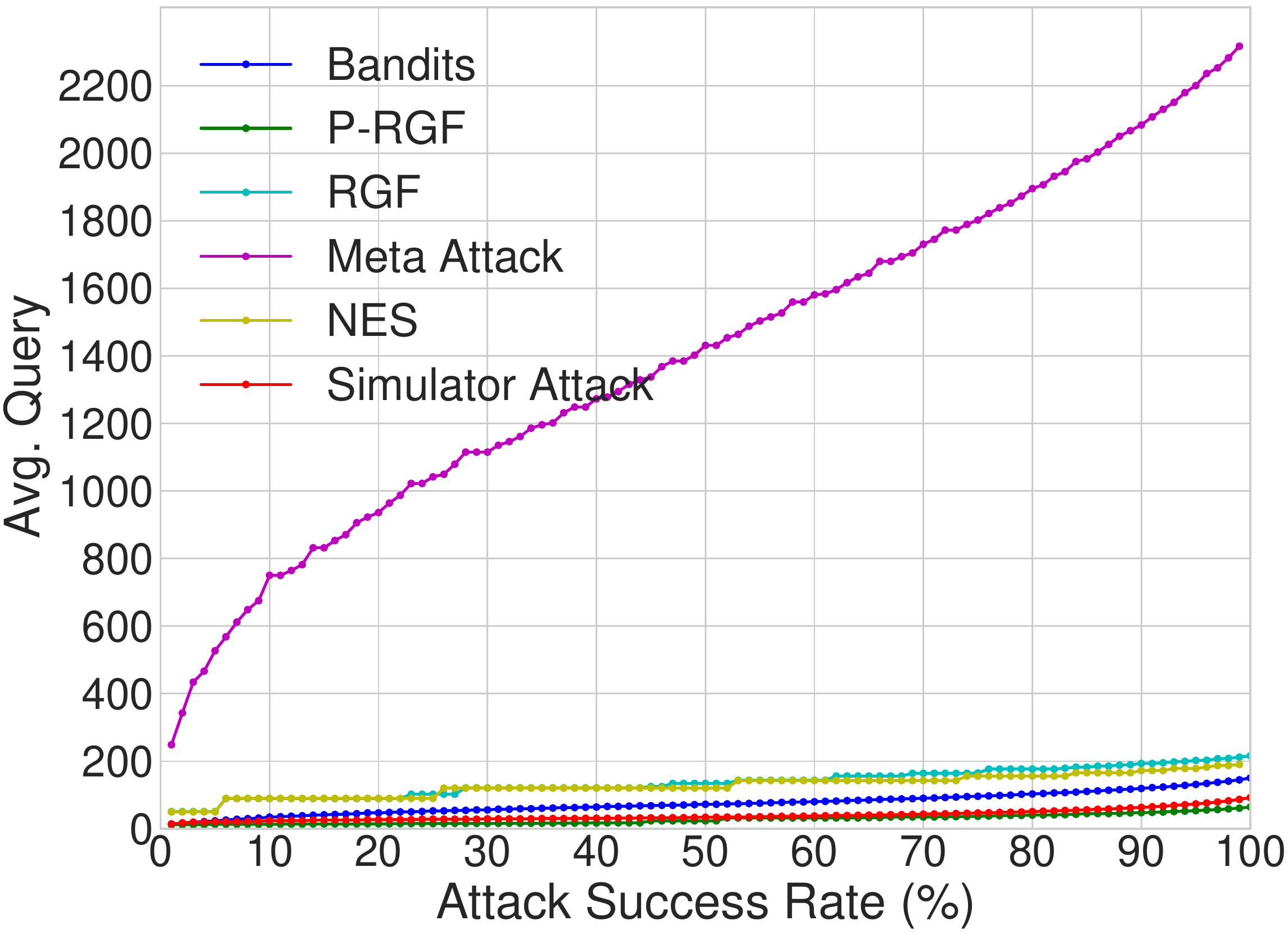}
		\subcaption{untargeted $\ell_2$ attack PyramidNet-272}
	\end{minipage}
	\begin{minipage}[b]{.245\textwidth}
		\includegraphics[width=\linewidth]{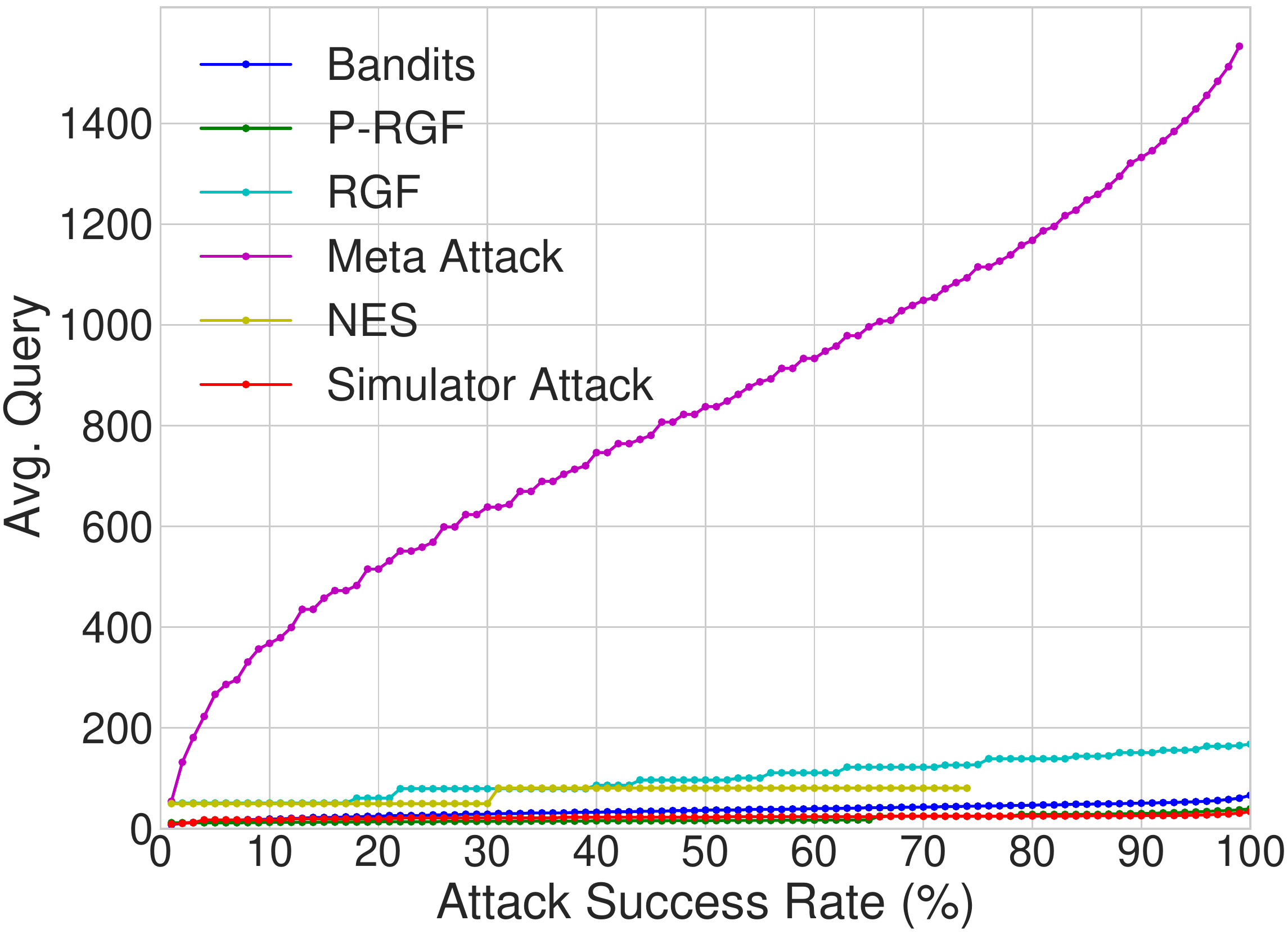}
		\subcaption{untargeted $\ell_2$ attack GDAS}
	\end{minipage}
	\begin{minipage}[b]{.245\textwidth}
		\includegraphics[width=\linewidth]{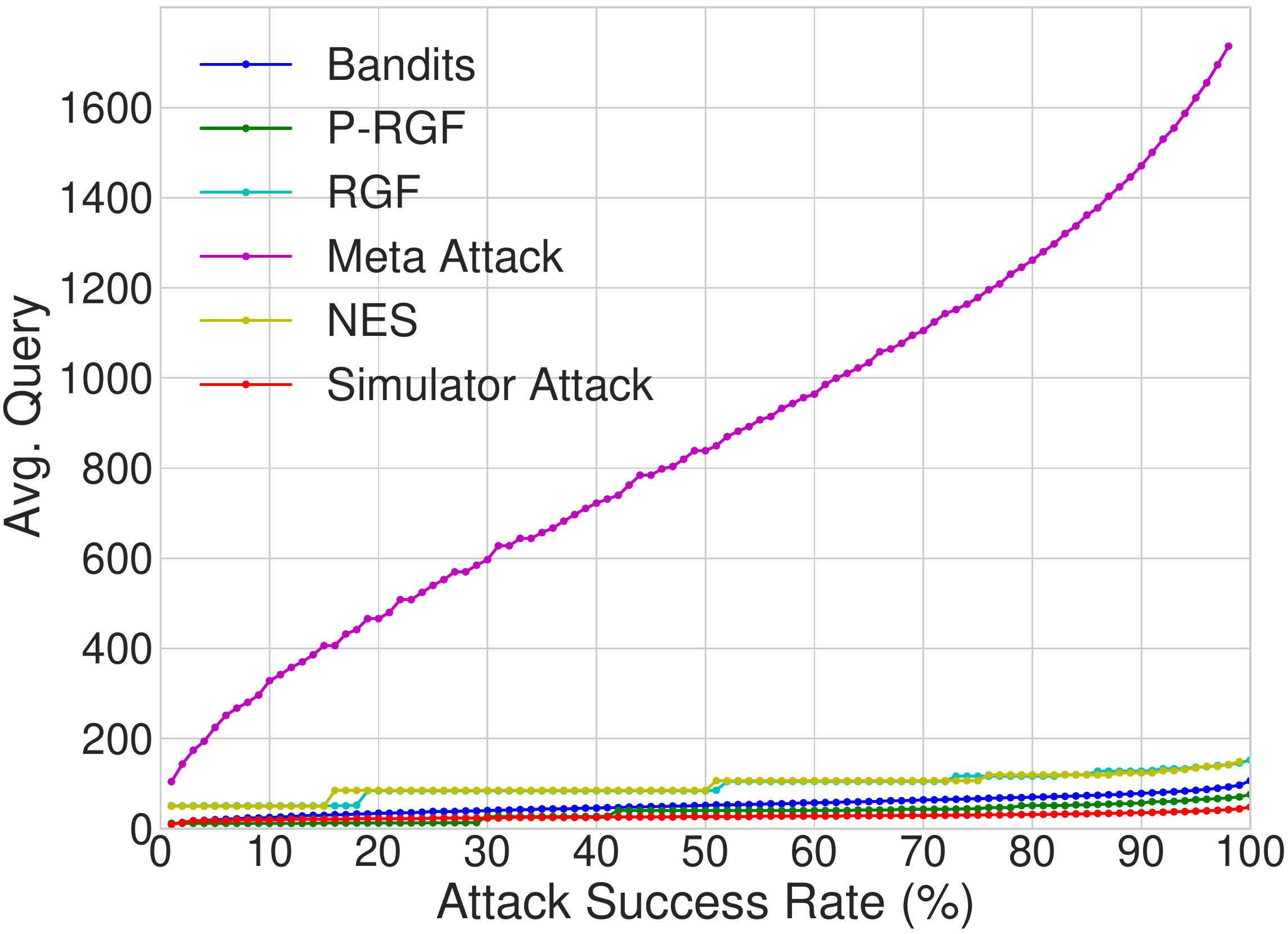}
		\subcaption{untargeted $\ell_2$ attack WRN-28}
	\end{minipage}
	\begin{minipage}[b]{.245\textwidth}
		\includegraphics[width=\linewidth]{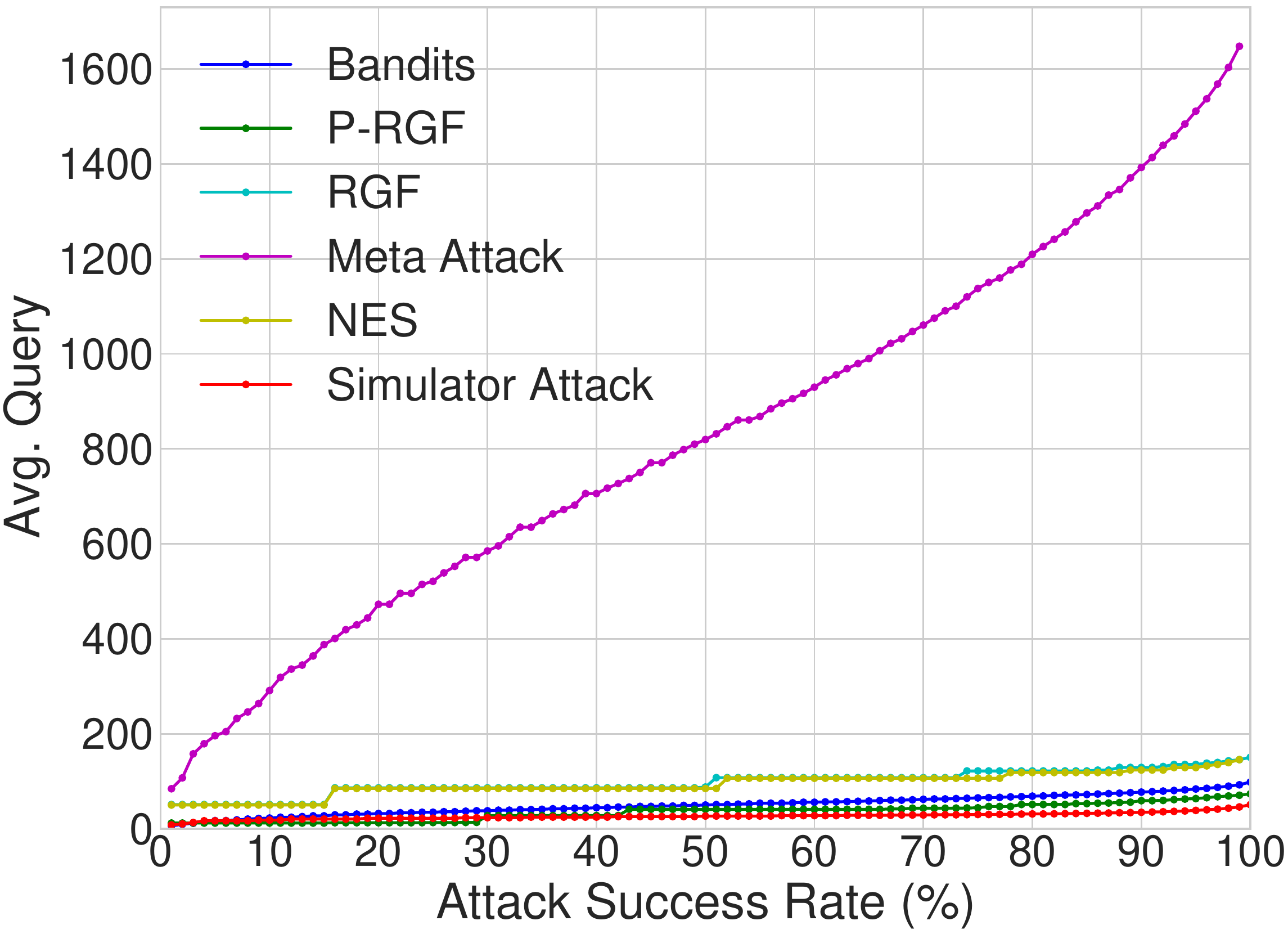}
		\subcaption{untargeted $\ell_2$ attack WRN-40}
	\end{minipage}
	\begin{minipage}[b]{.245\textwidth}
		\includegraphics[width=\linewidth]{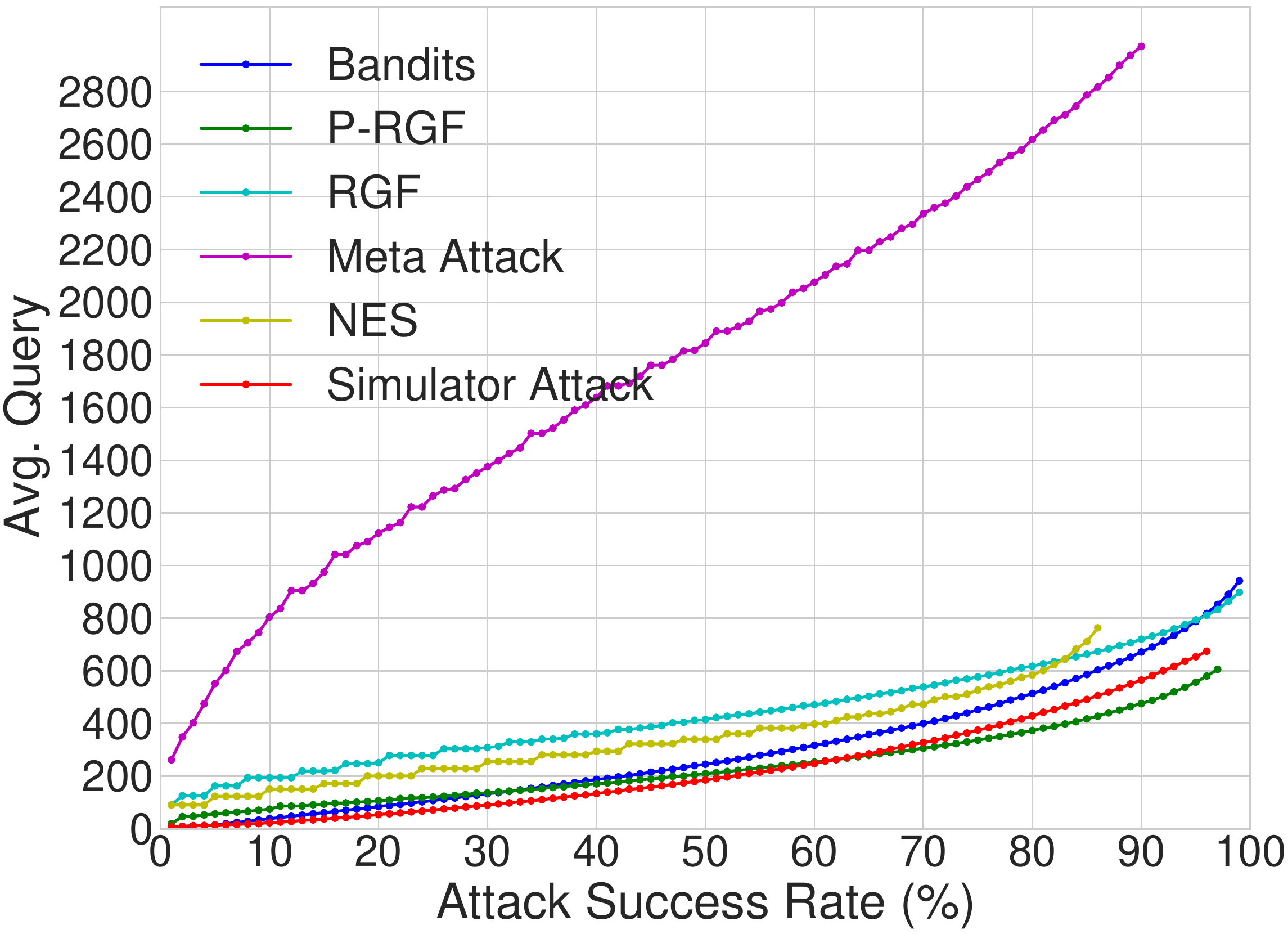}
		\subcaption{untargeted $\ell_\infty$ attack PyramidNet-272}
	\end{minipage}
	\begin{minipage}[b]{.245\textwidth}
		\includegraphics[width=\linewidth]{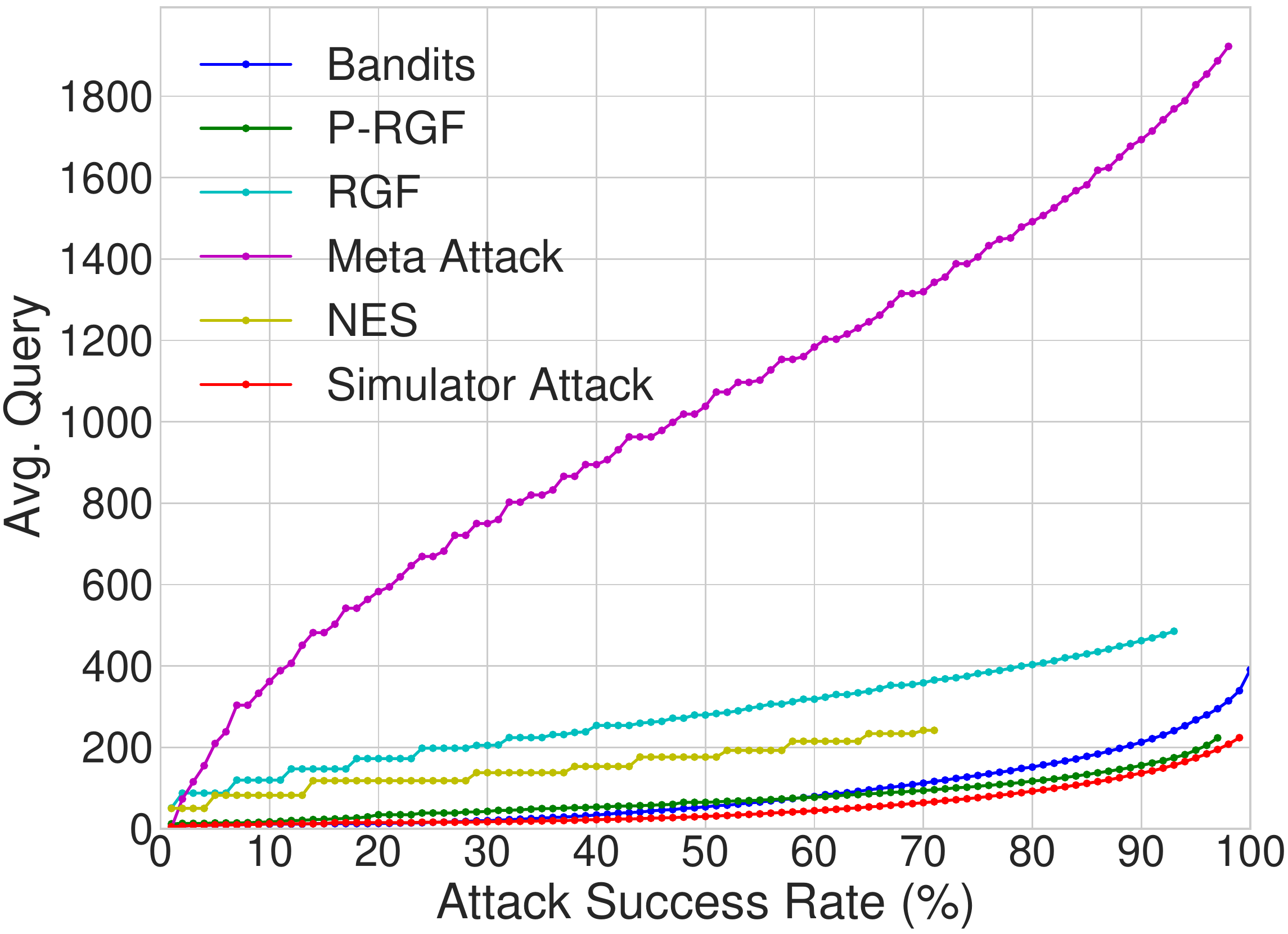}
		\subcaption{untargeted $\ell_\infty$ attack GDAS}
	\end{minipage}
	\begin{minipage}[b]{.245\textwidth}
		\includegraphics[width=\linewidth]{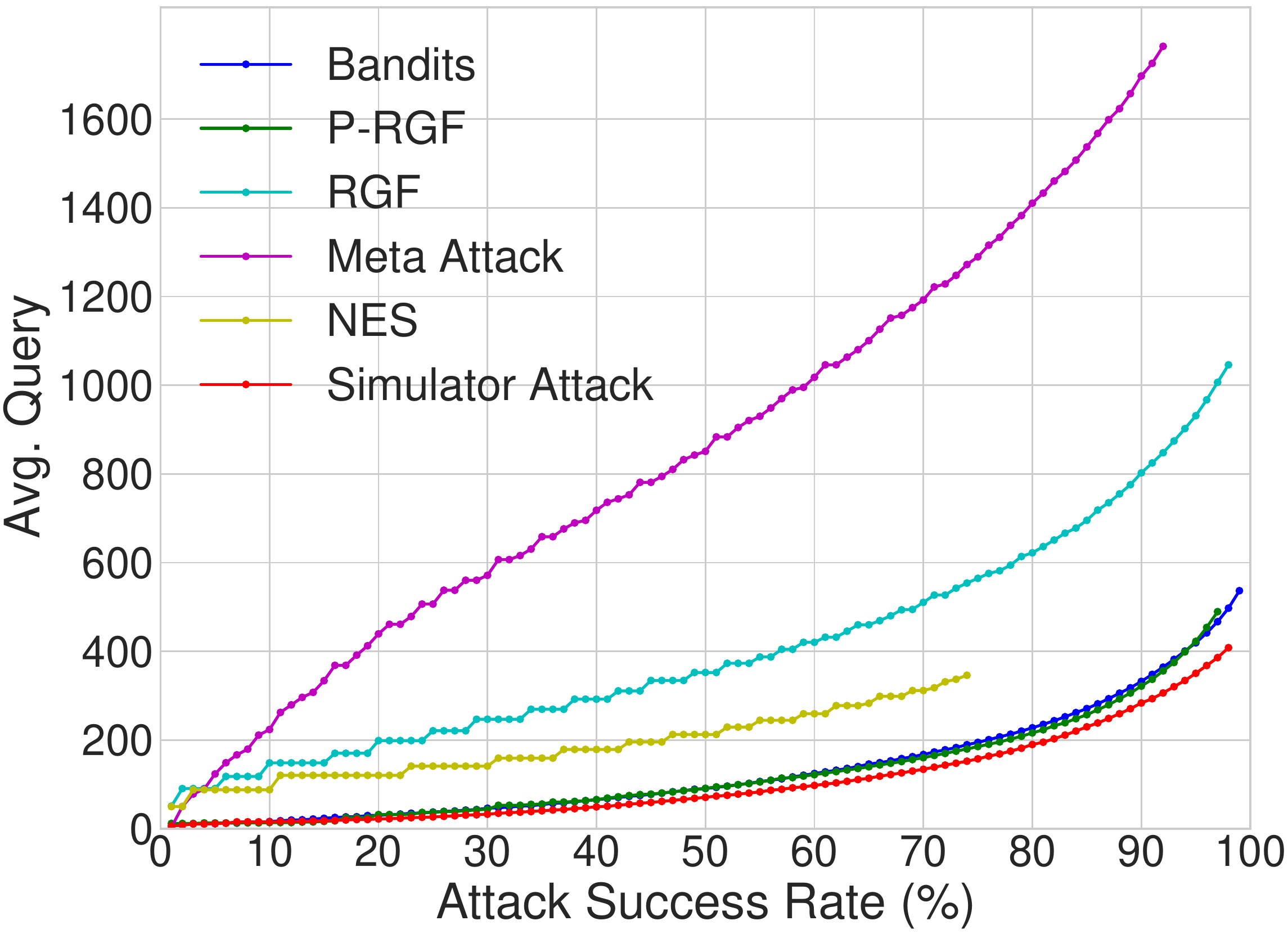}
		\subcaption{untargeted $\ell_\infty$ attack WRN-28}
	\end{minipage}
	\begin{minipage}[b]{.245\textwidth}
		\includegraphics[width=\linewidth]{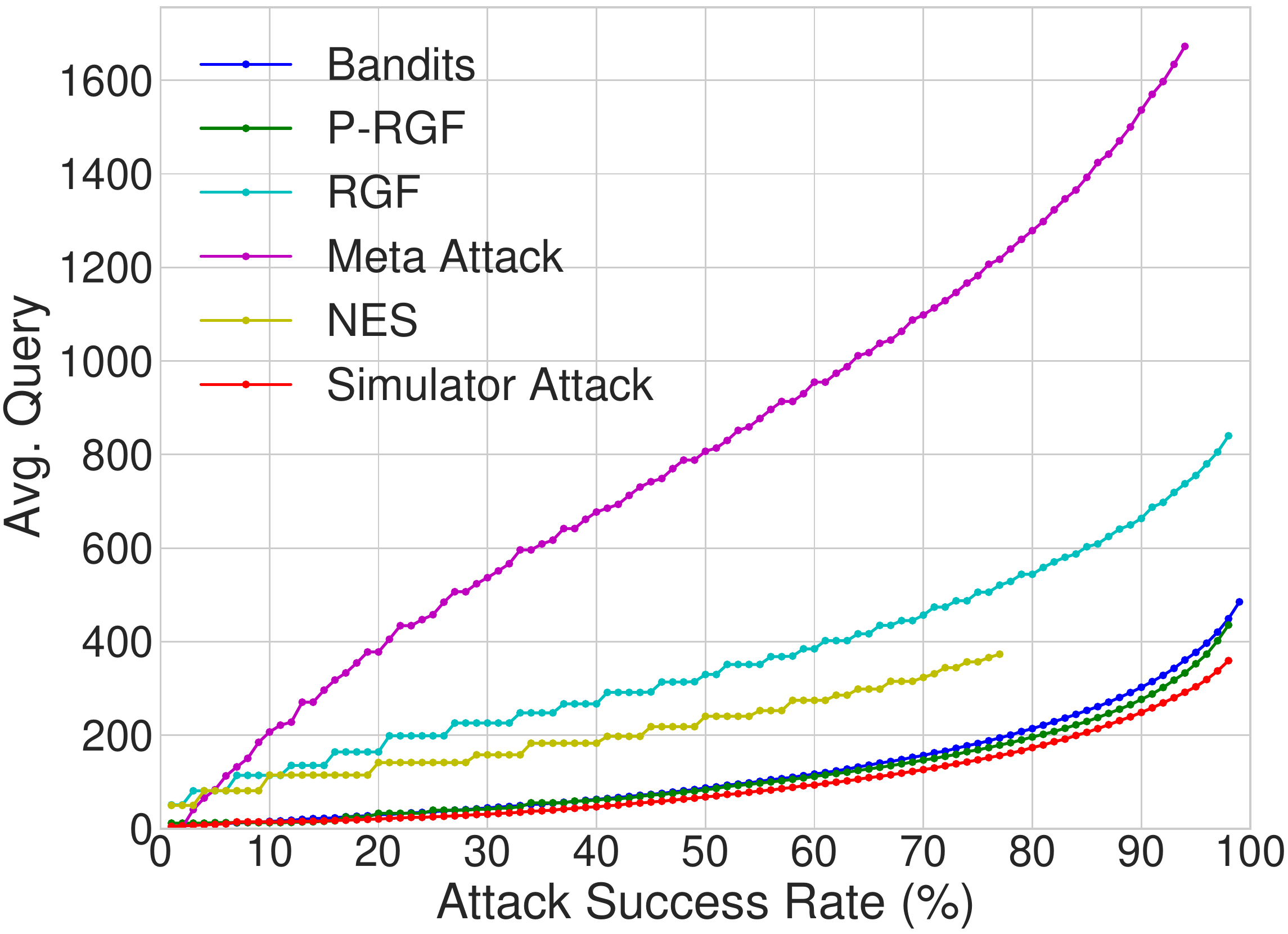}
		\subcaption{untargeted $\ell_\infty$ attack WRN-40}
	\end{minipage}
	\begin{minipage}[b]{.245\textwidth}
		\includegraphics[width=\linewidth]{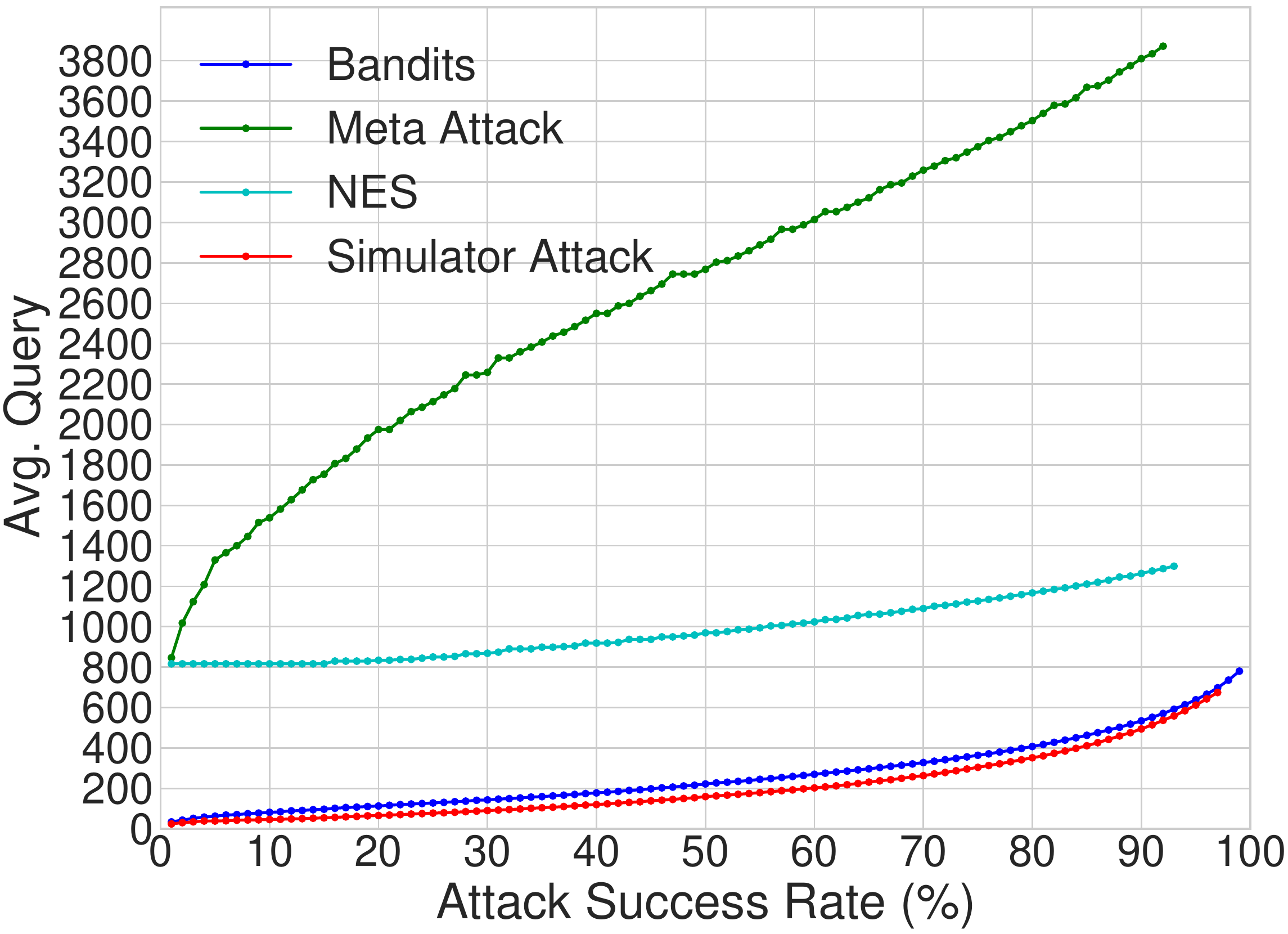}
		\subcaption{targeted $\ell_2$ attack PyramidNet-272}
	\end{minipage}
	\begin{minipage}[b]{.245\textwidth}
		\includegraphics[width=\linewidth]{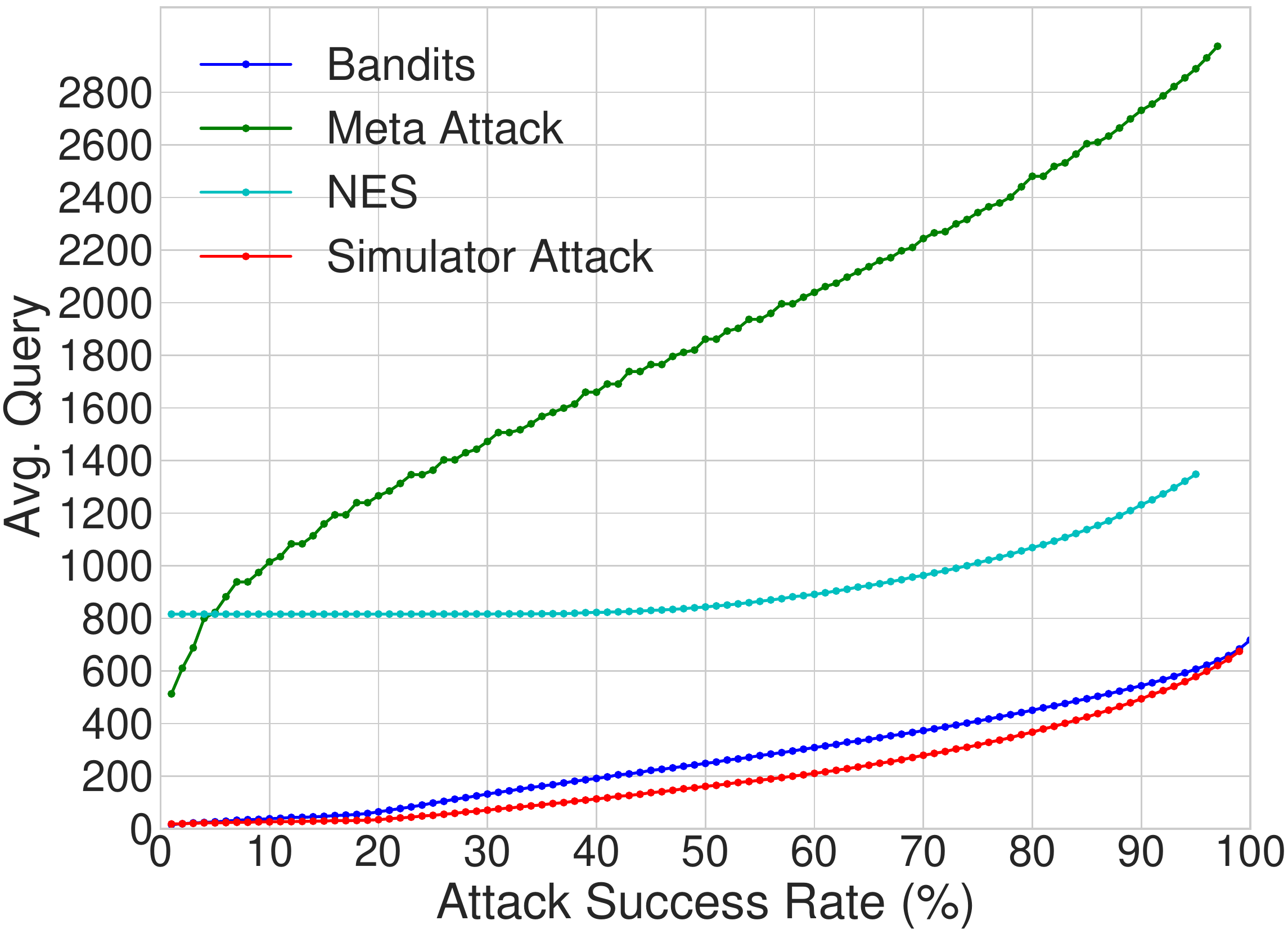}
		\subcaption{targeted $\ell_2$ attack GDAS}
	\end{minipage}
	\begin{minipage}[b]{.245\textwidth}
		\includegraphics[width=\linewidth]{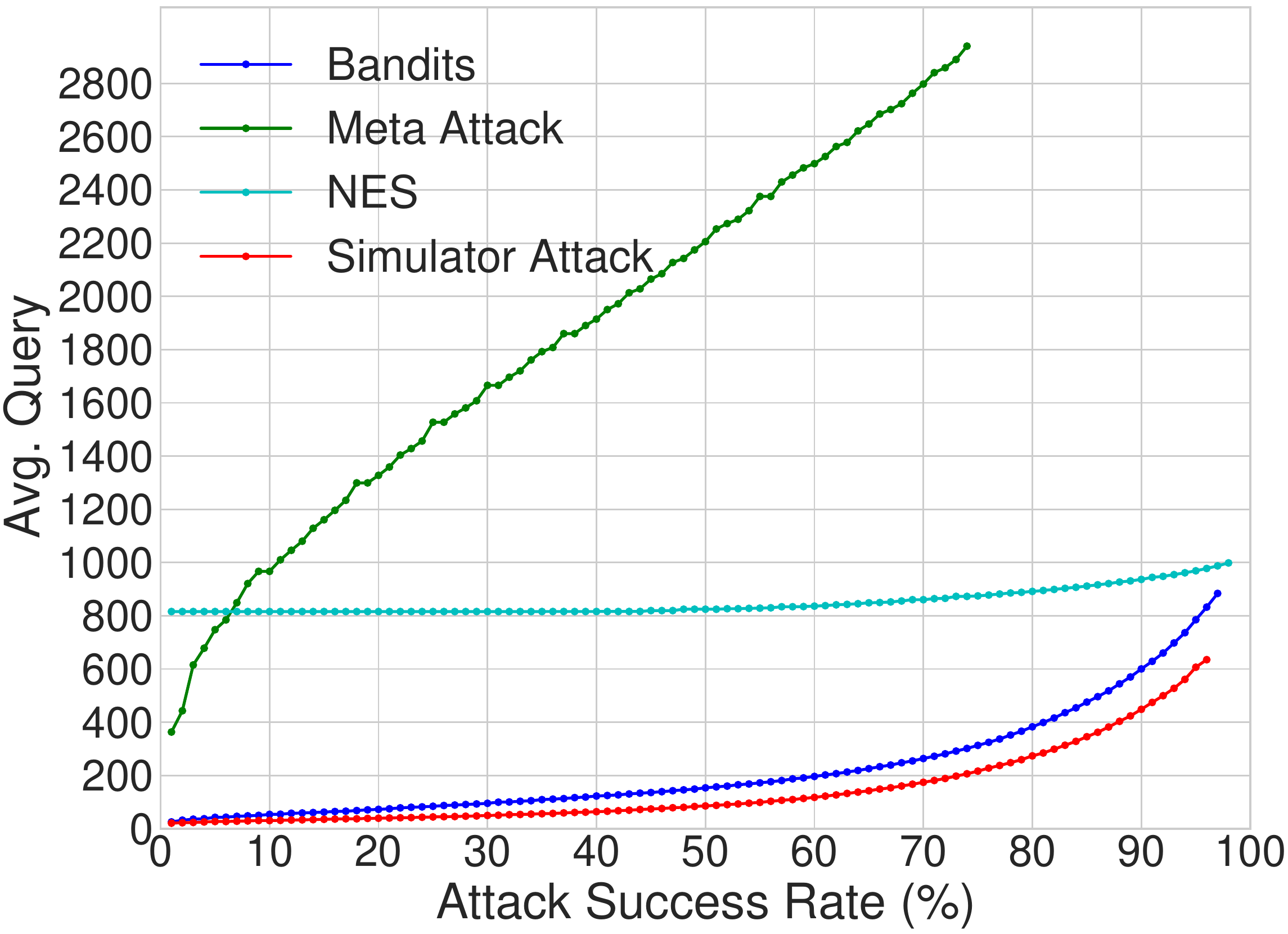}
		\subcaption{targeted $\ell_2$ attack WRN-28}
	\end{minipage}
	\begin{minipage}[b]{.245\textwidth}
		\includegraphics[width=\linewidth]{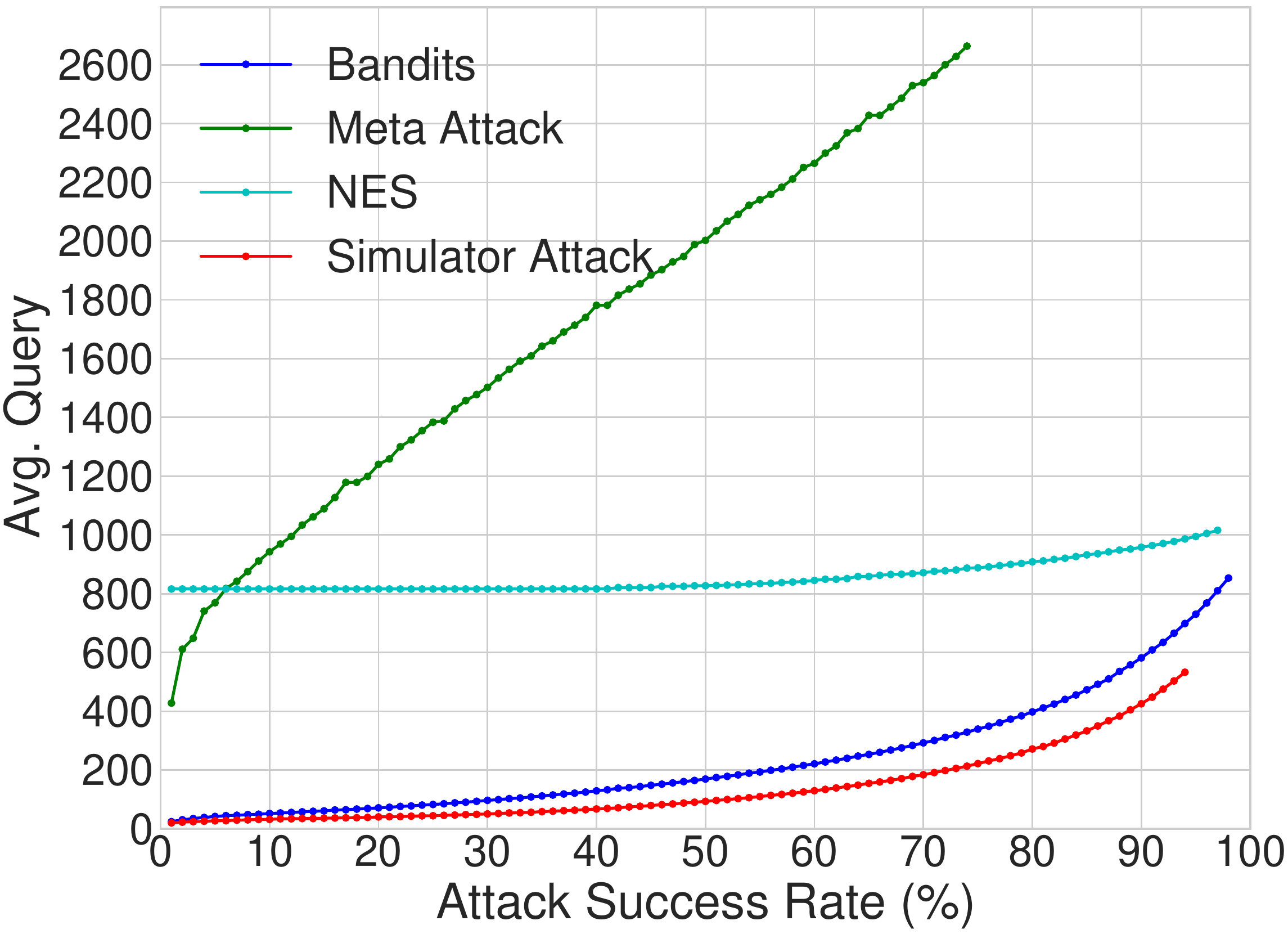}
		\subcaption{targeted $\ell_2$ attack WRN-40}
	\end{minipage}
	\begin{minipage}[b]{.245\textwidth}
		\includegraphics[width=\linewidth]{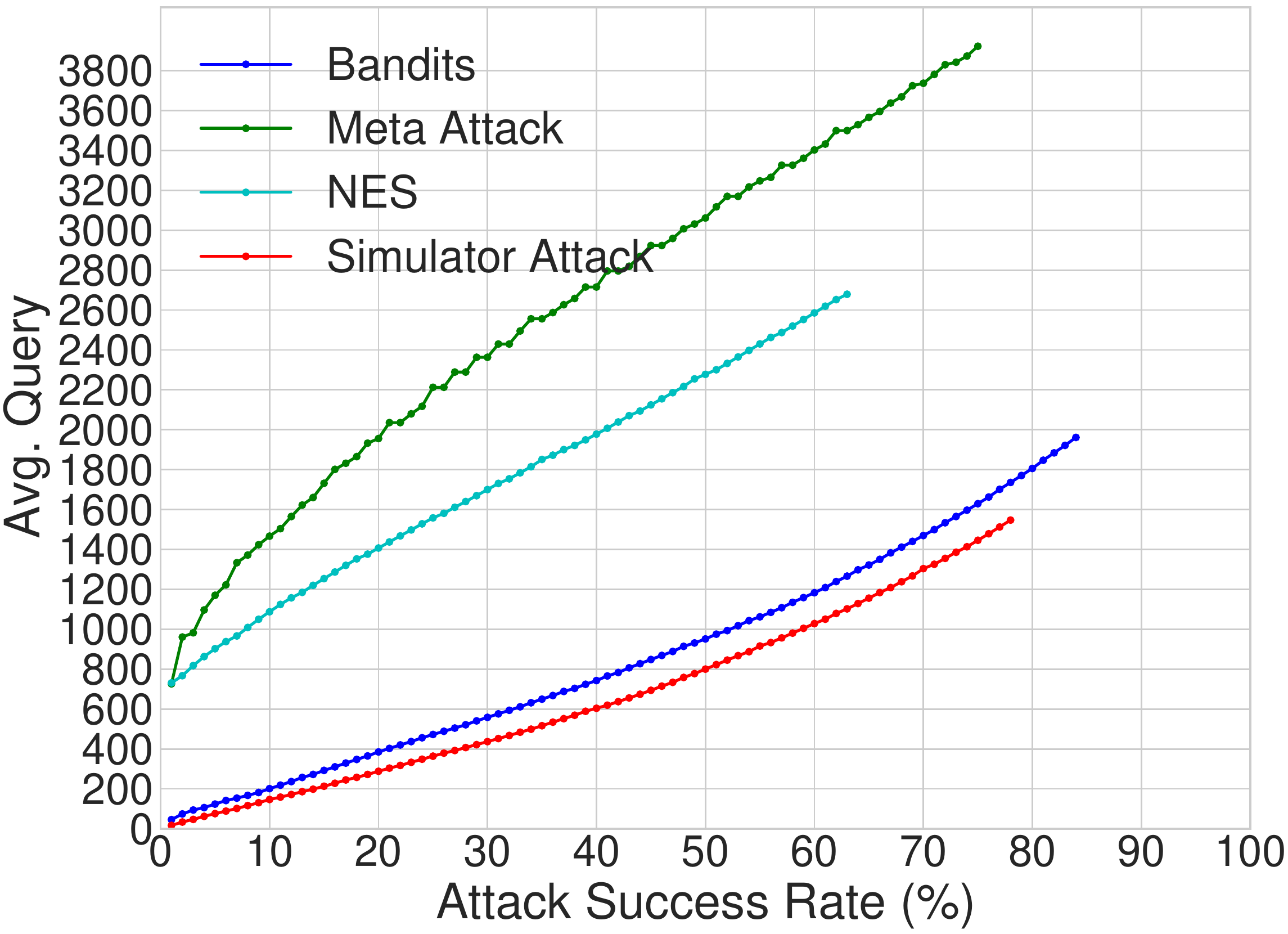}
		\subcaption{targeted $\ell_\infty$ attack PyramidNet-272}
	\end{minipage}
	\begin{minipage}[b]{.245\textwidth}
		\includegraphics[width=\linewidth]{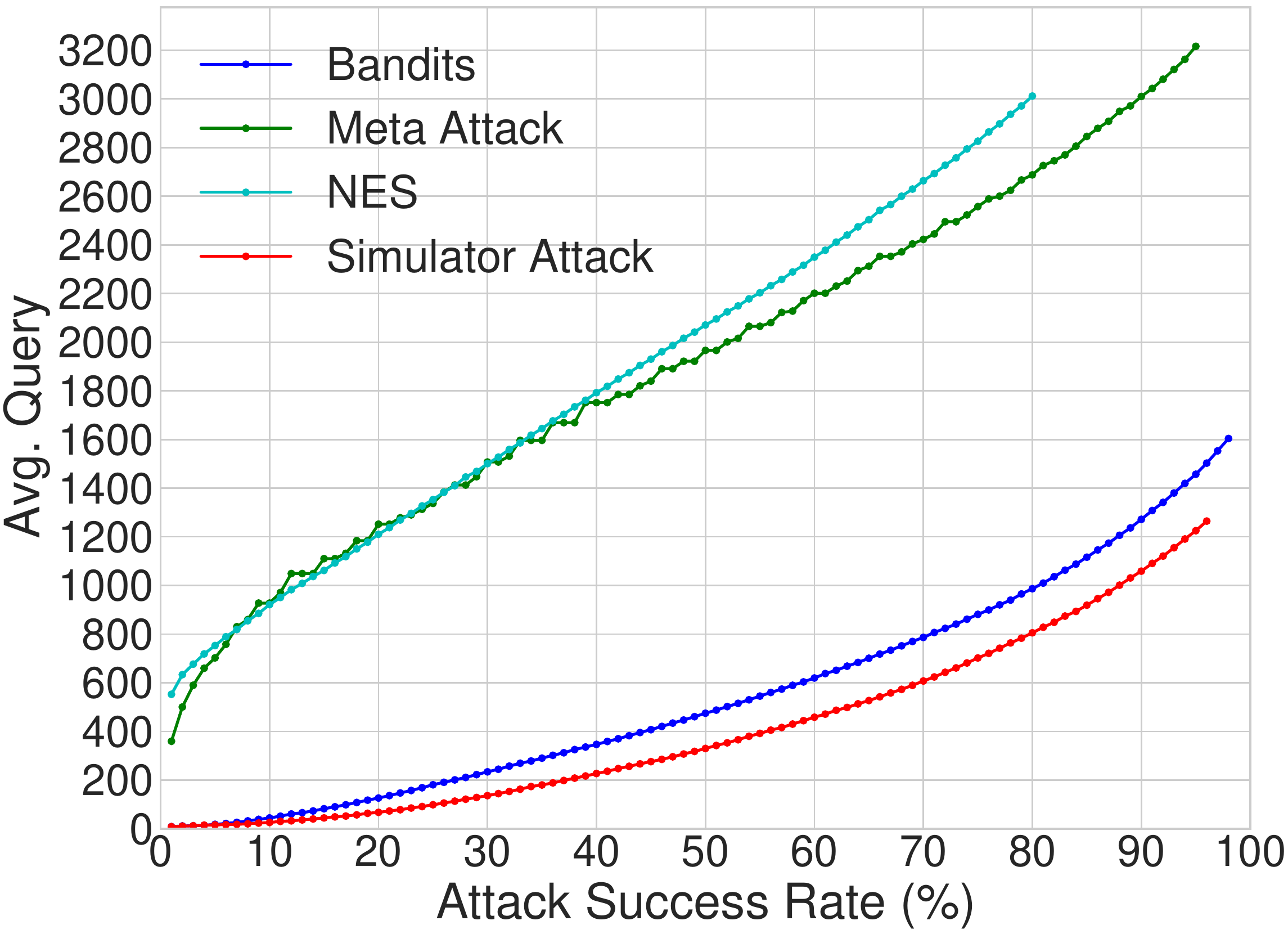}
		\subcaption{targeted $\ell_\infty$ attack GDAS}
	\end{minipage}
	\begin{minipage}[b]{.245\textwidth}
		\includegraphics[width=\linewidth]{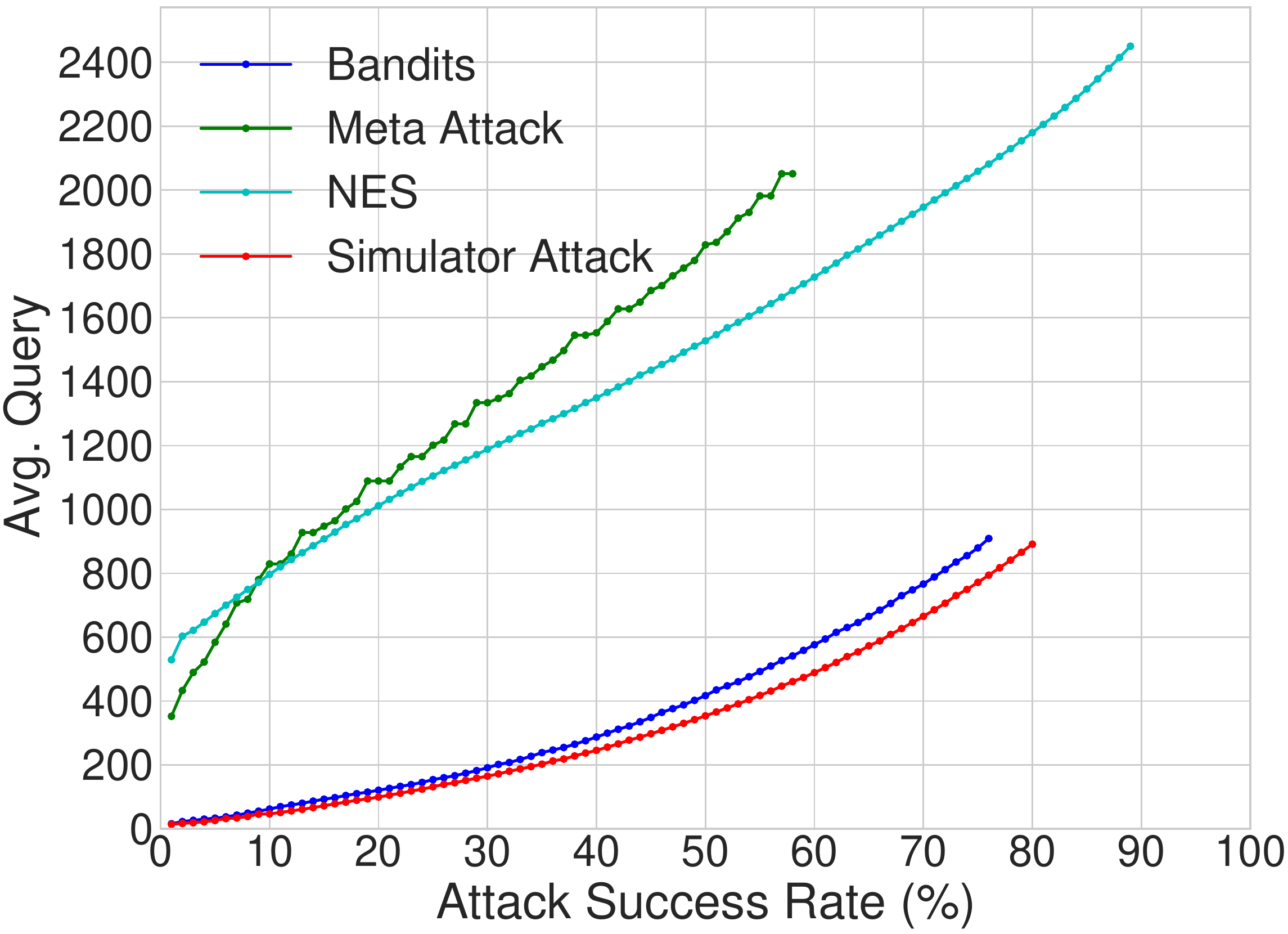}
		\subcaption{targeted $\ell_\infty$ attack WRN-28}
	\end{minipage}
	\begin{minipage}[b]{.245\textwidth}
		\includegraphics[width=\linewidth]{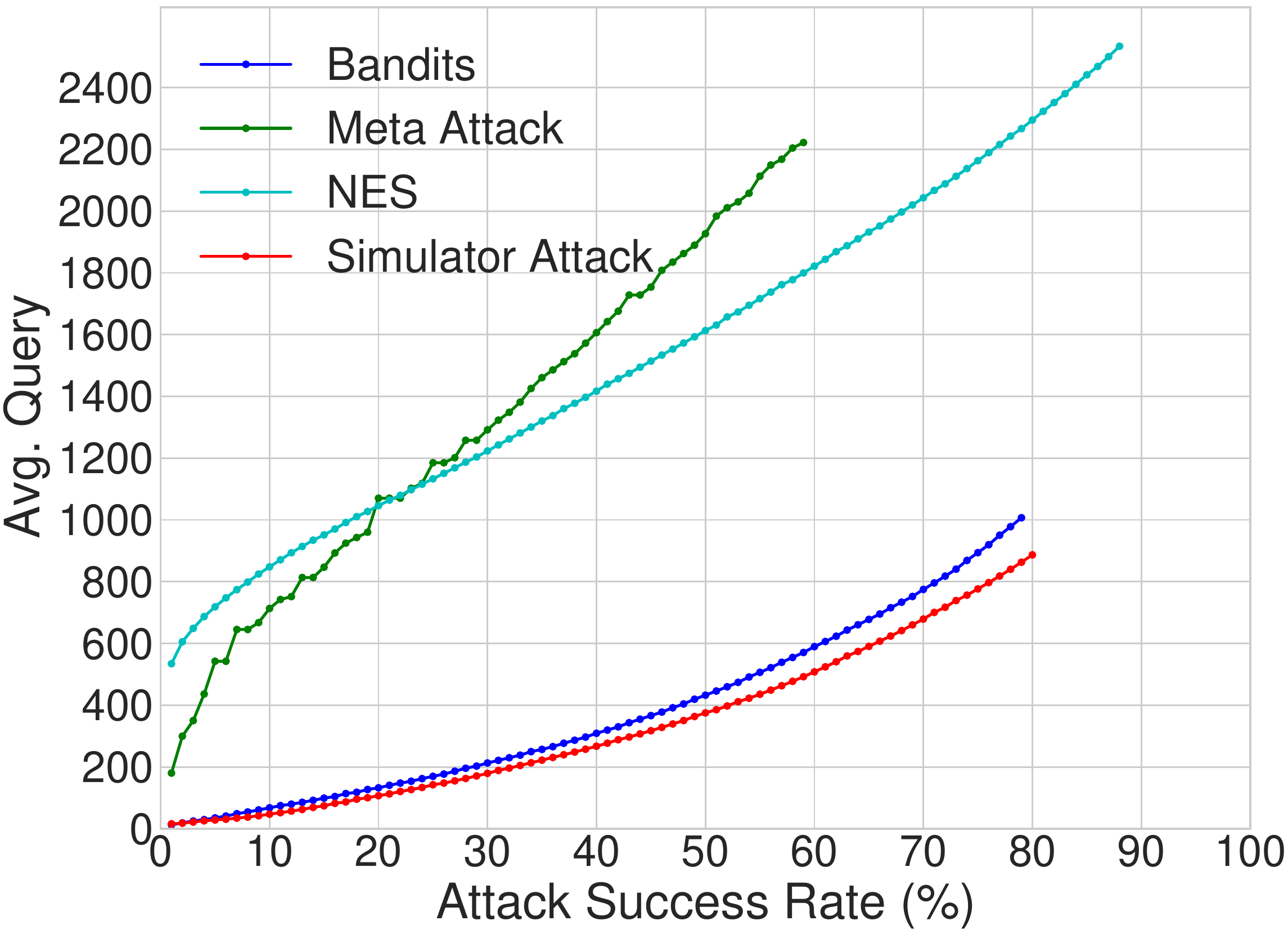}
		\subcaption{targeted $\ell_\infty$ attack WRN-40}
	\end{minipage}
	\caption{Comparisons of the average query per successful image at different desired success rates in CIFAR-10 dataset.}
	\label{fig:success_rate_to_avg_query_CIFAR-10}
\end{figure*}

\begin{figure*}[bp]
	\setlength{\abovecaptionskip}{0pt}
	\setlength{\belowcaptionskip}{0pt}
	\captionsetup[sub]{font={scriptsize}}
	\centering 
	\begin{minipage}[b]{.245\textwidth}
		\includegraphics[width=\linewidth]{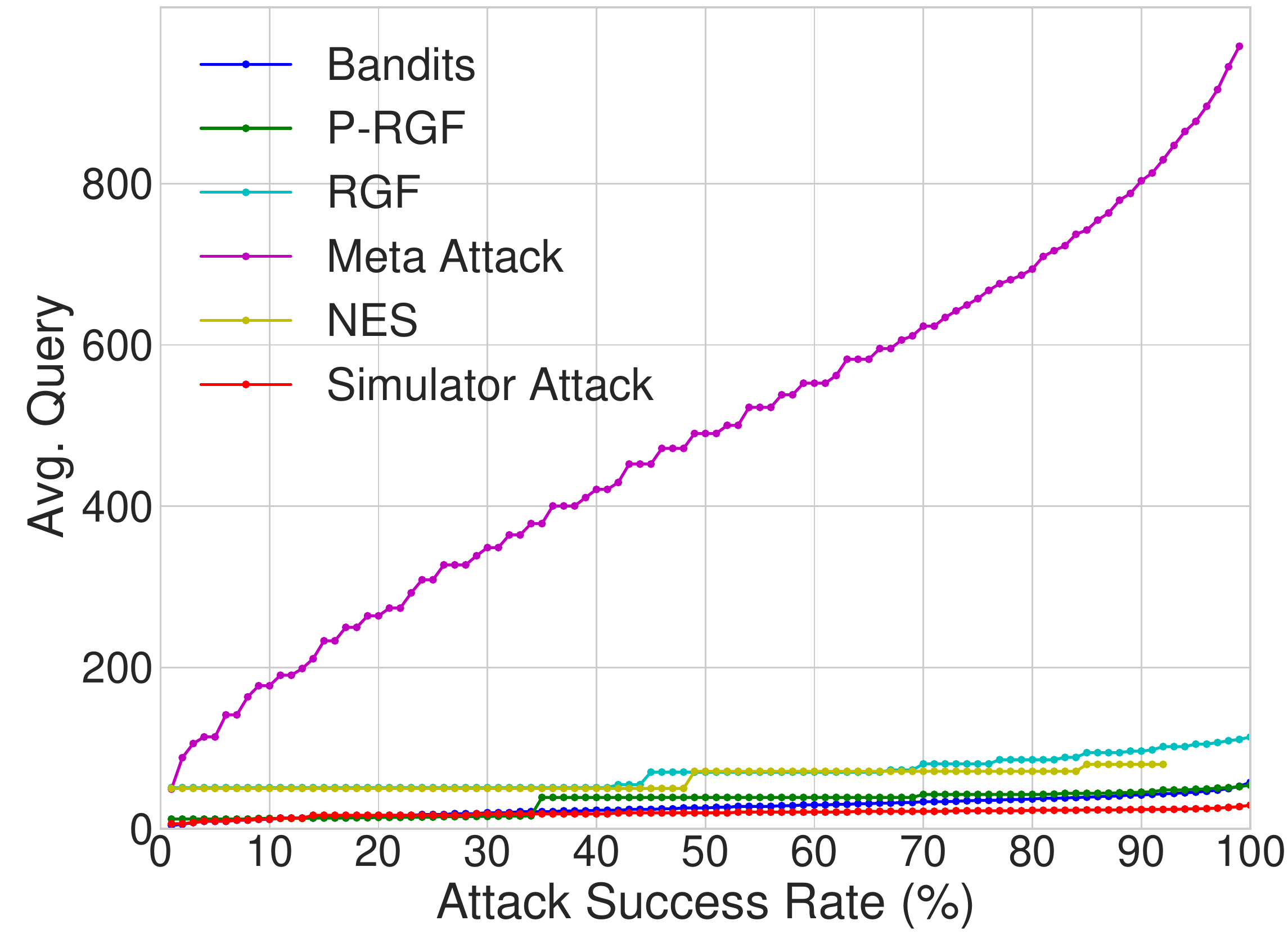}
		\subcaption{untargeted $\ell_2$ attack PyramidNet-272}
	\end{minipage}
	\begin{minipage}[b]{.245\textwidth}
		\includegraphics[width=\linewidth]{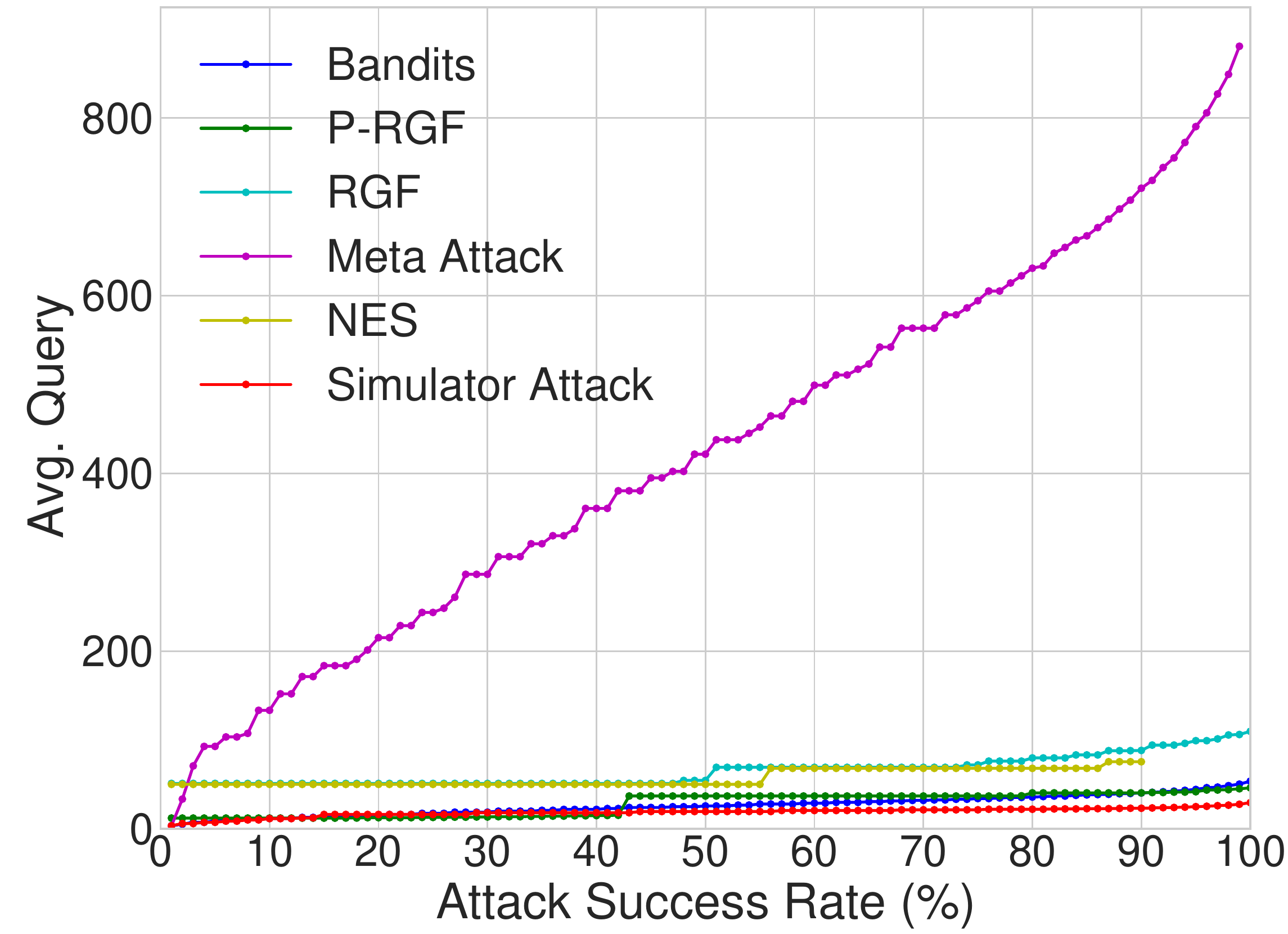}
		\subcaption{untargeted $\ell_2$ attack GDAS}
	\end{minipage}
	\begin{minipage}[b]{.245\textwidth}
		\includegraphics[width=\linewidth]{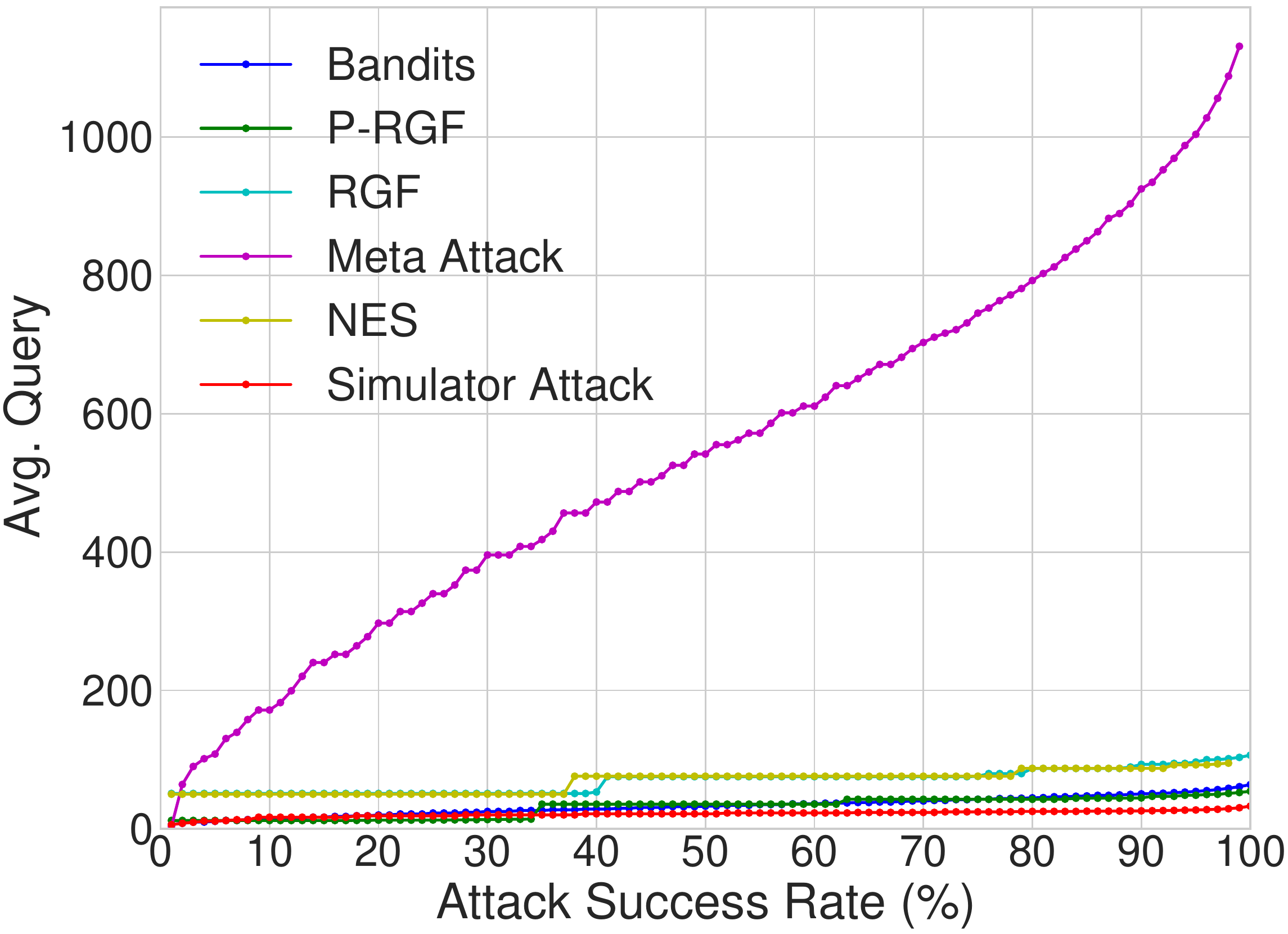}
		\subcaption{untargeted $\ell_2$ attack WRN-28}
	\end{minipage}
	\begin{minipage}[b]{.245\textwidth}
		\includegraphics[width=\linewidth]{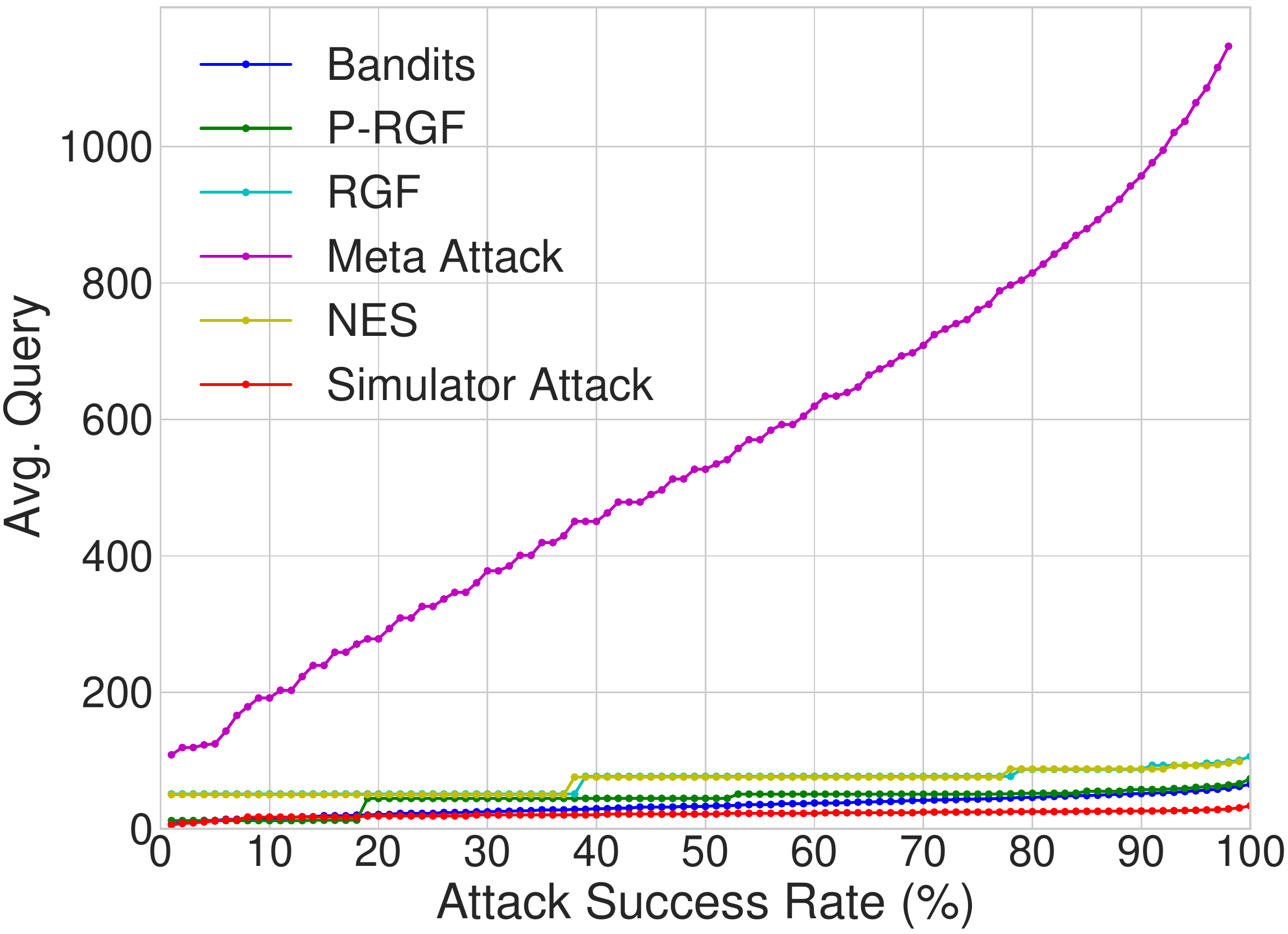}
		\subcaption{untargeted $\ell_2$ attack WRN-40}
	\end{minipage}
	\begin{minipage}[b]{.245\textwidth}
		\includegraphics[width=\linewidth]{figure/fig/success_rate_to_avg_query/CIFAR-100_pyramidnet272_linf_untargeted_attack_success_rate_to_avg_query.pdf}
		\subcaption{untargeted $\ell_\infty$ attack PyramidNet-272}
	\end{minipage}
	\begin{minipage}[b]{.245\textwidth}
		\includegraphics[width=\linewidth]{figure/fig/success_rate_to_avg_query/CIFAR-100_gdas_linf_untargeted_attack_success_rate_to_avg_query.pdf}
		\subcaption{untargeted $\ell_\infty$ attack GDAS}
	\end{minipage}
	\begin{minipage}[b]{.245\textwidth}
		\includegraphics[width=\linewidth]{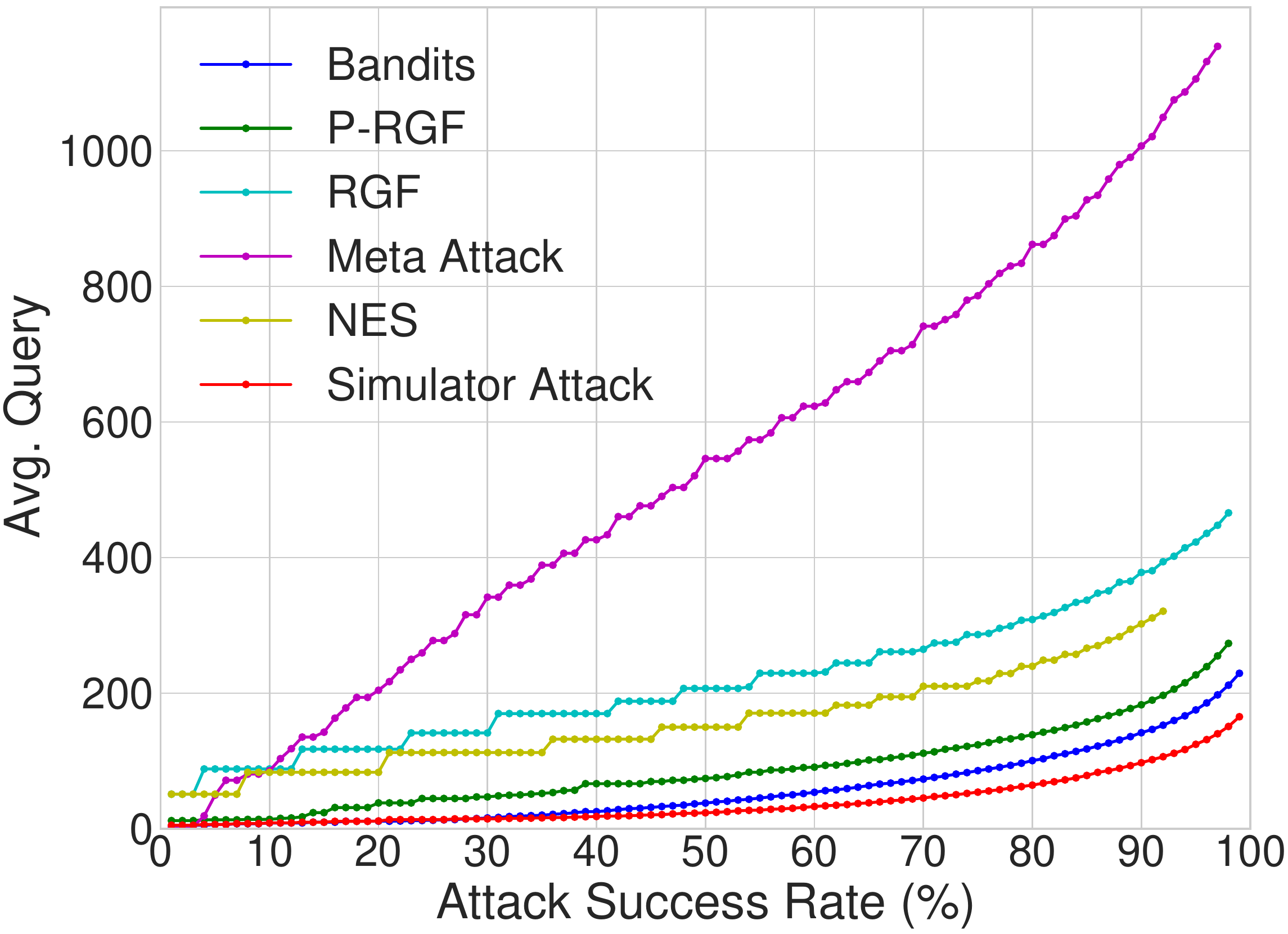}
		\subcaption{untargeted $\ell_\infty$ attack WRN-28}
	\end{minipage}
	\begin{minipage}[b]{.245\textwidth}
		\includegraphics[width=\linewidth]{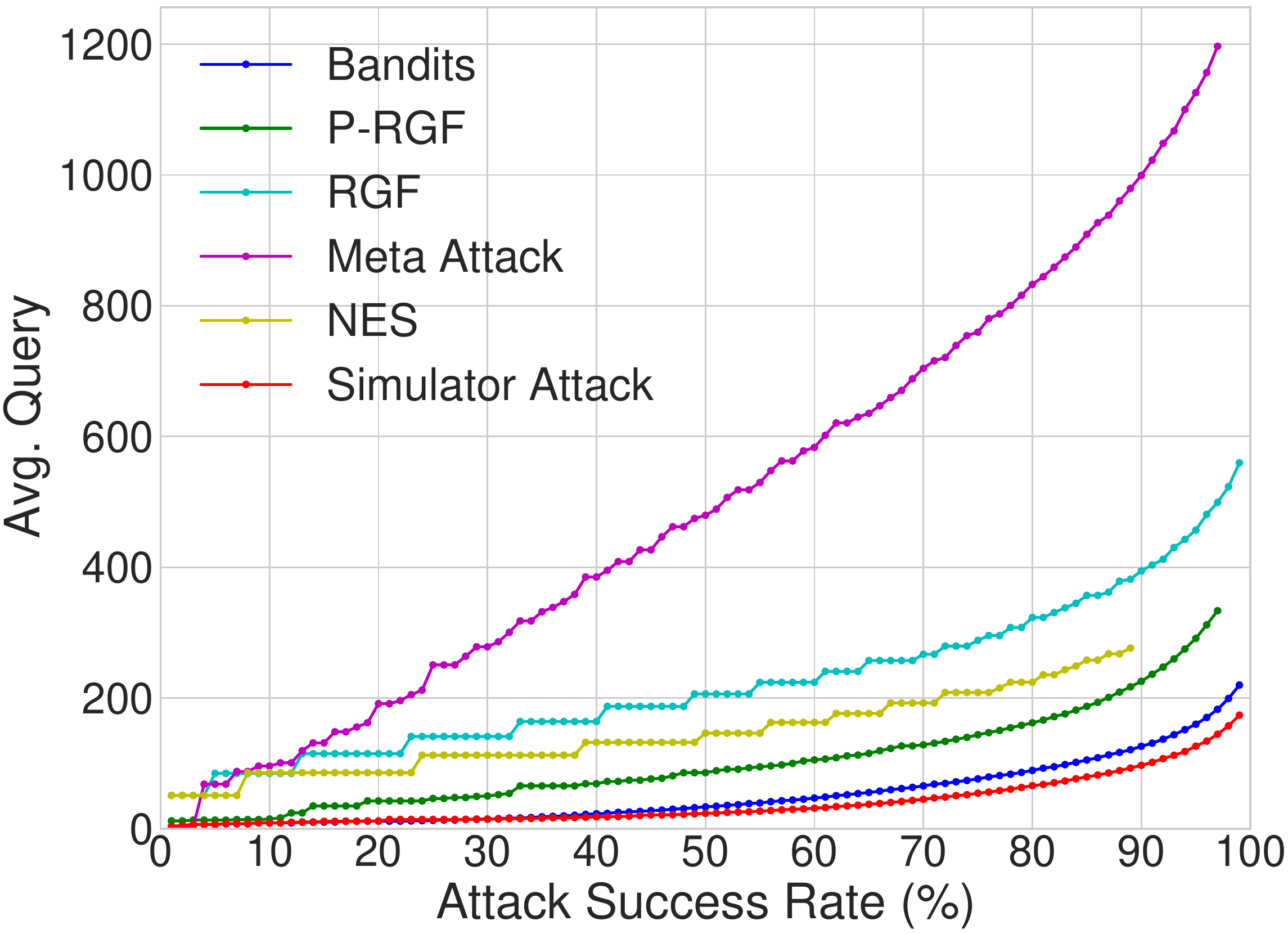}
		\subcaption{untargeted $\ell_\infty$ attack WRN-40}
	\end{minipage}
	\begin{minipage}[b]{.245\textwidth}
		\includegraphics[width=\linewidth]{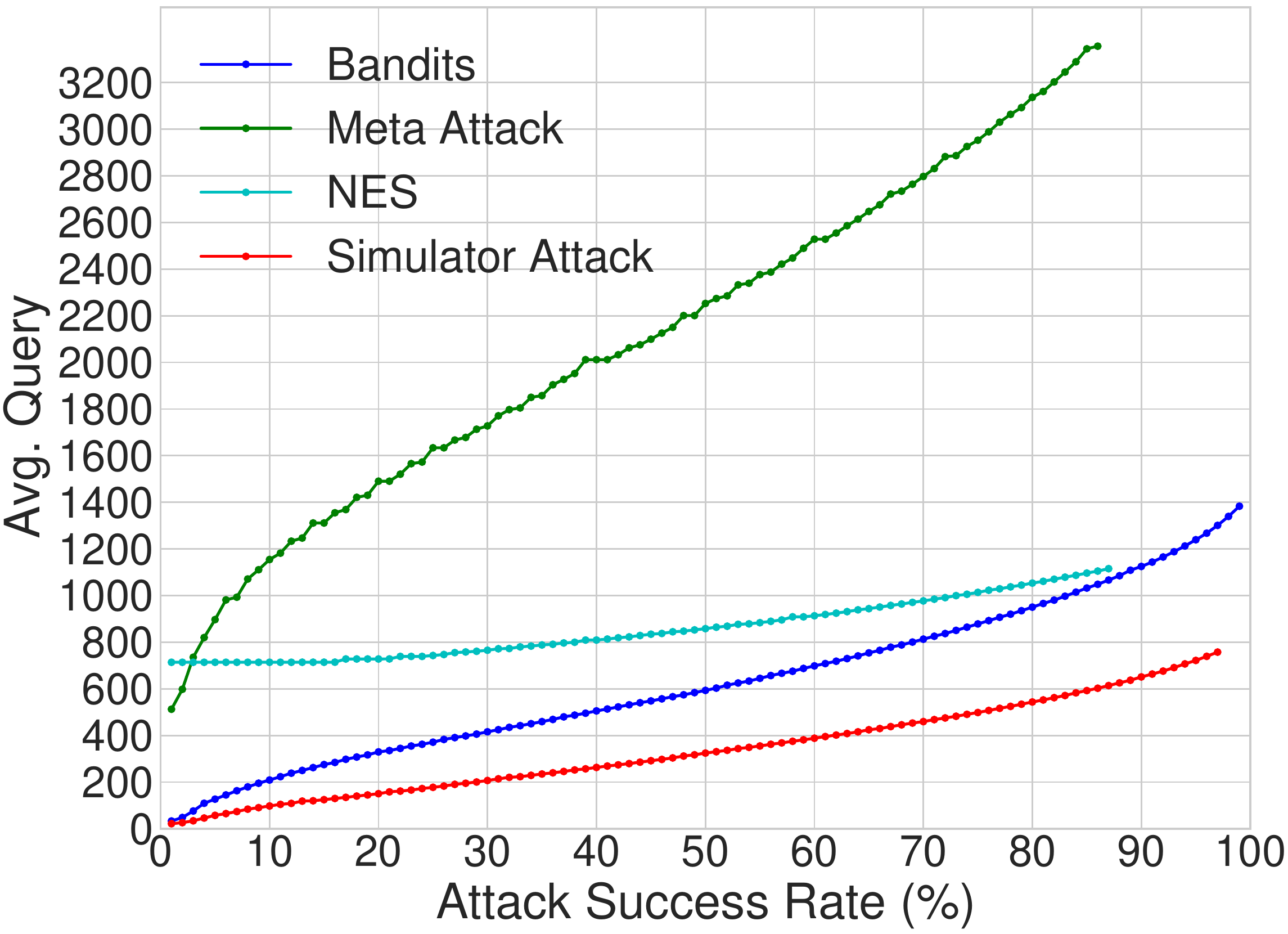}
		\subcaption{targeted $\ell_2$ attack PyramidNet-272}
	\end{minipage}
	\begin{minipage}[b]{.245\textwidth}
		\includegraphics[width=\linewidth]{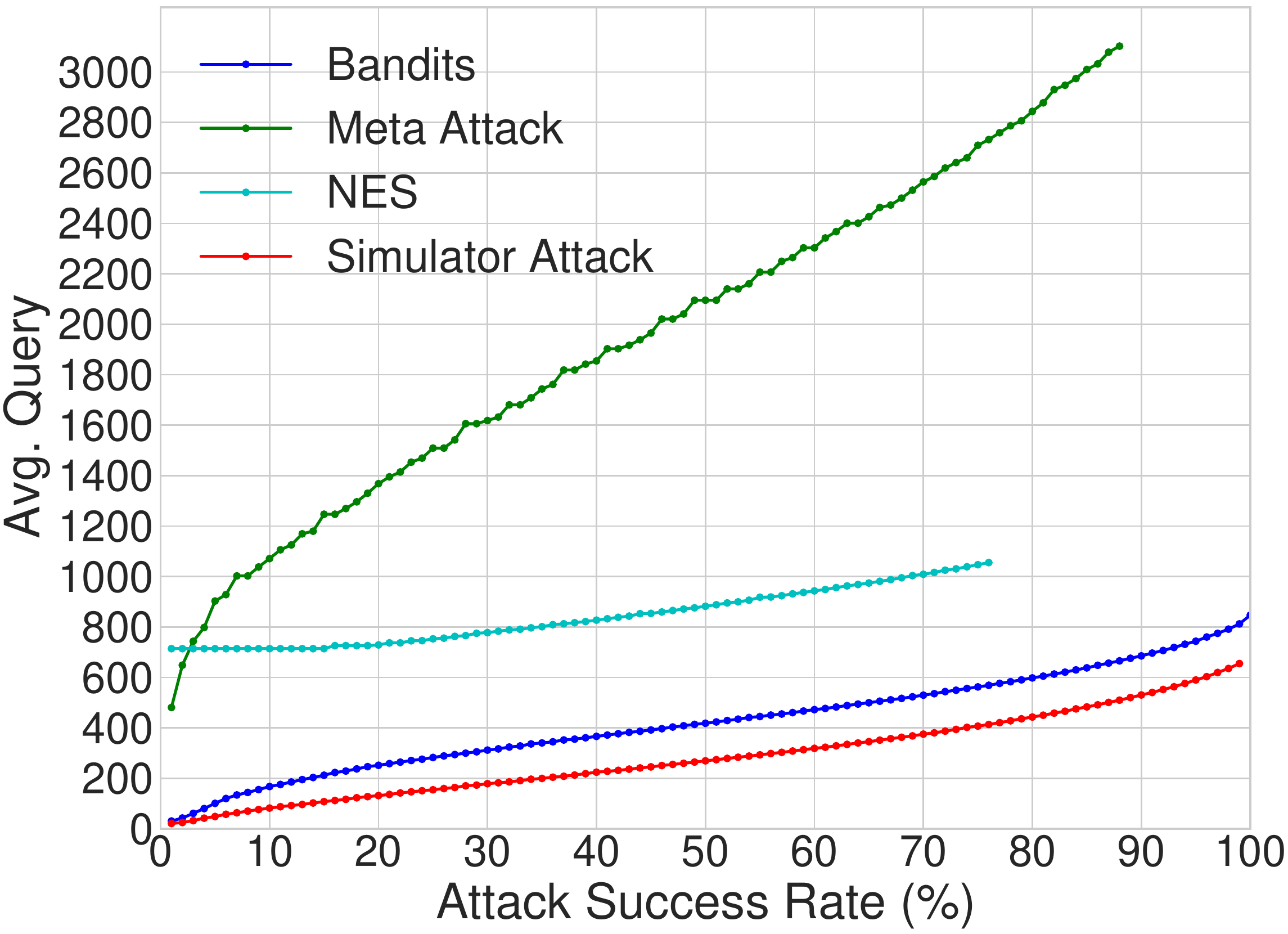}
		\subcaption{targeted $\ell_2$ attack GDAS}
	\end{minipage}
	\begin{minipage}[b]{.245\textwidth}
		\includegraphics[width=\linewidth]{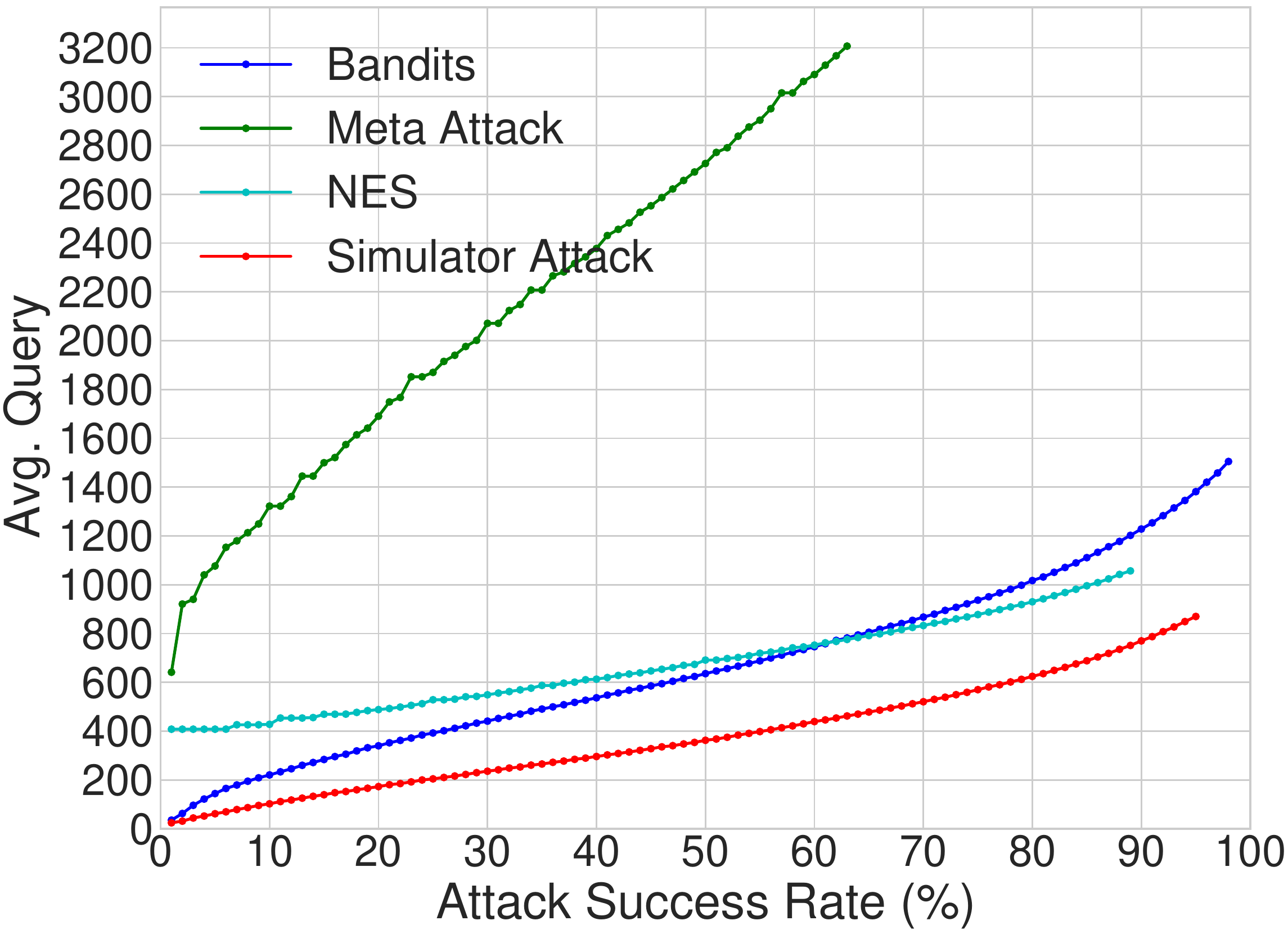}
		\subcaption{targeted $\ell_2$ attack WRN-28}
	\end{minipage}
	\begin{minipage}[b]{.245\textwidth}
		\includegraphics[width=\linewidth]{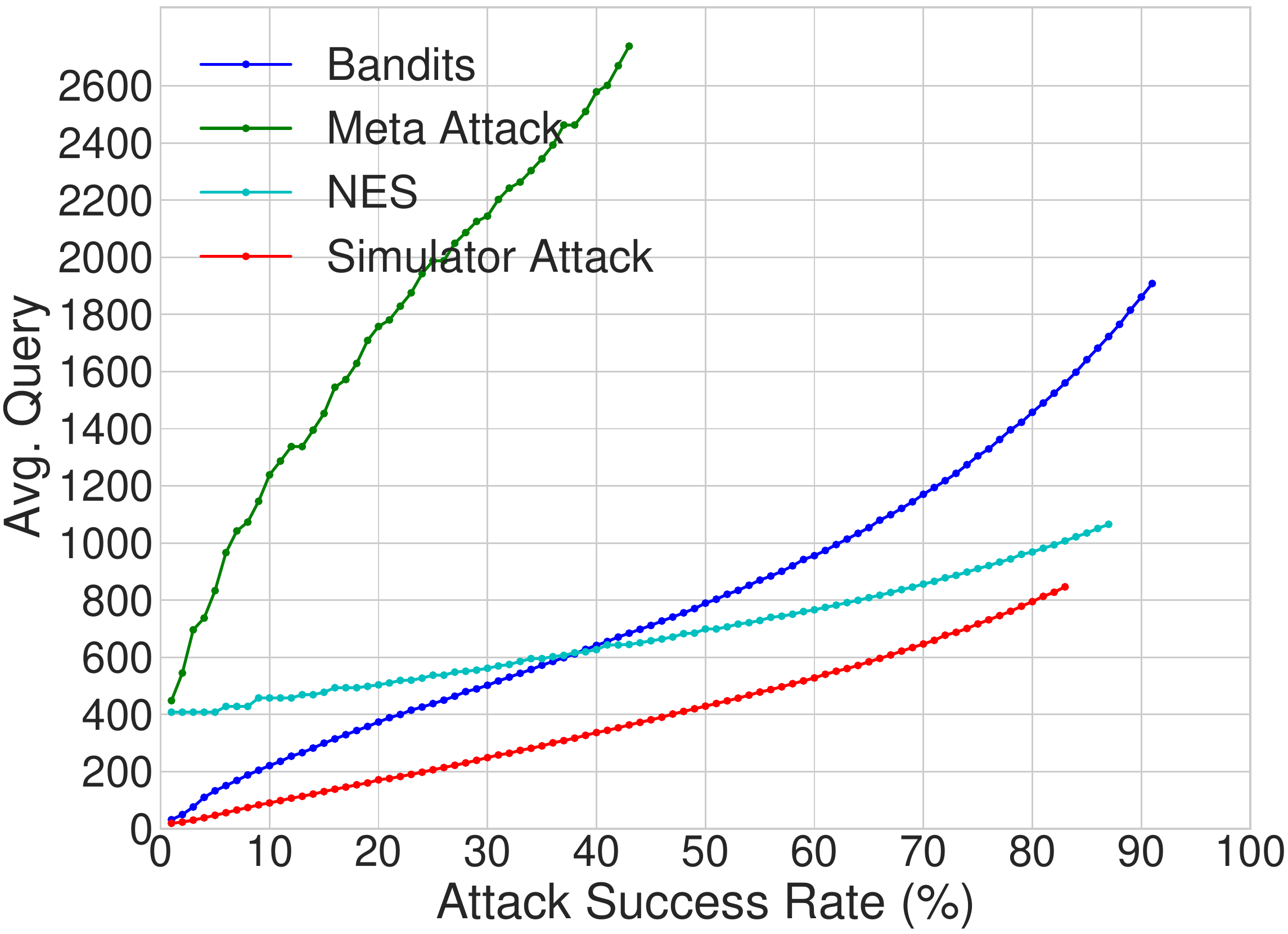}
		\subcaption{targeted $\ell_2$ attack WRN-40}
	\end{minipage}
	\begin{minipage}[b]{.245\textwidth}
		\includegraphics[width=\linewidth]{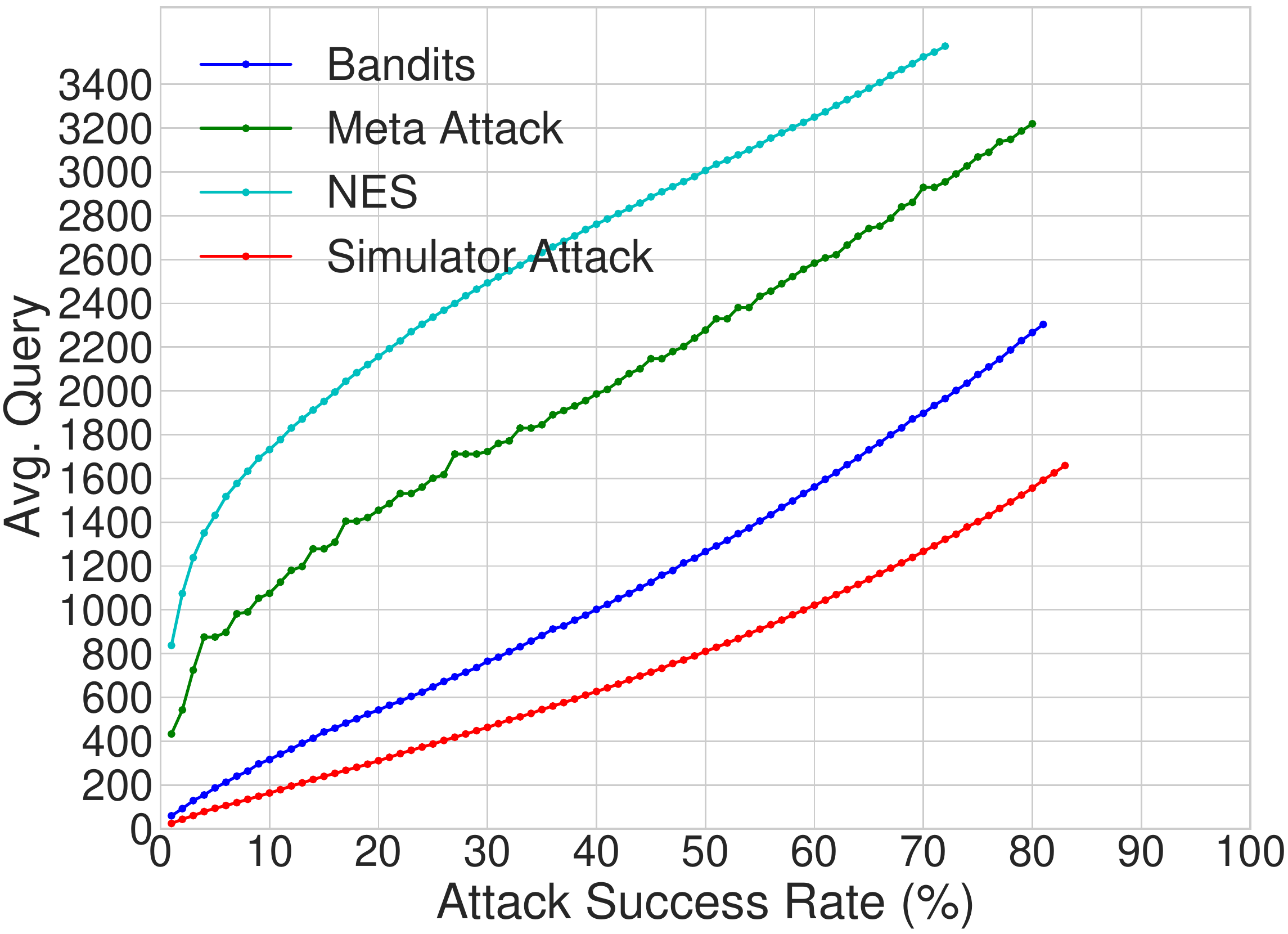}
		\subcaption{targeted $\ell_\infty$ attack PyramidNet-272}
	\end{minipage}
	\begin{minipage}[b]{.245\textwidth}
		\includegraphics[width=\linewidth]{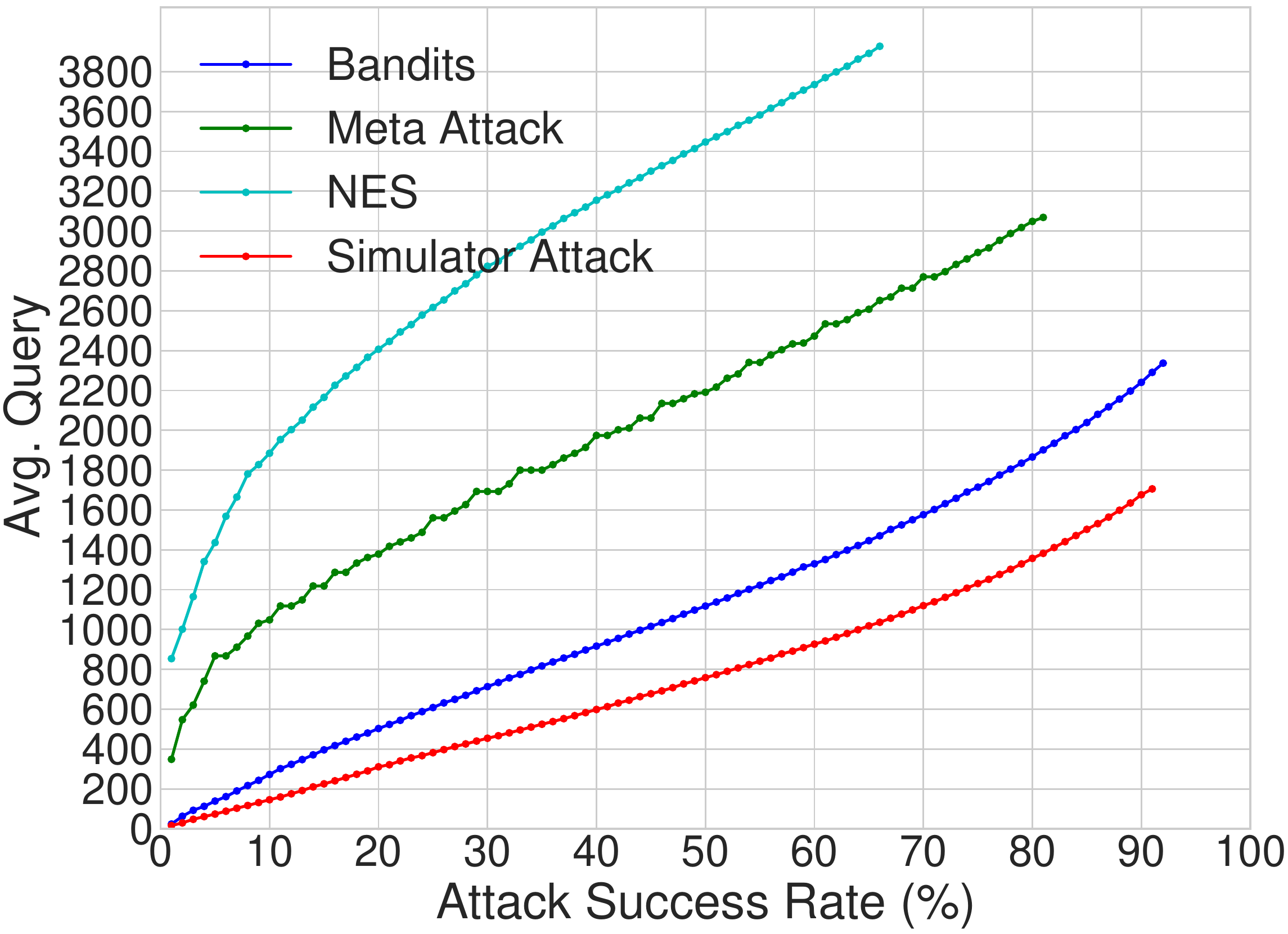}
		\subcaption{targeted $\ell_\infty$ attack GDAS}
	\end{minipage}
	\begin{minipage}[b]{.245\textwidth}
		\includegraphics[width=\linewidth]{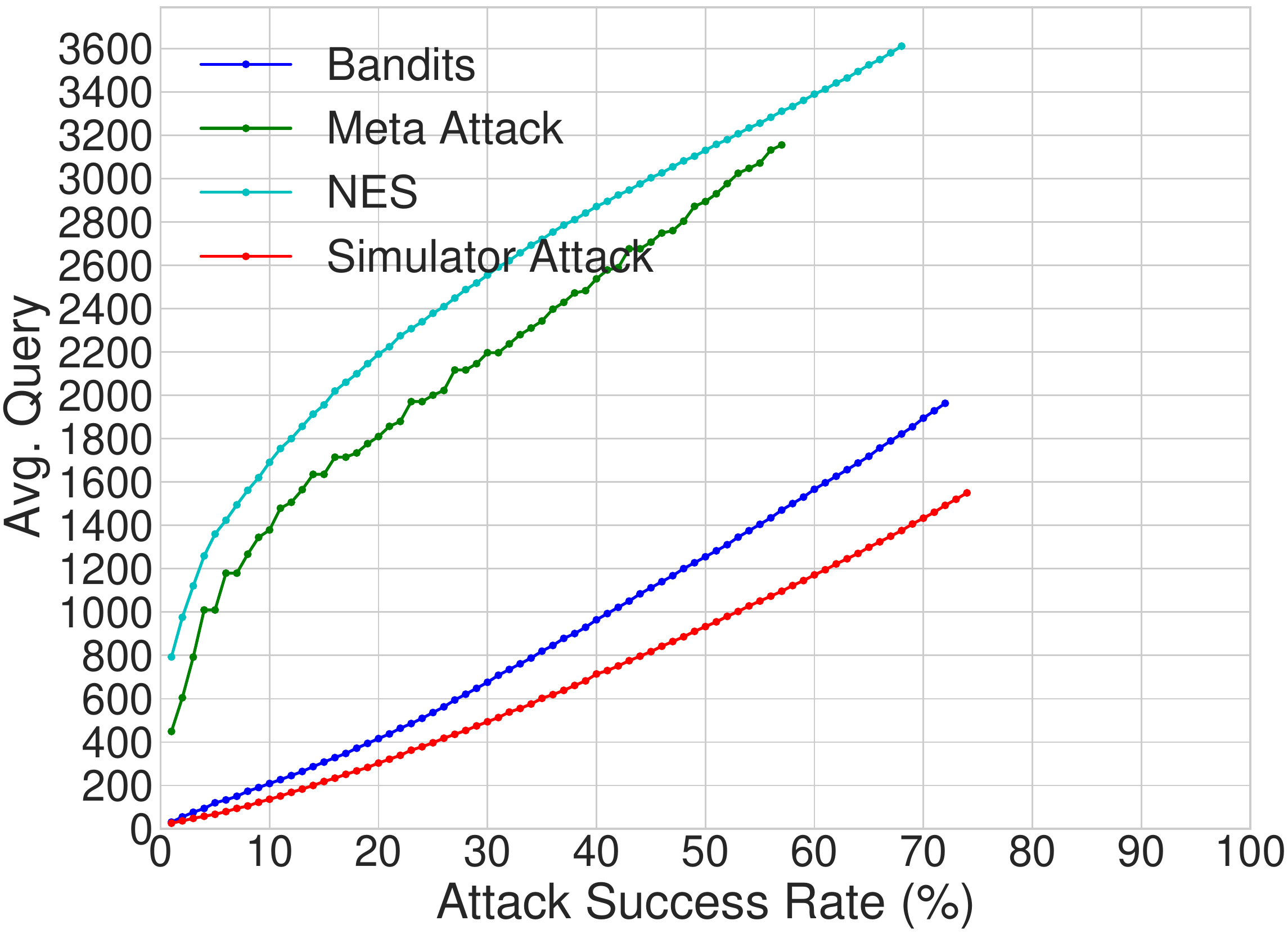}
		\subcaption{targeted $\ell_\infty$ attack WRN-28}
	\end{minipage}
	\begin{minipage}[b]{.245\textwidth}
		\includegraphics[width=\linewidth]{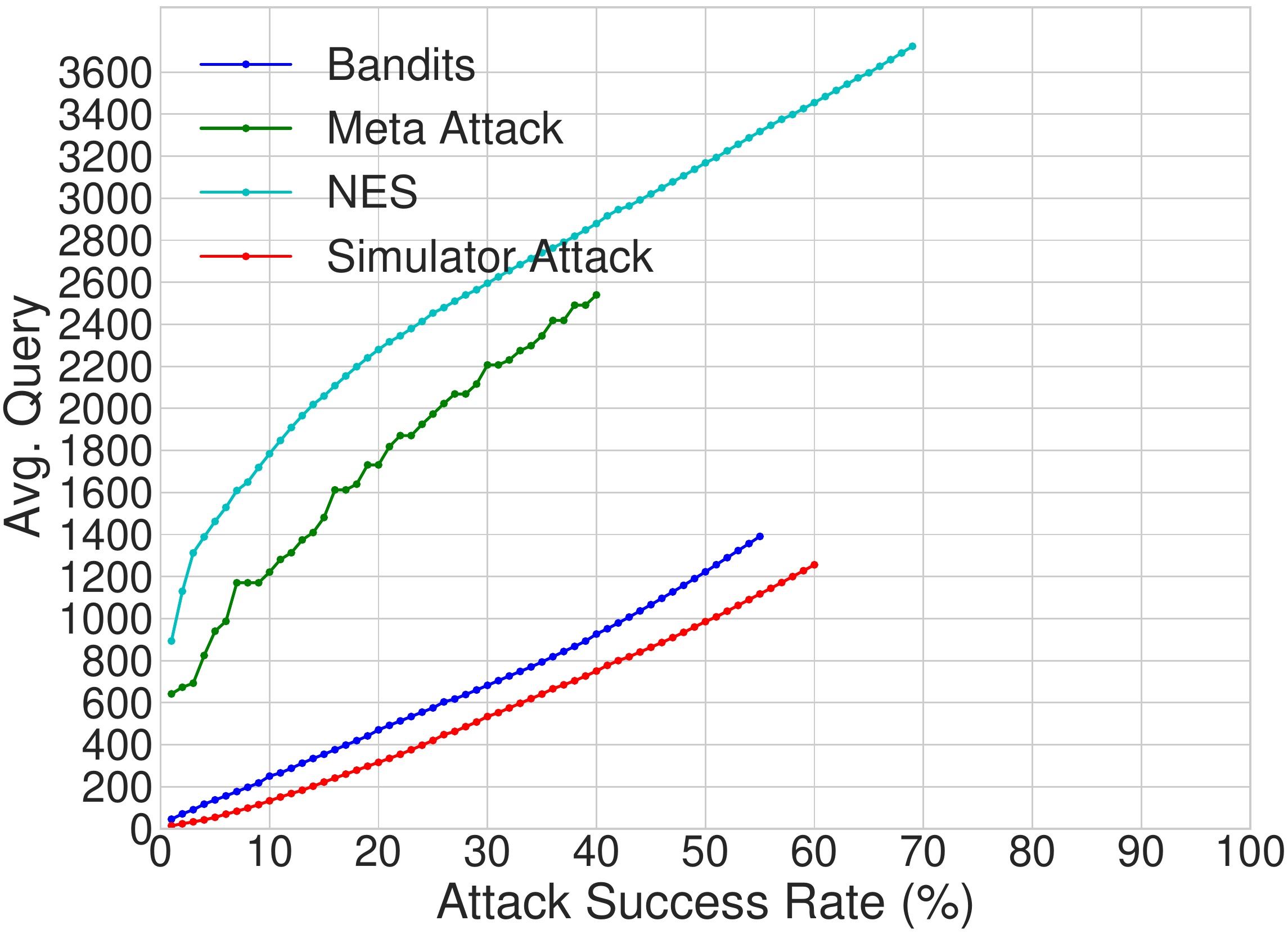}
		\subcaption{targeted $\ell_\infty$ attack WRN-40}
	\end{minipage}
	\caption{Comparisons of the average query per successful image at different desired success rates in CIFAR-100 dataset.}
	\label{fig:success_rate_to_avg_query_CIFAR-100}
\end{figure*}

\begin{figure*}[bp]
	\setlength{\abovecaptionskip}{0pt}
	\setlength{\belowcaptionskip}{0pt}
	\captionsetup[sub]{font={scriptsize}}
	\centering 
	\begin{minipage}[b]{.3\textwidth}
		\includegraphics[width=\linewidth]{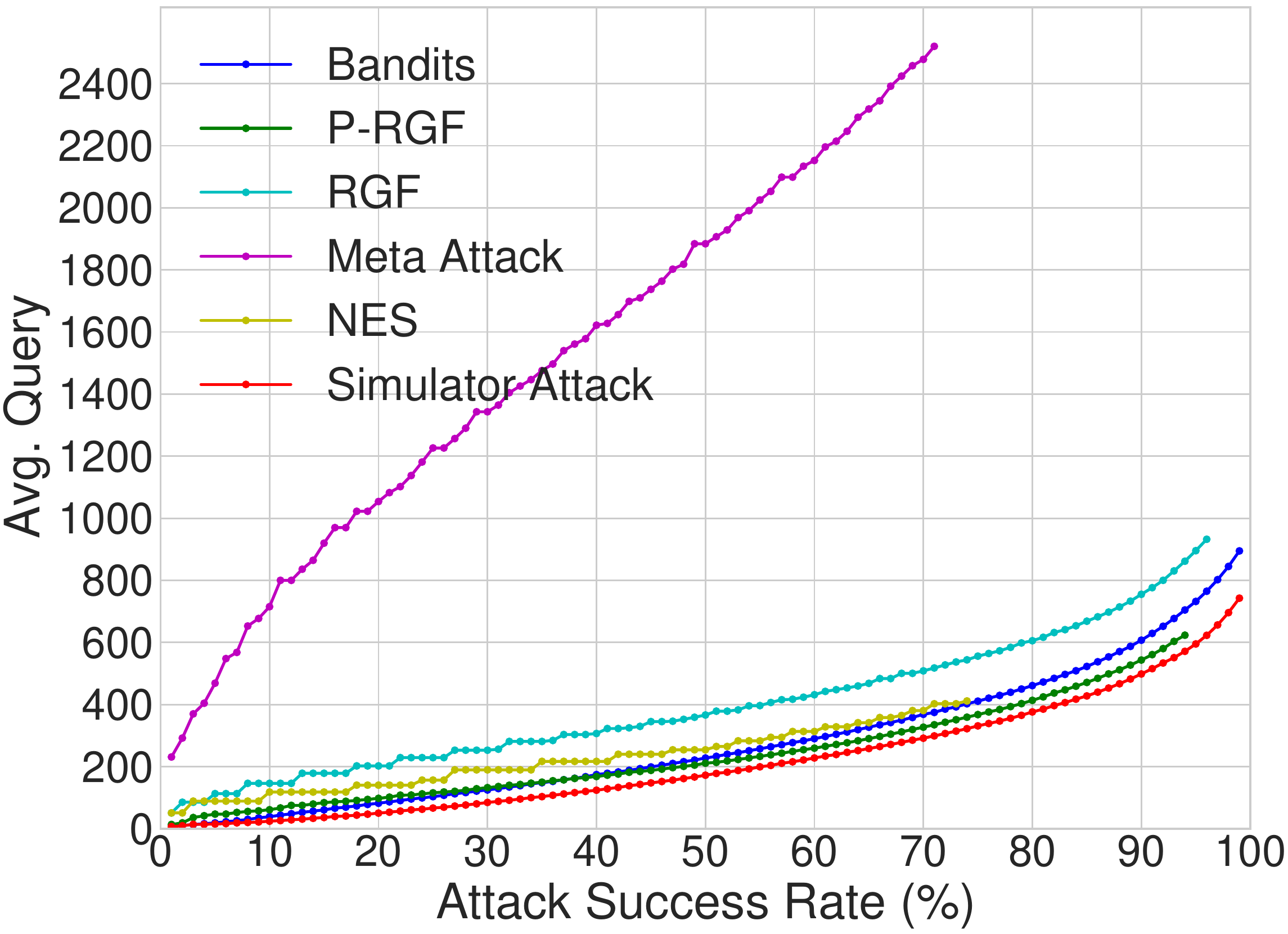}
		\subcaption{untargeted $\ell_\infty$ attack DenseNet-121}
	\end{minipage}
	\begin{minipage}[b]{.3\textwidth}
		\includegraphics[width=\linewidth]{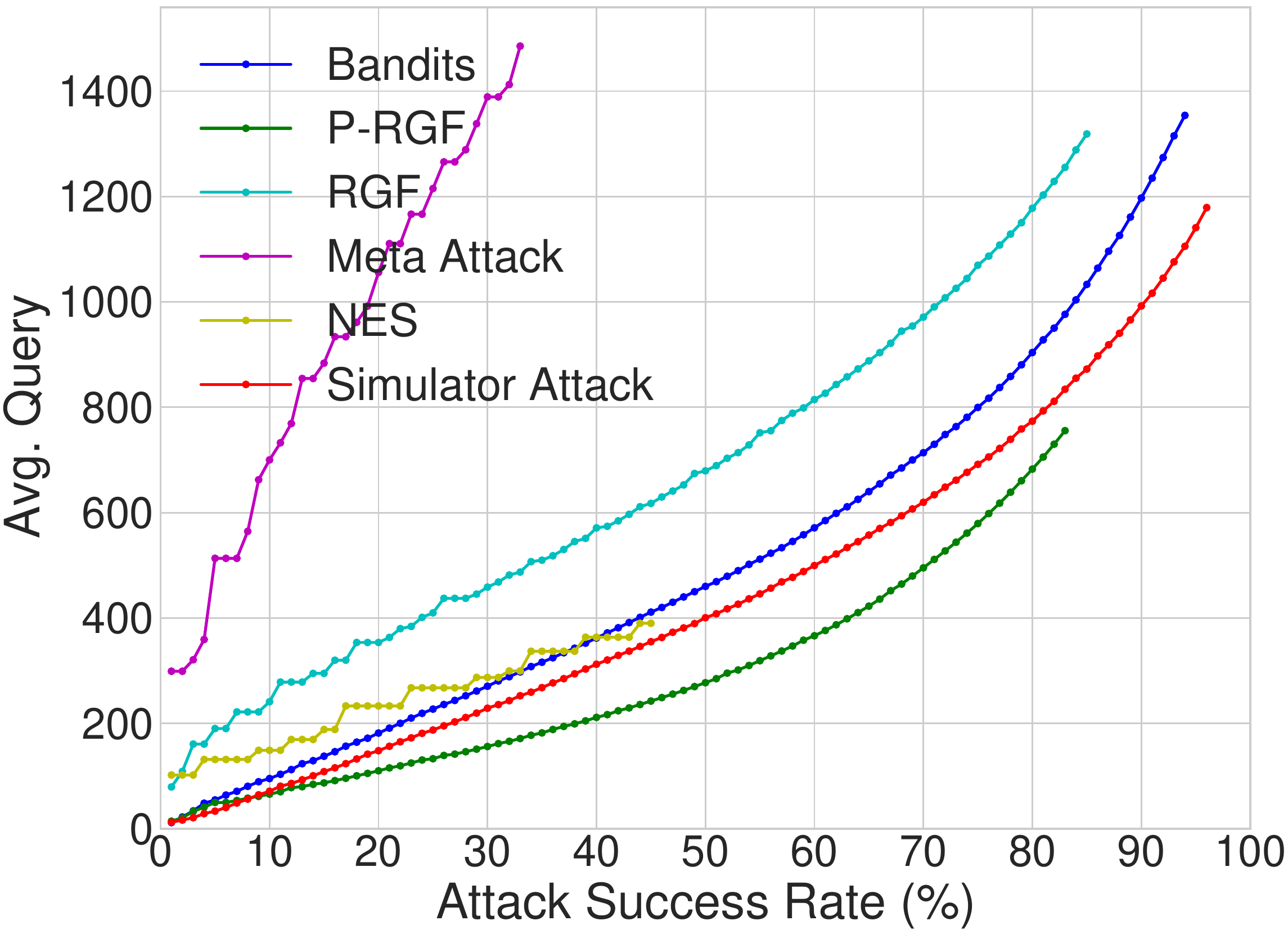}
		\subcaption{untargeted $\ell_\infty$ attack ResNeXt-101(32$\times$4d)}
	\end{minipage}
	\begin{minipage}[b]{.3\textwidth}
		\includegraphics[width=\linewidth]{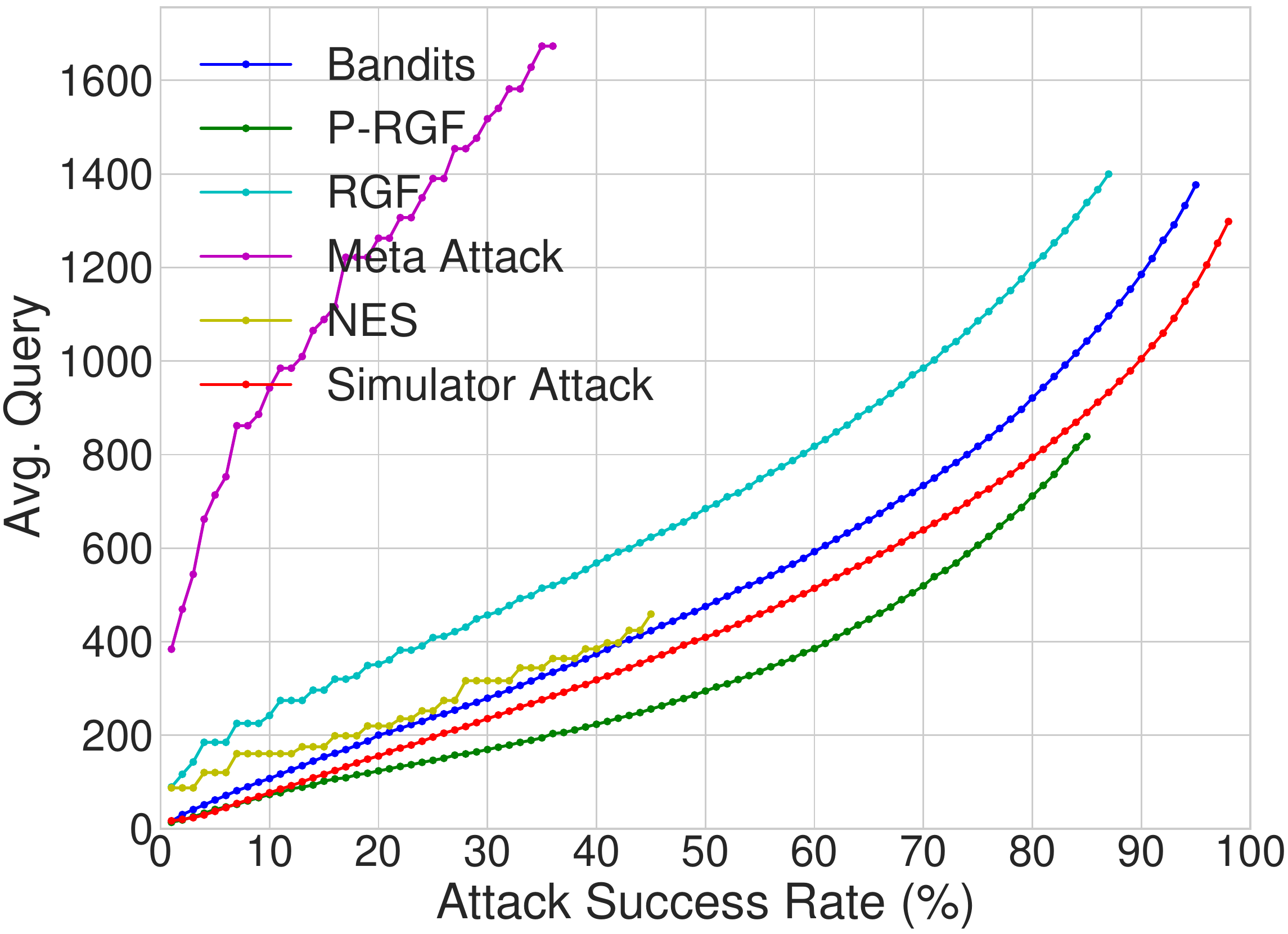}
		\subcaption{untargeted $\ell_\infty$ attack ResNeXt-101(64$\times$4d)}
	\end{minipage}
	\begin{minipage}[b]{.3\textwidth}
		\includegraphics[width=\linewidth]{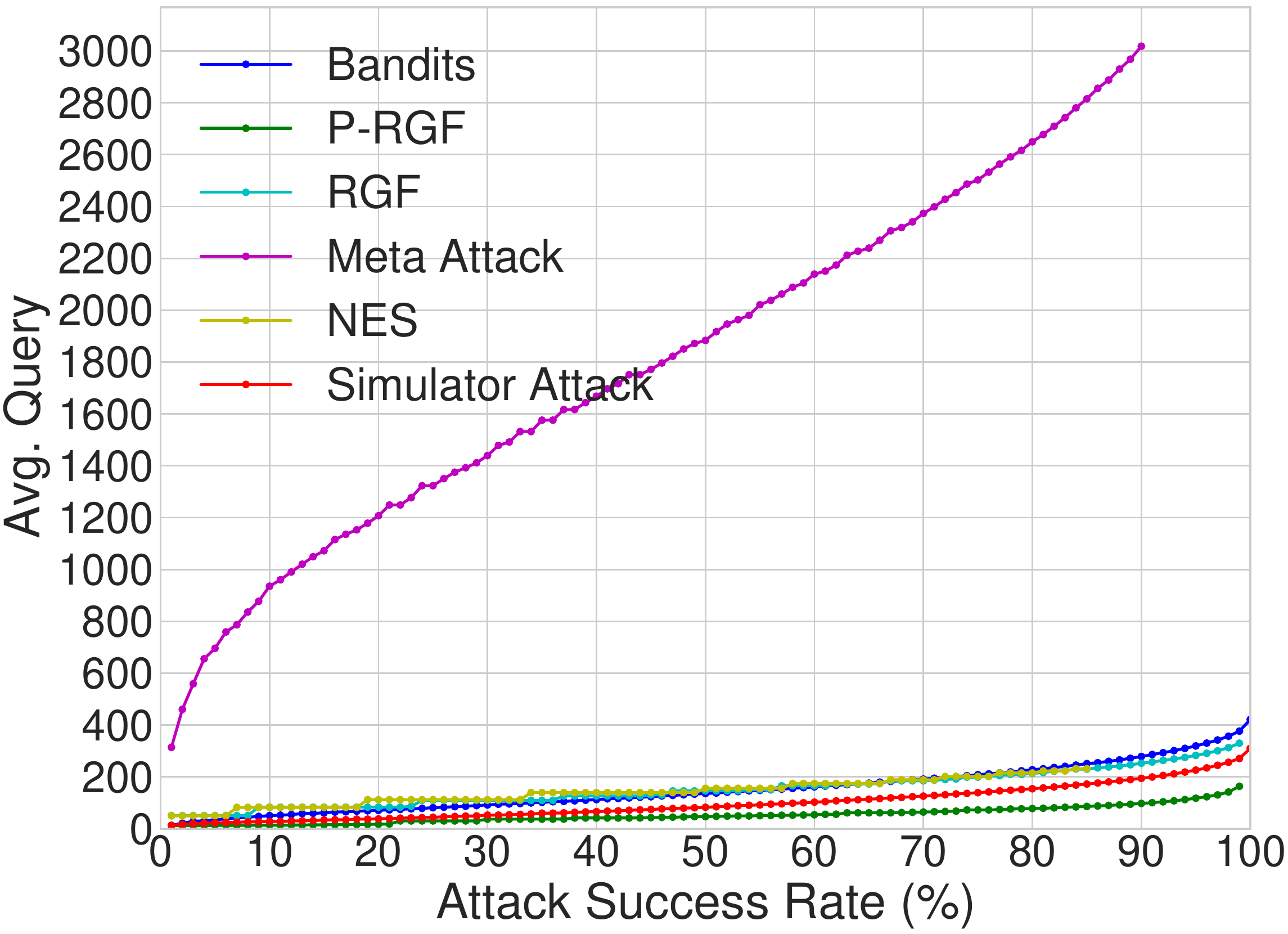}
		\subcaption{untargeted $\ell_2$ attack DenseNet-121}
	\end{minipage}
	\begin{minipage}[b]{.3\textwidth}
		\includegraphics[width=\linewidth]{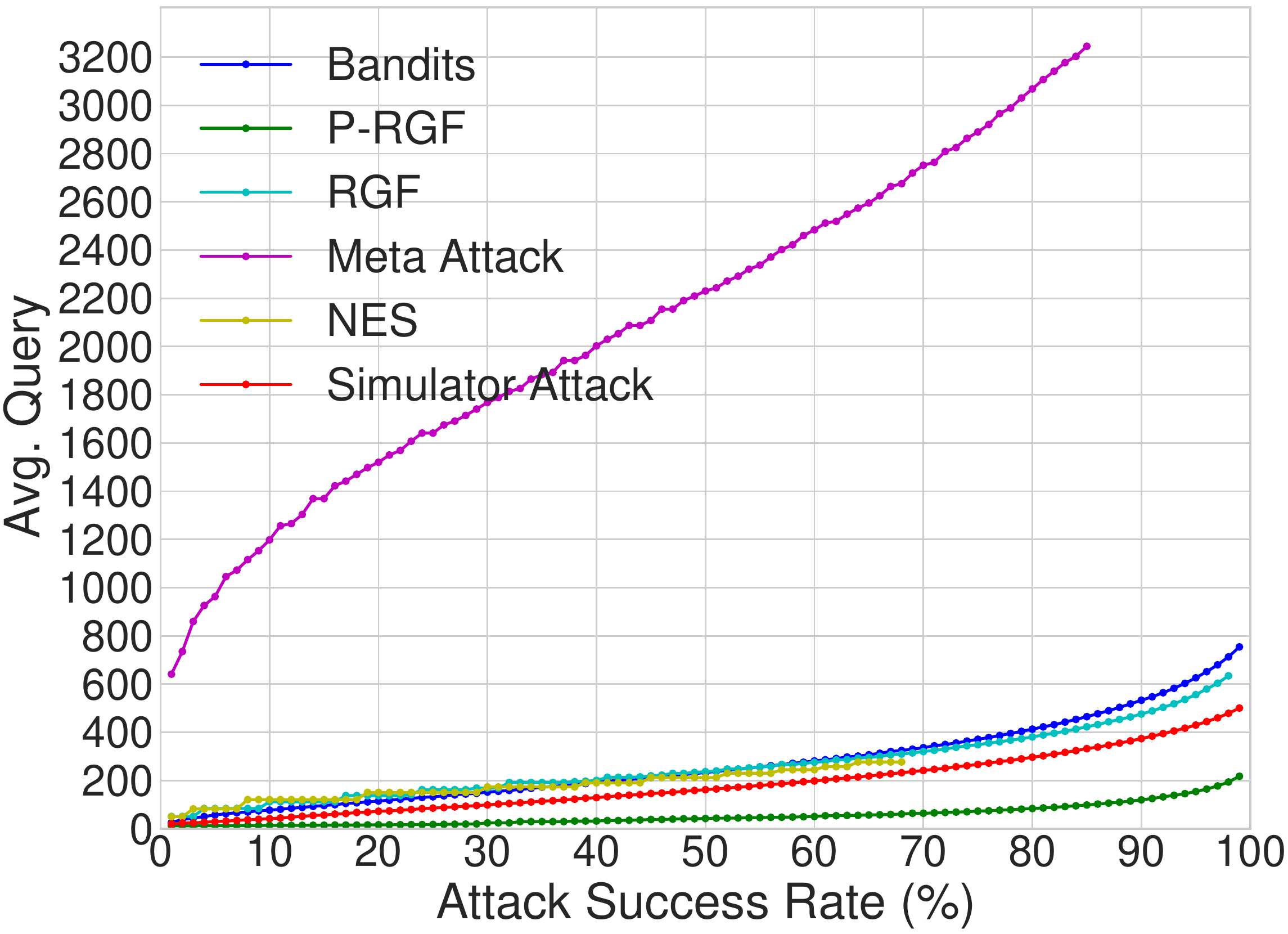}
		\subcaption{untargeted $\ell_2$ attack ResNeXt-101(32$\times$4d)}
	\end{minipage}
	\begin{minipage}[b]{.3\textwidth}
		\includegraphics[width=\linewidth]{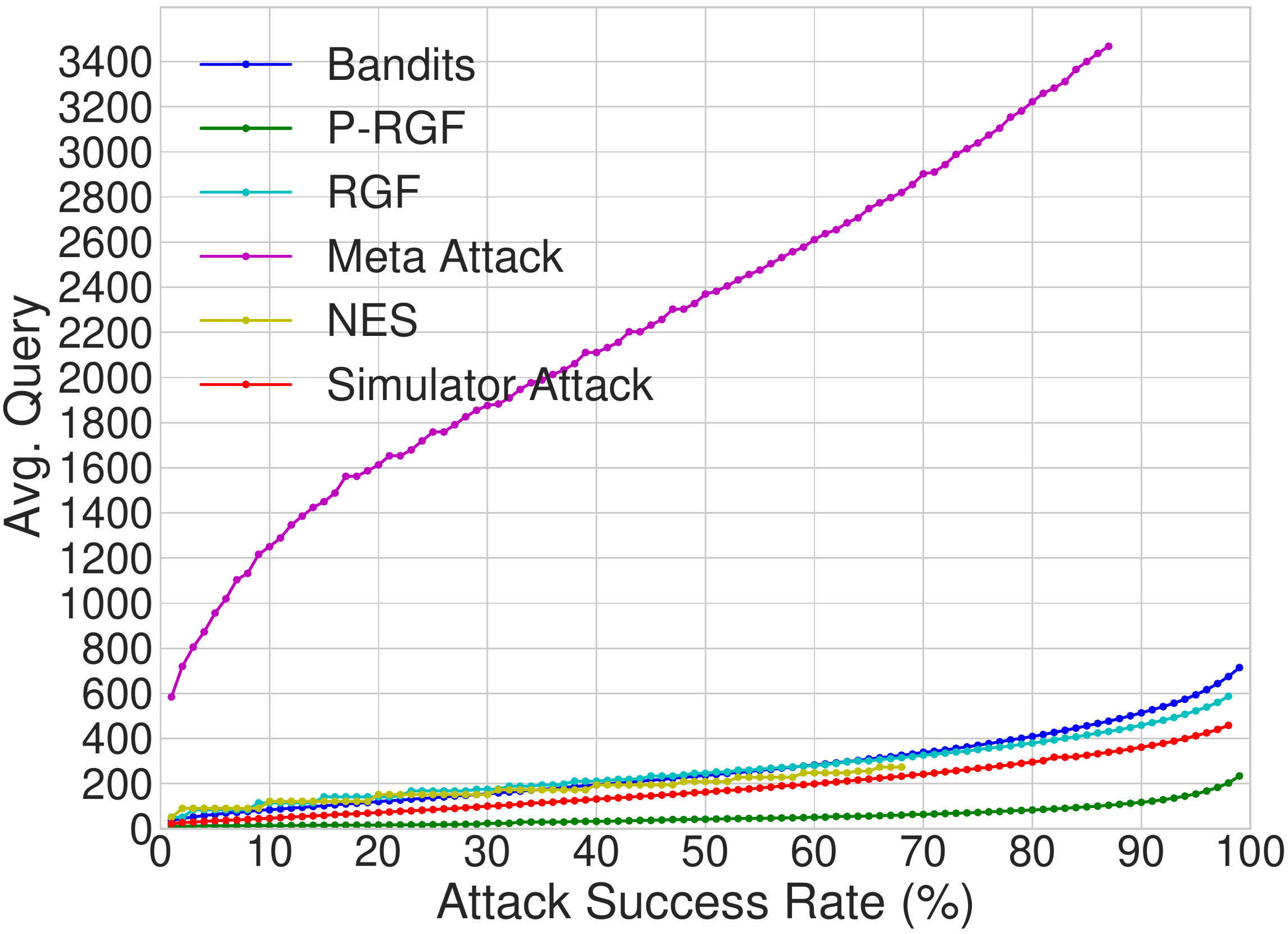}
		\subcaption{untargeted $\ell_2$ attack ResNeXt-101(64$\times$4d)}
	\end{minipage}
	\begin{minipage}[b]{.3\textwidth}
		\includegraphics[width=\linewidth]{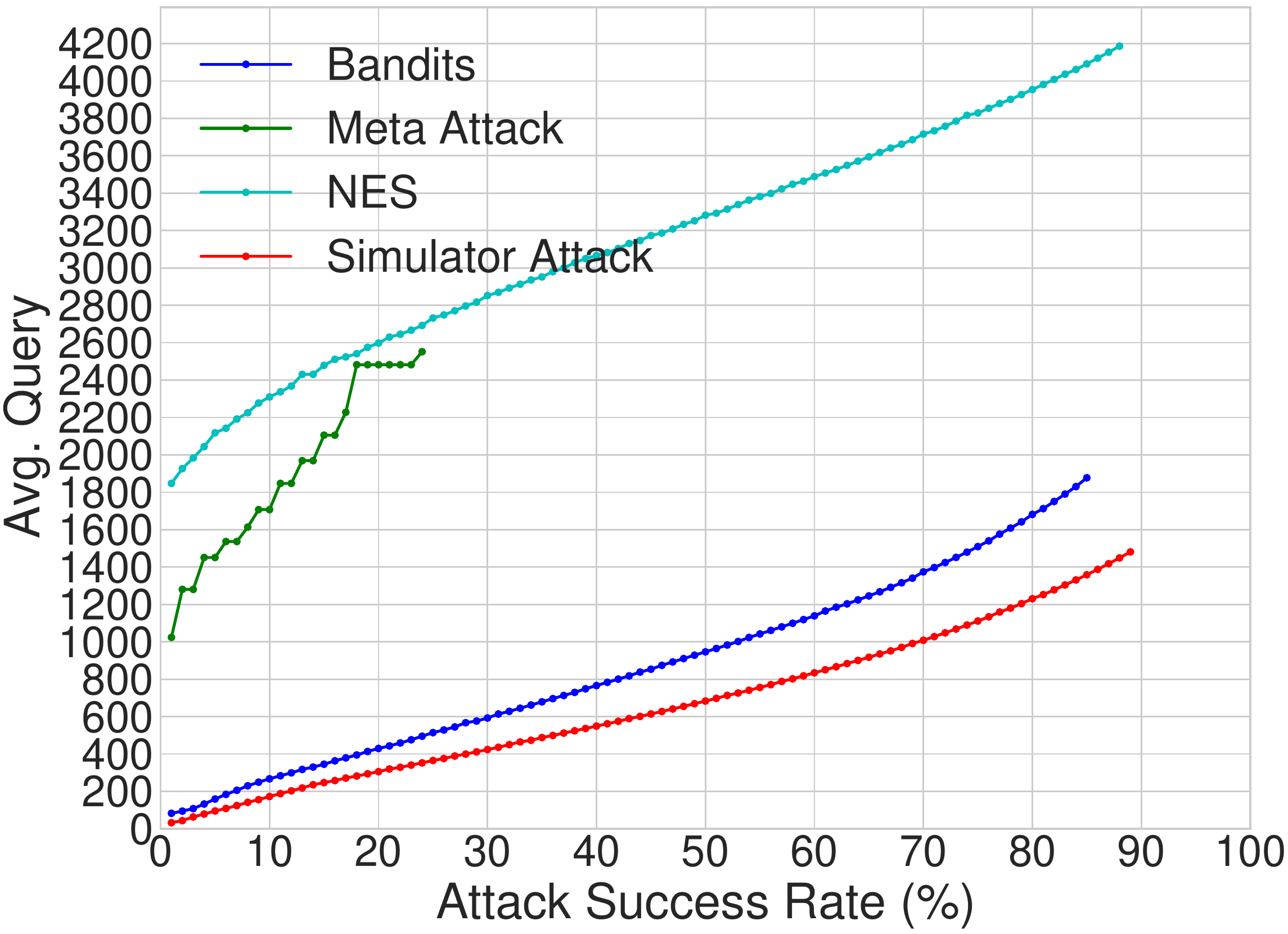}
		\subcaption{targeted $\ell_2$ attack DenseNet-121}
	\end{minipage}
	\begin{minipage}[b]{.3\textwidth}
		\includegraphics[width=\linewidth]{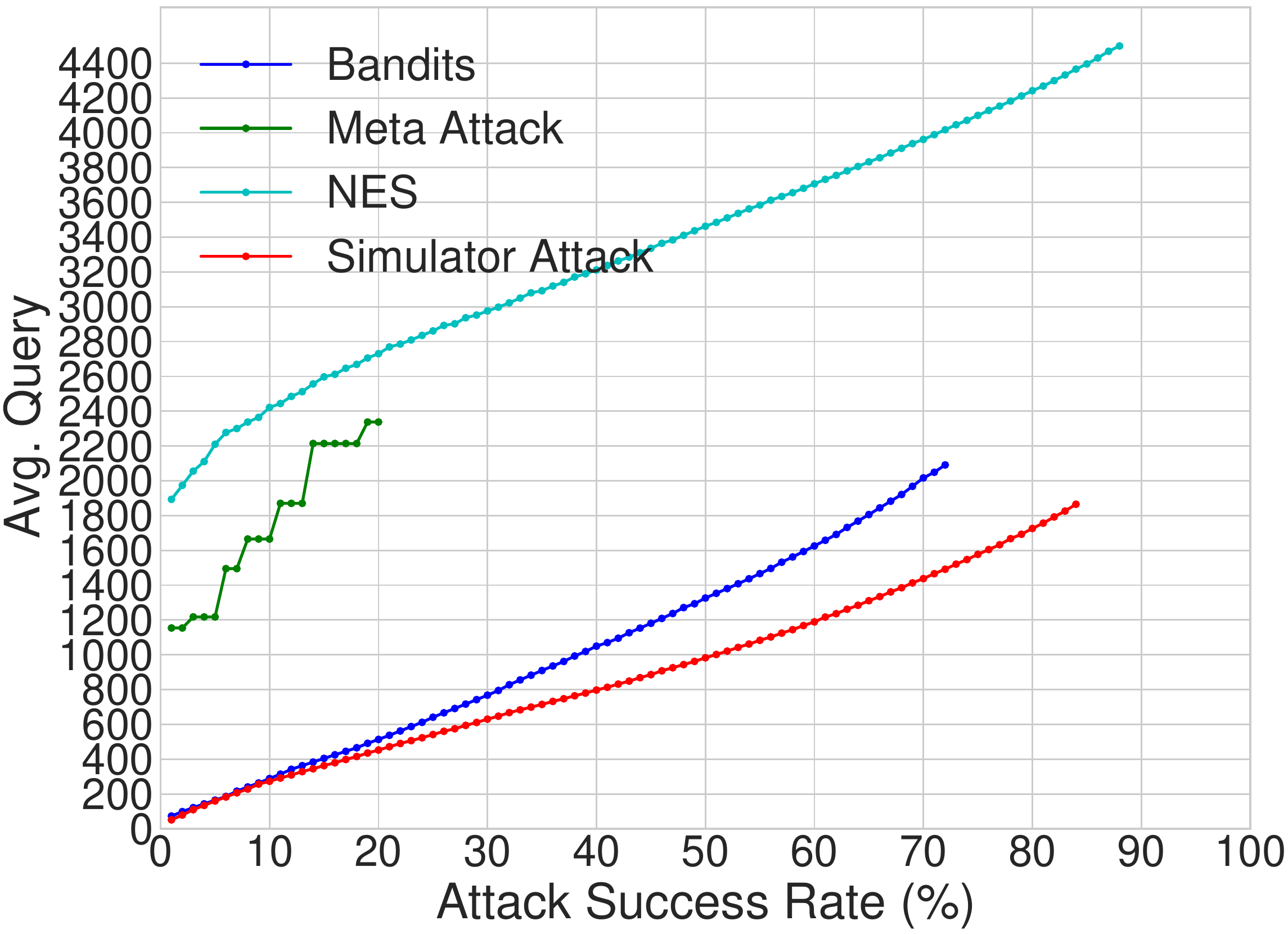}
		\subcaption{targeted $\ell_2$ attack ResNeXt-101(32$\times$4d)}
	\end{minipage}
	\begin{minipage}[b]{.3\textwidth}
		\includegraphics[width=\linewidth]{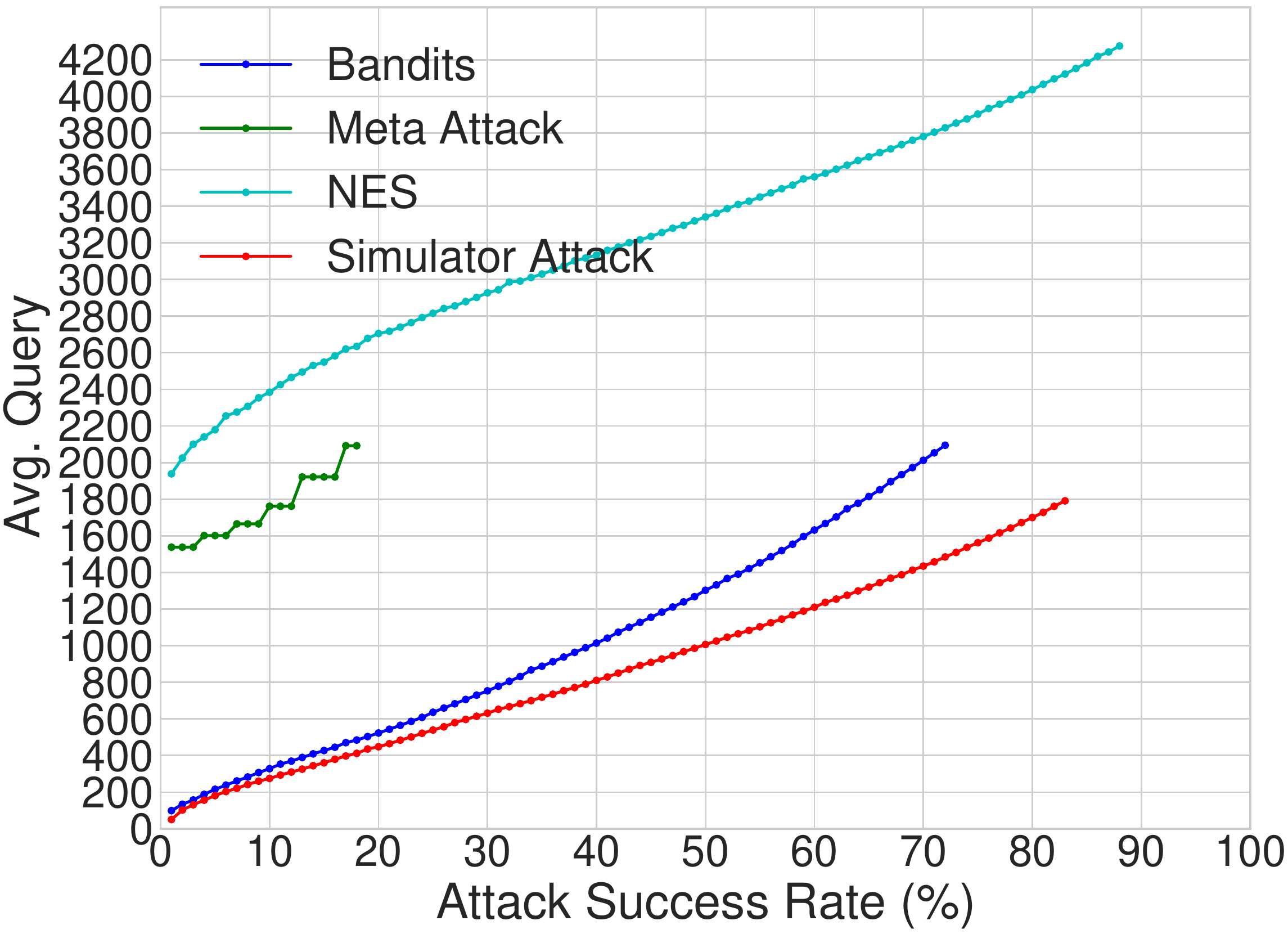}
		\subcaption{targeted $\ell_2$ attack ResNeXt-101(64$\times$4d)}
	\end{minipage}
	\caption{Comparisons of the average query per successful image at different desired success rates in TinyImageNet dataset.}
	\label{fig:success_rate_to_avg_query_TinyImageNet}
\end{figure*}

\begin{figure*}[htbp]
	\setlength{\abovecaptionskip}{0pt}
	\setlength{\belowcaptionskip}{0pt}
	\captionsetup[sub]{font={scriptsize}}
	\centering 
	\begin{minipage}[b]{.3\textwidth}
		\includegraphics[width=\linewidth]{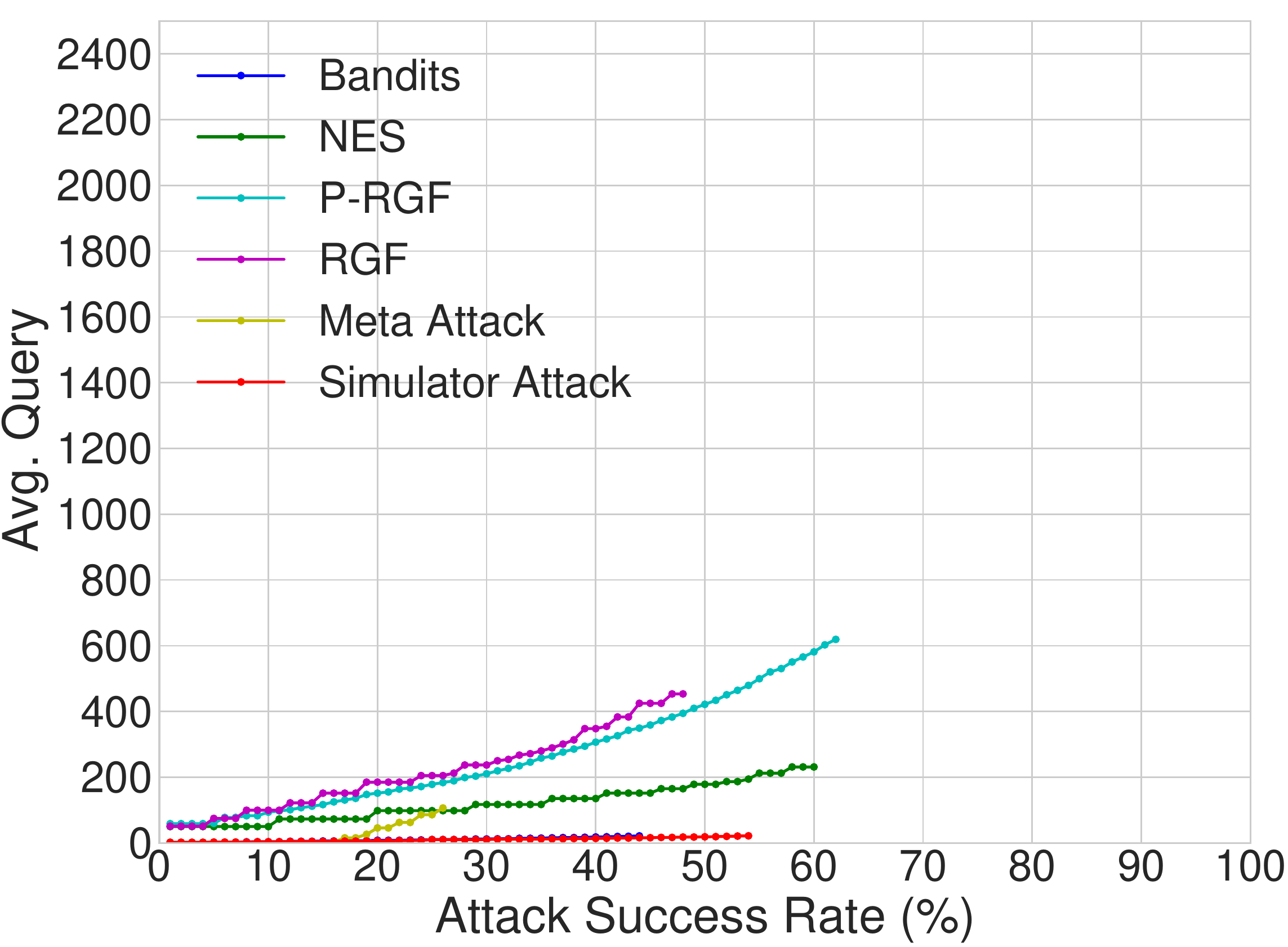}
		\subcaption{attack ComDefend in CIFAR-10}
	\end{minipage}
	\begin{minipage}[b]{.3\textwidth}
		\includegraphics[width=\linewidth]{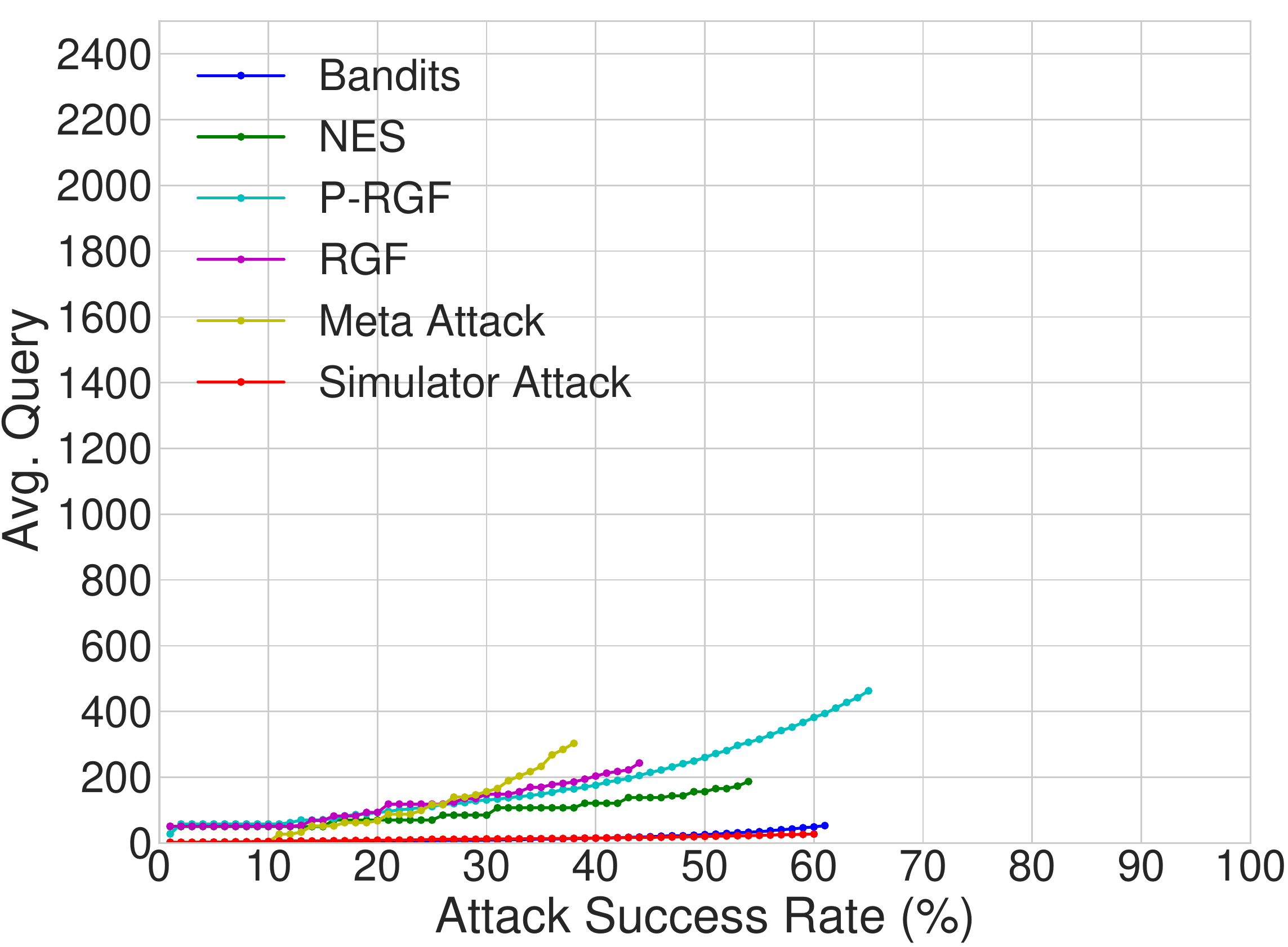}
		\subcaption{attack Feature Distillation in CIFAR-10}
	\end{minipage}
	\begin{minipage}[b]{.3\textwidth}
		\includegraphics[width=\linewidth]{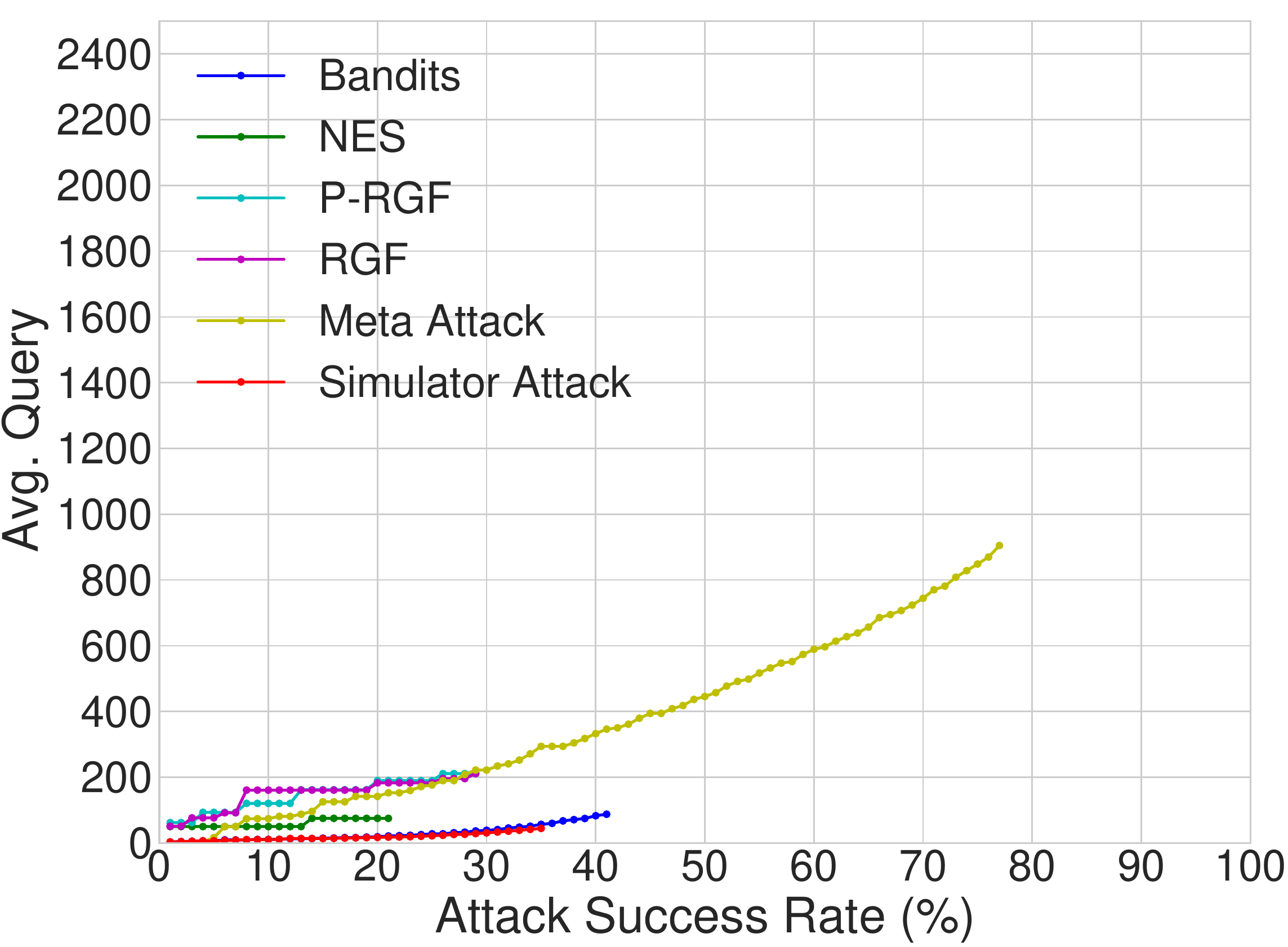}
		\subcaption{attack PCL in CIFAR-10}
	\end{minipage}
	\begin{minipage}[b]{.3\textwidth}
		\includegraphics[width=\linewidth]{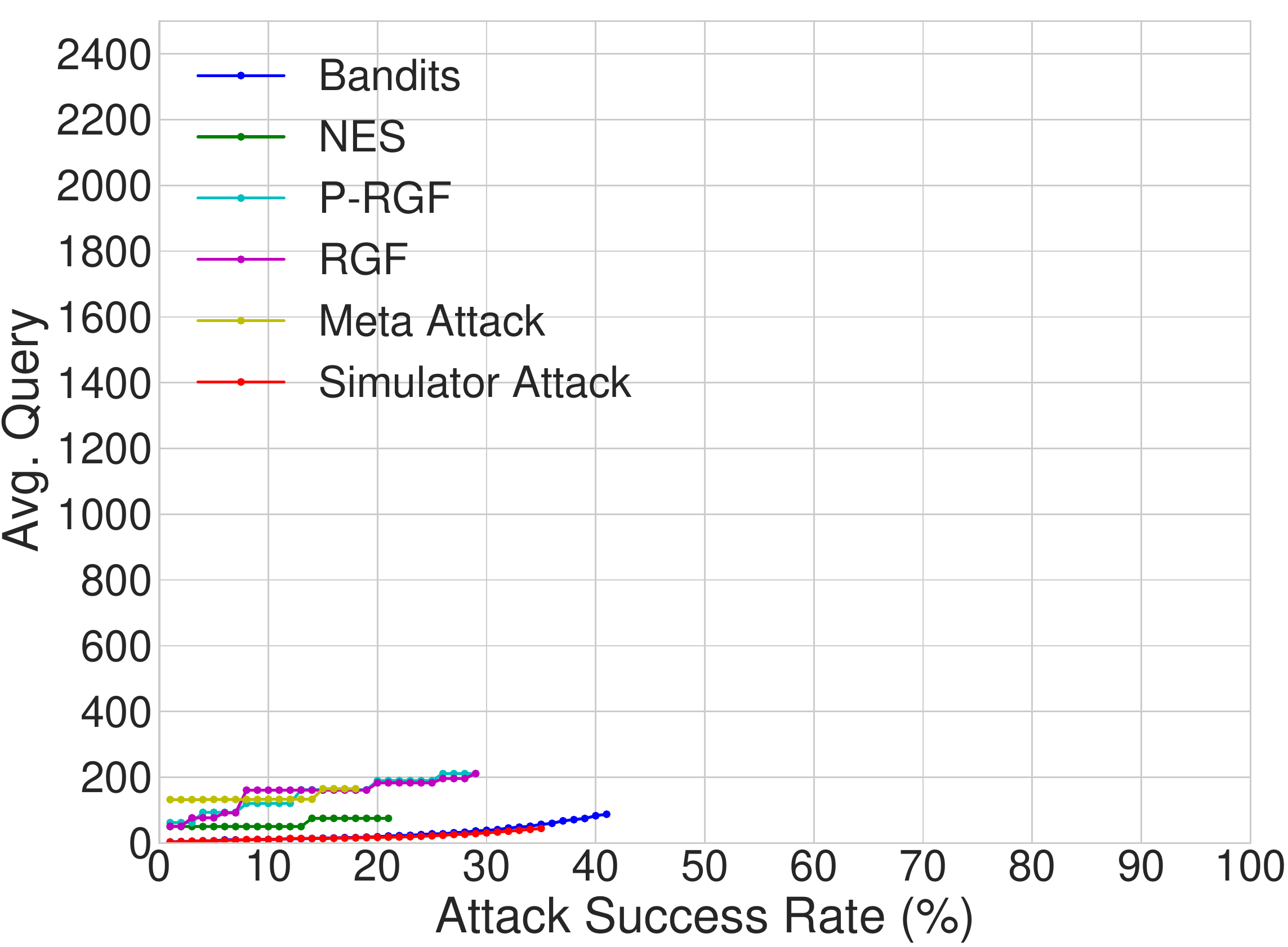}
		\subcaption{attack Adv Train in CIFAR-10}
	\end{minipage}
	\begin{minipage}[b]{.3\textwidth}
		\includegraphics[width=\linewidth]{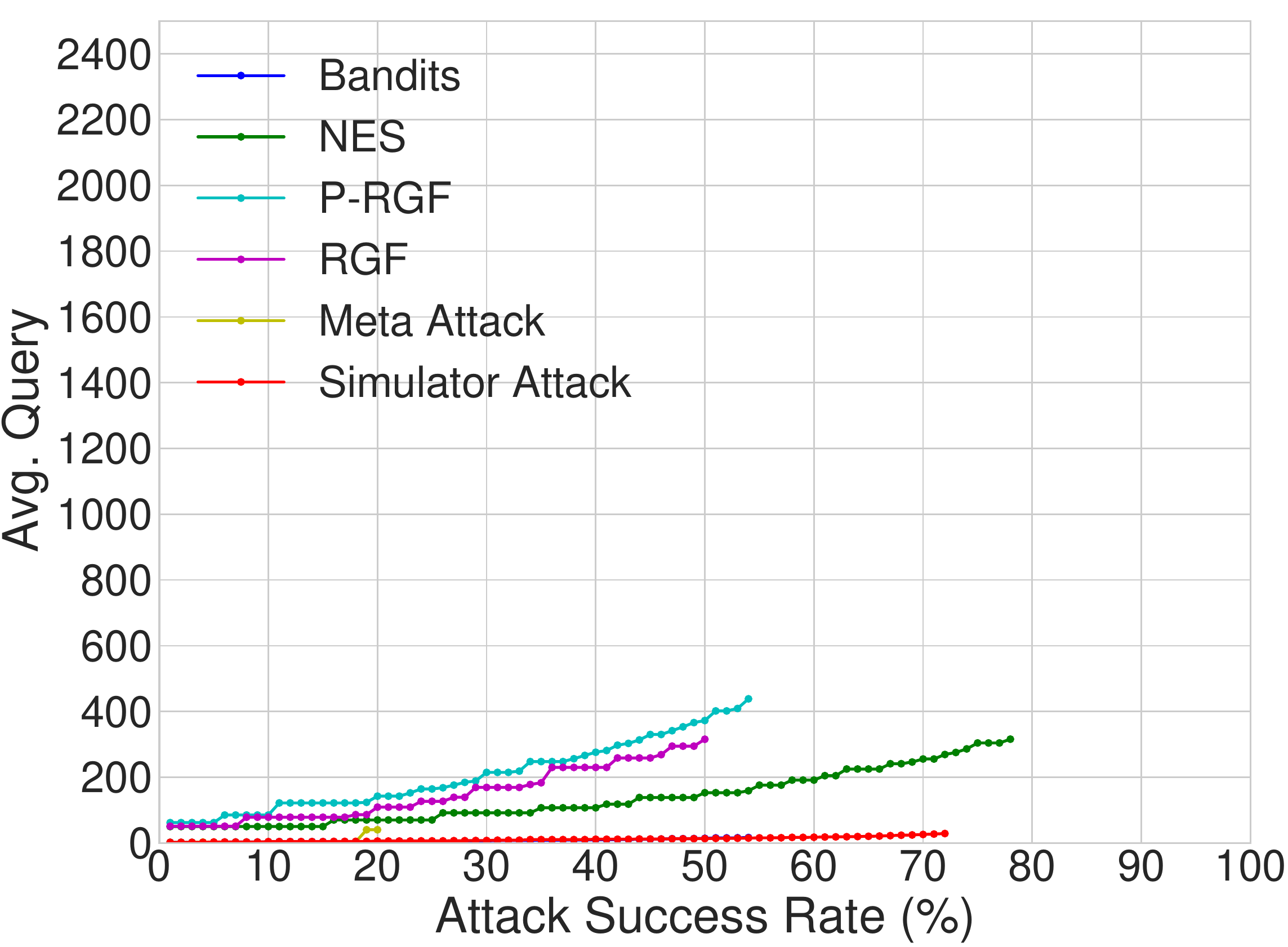}
		\subcaption{attack ComDefend in CIFAR-100}
	\end{minipage}
	\begin{minipage}[b]{.3\textwidth}
		\includegraphics[width=\linewidth]{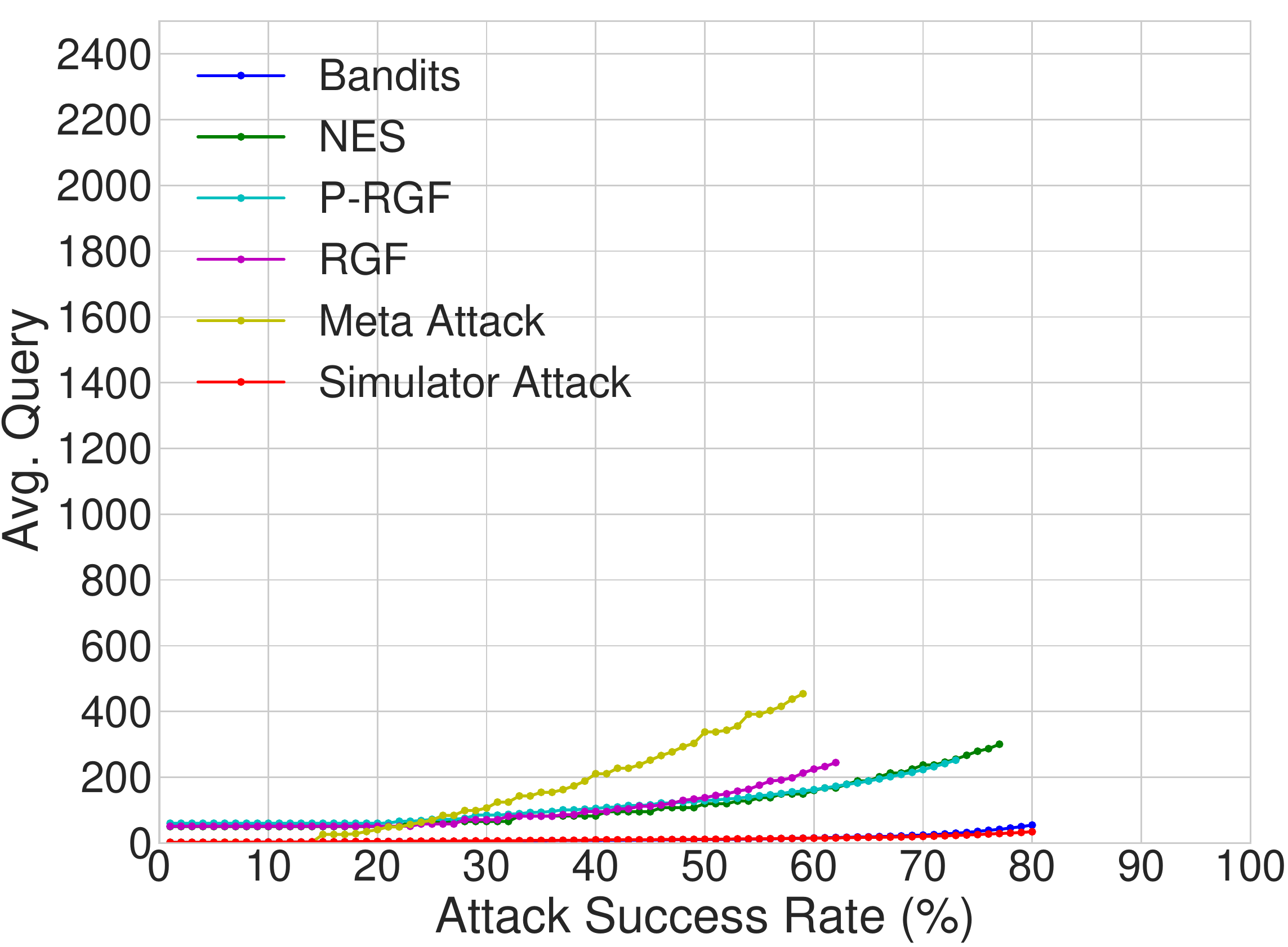}
		\subcaption{attack Feature Distillation in CIFAR-100}
	\end{minipage}
	\begin{minipage}[b]{.3\textwidth}
		\includegraphics[width=\linewidth]{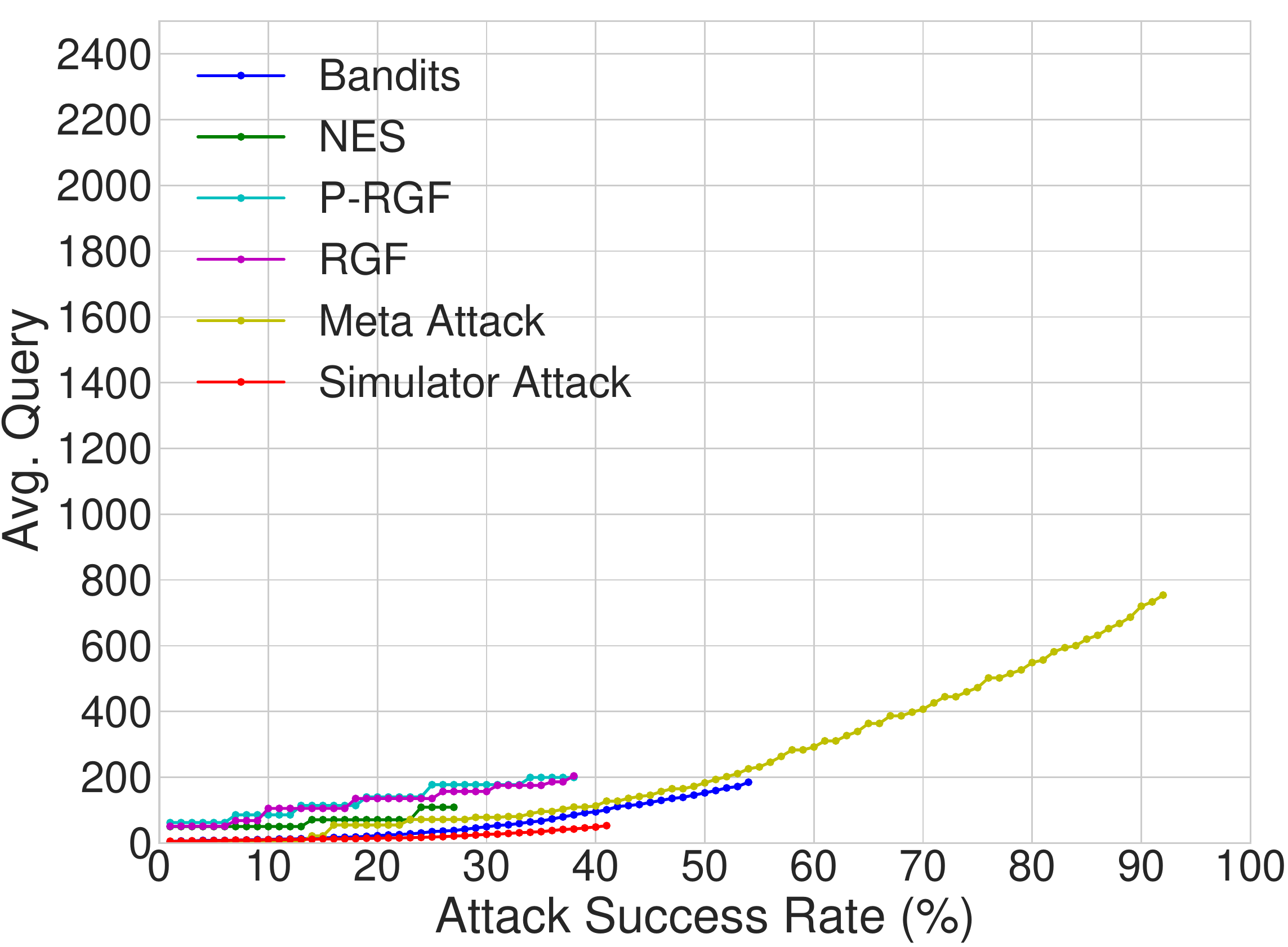}
		\subcaption{attack PCL in CIFAR-100}
	\end{minipage}
	\begin{minipage}[b]{.3\textwidth}
		\includegraphics[width=\linewidth]{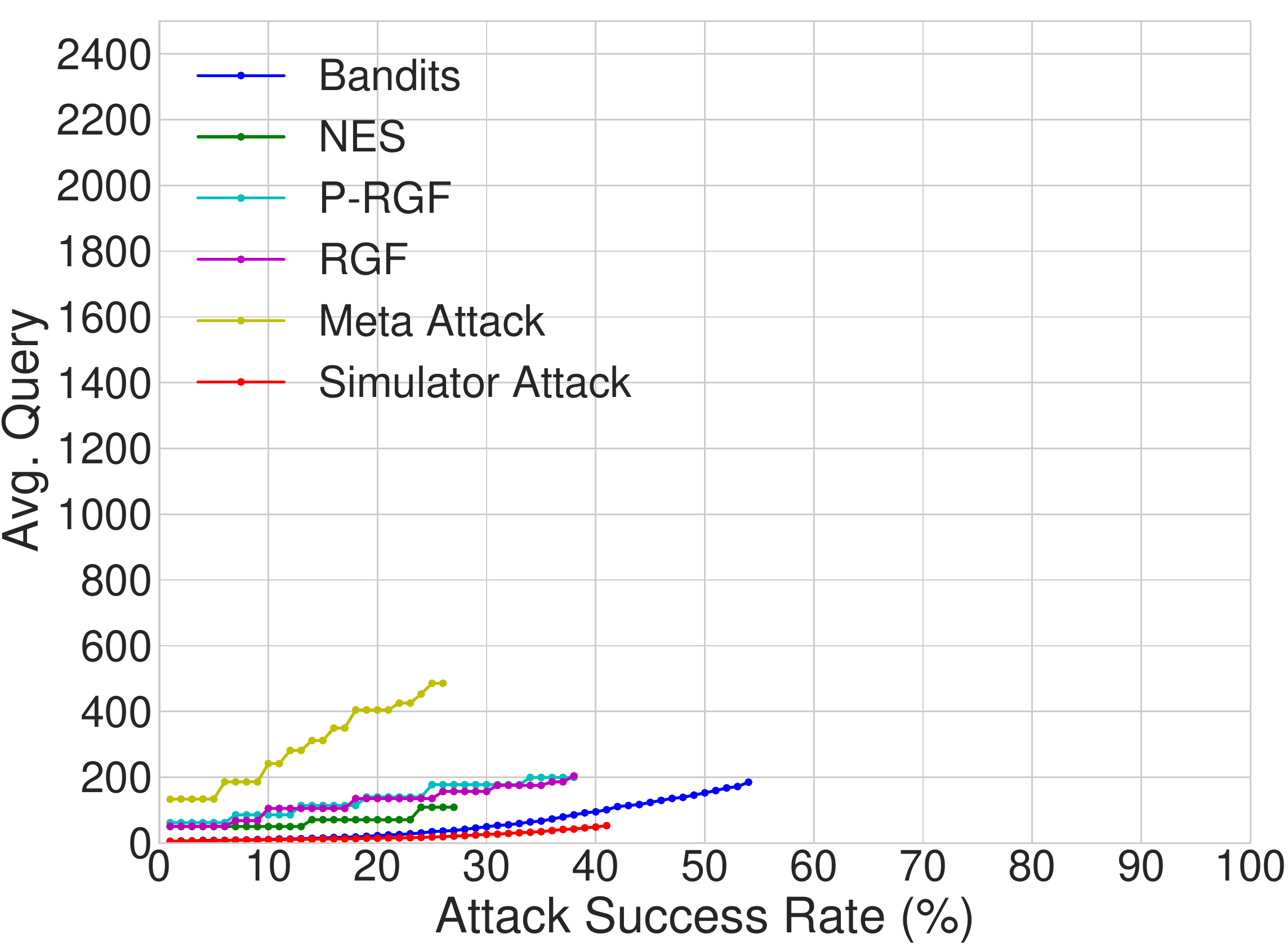}
		\subcaption{attack Adv Train in CIFAR-100}
	\end{minipage}
	\begin{minipage}[b]{.3\textwidth}
		\includegraphics[width=\linewidth]{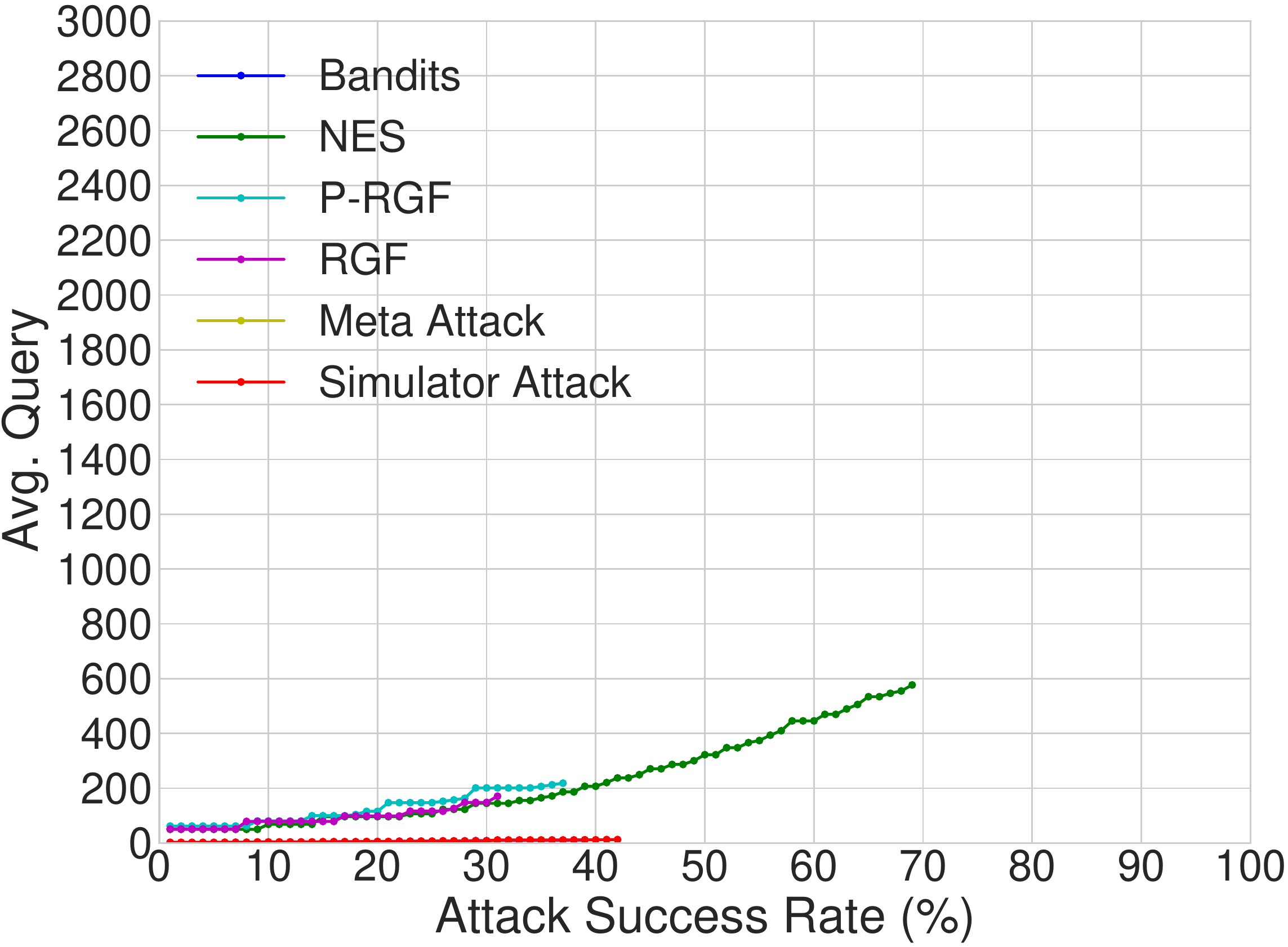}
		\subcaption{attack ComDefend in TinyImageNet}
	\end{minipage}
	\begin{minipage}[b]{.3\textwidth}
		\includegraphics[width=\linewidth]{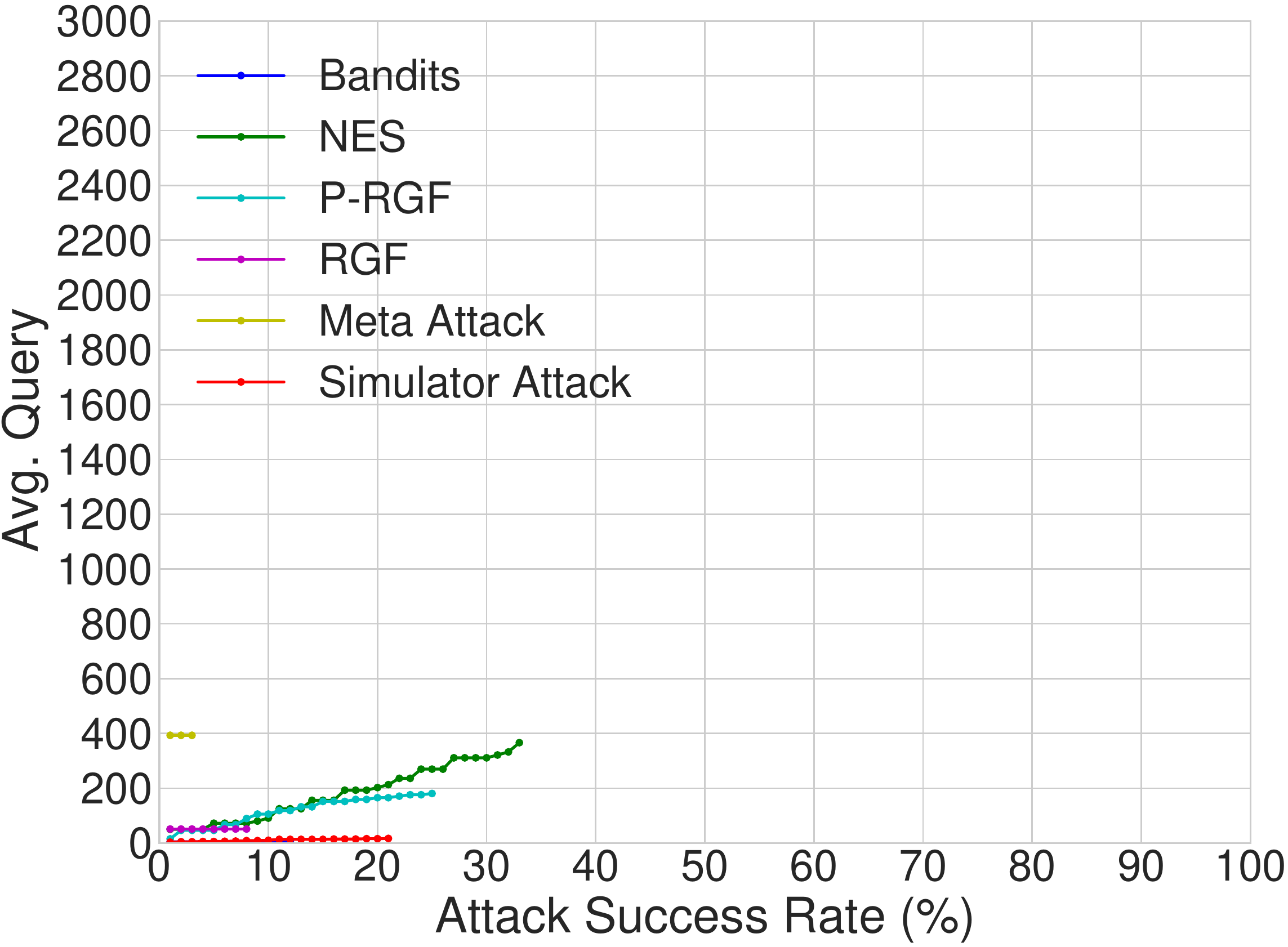}
		\subcaption{attack Feature Distillation in TinyImageNet}
	\end{minipage}
	\begin{minipage}[b]{.3\textwidth}
		\includegraphics[width=\linewidth]{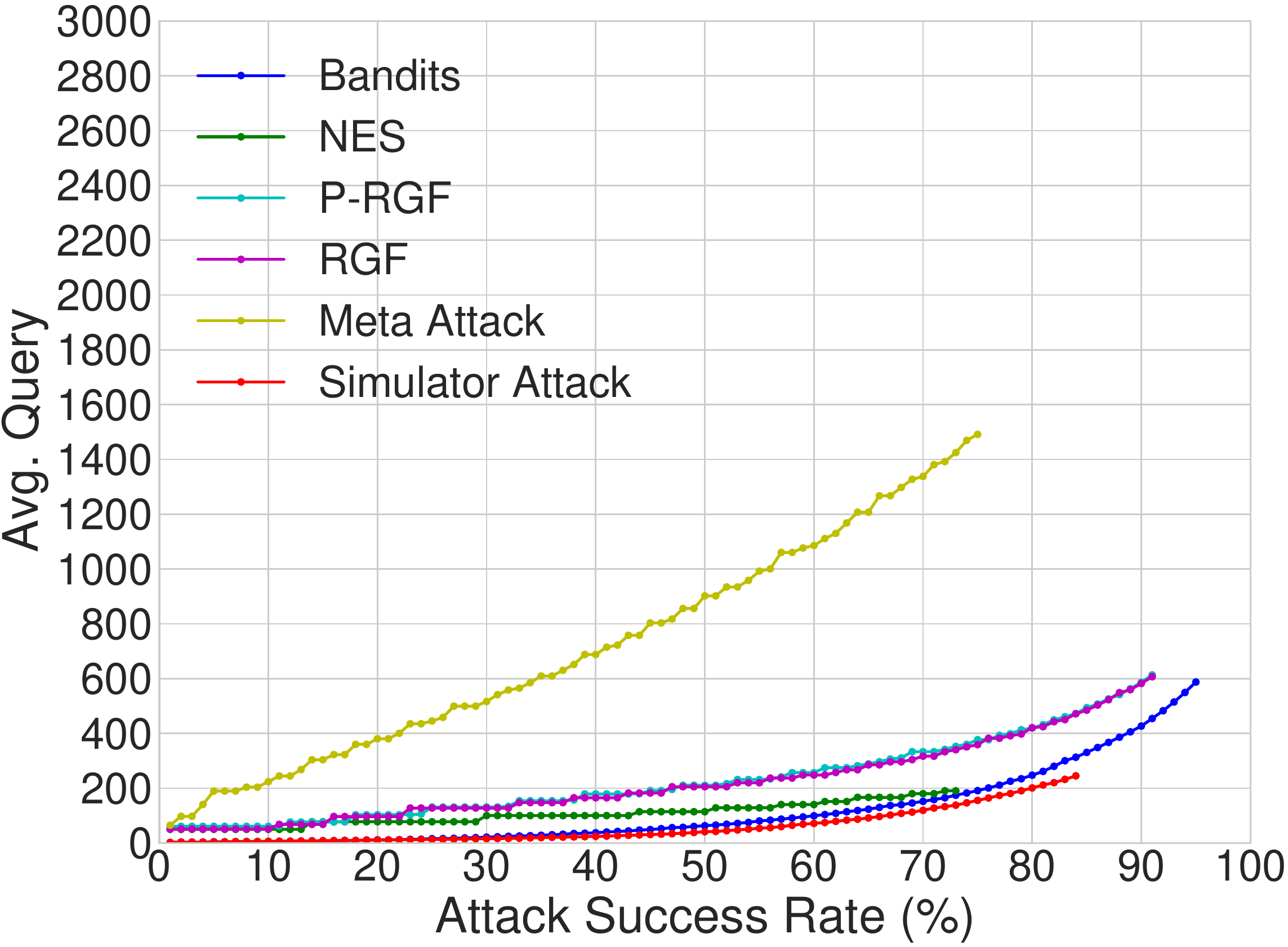}
		\subcaption{attack PCL in TinyImageNet}
	\end{minipage}
	\begin{minipage}[b]{.3\textwidth}
		\includegraphics[width=\linewidth]{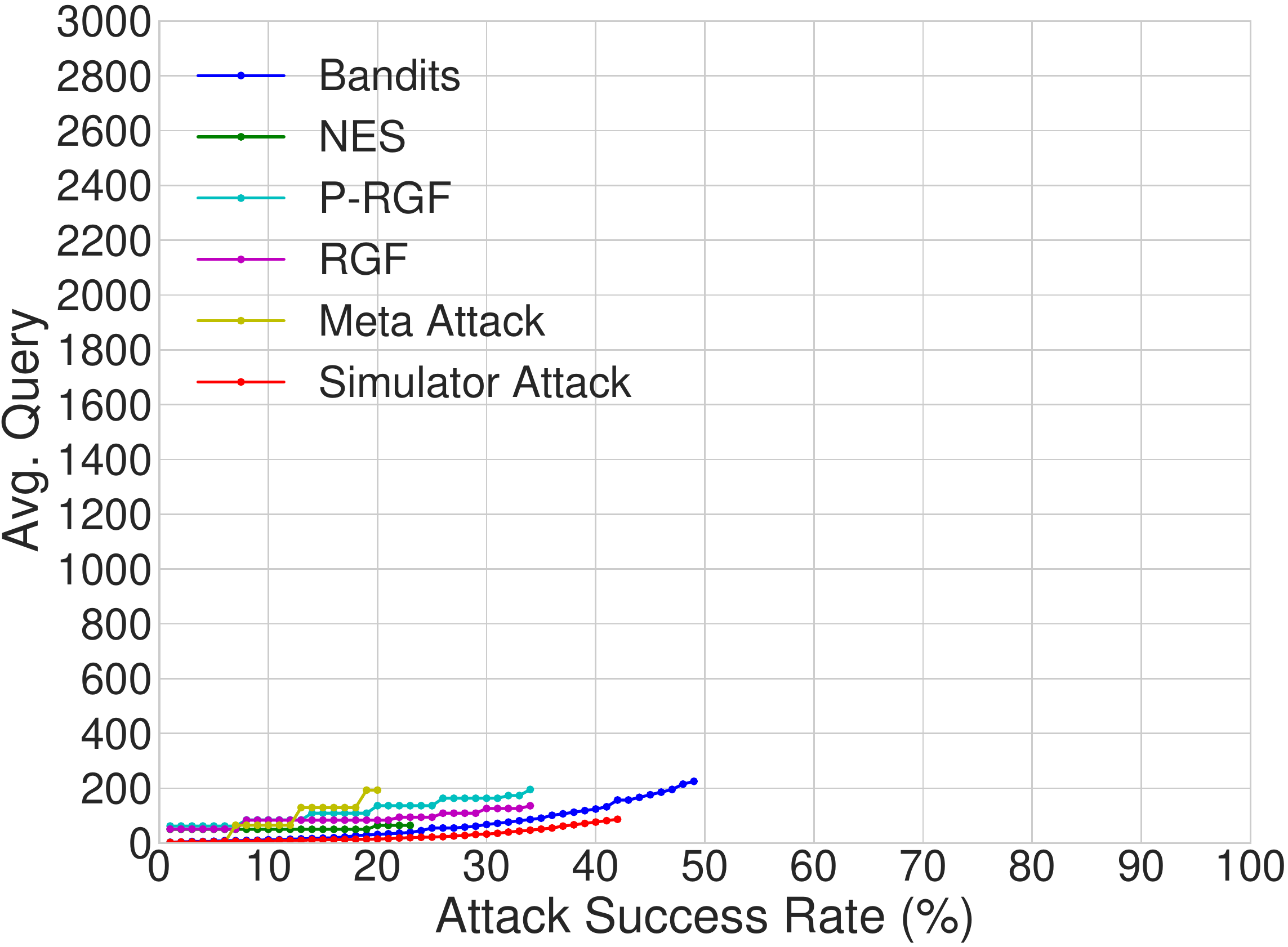}
		\subcaption{attack Adv Train in TinyImageNet}
	\end{minipage}
	\caption{Comparisons of the average query per successful image at different desired success rates on defensive models with the backbone of ResNet-50. The experimental results are obtained by performing the untargeted attacks under $\ell_\infty$ norm.}
	\label{fig:success_rate_to_avg_query_on_defensive_model}
	\vspace{-0.3cm}
\end{figure*}

\noindent\textbf{Average Queries at Different Success Rates.}  The second type of figure measures the average number of queries that reaches different desired success rates. It demonstrates the relation between the query number and attack success rate from a different angle. Specifically, given a desired success rate $a$ and the query list $Q$ of all successful attacked samples, the average query (Avg. $Q_a$) is defined as follows:
\begin{equation}
\text{Avg. }Q_a = \frac{\sum_{i=1}^{N}\hat{Q_i}}{N}, \qquad\text{where } \hat{Q} = Q[Q\leq P_a],
\end{equation}
where $P_a$ is the $a$-th percentile value of $Q$ and $N$ is the length of $\hat{Q}$.
Figs.~\ref{fig:success_rate_to_avg_query_CIFAR-10}, \ref{fig:success_rate_to_avg_query_CIFAR-100}, \ref{fig:success_rate_to_avg_query_TinyImageNet}, and \ref{fig:success_rate_to_avg_query_on_defensive_model} show the results. All the experimental results demonstrate that the Simulator Attack requires the lowest queries and achieves the highest attack success rate, so the superior performance of the Simulator Attack is verified. 

\begin{figure*}[htbp]
	\setlength{\abovecaptionskip}{0pt}
	\setlength{\belowcaptionskip}{0pt}
	\captionsetup[sub]{font={scriptsize}}
	\centering 
	\begin{minipage}[b]{.245\textwidth}
		\includegraphics[width=\linewidth]{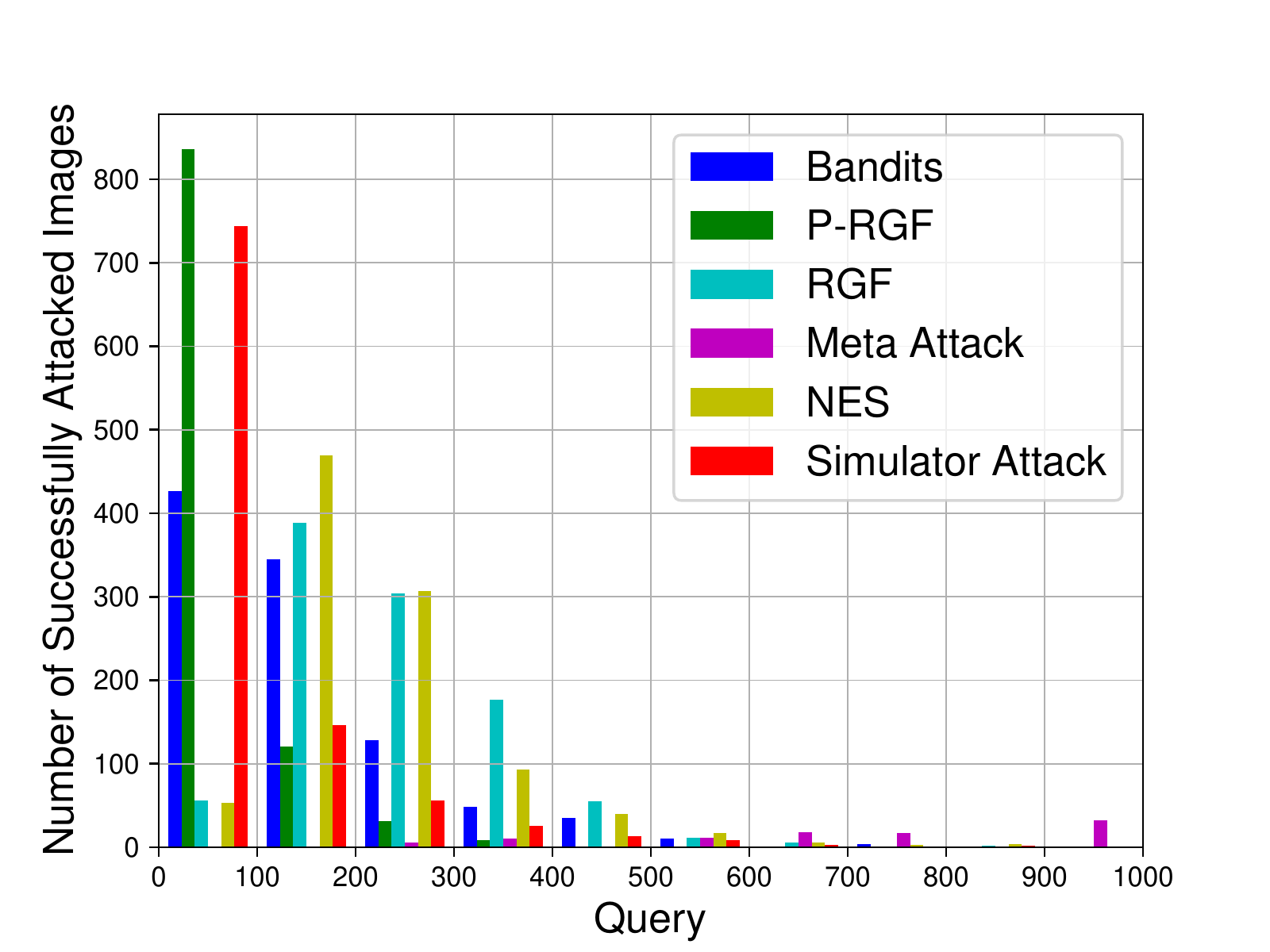}
		\subcaption{untargeted $\ell_2$ attack PyramidNet-272}
	\end{minipage}
	\begin{minipage}[b]{.245\textwidth}
		\includegraphics[width=\linewidth]{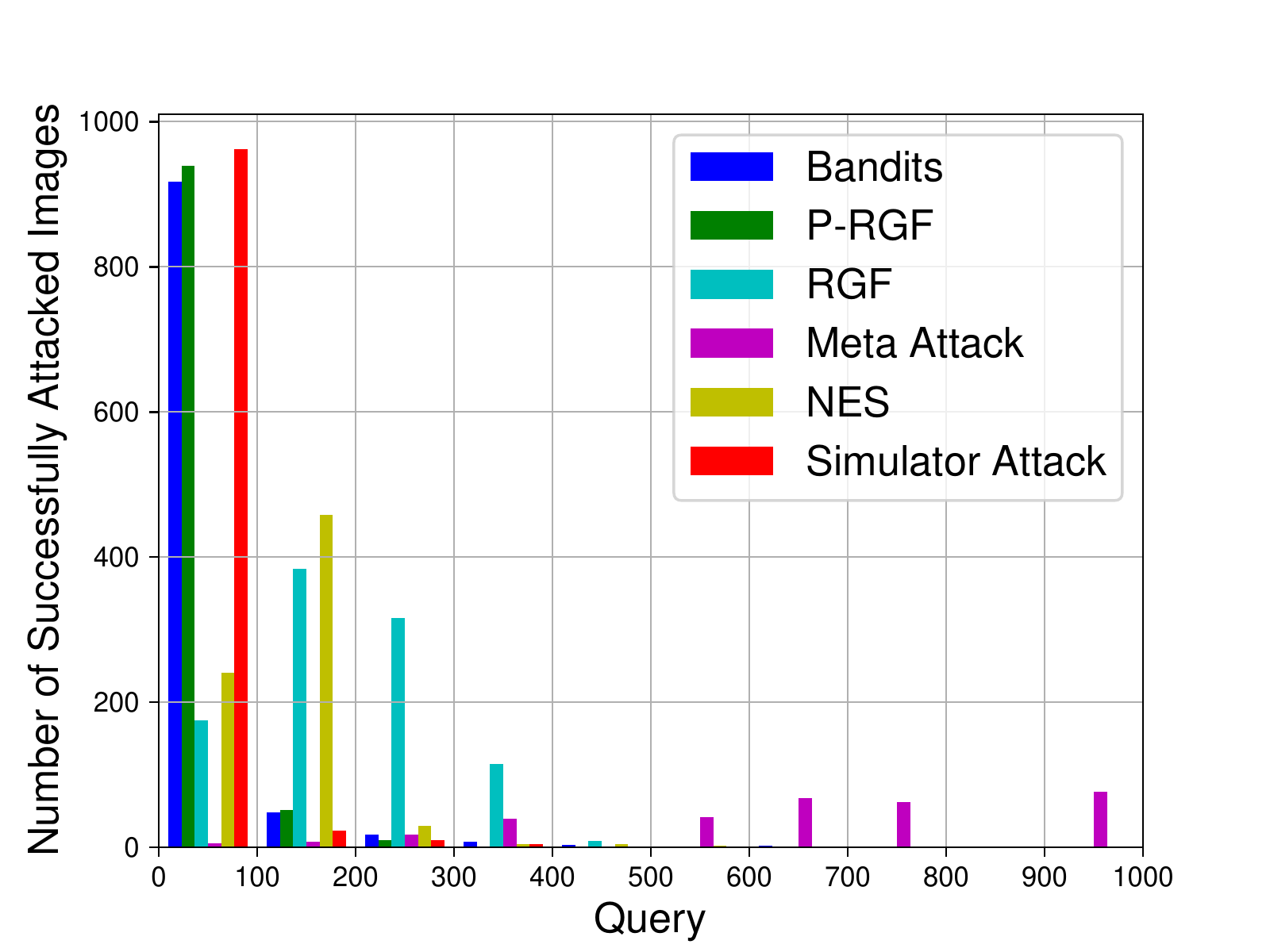}
		\subcaption{untargeted $\ell_2$ attack GDAS}
	\end{minipage}
	\begin{minipage}[b]{.245\textwidth}
		\includegraphics[width=\linewidth]{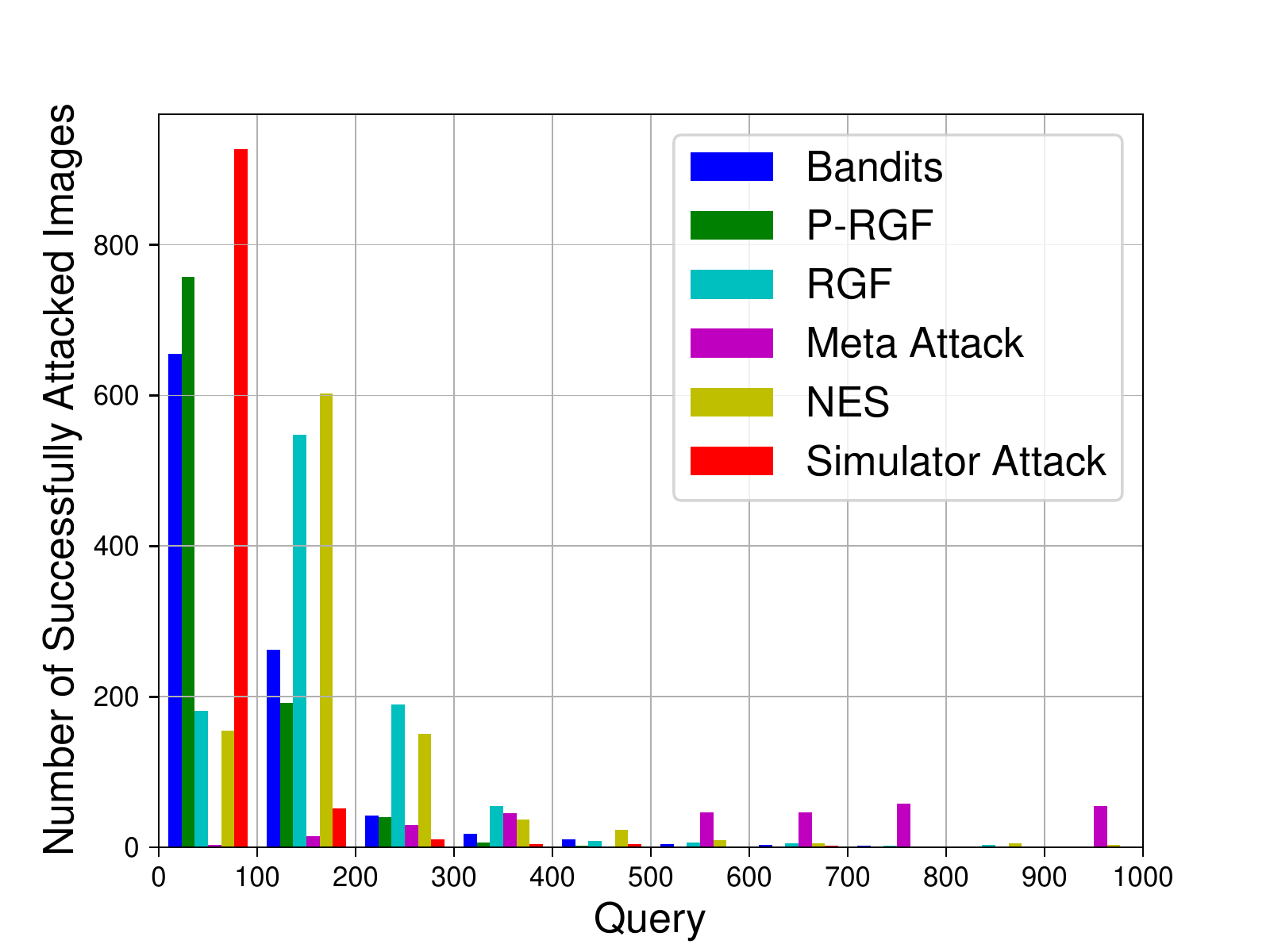}
		\subcaption{untargeted $\ell_2$ attack WRN-28}
	\end{minipage}
	\begin{minipage}[b]{.245\textwidth}
		\includegraphics[width=\linewidth]{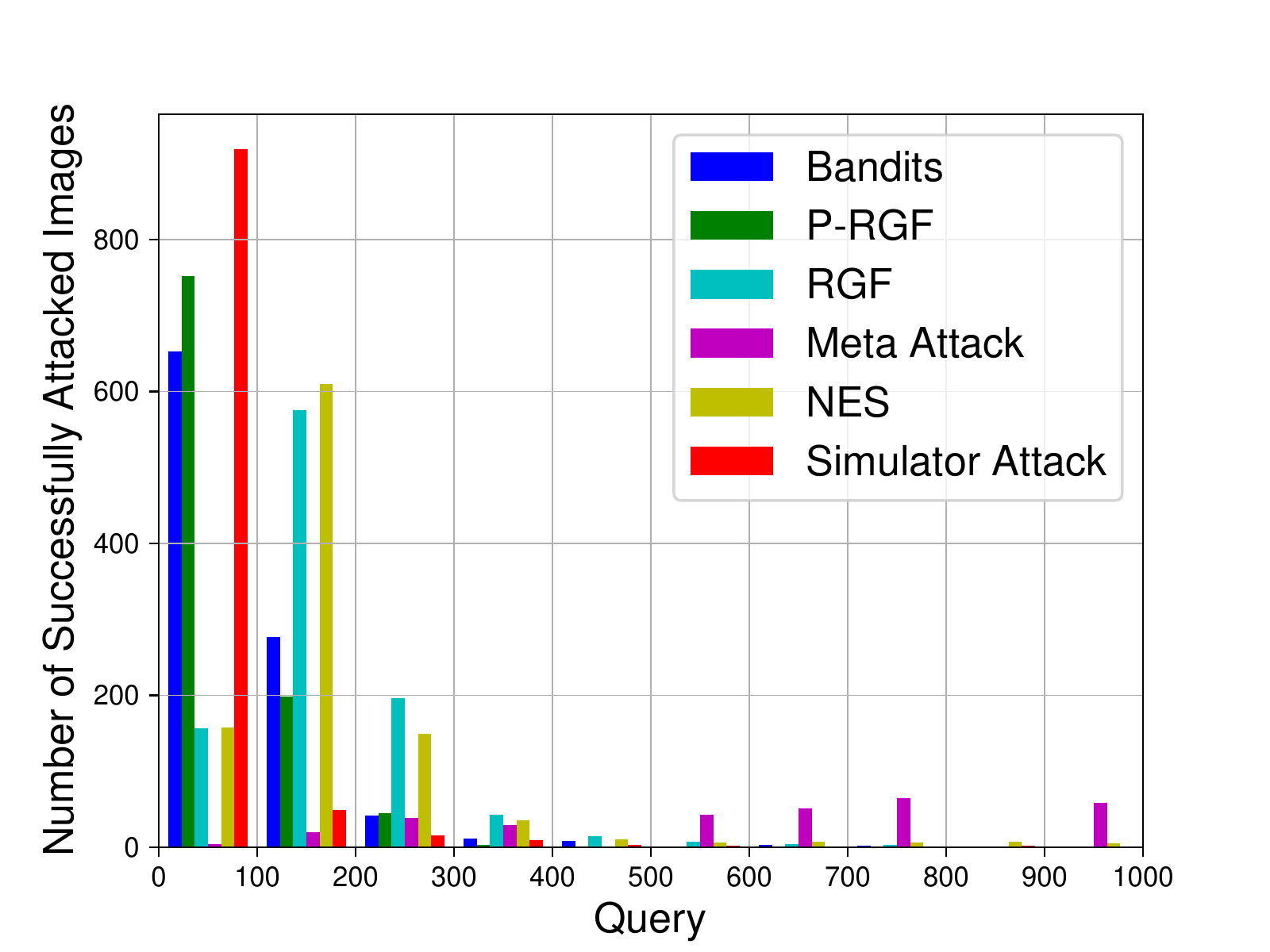}
		\subcaption{untargeted $\ell_2$ attack WRN-40}
	\end{minipage}
	\begin{minipage}[b]{.245\textwidth}
		\includegraphics[width=\linewidth]{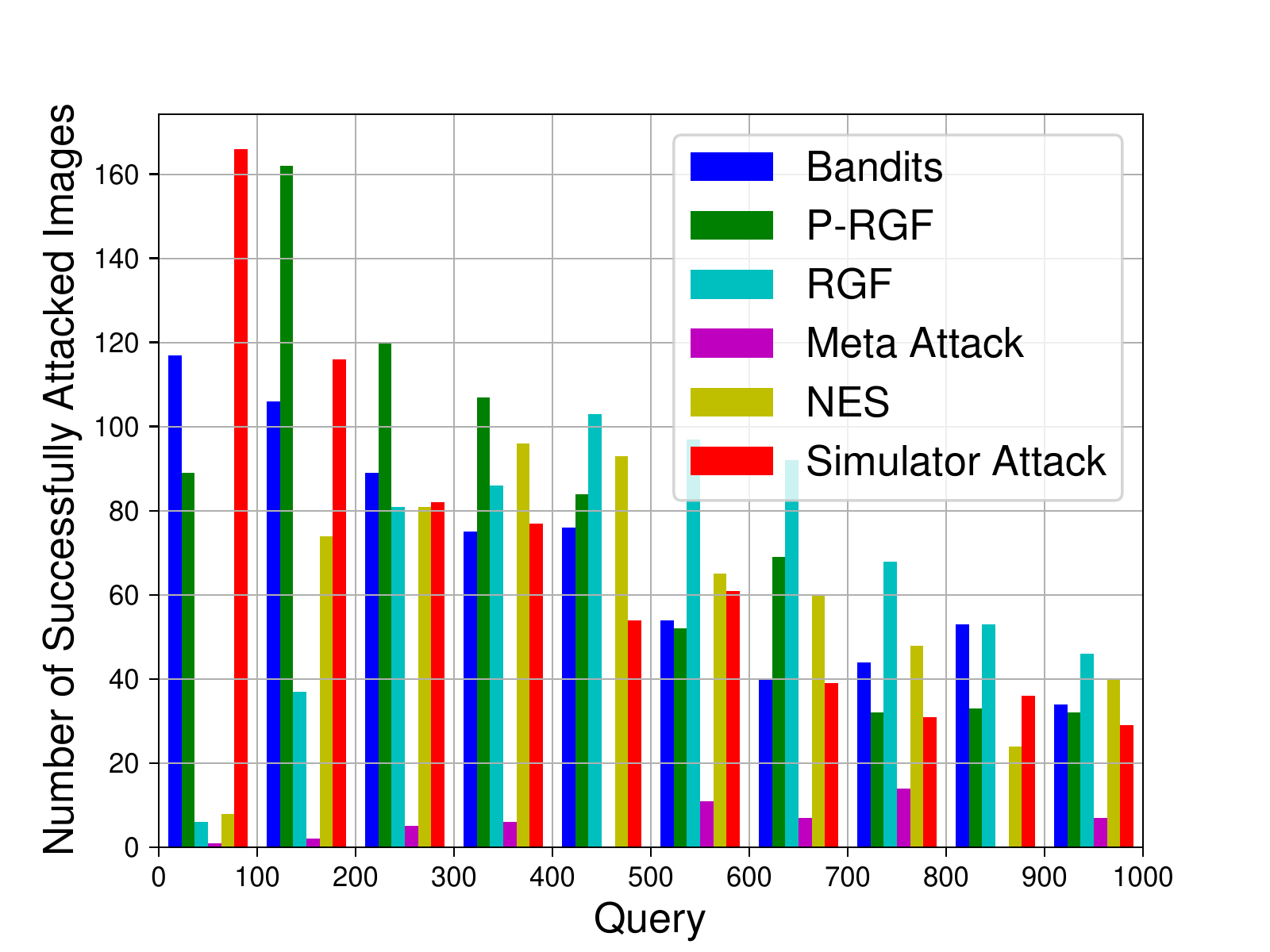}
		\subcaption{untargeted $\ell_\infty$ attack PyramidNet-272}
	\end{minipage}
	\begin{minipage}[b]{.245\textwidth}
		\includegraphics[width=\linewidth]{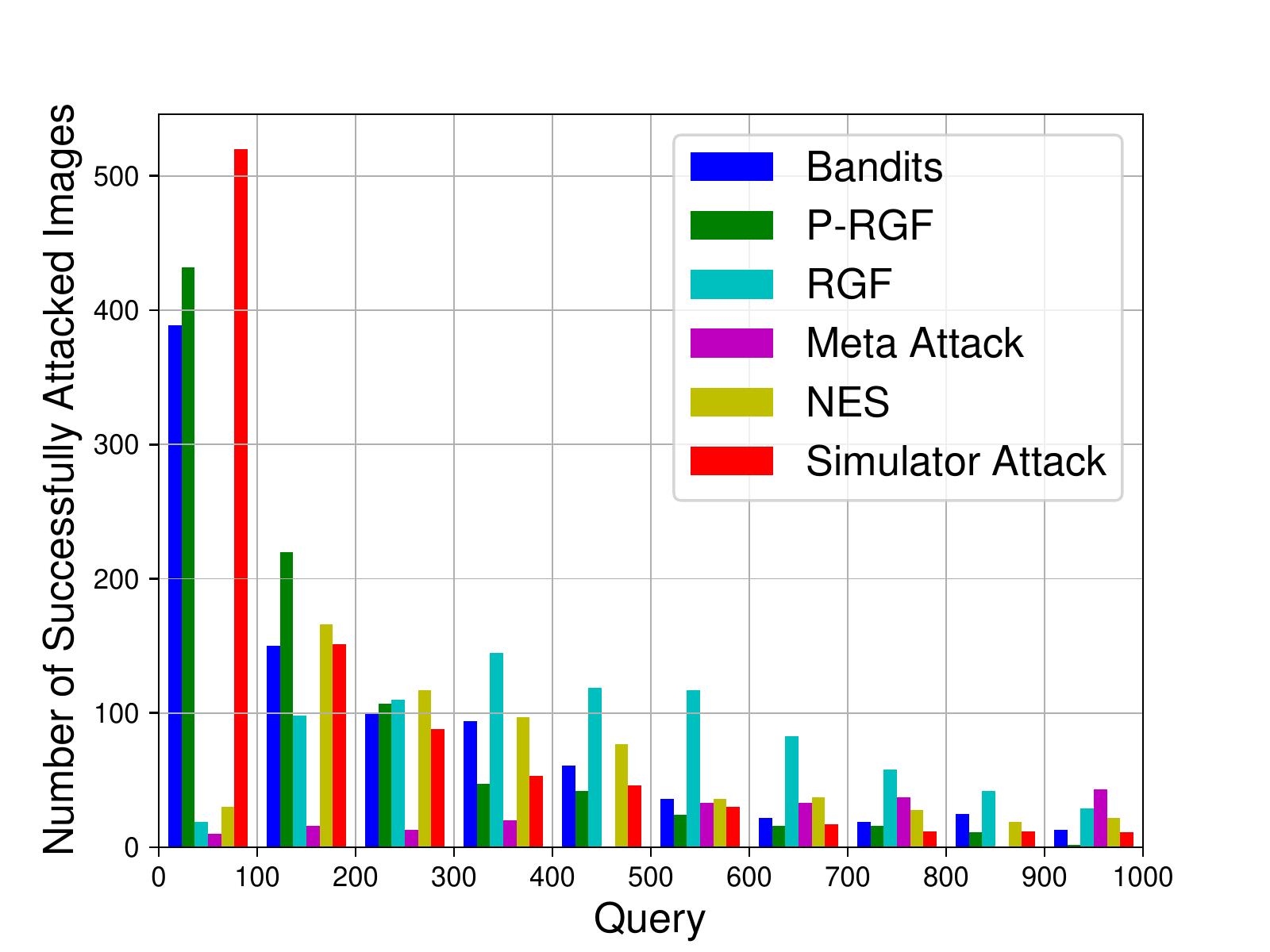}
		\subcaption{untargeted $\ell_\infty$ attack GDAS}
	\end{minipage}
	\begin{minipage}[b]{.245\textwidth}
		\includegraphics[width=\linewidth]{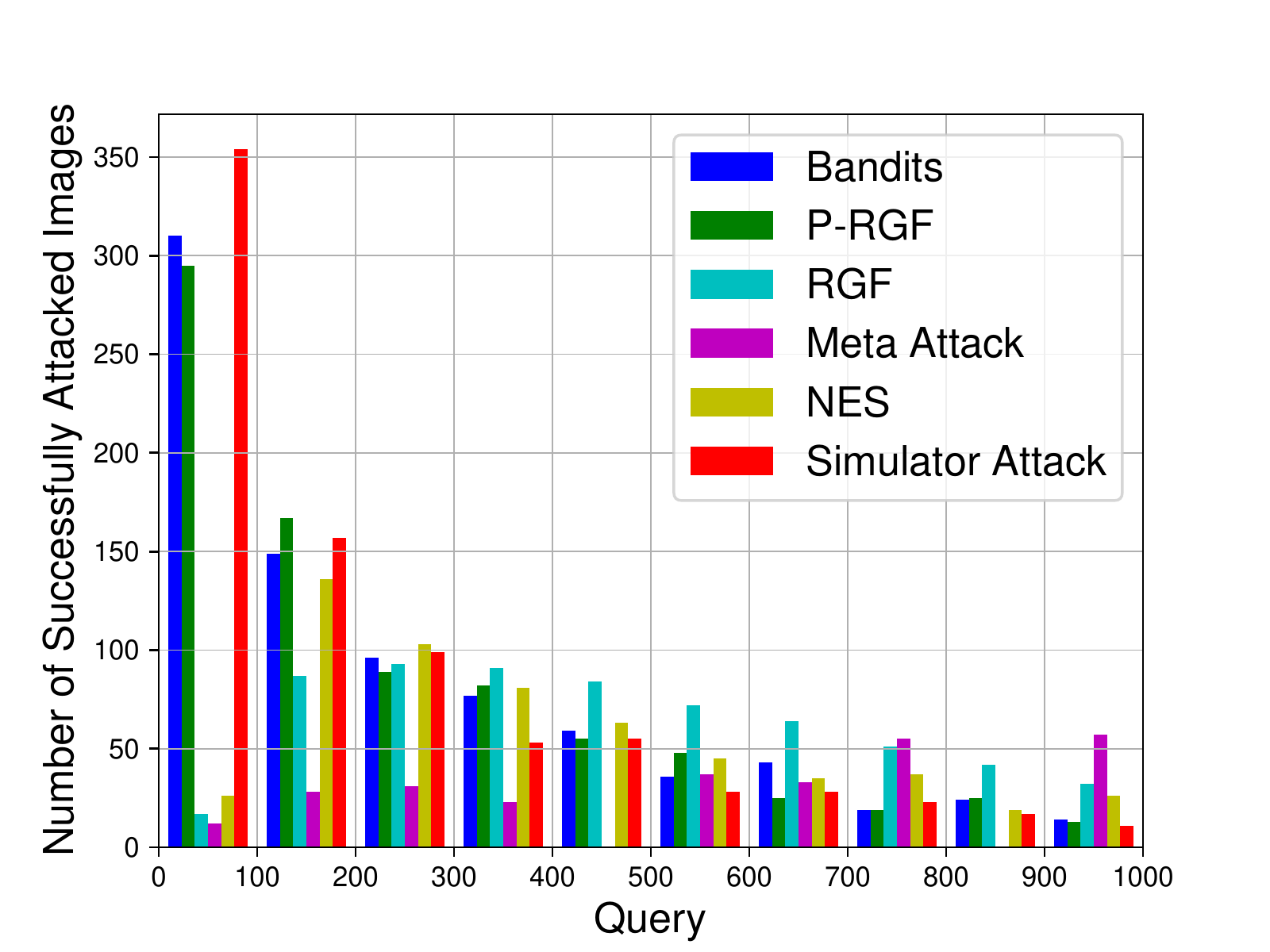}
		\subcaption{untargeted $\ell_\infty$ attack WRN-28}
	\end{minipage}
	\begin{minipage}[b]{.245\textwidth}
		\includegraphics[width=\linewidth]{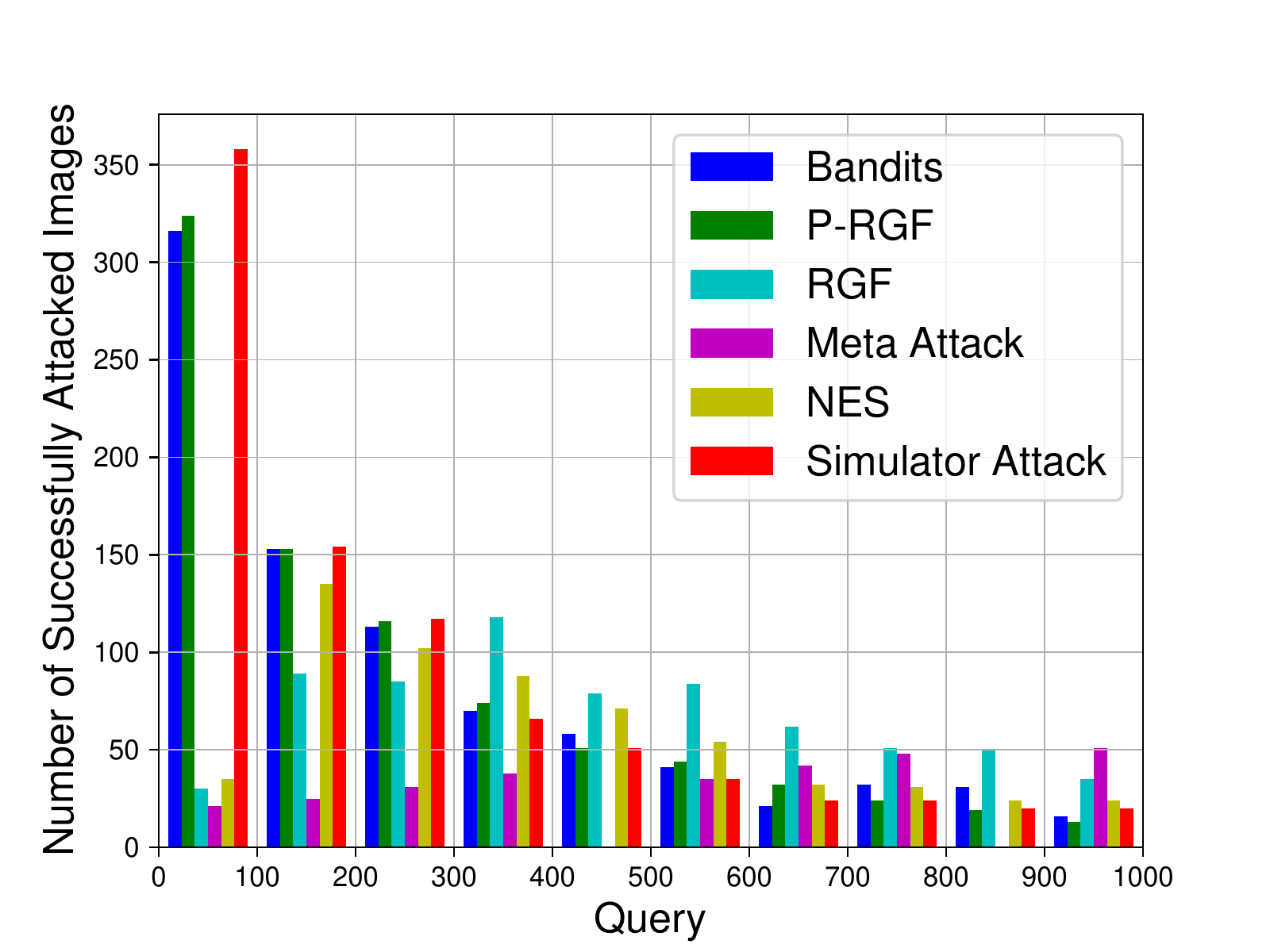}
		\subcaption{untargeted $\ell_\infty$ attack WRN-40}
	\end{minipage}
	\begin{minipage}[b]{.245\textwidth}
		\includegraphics[width=\linewidth]{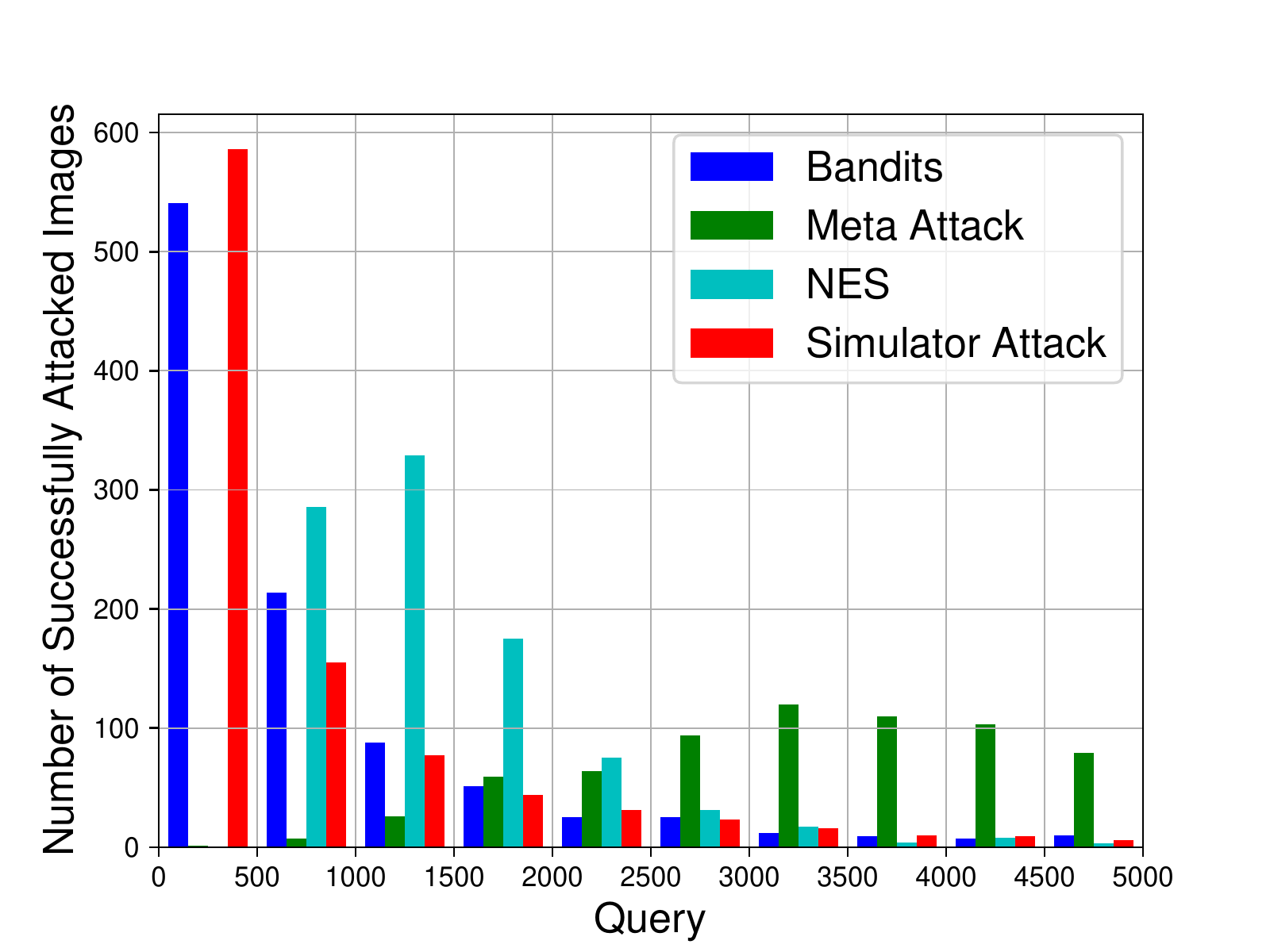}
		\subcaption{targeted $\ell_2$ attack PyramidNet-272}
	\end{minipage}
	\begin{minipage}[b]{.245\textwidth}
		\includegraphics[width=\linewidth]{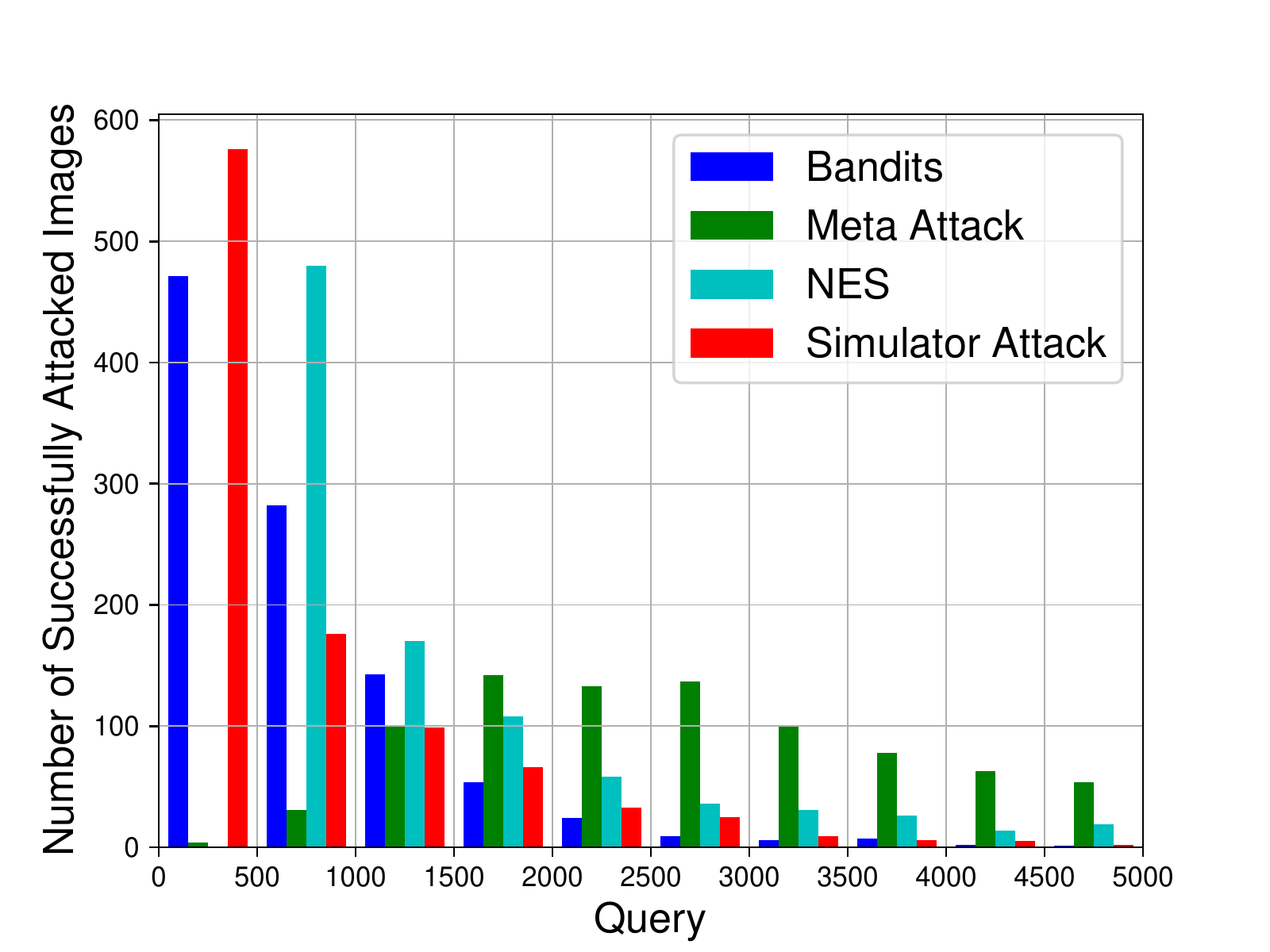}
		\subcaption{targeted $\ell_2$ attack GDAS}
	\end{minipage}
	\begin{minipage}[b]{.245\textwidth}
		\includegraphics[width=\linewidth]{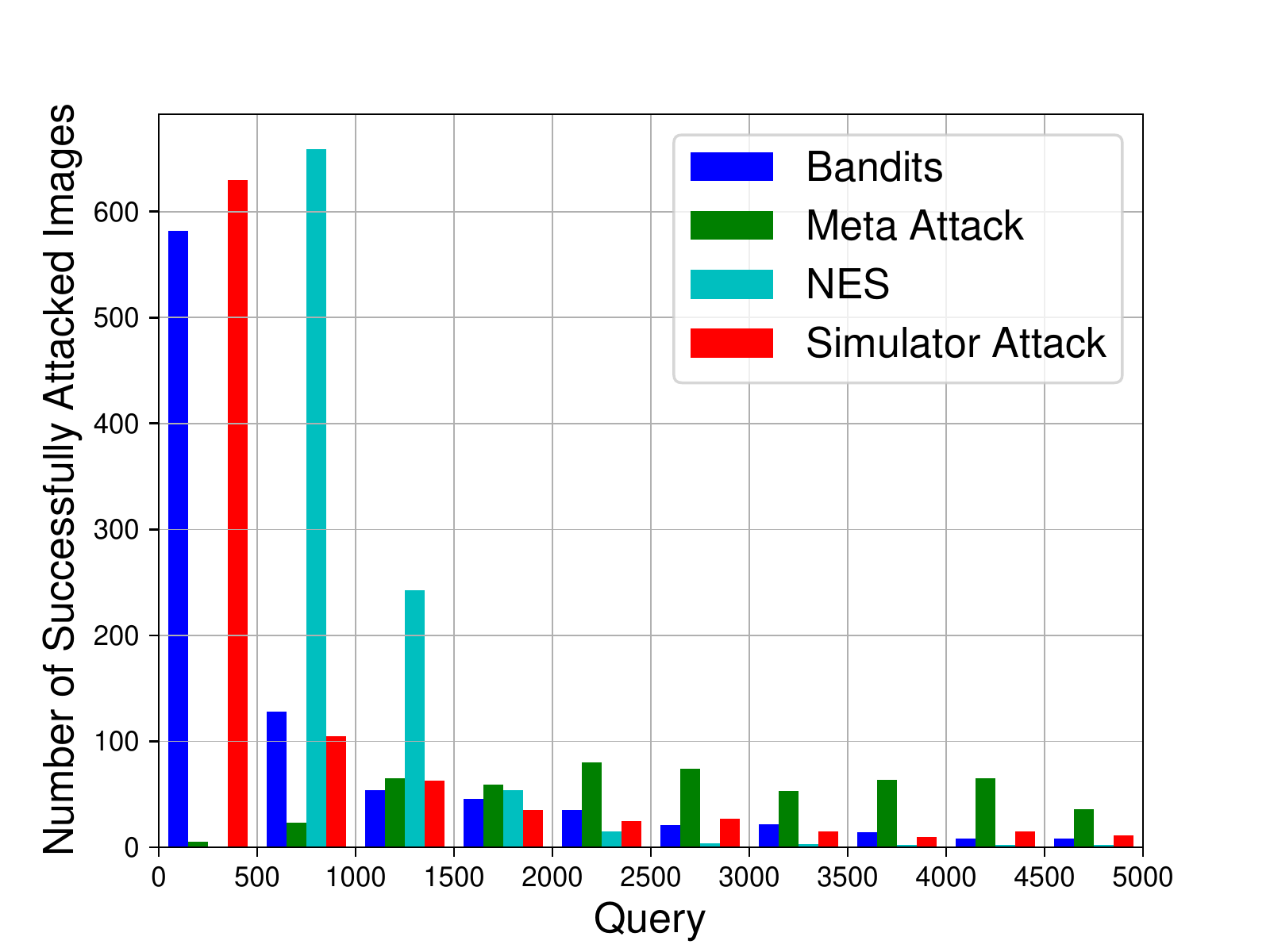}
		\subcaption{targeted $\ell_2$ attack WRN-28}
	\end{minipage}
	\begin{minipage}[b]{.245\textwidth}
		\includegraphics[width=\linewidth]{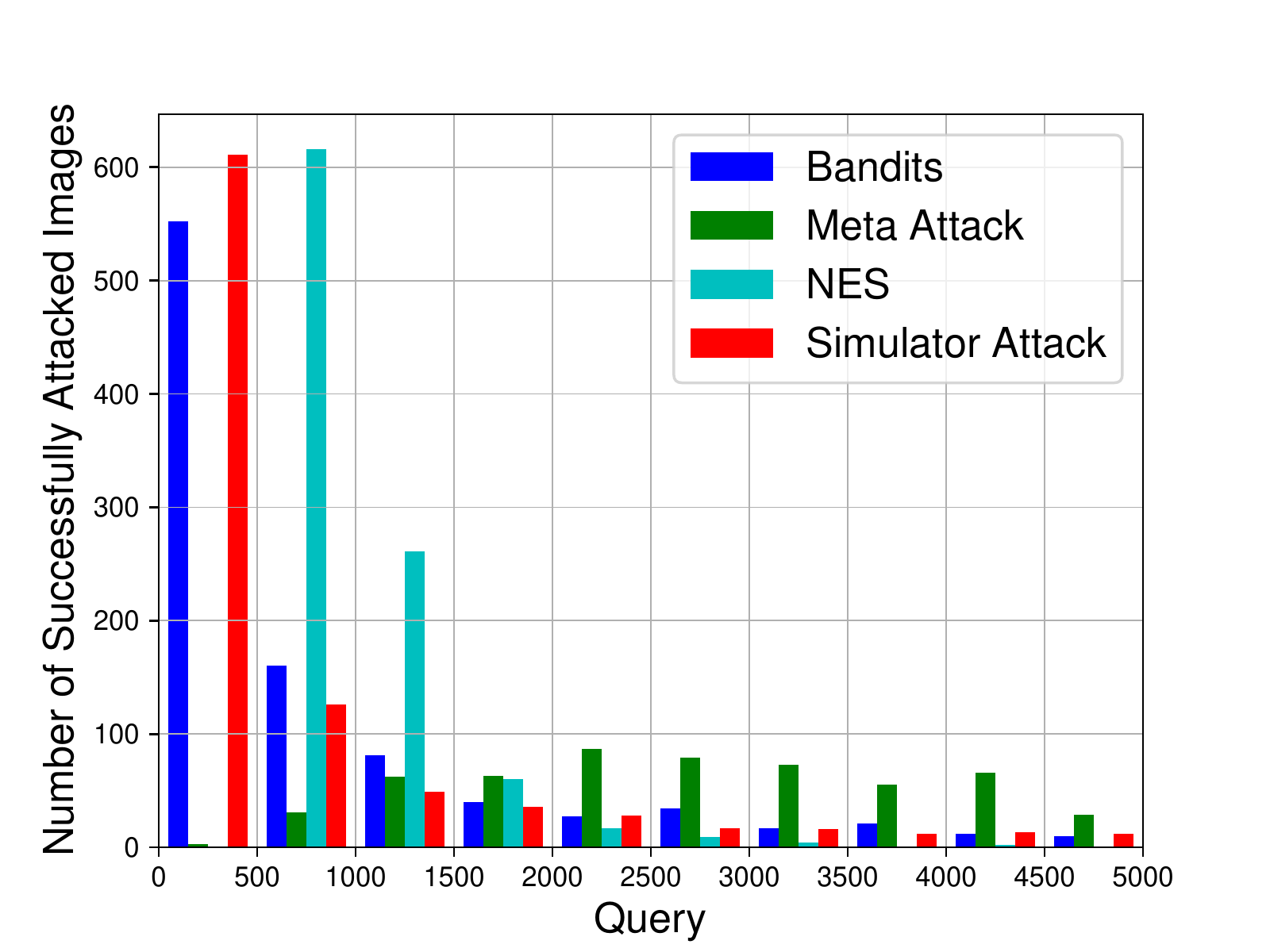}
		\subcaption{targeted $\ell_2$ attack WRN-40}
	\end{minipage}
	\begin{minipage}[b]{.245\textwidth}
		\includegraphics[width=\linewidth]{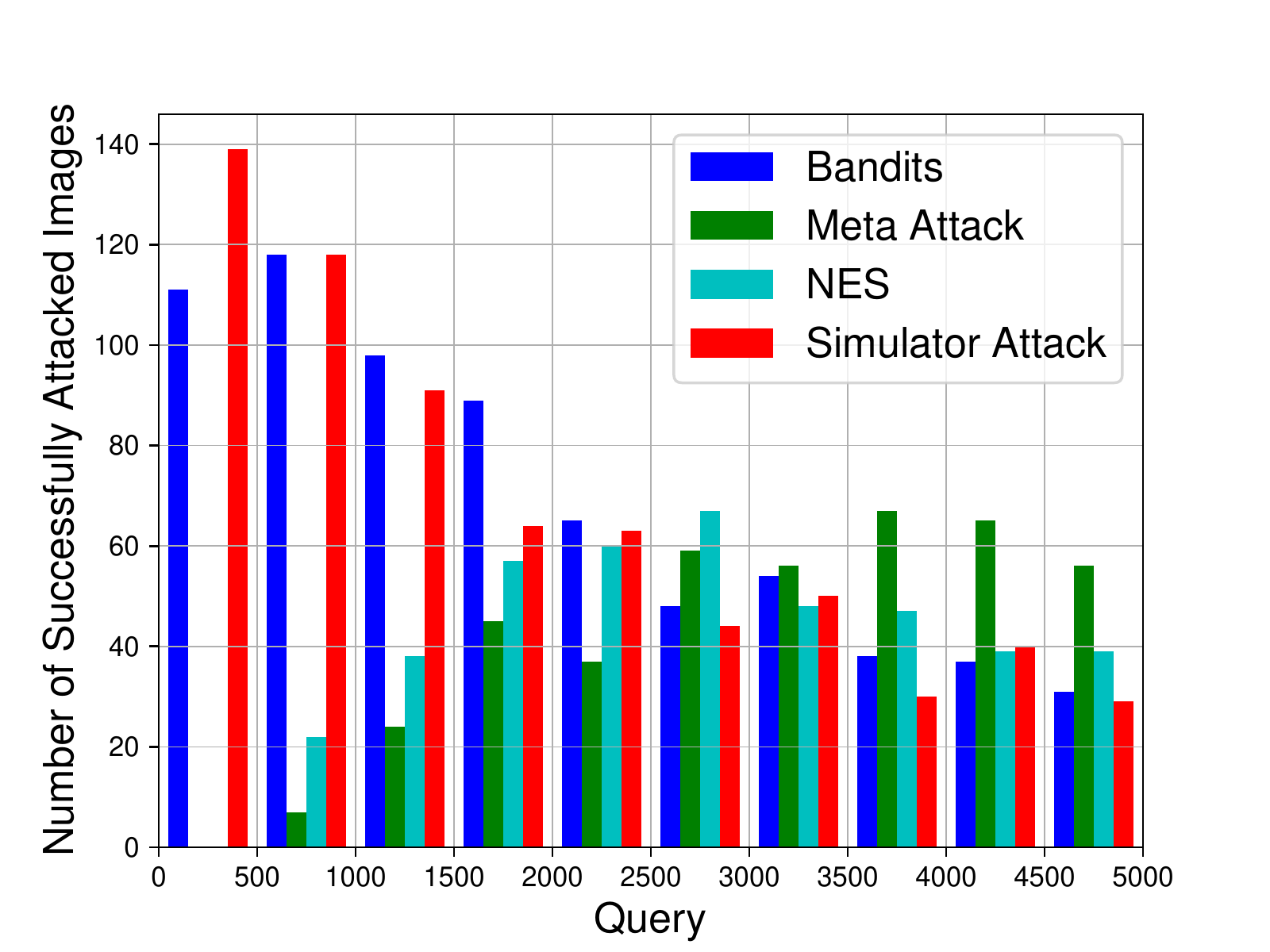}
		\subcaption{targeted $\ell_\infty$ attack PyramidNet-272}
	\end{minipage}
	\begin{minipage}[b]{.245\textwidth}
		\includegraphics[width=\linewidth]{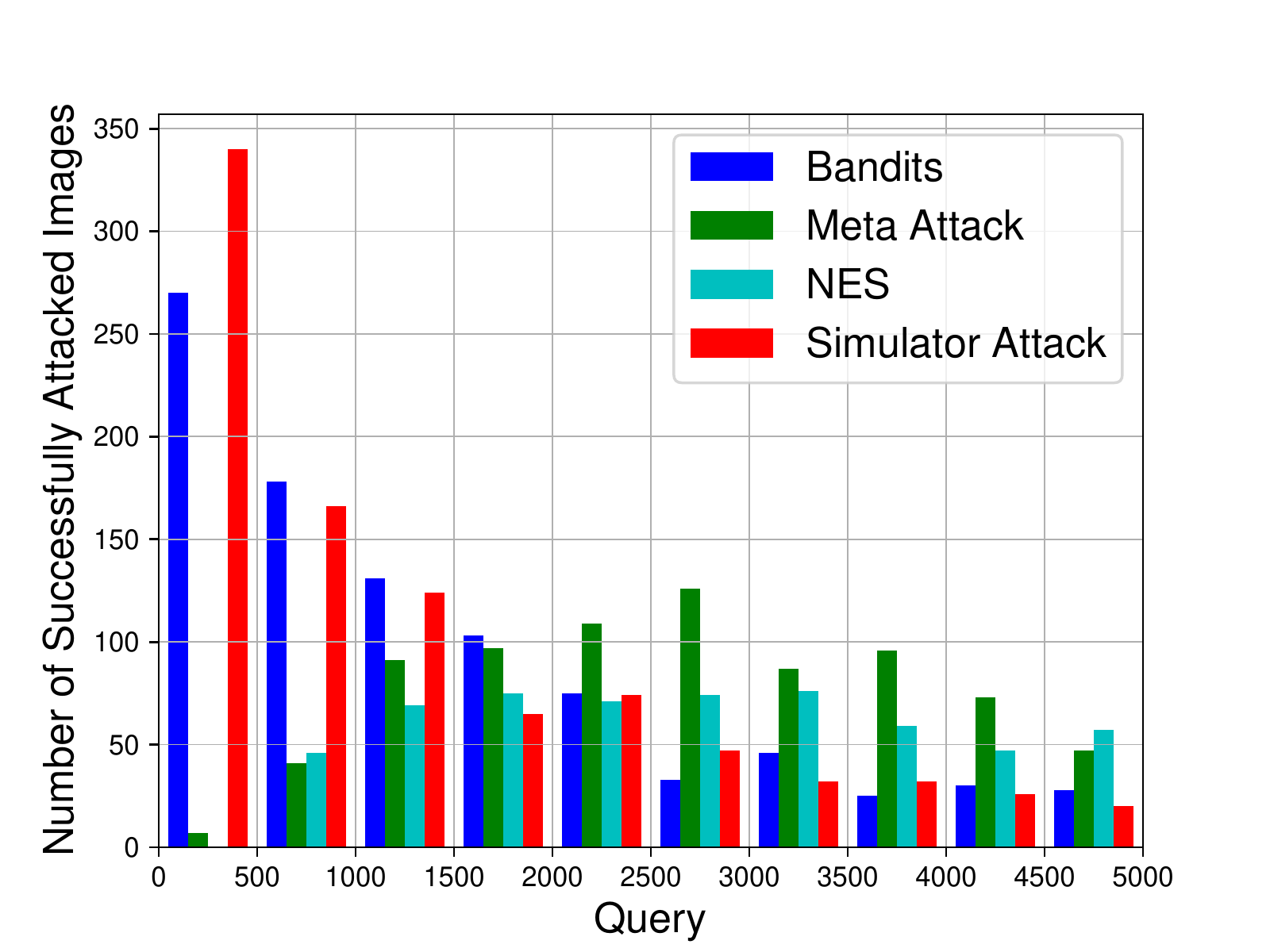}
		\subcaption{targeted $\ell_\infty$ attack GDAS}
	\end{minipage}
	\begin{minipage}[b]{.245\textwidth}
		\includegraphics[width=\linewidth]{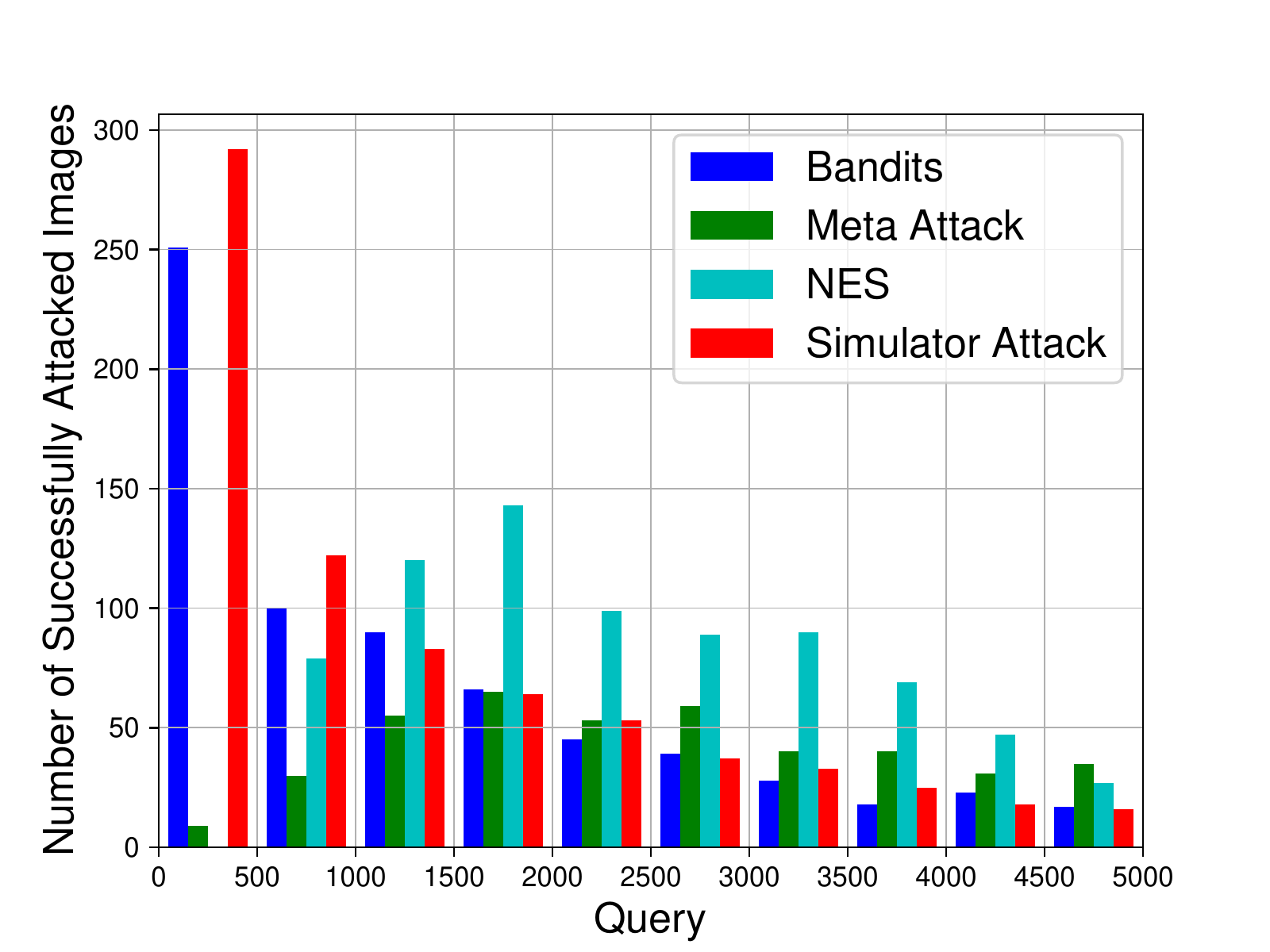}
		\subcaption{targeted $\ell_\infty$ attack WRN-28}
	\end{minipage}
	\begin{minipage}[b]{.245\textwidth}
		\includegraphics[width=\linewidth]{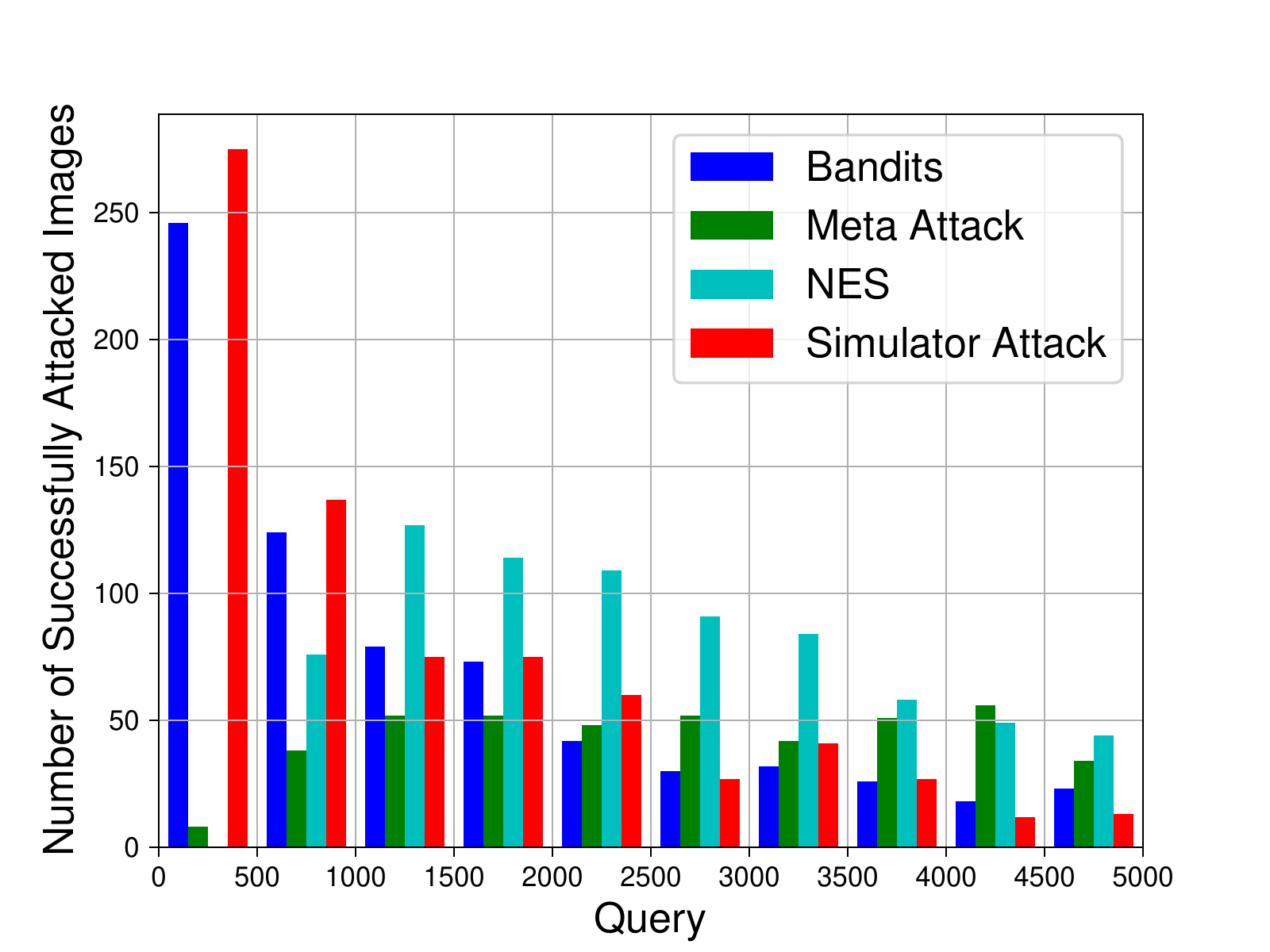}
		\subcaption{targeted $\ell_\infty$ attack WRN-40}
	\end{minipage}
	\caption{The histogram of query number in the CIFAR-10 dataset.}
	\label{fig:histogram_CIFAR-10}
\end{figure*}

\begin{figure*}[t]
	\setlength{\abovecaptionskip}{0pt}
	\setlength{\belowcaptionskip}{0pt}
	\captionsetup[sub]{font={scriptsize}}
	\centering 
	\begin{minipage}[b]{.245\textwidth}
		\includegraphics[width=\linewidth]{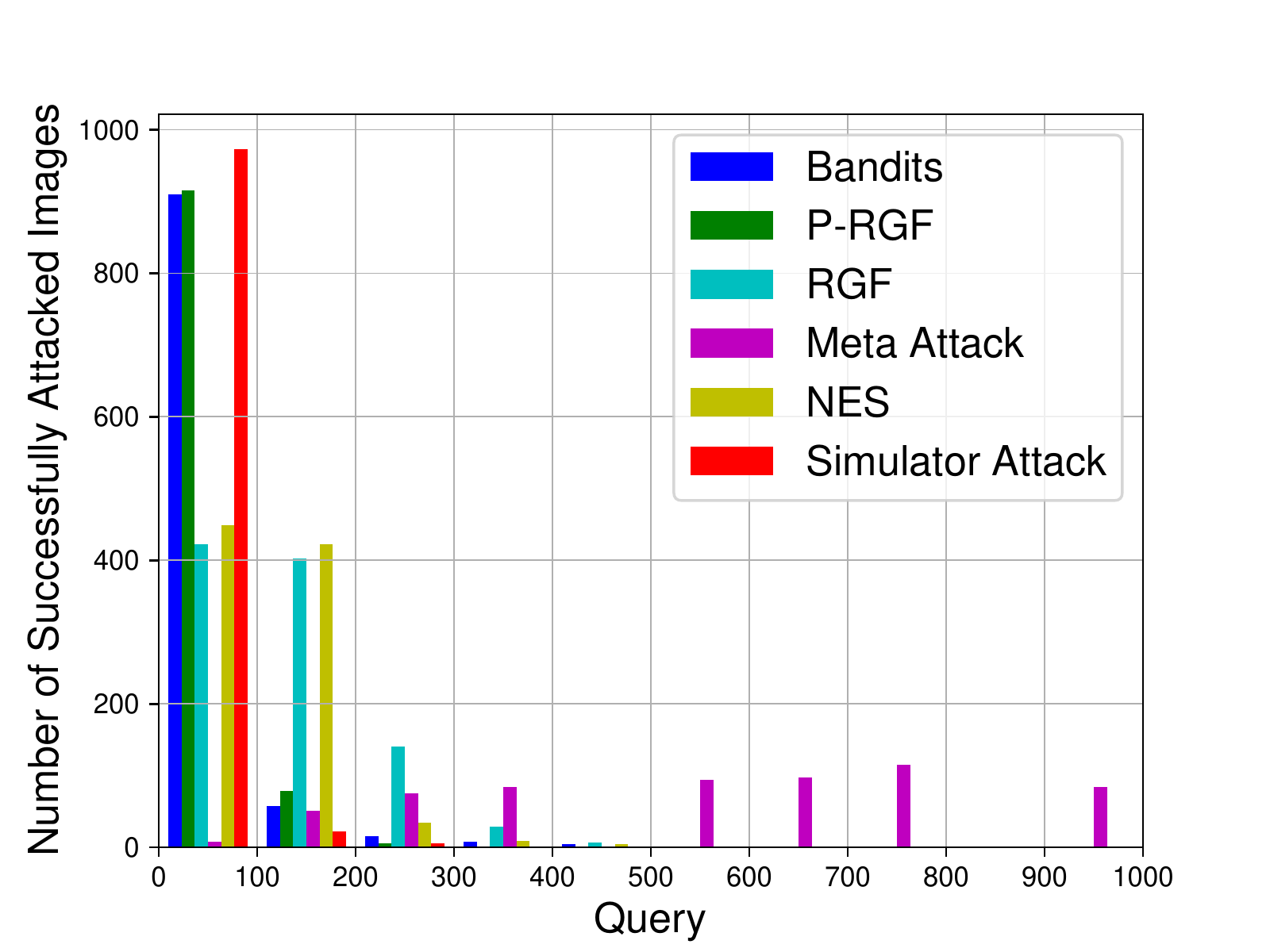}
		\subcaption{untargeted $\ell_2$ attack PyramidNet-272}
	\end{minipage}
	\begin{minipage}[b]{.245\textwidth}
		\includegraphics[width=\linewidth]{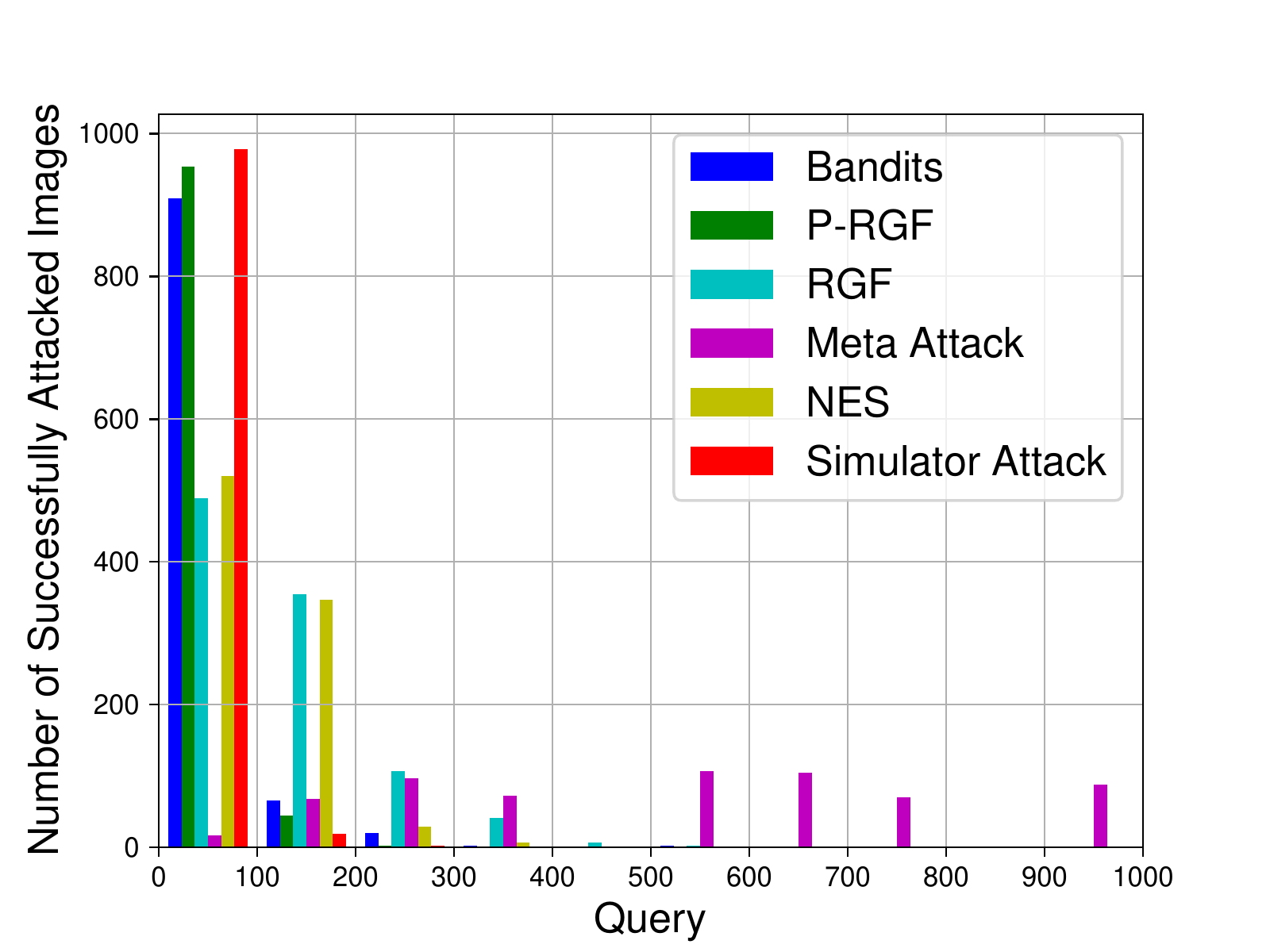}
		\subcaption{untargeted $\ell_2$ attack GDAS}
	\end{minipage}
	\begin{minipage}[b]{.245\textwidth}
		\includegraphics[width=\linewidth]{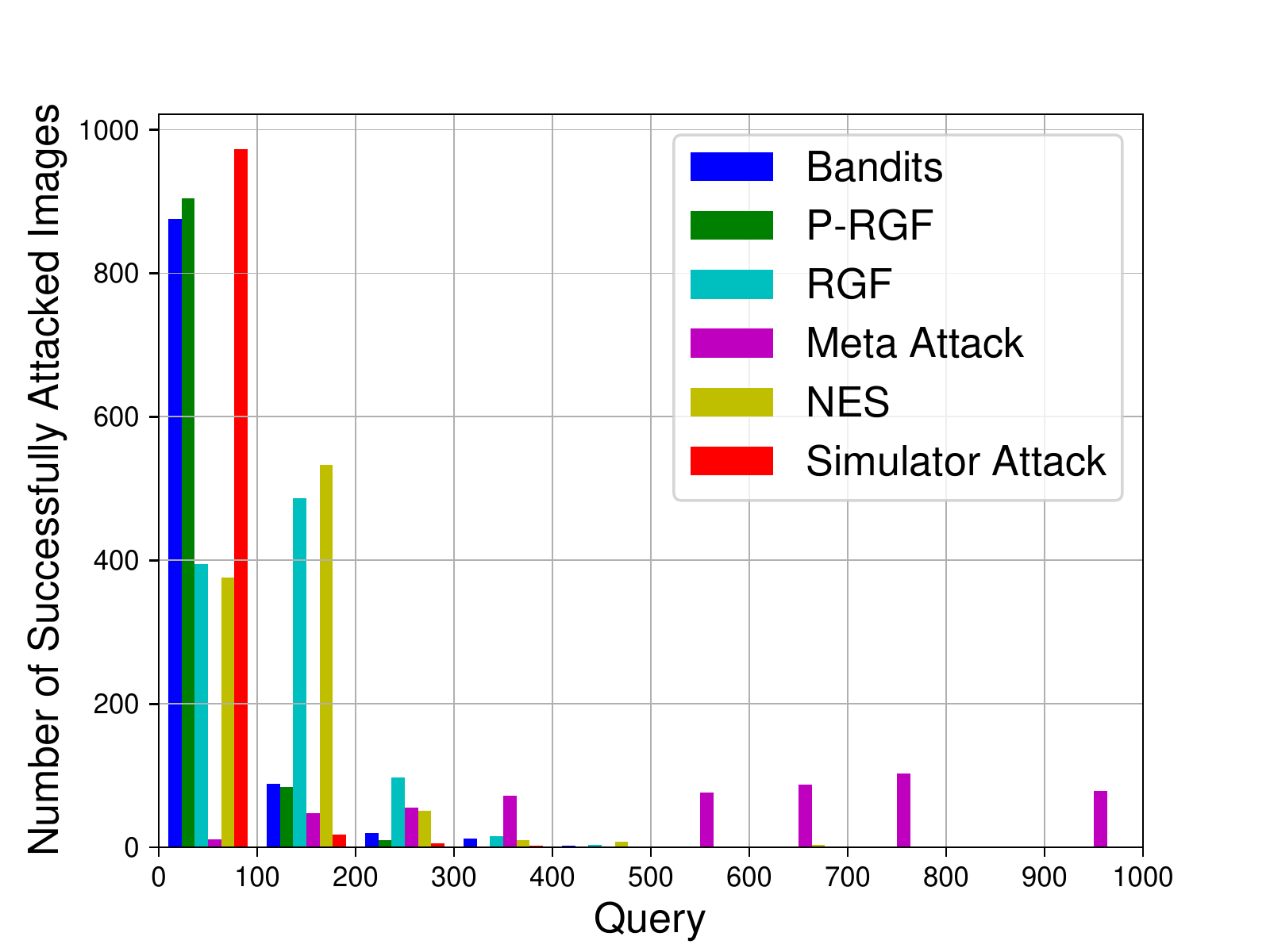}
		\subcaption{untargeted $\ell_2$ attack WRN-28}
	\end{minipage}
	\begin{minipage}[b]{.245\textwidth}
		\includegraphics[width=\linewidth]{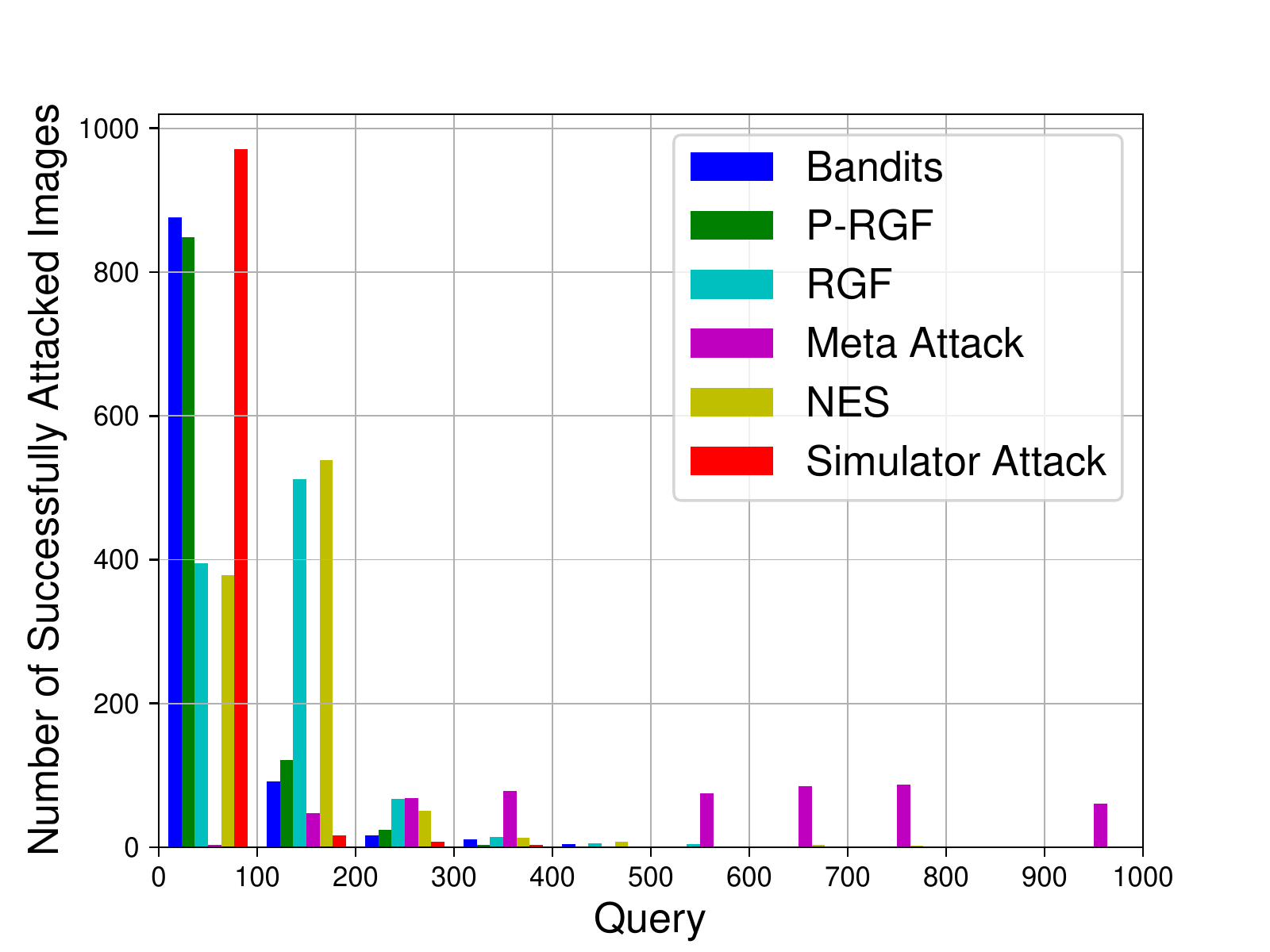}
		\subcaption{untargeted $\ell_2$ attack WRN-40}
	\end{minipage}
	\begin{minipage}[b]{.245\textwidth}
		\includegraphics[width=\linewidth]{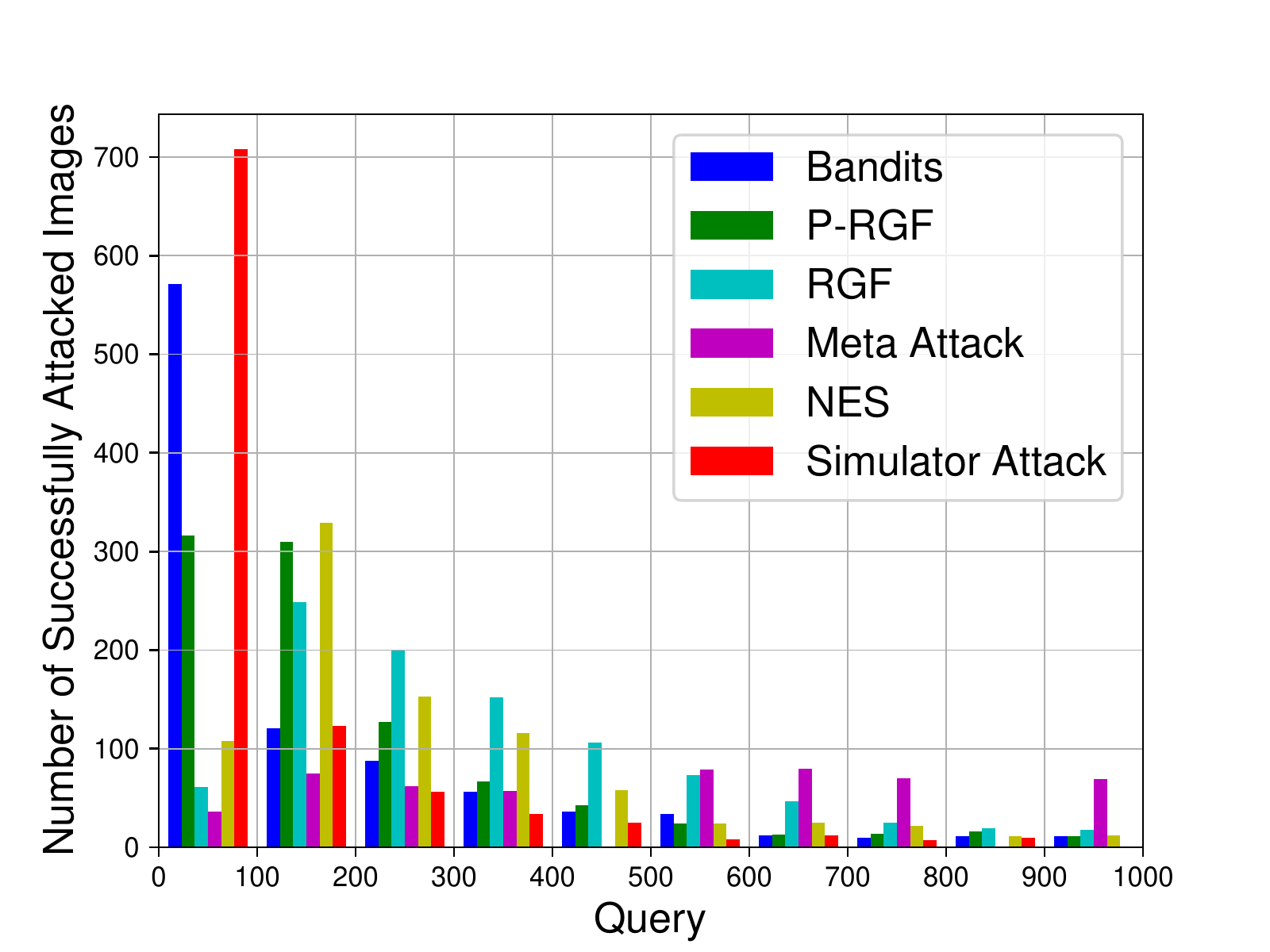}
		\subcaption{untargeted $\ell_\infty$ attack PyramidNet-272}
	\end{minipage}
	\begin{minipage}[b]{.245\textwidth}
		\includegraphics[width=\linewidth]{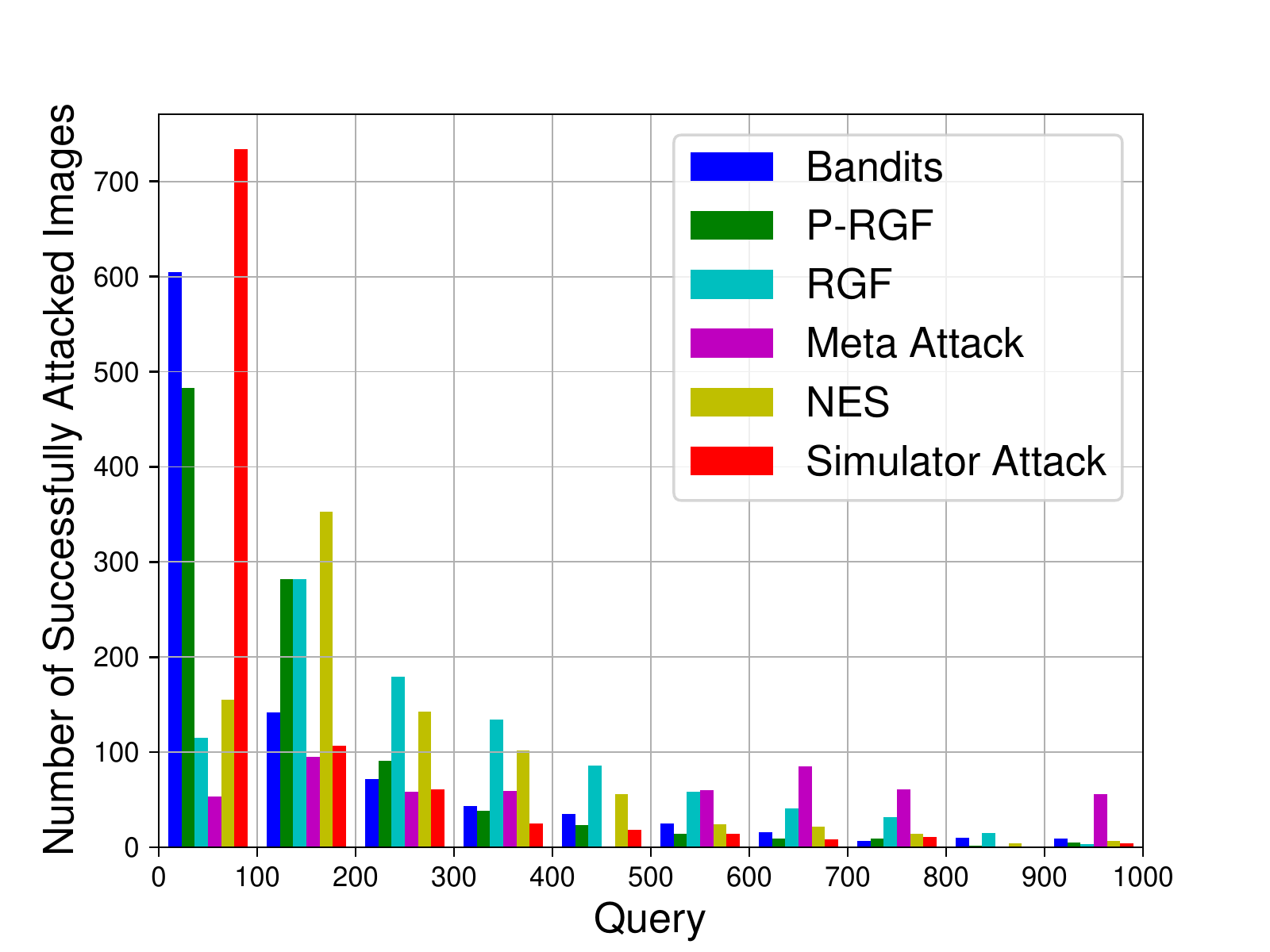}
		\subcaption{untargeted $\ell_\infty$ attack GDAS}
	\end{minipage}
	\begin{minipage}[b]{.245\textwidth}
		\includegraphics[width=\linewidth]{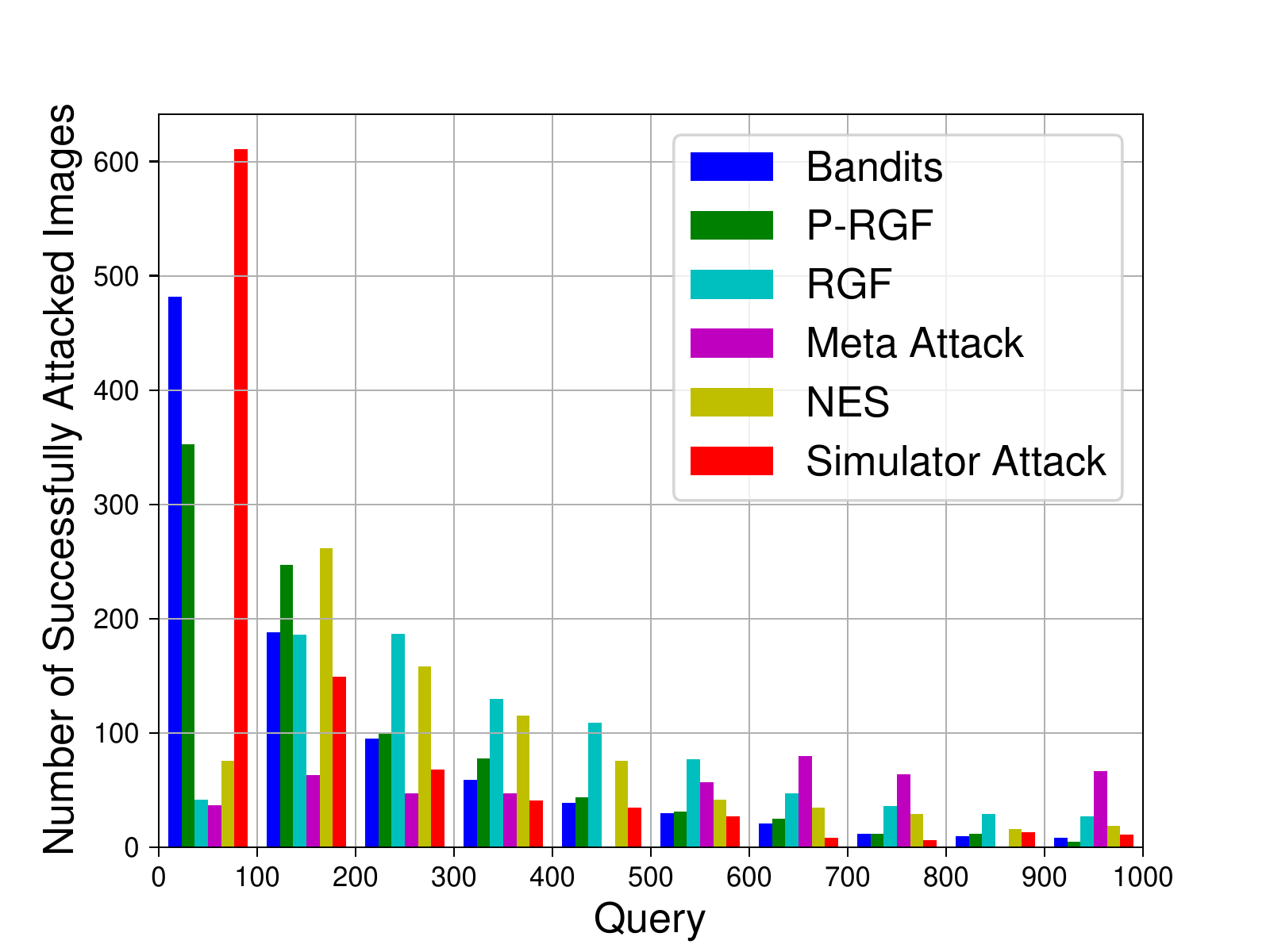}
		\subcaption{untargeted $\ell_\infty$ attack WRN-28}
	\end{minipage}
	\begin{minipage}[b]{.245\textwidth}
		\includegraphics[width=\linewidth]{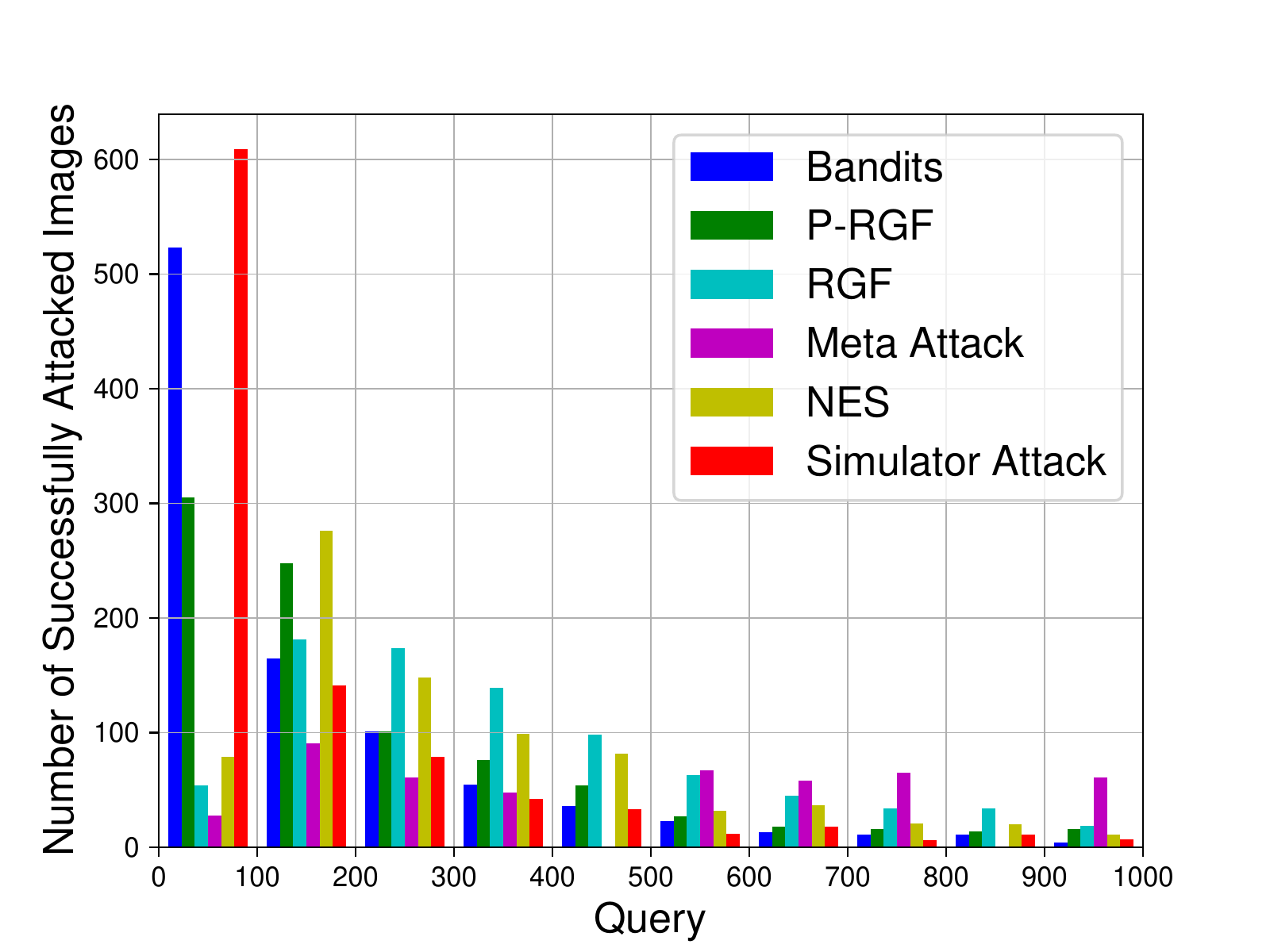}
		\subcaption{untargeted $\ell_\infty$ attack WRN-40}
	\end{minipage}
	\begin{minipage}[b]{.245\textwidth}
		\includegraphics[width=\linewidth]{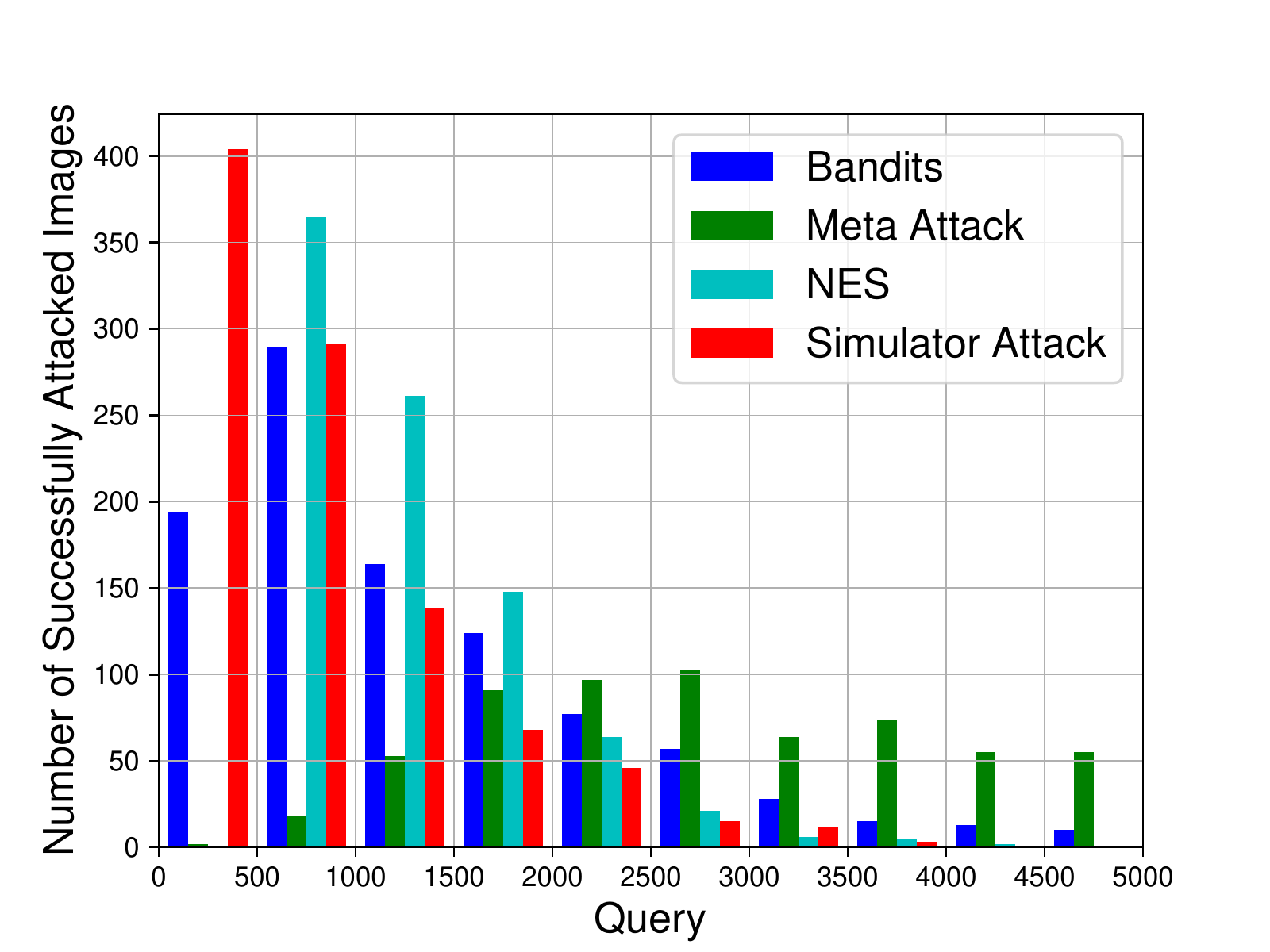}
		\subcaption{targeted $\ell_2$ attack PyramidNet-272}
	\end{minipage}
	\begin{minipage}[b]{.245\textwidth}
		\includegraphics[width=\linewidth]{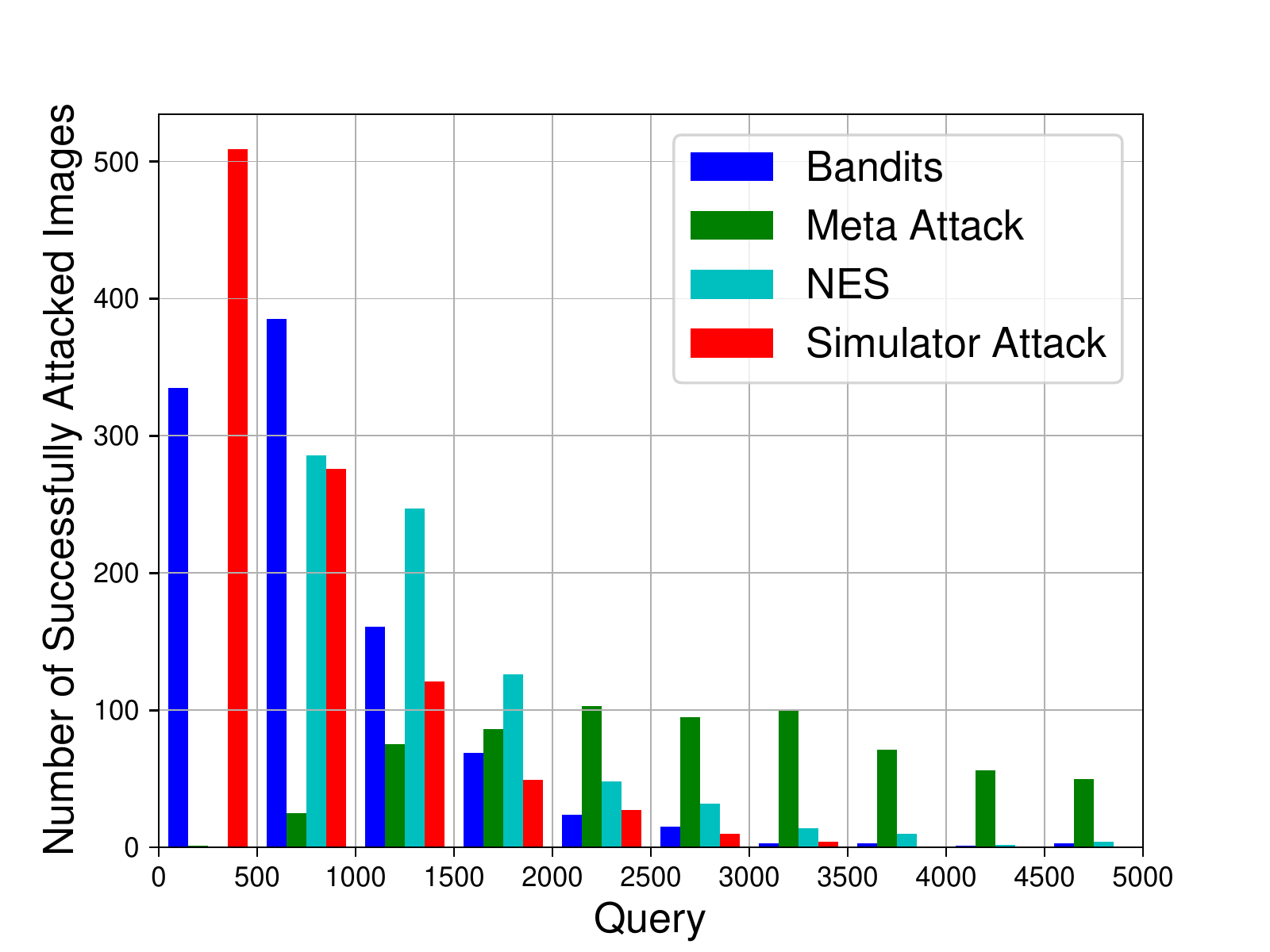}
		\subcaption{targeted $\ell_2$ attack GDAS}
	\end{minipage}
	\begin{minipage}[b]{.245\textwidth}
		\includegraphics[width=\linewidth]{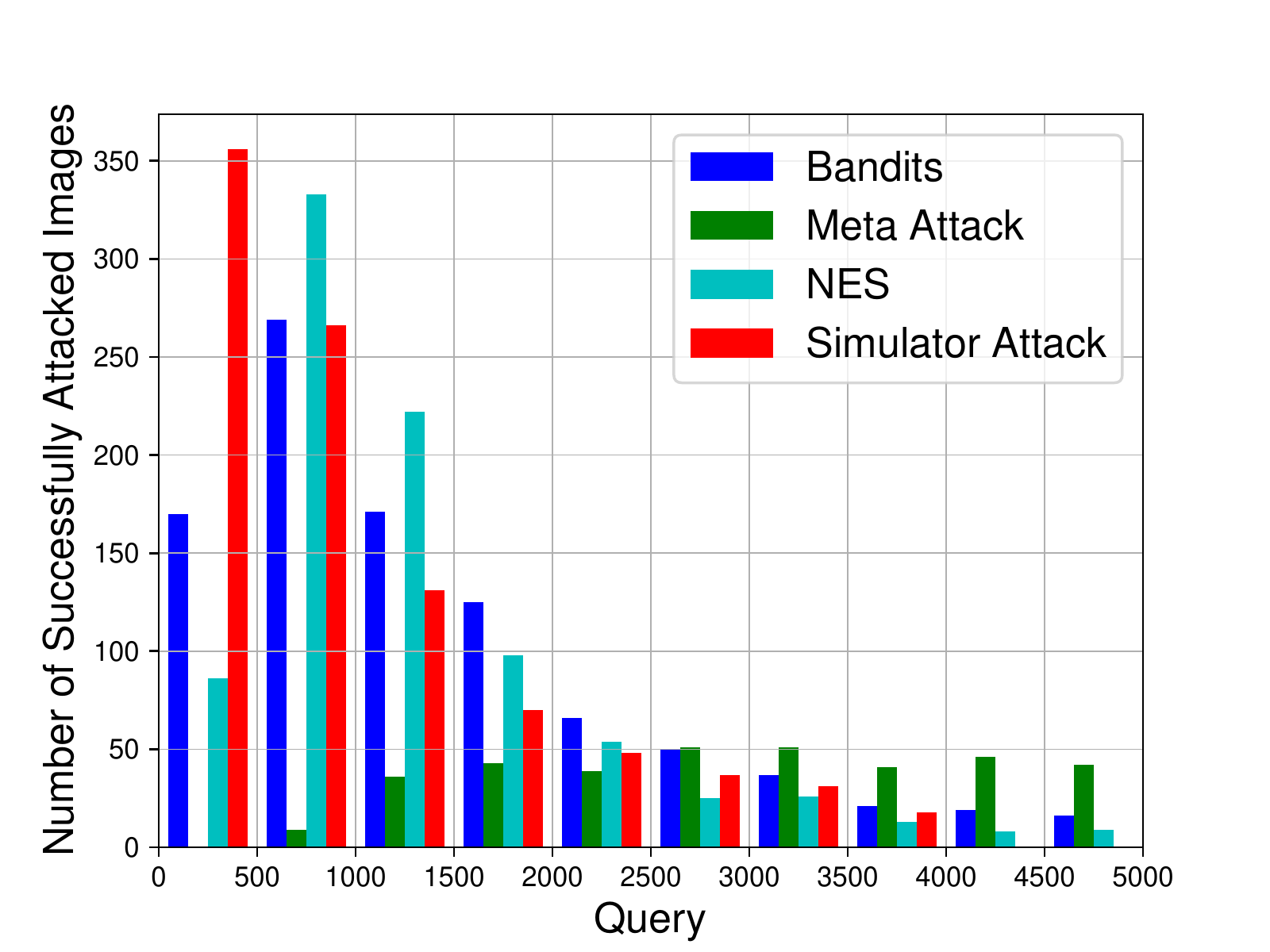}
		\subcaption{targeted $\ell_2$ attack WRN-28}
	\end{minipage}
	\begin{minipage}[b]{.245\textwidth}
		\includegraphics[width=\linewidth]{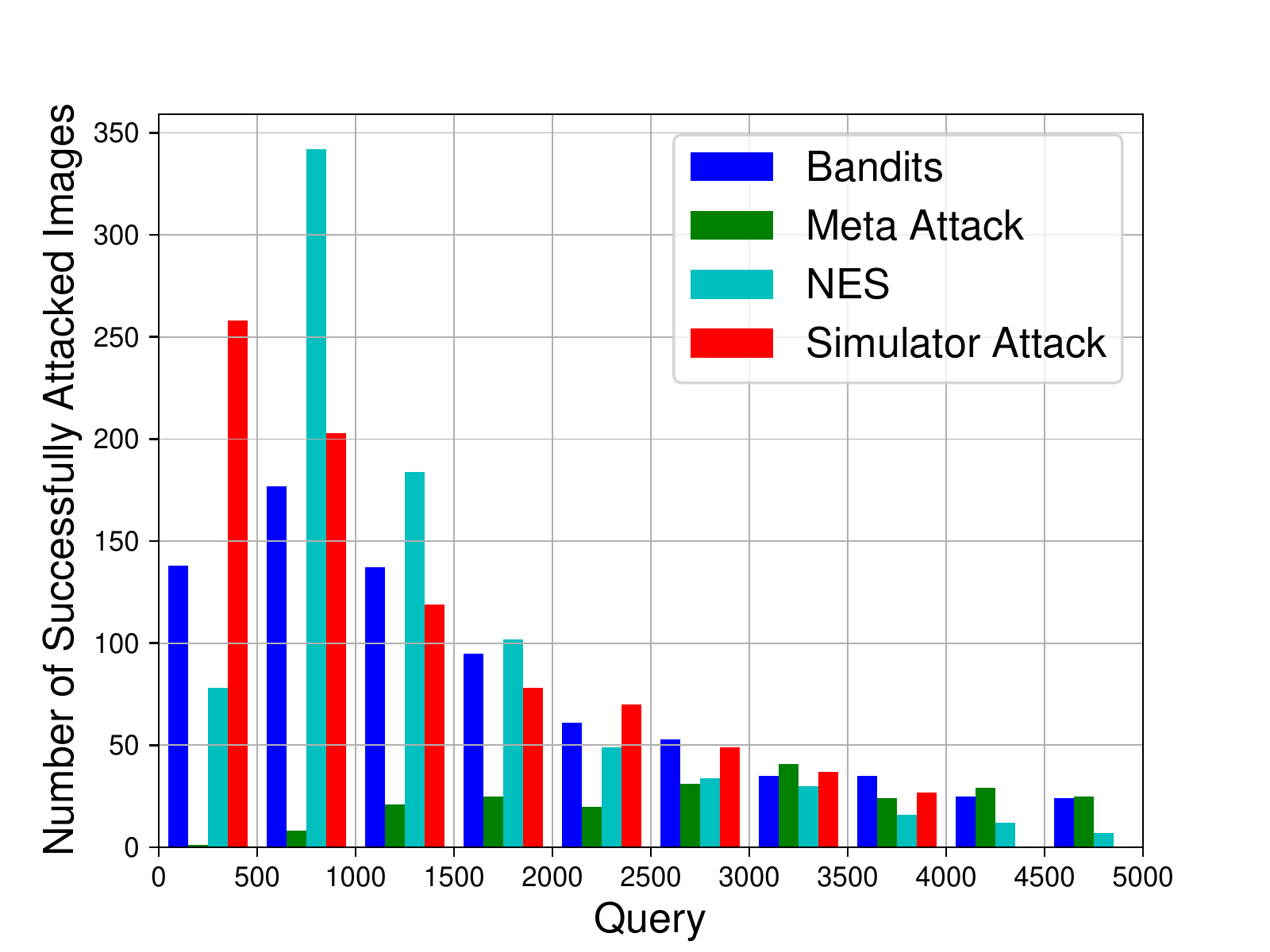}
		\subcaption{targeted $\ell_2$ attack WRN-40}
	\end{minipage}
	\begin{minipage}[b]{.245\textwidth}
		\includegraphics[width=\linewidth]{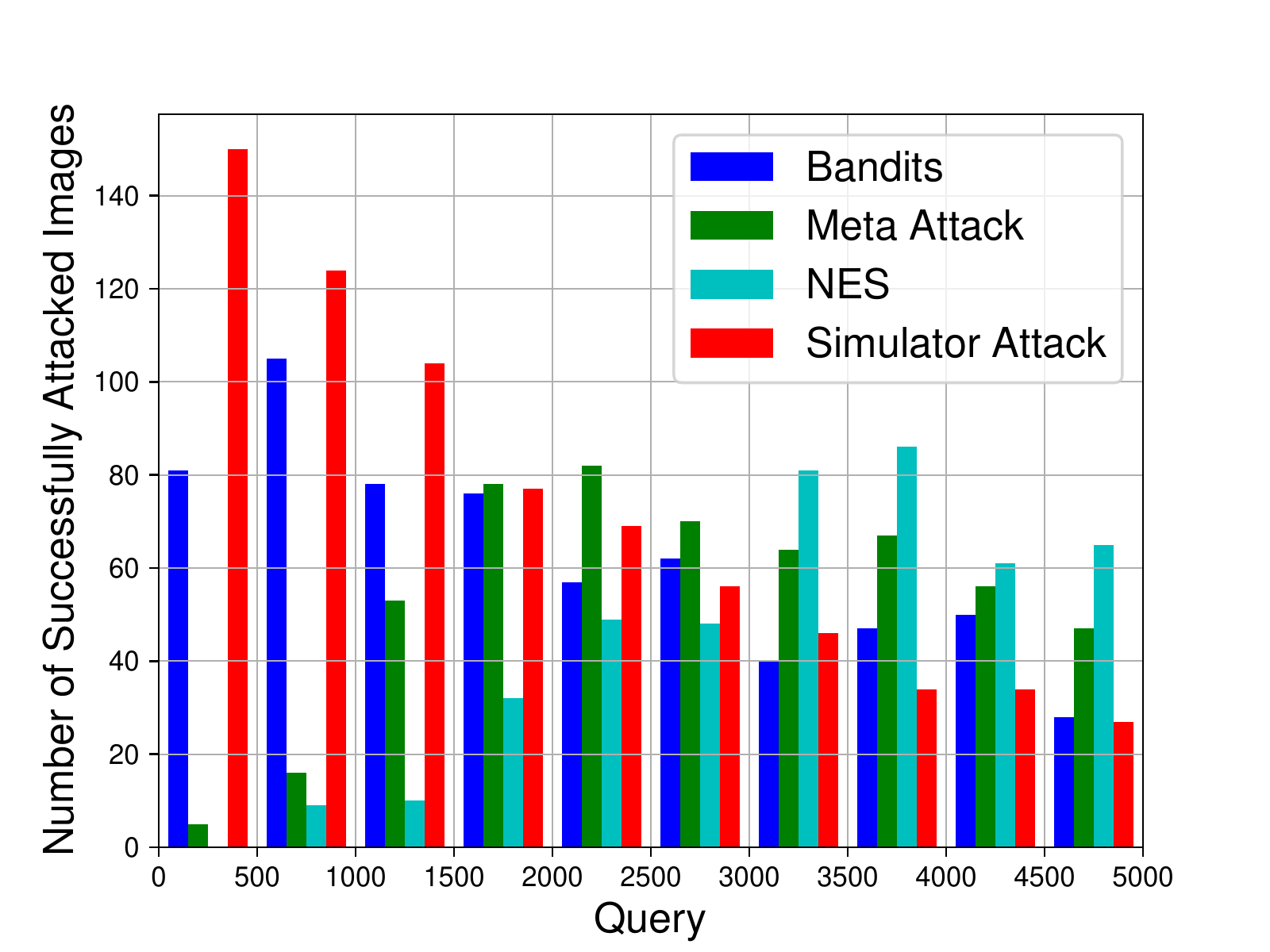}
		\subcaption{targeted $\ell_\infty$ attack PyramidNet-272}
	\end{minipage}
	\begin{minipage}[b]{.245\textwidth}
		\includegraphics[width=\linewidth]{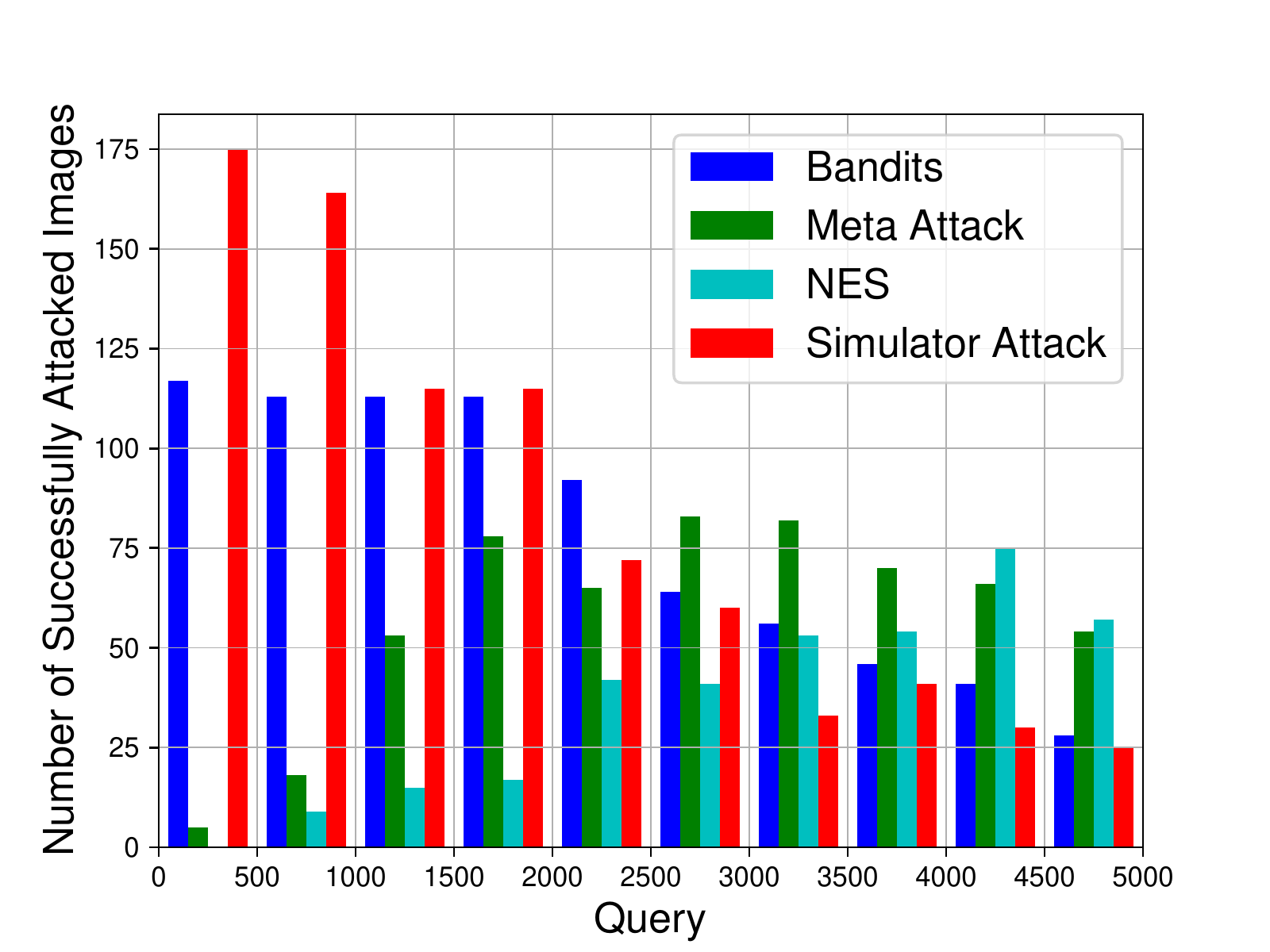}
		\subcaption{targeted $\ell_\infty$ attack GDAS}
	\end{minipage}
	\begin{minipage}[b]{.245\textwidth}
		\includegraphics[width=\linewidth]{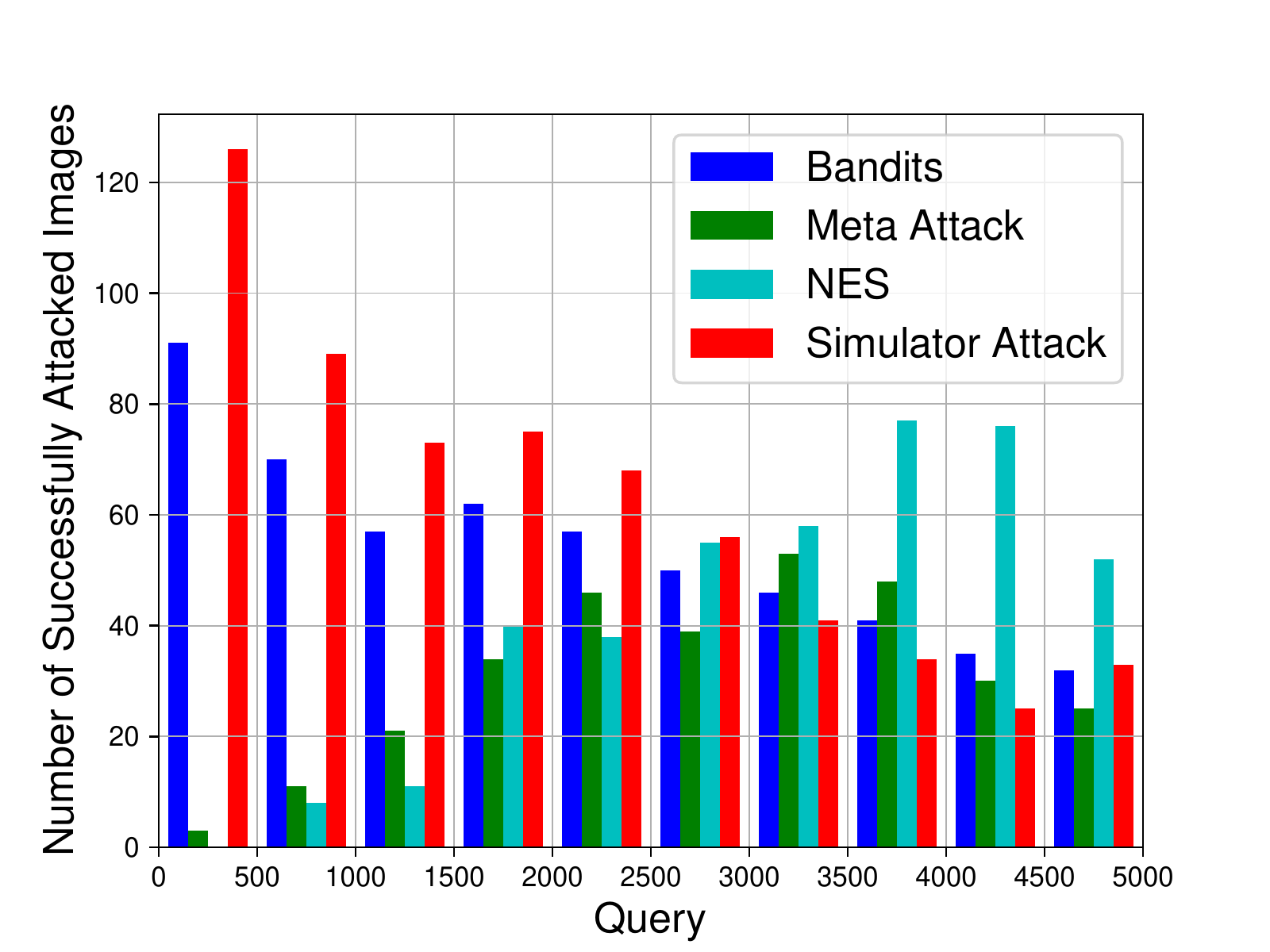}
		\subcaption{targeted $\ell_\infty$ attack WRN-28}
	\end{minipage}
	\begin{minipage}[b]{.245\textwidth}
		\includegraphics[width=\linewidth]{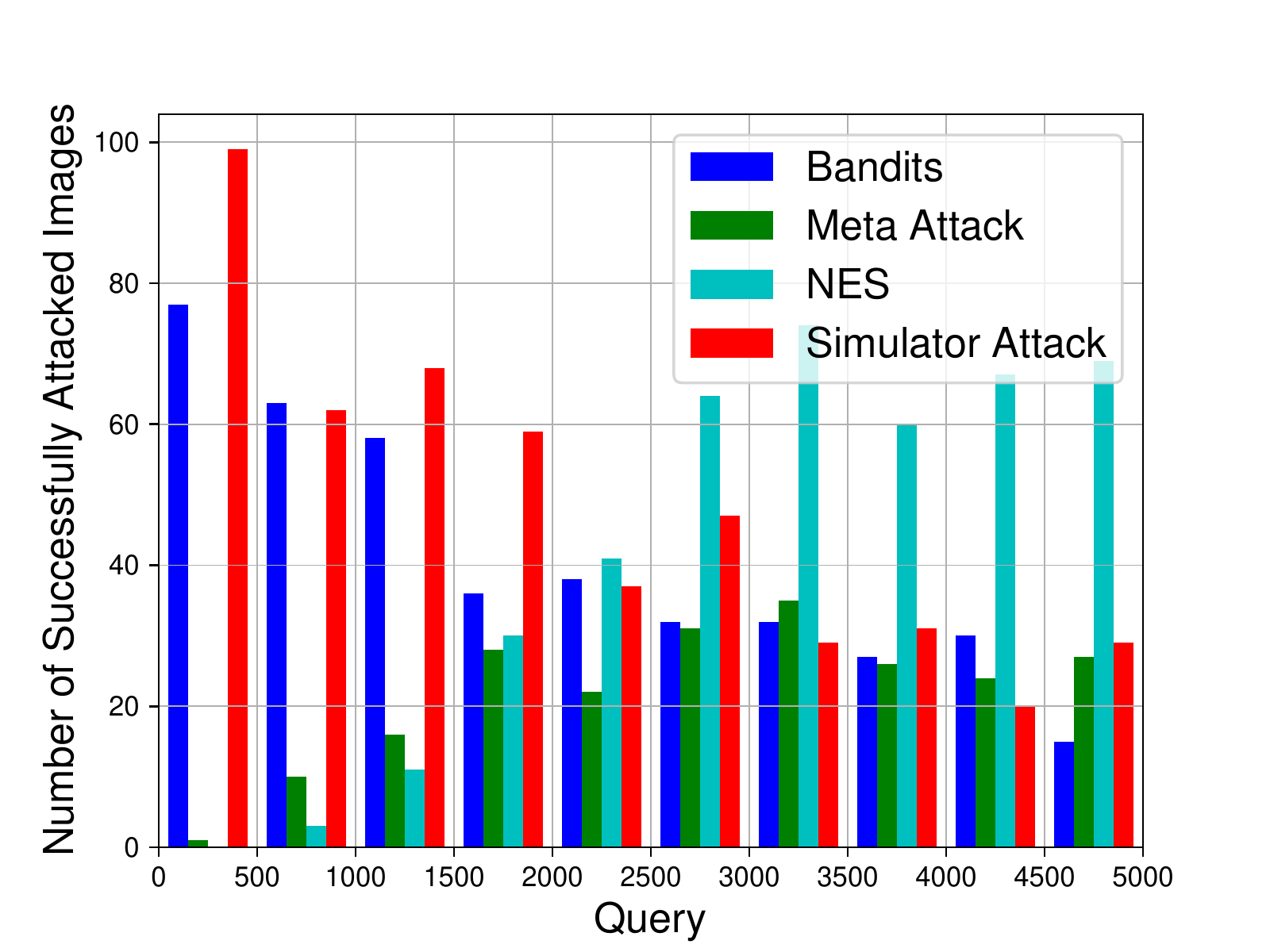}
		\subcaption{targeted $\ell_\infty$ attack WRN-40}
	\end{minipage}
	\caption{The histogram of query number in the CIFAR-100 dataset.}
	\label{fig:histogram_CIFAR-100}
\end{figure*}

\begin{figure*}[t]
	\setlength{\abovecaptionskip}{0pt}
	\setlength{\belowcaptionskip}{0pt}
	\captionsetup[sub]{font={scriptsize}}
	\centering 
	\begin{minipage}[b]{.3\textwidth}
		\includegraphics[width=\linewidth]{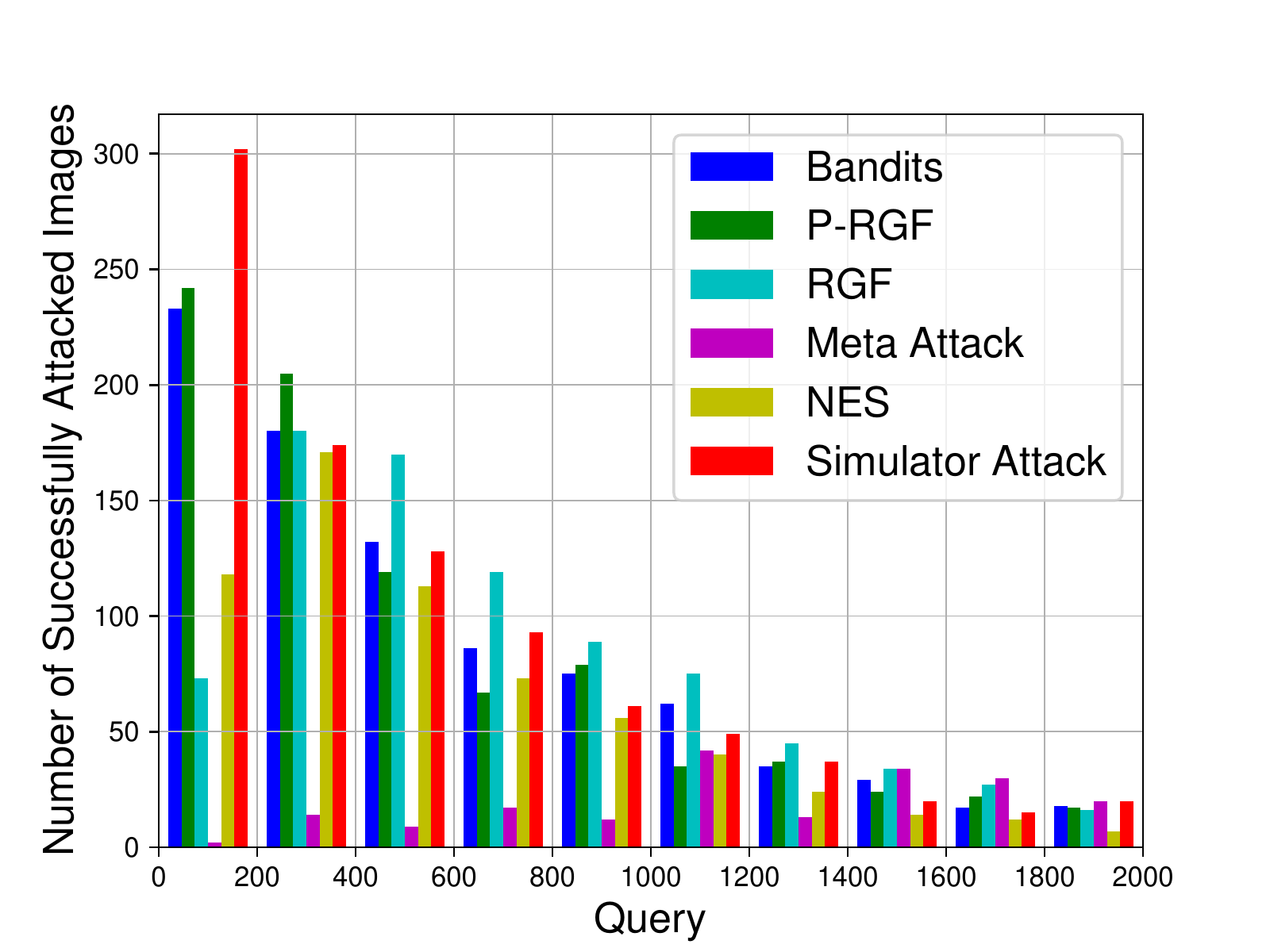}
		\subcaption{untargeted $\ell_\infty$ attack DenseNet-121}
	\end{minipage}
	\begin{minipage}[b]{.3\textwidth}
		\includegraphics[width=\linewidth]{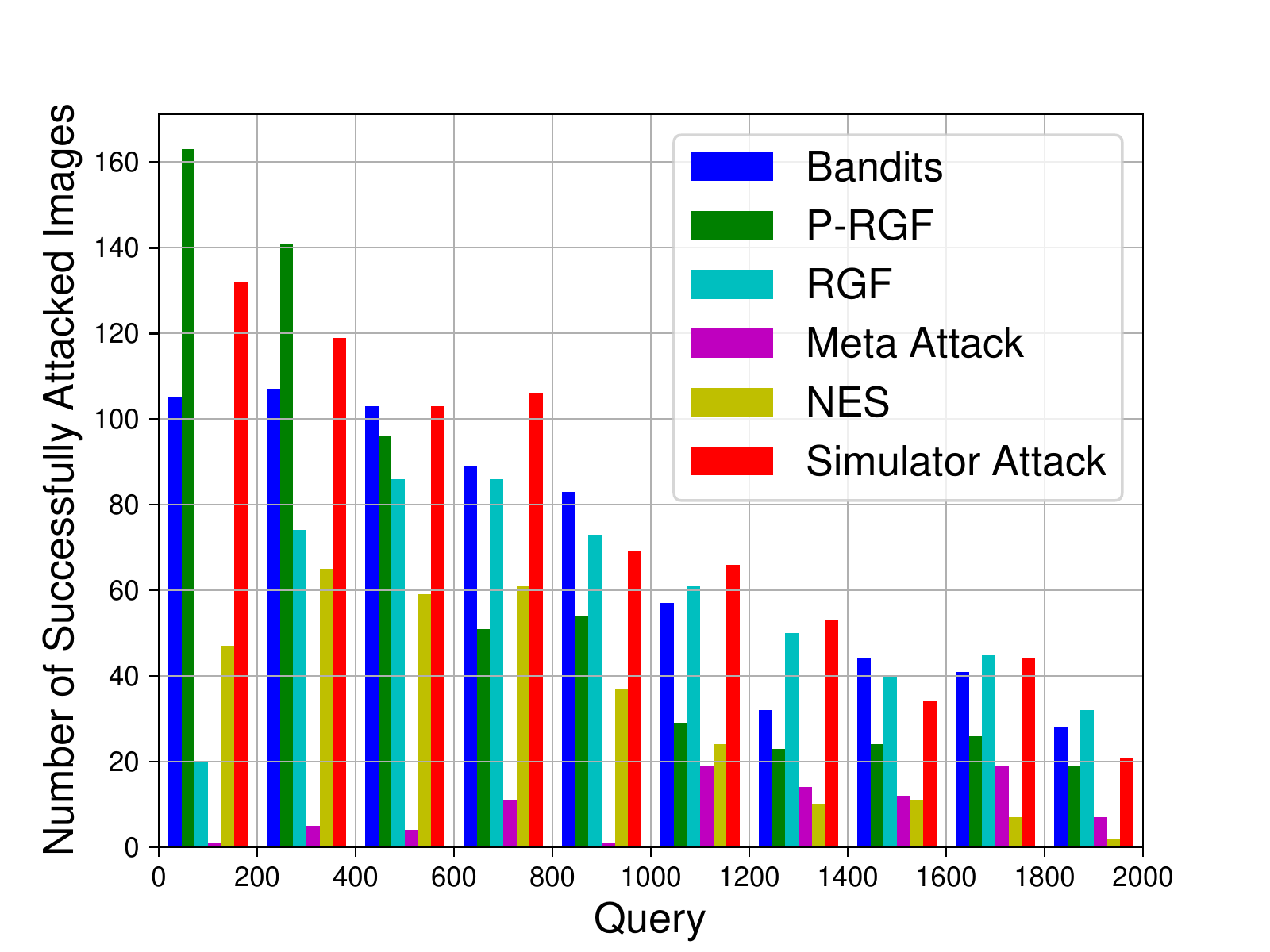}
		\subcaption{untargeted $\ell_\infty$ attack ResNext-101(32$\times$4d)}
	\end{minipage}
	\begin{minipage}[b]{.3\textwidth}
		\includegraphics[width=\linewidth]{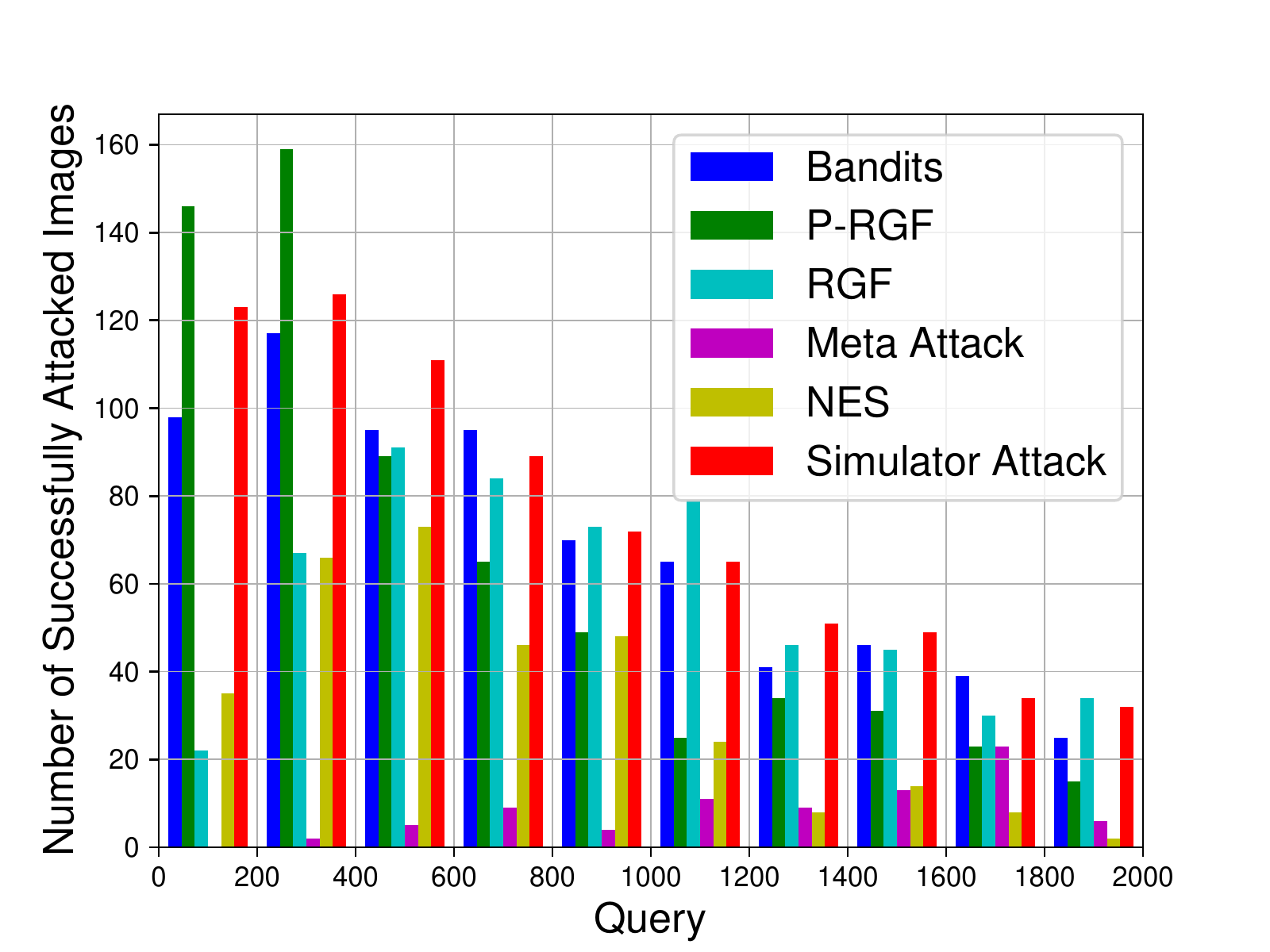}
		\subcaption{untargeted $\ell_\infty$ attack ResNext-101(64$\times$4d)}
	\end{minipage}
	\begin{minipage}[b]{.3\textwidth}
		\includegraphics[width=\linewidth]{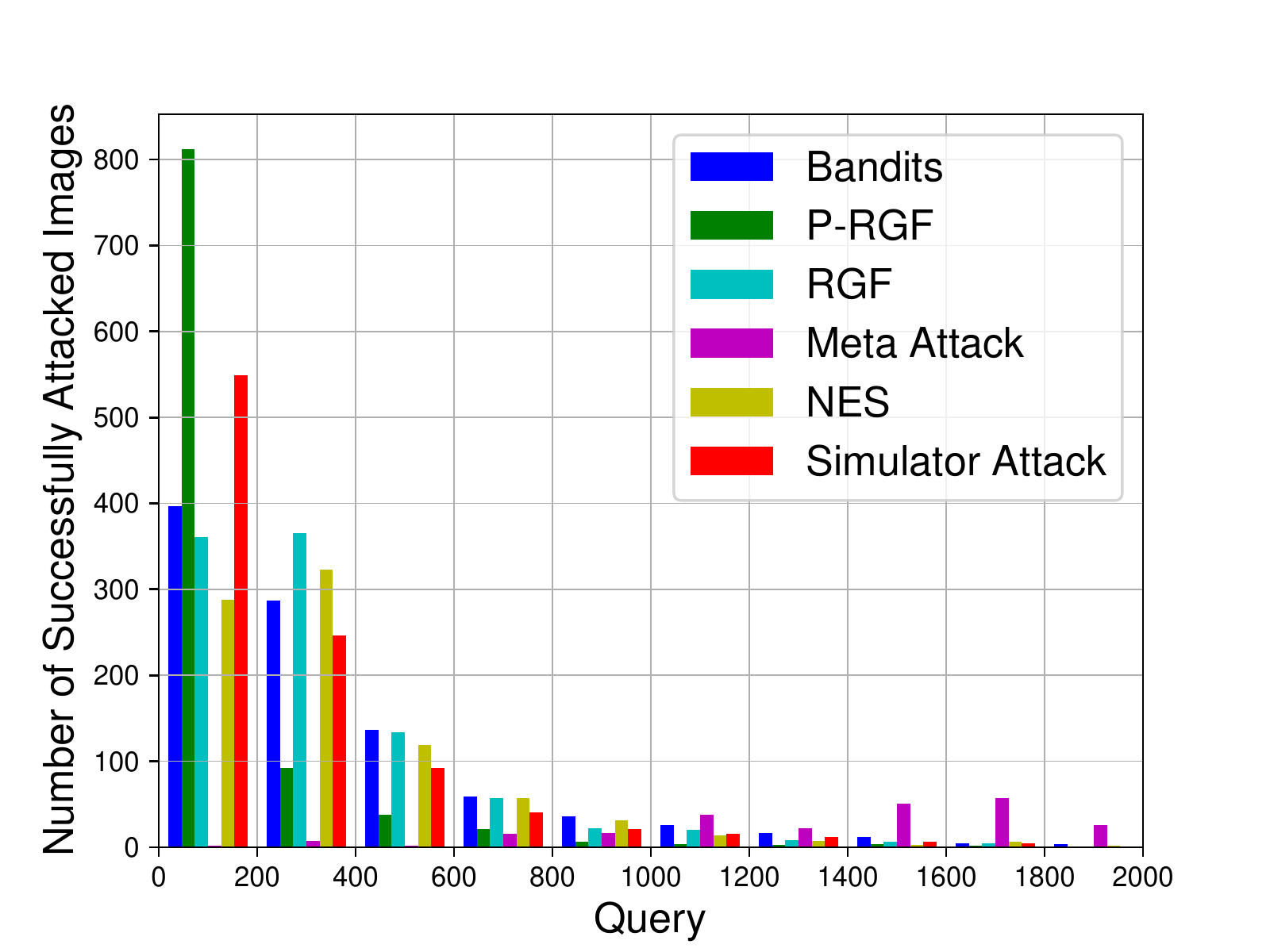}
		\subcaption{untargeted $\ell_2$ attack DenseNet-121}
	\end{minipage}
	\begin{minipage}[b]{.3\textwidth}
		\includegraphics[width=\linewidth]{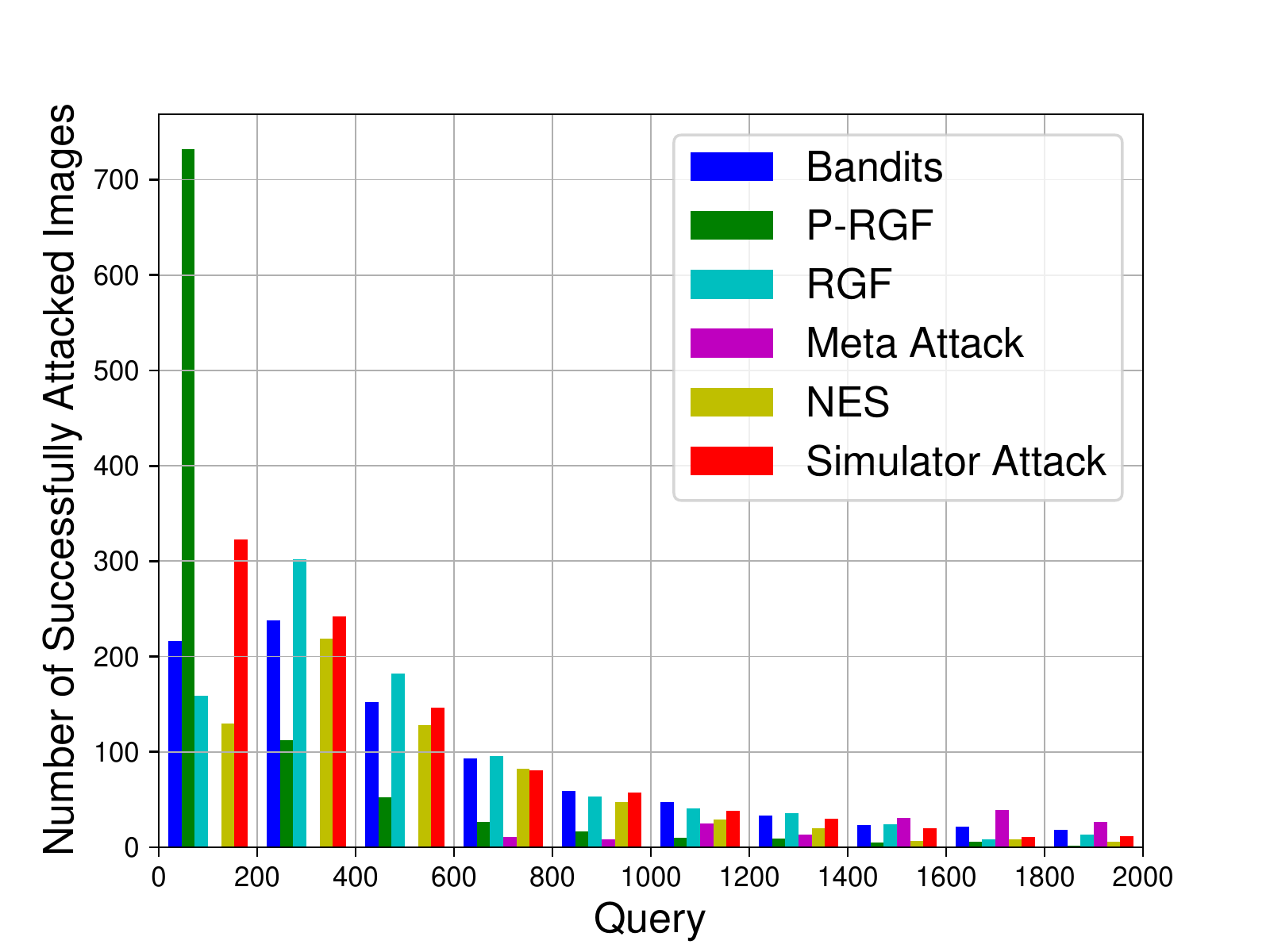}
		\subcaption{untargeted $\ell_2$ attack ResNext-101(32$\times$4d)}
	\end{minipage}
	\begin{minipage}[b]{.3\textwidth}
		\includegraphics[width=\linewidth]{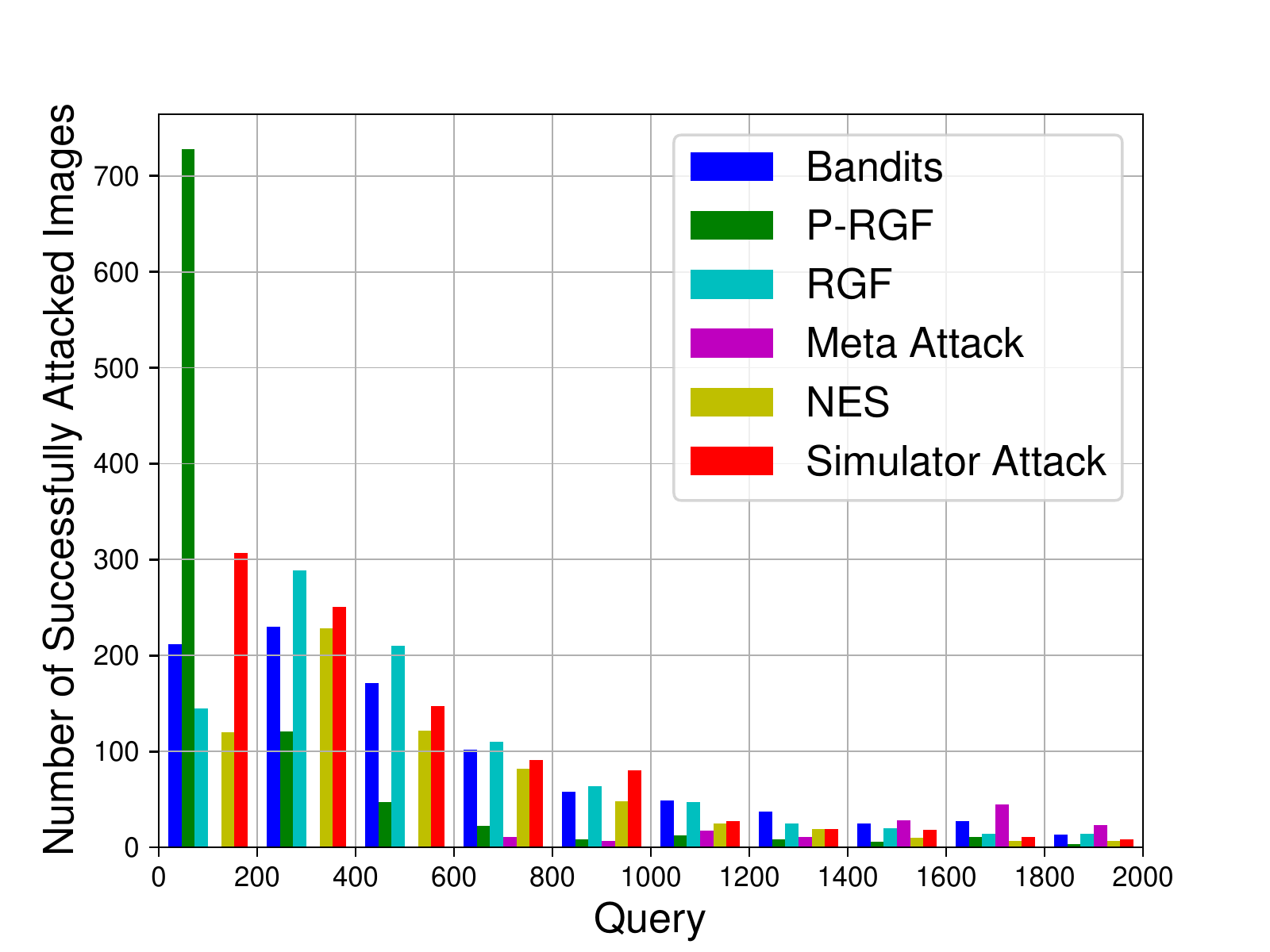}
		\subcaption{untargeted $\ell_2$ attack ResNext-101(64$\times$4d)}
	\end{minipage}
	\begin{minipage}[b]{.3\textwidth}
		\includegraphics[width=\linewidth]{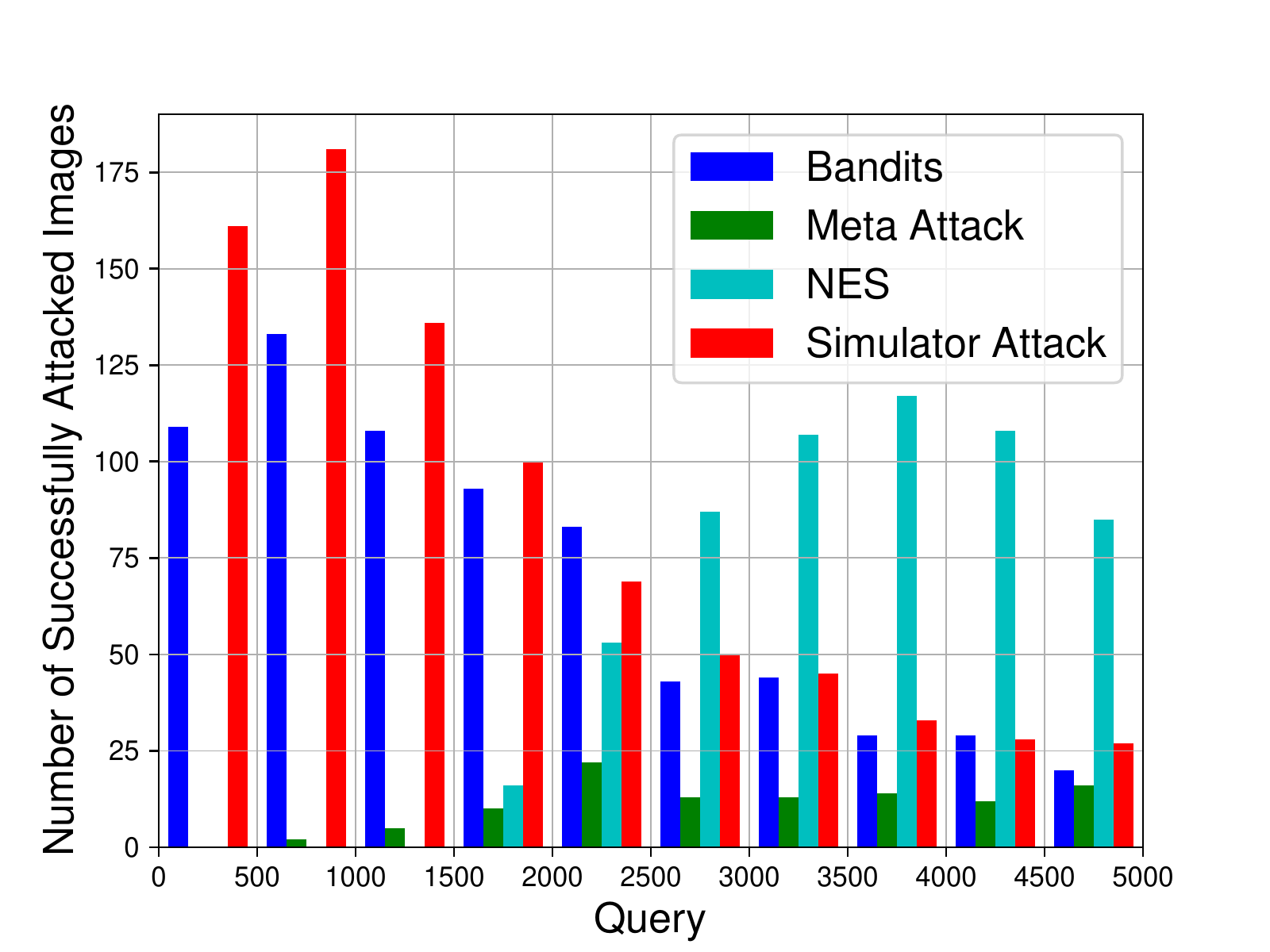}
		\subcaption{targeted $\ell_2$ attack DenseNet-121}
	\end{minipage}
	\begin{minipage}[b]{.3\textwidth}
		\includegraphics[width=\linewidth]{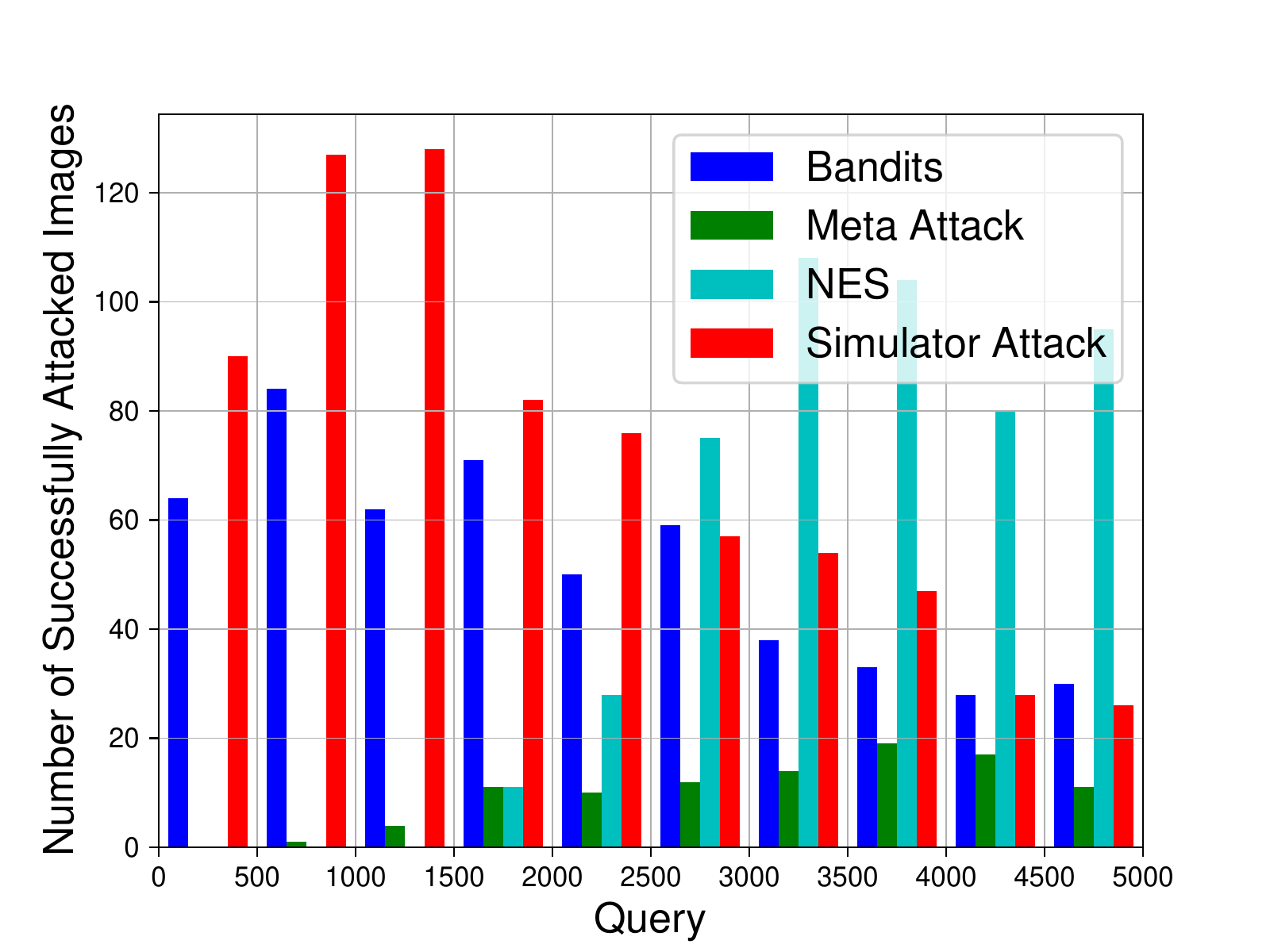}
		\subcaption{targeted $\ell_2$ attack ResNext-101(32$\times$4d)}
	\end{minipage}
	\begin{minipage}[b]{.3\textwidth}
		\includegraphics[width=\linewidth]{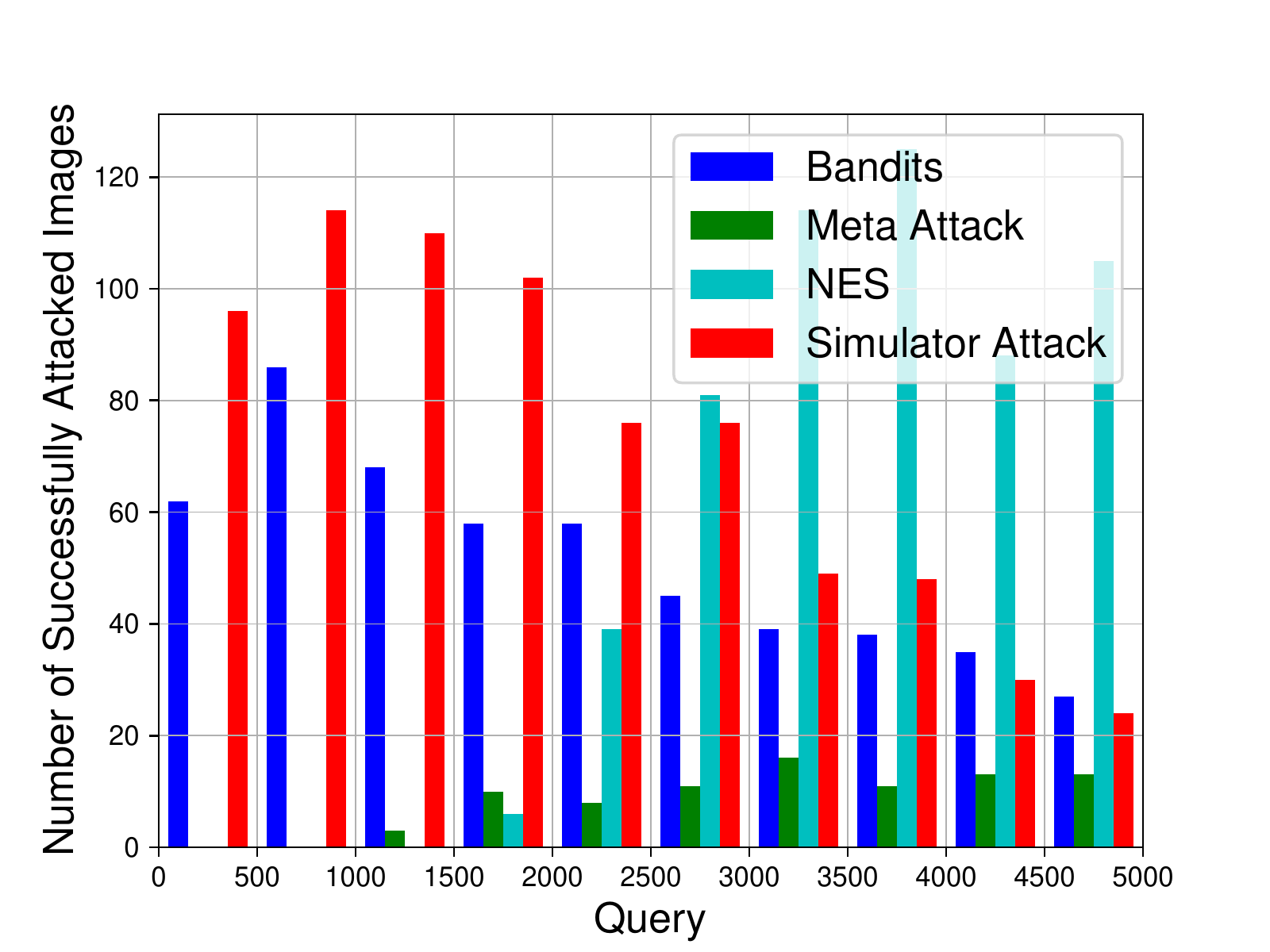}
		\subcaption{targeted $\ell_2$ attack ResNext-101(64$\times$4d)}
	\end{minipage}
	\caption{The histogram of query number in the TinyImageNet dataset.}
	\label{fig:histogram_TinyImageNet}
\end{figure*}

\begin{figure*}[t]
	\setlength{\abovecaptionskip}{0pt}
	\setlength{\belowcaptionskip}{0pt}
	\captionsetup[sub]{font={scriptsize}}
	\centering 
	\begin{minipage}[b]{.3\textwidth}
		\includegraphics[width=\linewidth]{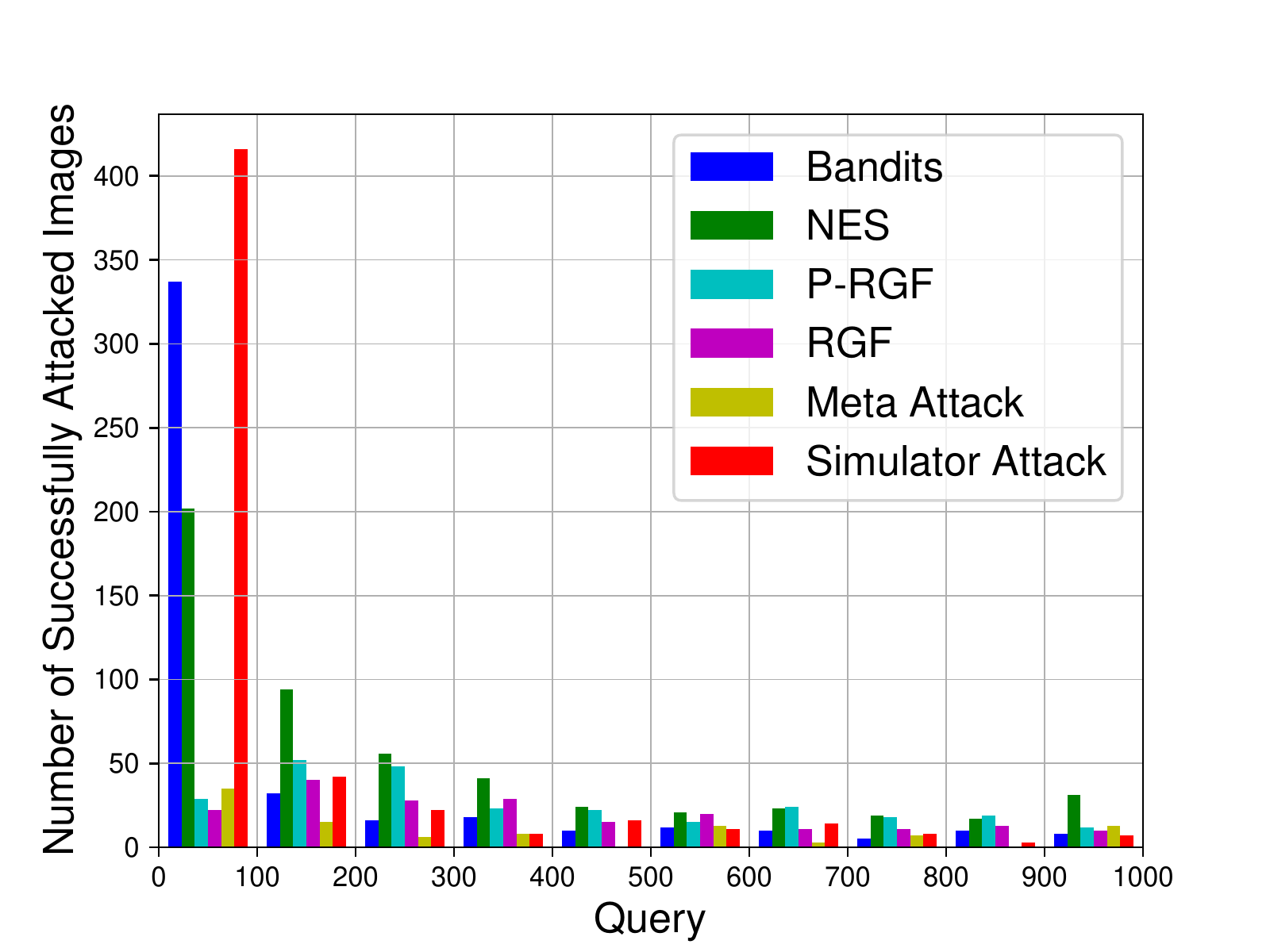}
		\subcaption{attack ComDefend in CIFAR-10}
	\end{minipage}
	\begin{minipage}[b]{.3\textwidth}
		\includegraphics[width=\linewidth]{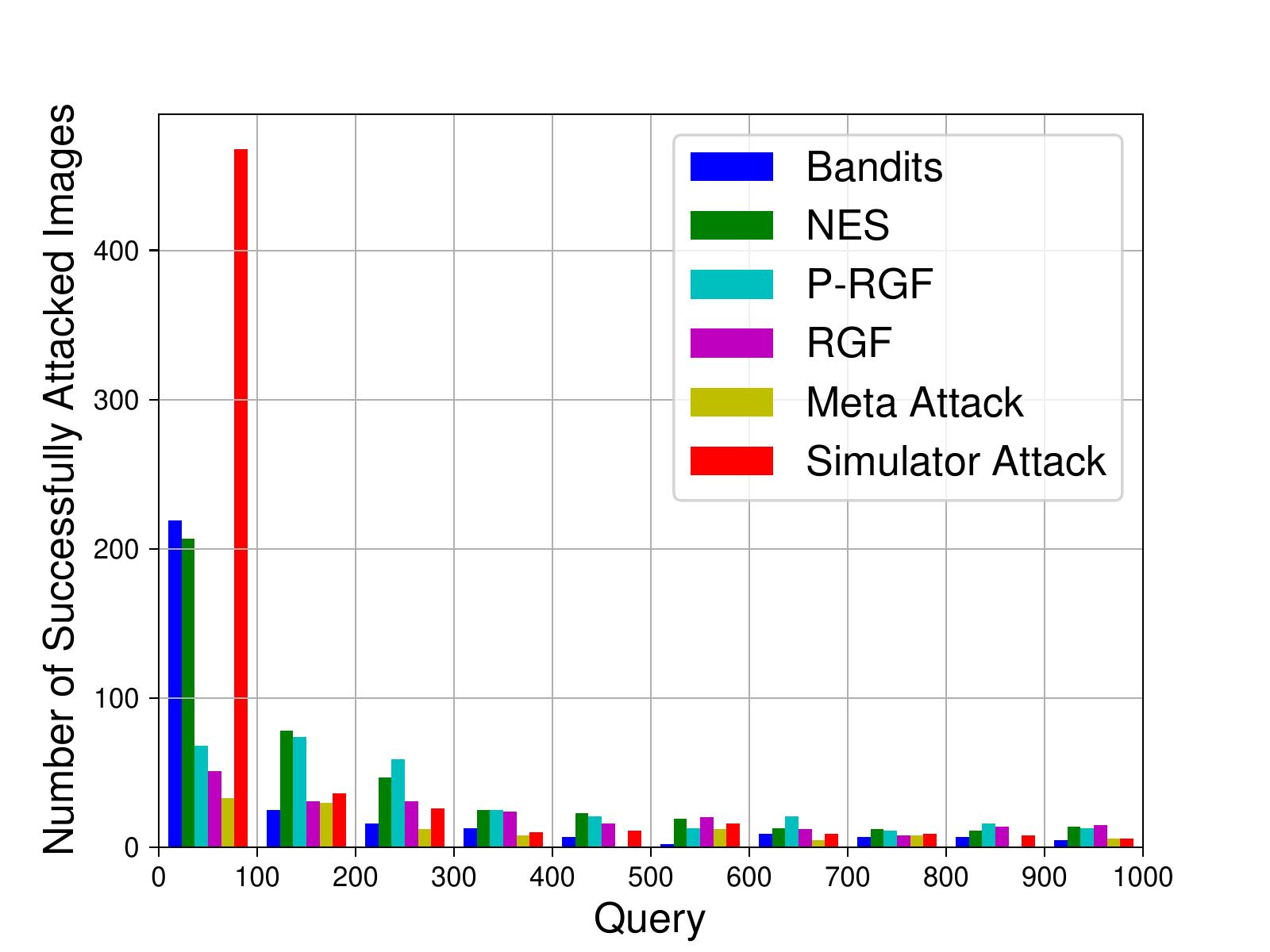}
		\subcaption{attack Feature Distillation in CIFAR-10}
	\end{minipage}
	\begin{minipage}[b]{.3\textwidth}
		\includegraphics[width=\linewidth]{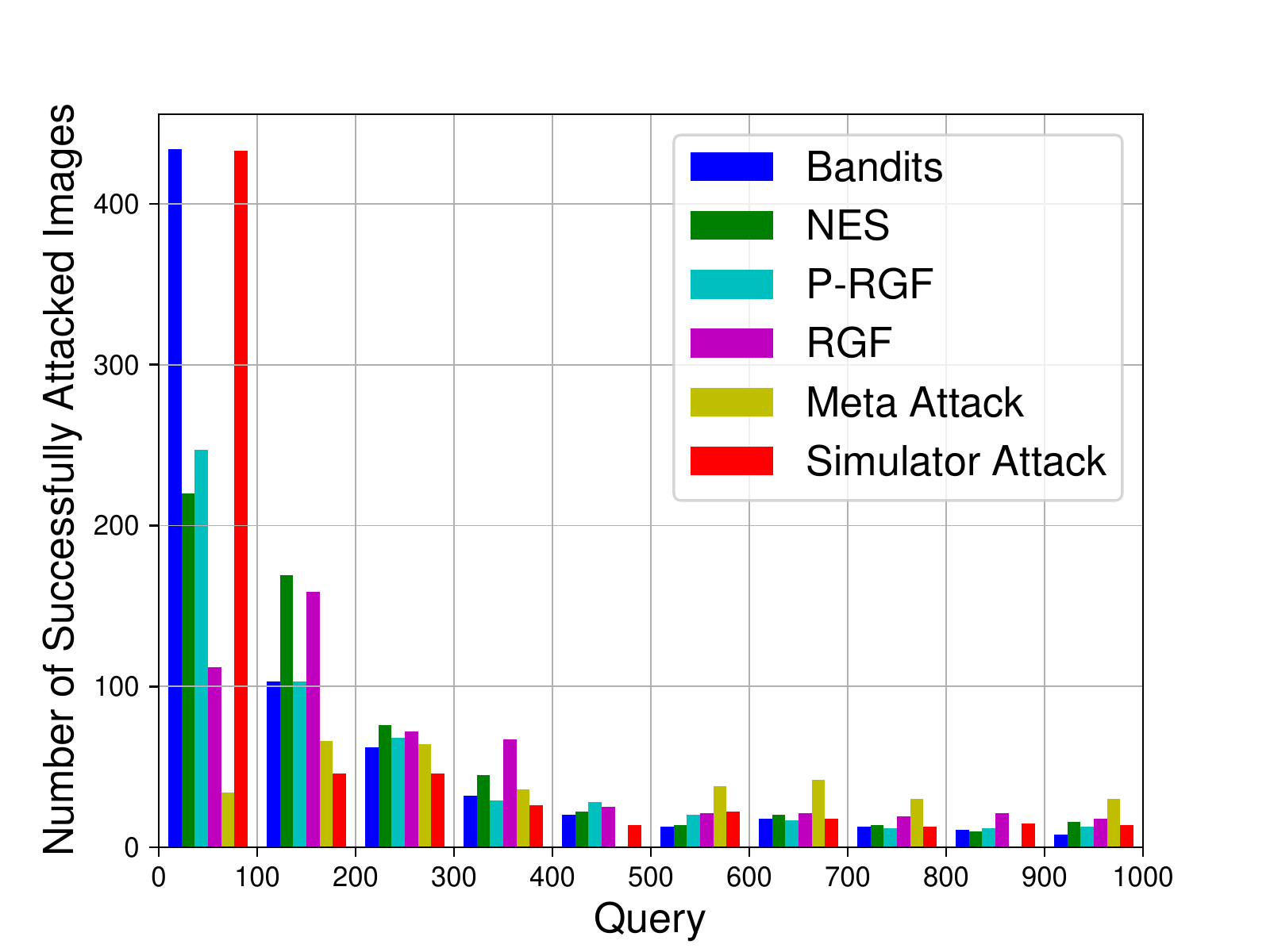}
		\subcaption{attack PCL in CIFAR-10}
	\end{minipage}
	\begin{minipage}[b]{.3\textwidth}
		\includegraphics[width=\linewidth]{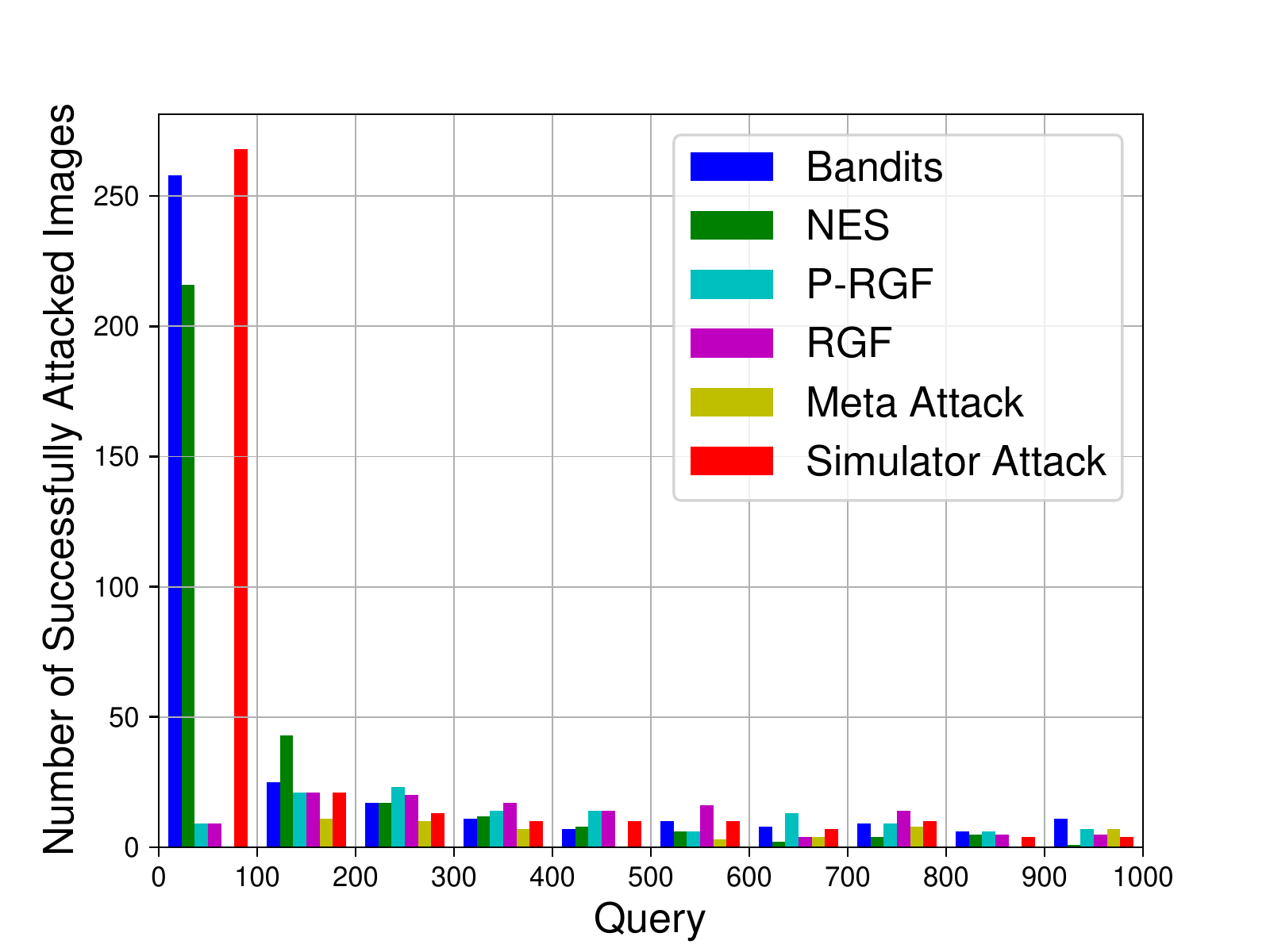}
		\subcaption{attack Adv Train in CIFAR-10}
	\end{minipage}
	\begin{minipage}[b]{.3\textwidth}
		\includegraphics[width=\linewidth]{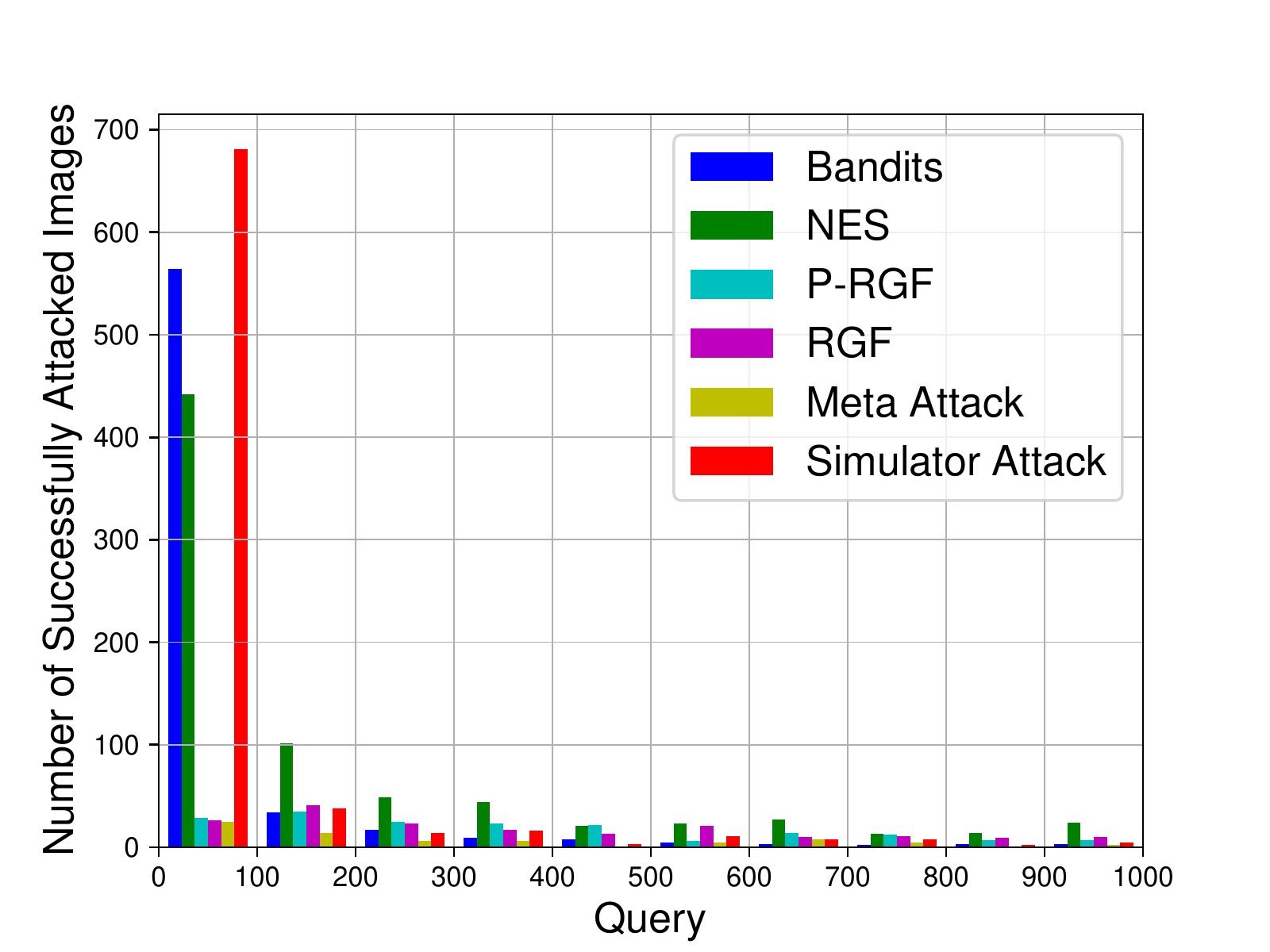}
		\subcaption{attack ComDefend in CIFAR-100}
	\end{minipage}
	\begin{minipage}[b]{.3\textwidth}
		\includegraphics[width=\linewidth]{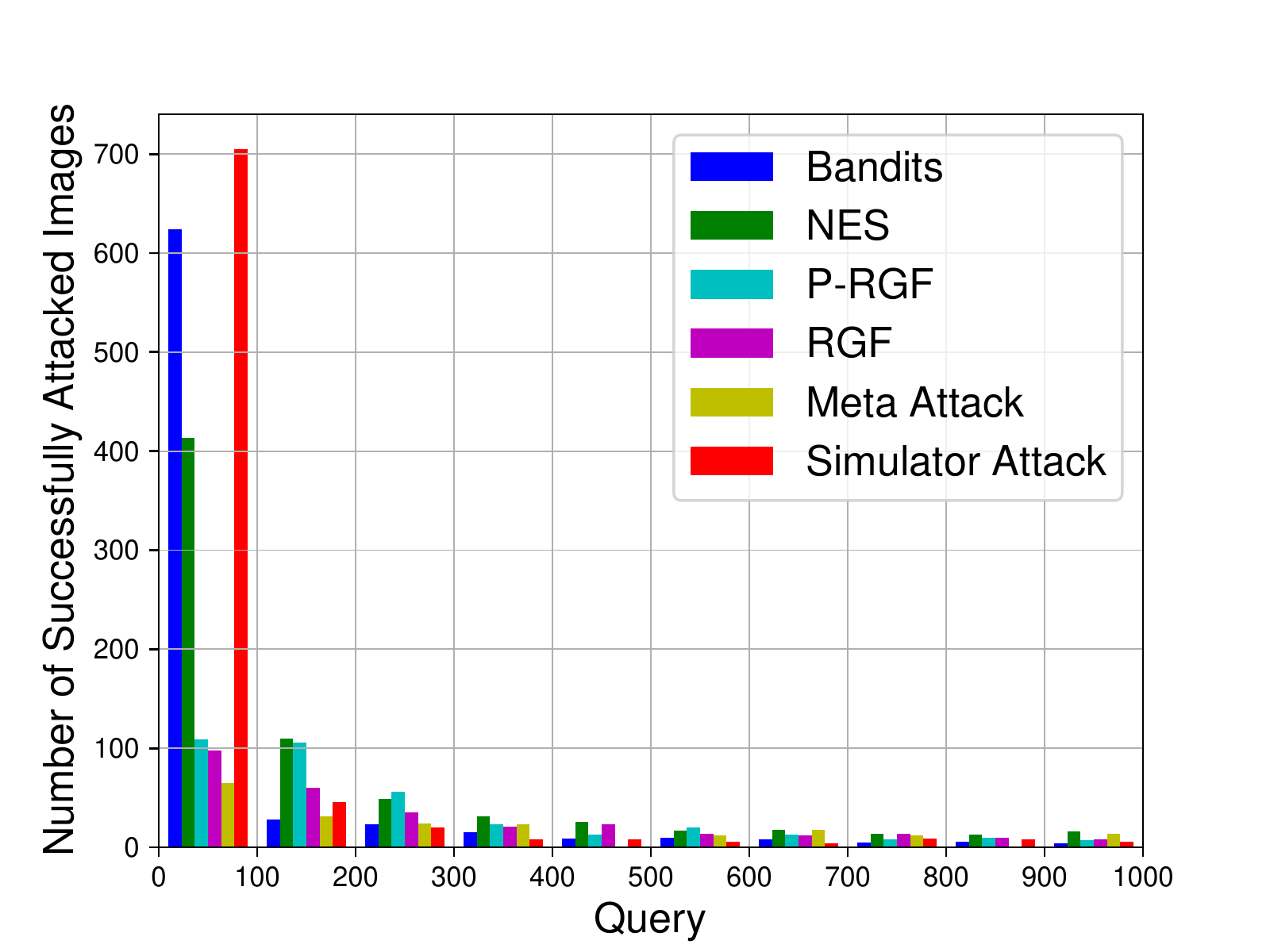}
		\subcaption{attack Feature Distillation in CIFAR-100}
	\end{minipage}
	\begin{minipage}[b]{.3\textwidth}
		\includegraphics[width=\linewidth]{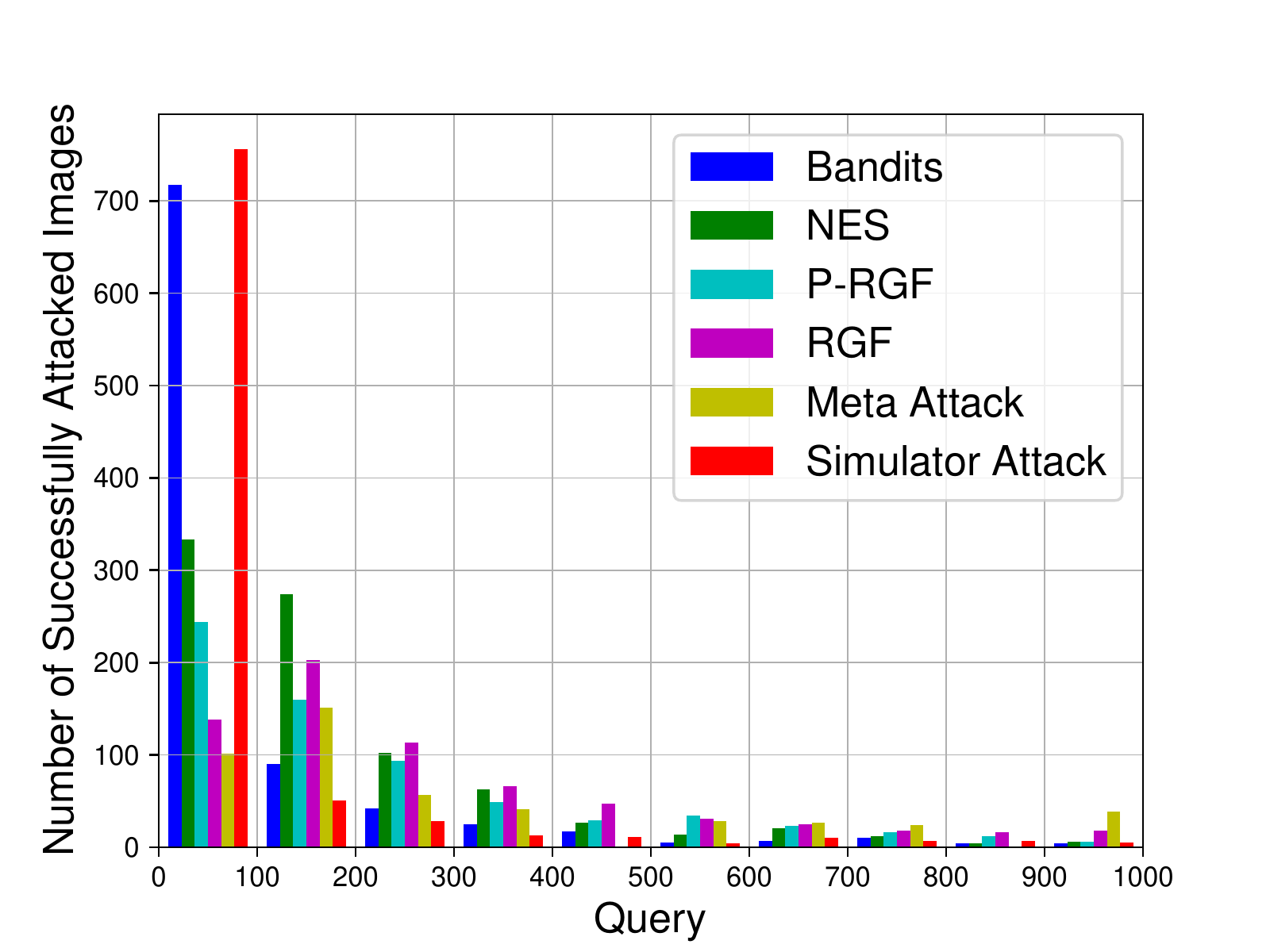}
		\subcaption{attack PCL in CIFAR-100}
	\end{minipage}
	\begin{minipage}[b]{.3\textwidth}
		\includegraphics[width=\linewidth]{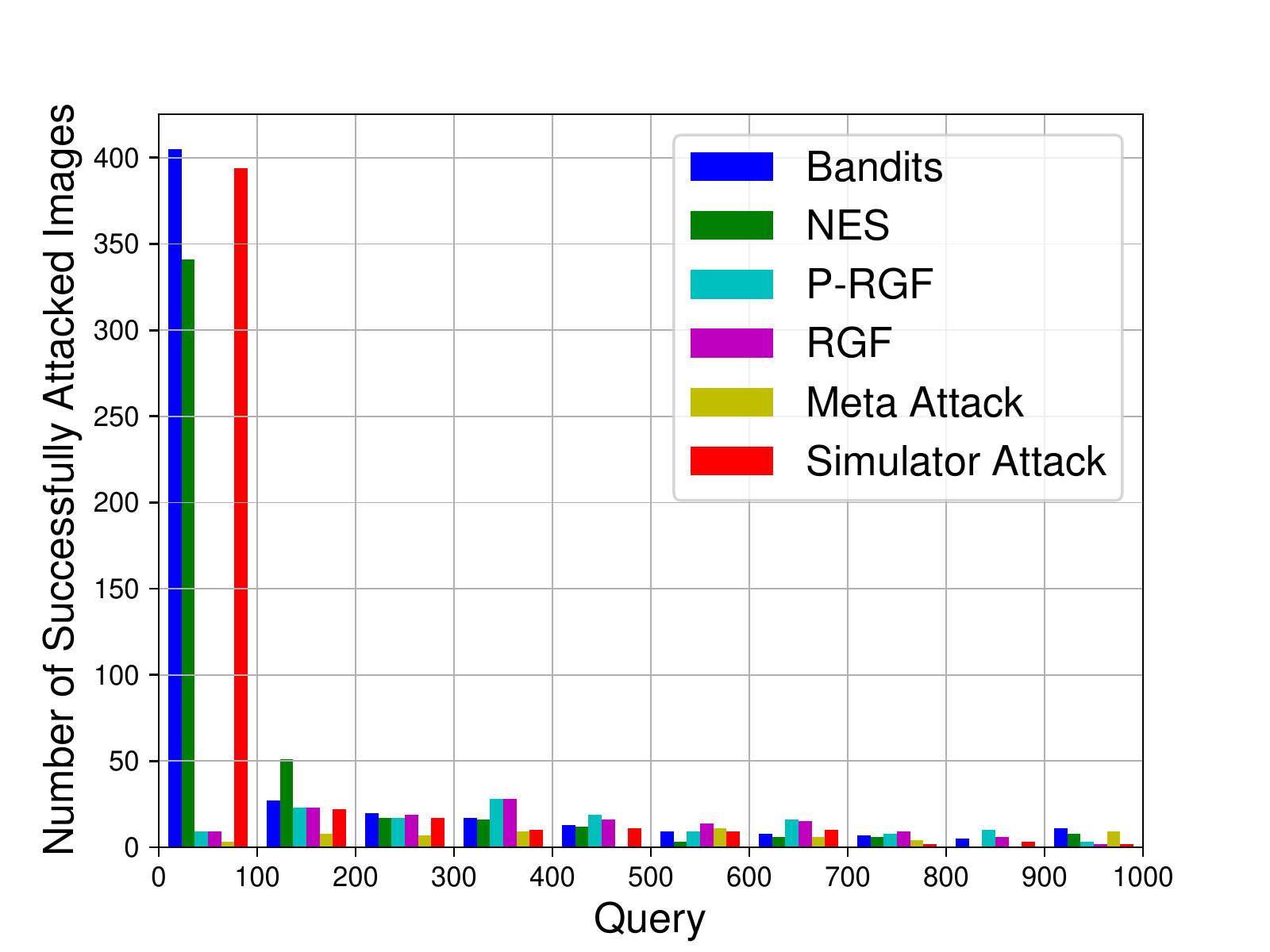}
		\subcaption{attack Adv Train in CIFAR-100}
	\end{minipage}
	\begin{minipage}[b]{.3\textwidth}
		\includegraphics[width=\linewidth]{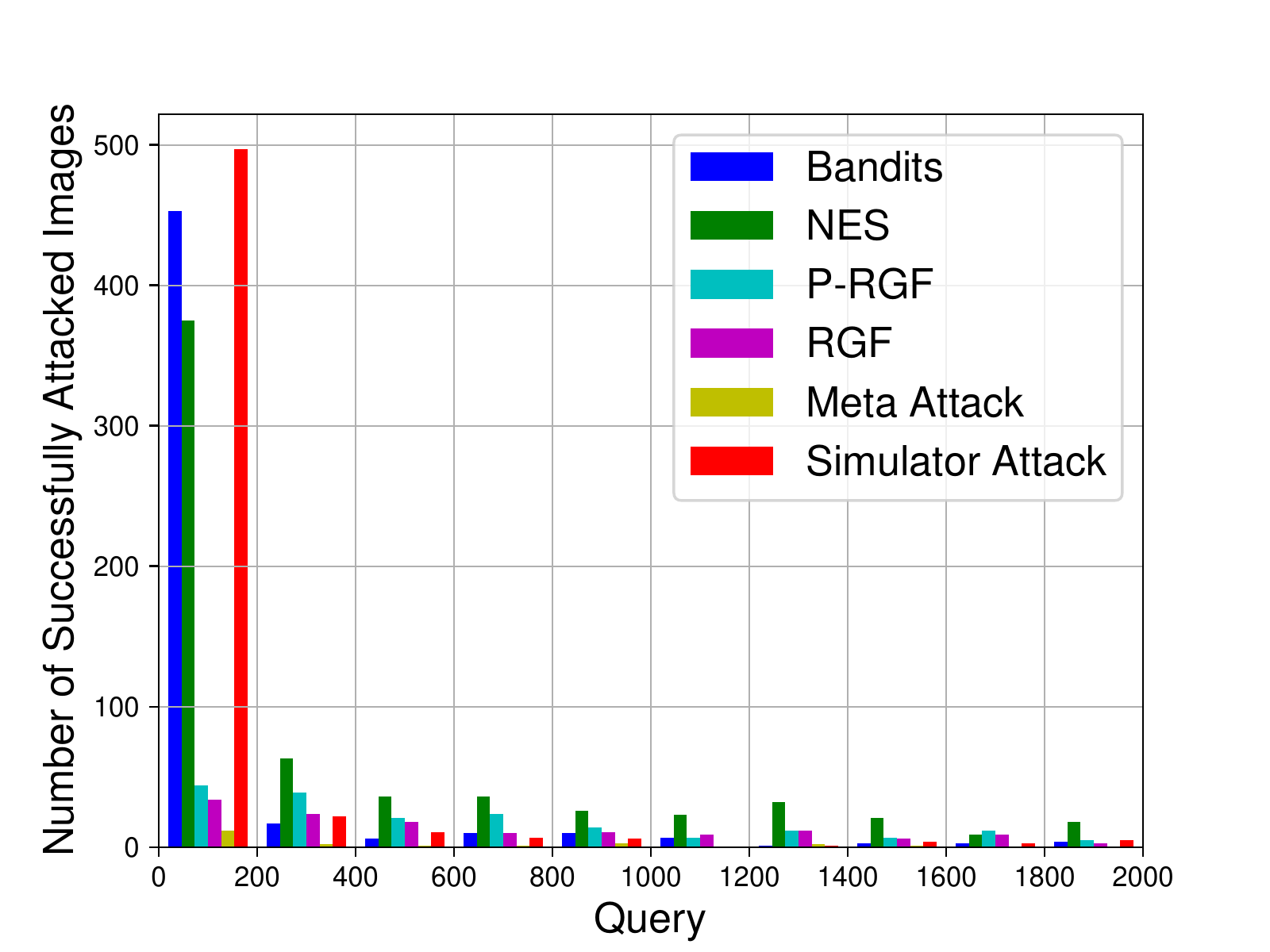}
		\subcaption{attack ComDefend in TinyImageNet}
	\end{minipage}
	\begin{minipage}[b]{.3\textwidth}
		\includegraphics[width=\linewidth]{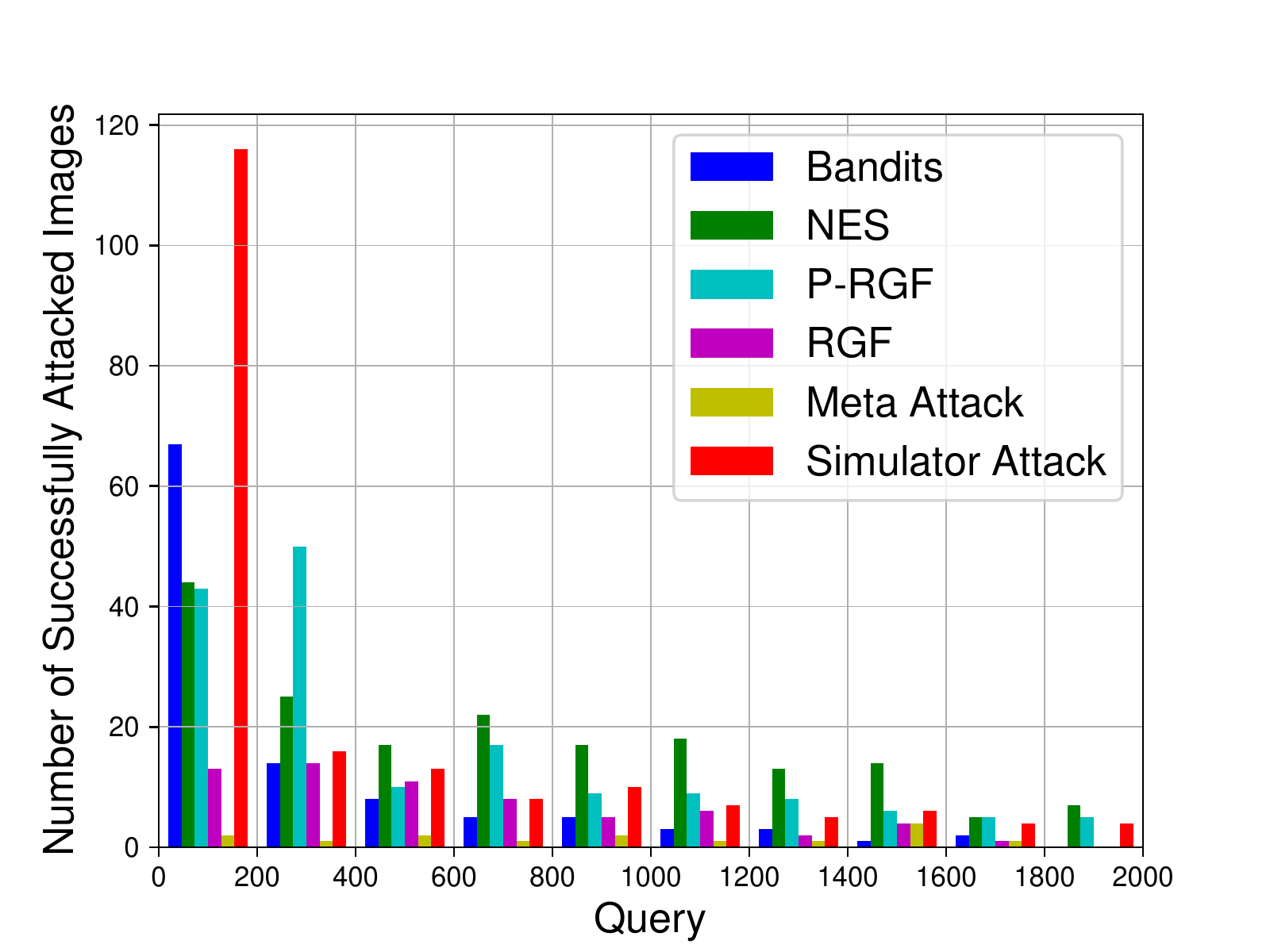}
		\subcaption{attack Feature Distillation in TinyImageNet}
	\end{minipage}
	\begin{minipage}[b]{.3\textwidth}
		\includegraphics[width=\linewidth]{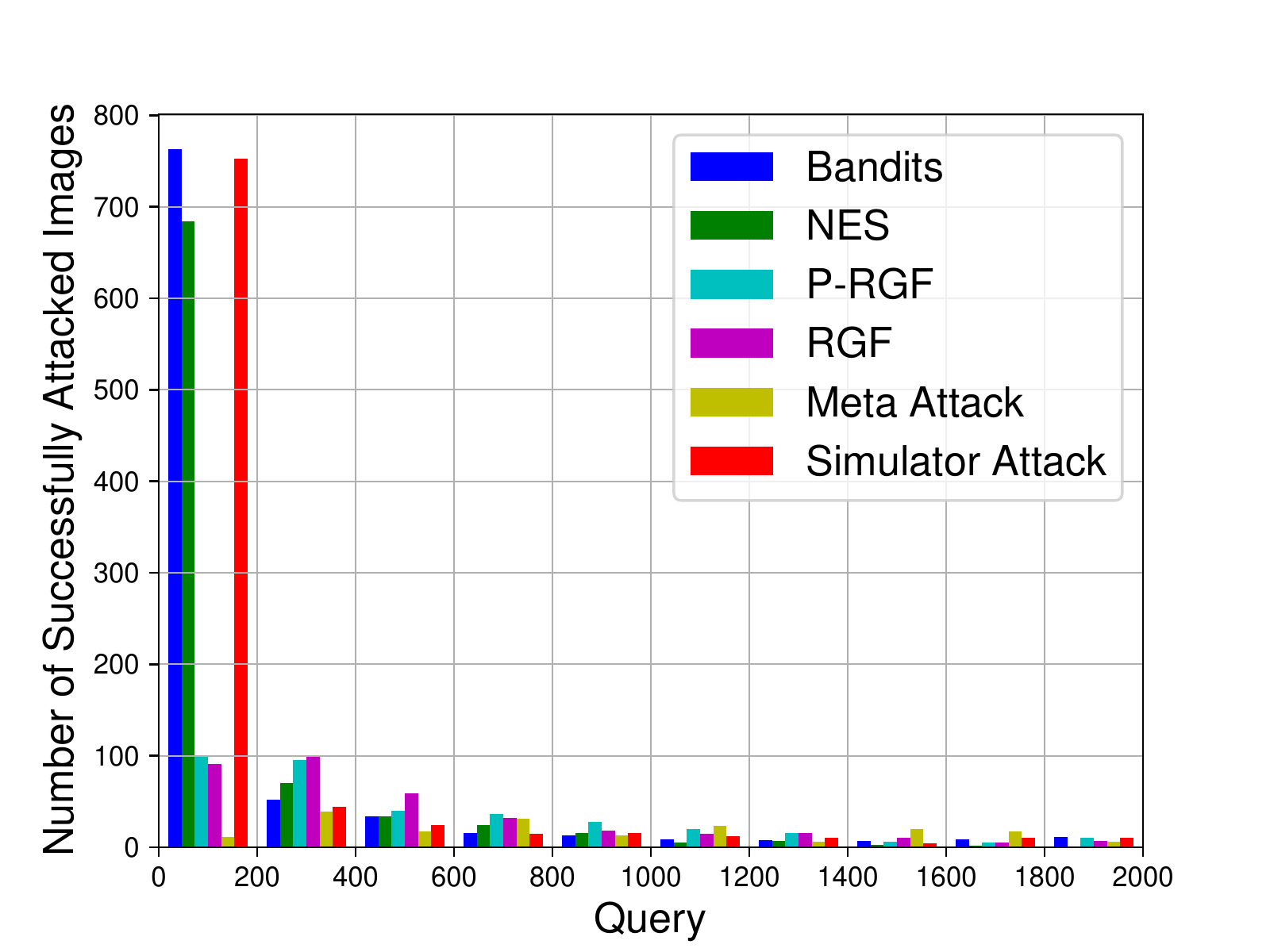}
		\subcaption{attack PCL in TinyImageNet}
	\end{minipage}
	\begin{minipage}[b]{.3\textwidth}
		\includegraphics[width=\linewidth]{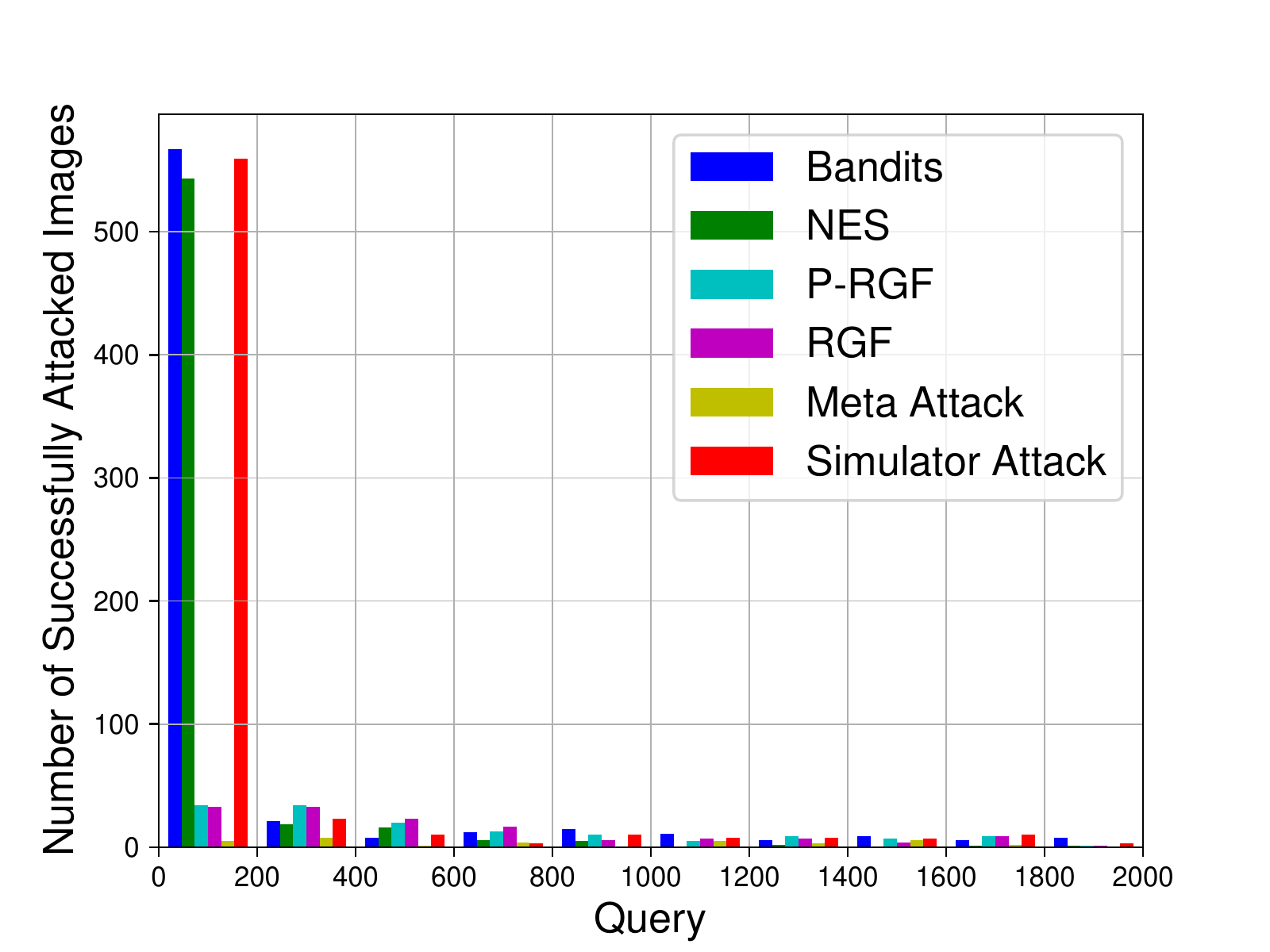}
		\subcaption{attack Adv Train in TinyImageNet}
	\end{minipage}
	\caption{The histogram of query number on defensive models with the backbone of ResNet-50. The experimental results are obtained by performing the untargeted attacks under $\ell_\infty$ norm.}
	\label{fig:histogram_defensive_models}
\end{figure*}

\noindent\textbf{Histogram of Query Numbers.} To observe the distribution of query numbers in detail, we collect the query number of each adversarial example to draw the histogram figures. Specifically, we divide the range of query number into 10 intervals, and then count the number of samples in each interval. These intervals are separated by the vertical lines of figures. Each bar indicates one attack, and its height indicates the number of samples with the queries belong to this query interval. Figs. \ref{fig:histogram_CIFAR-10}, \ref{fig:histogram_CIFAR-100}, \ref{fig:histogram_TinyImageNet}, and \ref{fig:histogram_defensive_models} show the histograms of query numbers in the CIFAR-10, CIFAR-100, and TinyImageNet datasets, respectively. The results demonstrates that the highest red bars (the Simulator Attack) are located in the area with low number of queries, which confirms that most adversarial examples of the Simulator Attack have the minimum number of queries.

\end{document}